%% file: main.tex
\pdfoutput=1

\documentclass[letterpaper, 10 pt, conference]{ieeeconf}  

\IEEEoverridecommandlockouts                                                                                        
\input{symbols/preamble_packages.tex}
\usepackage{moreverb,url}
\PassOptionsToPackage{hyphens}{url}
\usepackage[colorlinks,bookmarksopen,bookmarksnumbered,bookmarksdepth=3,citecolor=red,urlcolor=red]{hyperref}
\usepackage{prettyref}
\newcommand\BibTeX{{\rmfamily B\kern-.05em \textsc{i\kern-.025em b}\kern-.08em
T\kern-.1667em\lower.7ex\hbox{E}\kern-.125emX}}
\input{symbols/preamble_symbols.tex}

\input{symbols/shortcuts}

\newfloatcommand{capbtabbox}{table}[][\FBwidth]
\makeatletter
\def\thm@space@setup{  \thm@preskip=\parskip \thm@postskip=0mm}
\makeatother

\graphicspath{{./figures/},{./figures/resultsIROS/paperFigures/},{./figures/resultsIROS/}}

\title{\LARGE \textbf {
\\Sparse Depth Sensing for Resource-Constrained Robots
}}

\author{Fangchang Ma, Luca Carlone, Ulas Ayaz, Sertac Karaman\forFinalVersion{\thanks{This work was partially funded \red{????}}}\thanks{F. Ma, L. Carlone, U. Ayaz, and S. Karaman are with 
  the Laboratory for Information \& Decision Systems, Massachusetts 
  Institute of Technology, Cambridge, MA, USA, 
   {\sf \{fcma,lcarlone,uayaz,sertac\}@mit.edu}}}

\begin{document}

\maketitle
\thispagestyle{plain}
\pagestyle{plain}
\input{abstract}
\IEEEpeerreviewmaketitle
\input{supplementalInfo}
\input{introduction}

\input{related_work}

\input{preliminaries}

\input{formulation/problemFormulation}

  \input{formulation/problemFormulation_2D}
\input{formulation/problemFormulation_3D}

  \input{formulation/problemFormulation_unified}

\input{analysis/analysisIntro}

  \input{analysis/BP_exactRecovery}
  \input{analysis/BP_noiseless}
    \input{analysis/BP_noiseless_algebraic}

    \input{analysis/BP_noiseless_2D}

    \input{analysis/BP_noiseless_3D}

\input{analysis/BP_noisy}

    \input{analysis/BP_noisy_algebraic}

    \input{analysis/BP_noisy_2D}
    \input{analysis/BP_noisy_3D}
\input{algorithm/algorithms}
  \input{algorithm/algorithmicVariants}  
  \input{algorithm/solvers}

\input{experimentTex/experiments}

  \input{experimentTex/experiments_2d}

  \input{experimentTex/experiments_preliminary}

  \input{experimentTex/experiments_3dSparseReconstruction}

  \input{experimentTex/experiments_3dSparseReconstruction_multi}
  \input{experimentTex/experiments_dataCompression}
  \input{experimentTex/experiments_superResolution}
\input{conclusion}

\bibliographystyle{IEEEtran}
\bibliography{bib/refs_arxiv,bib/refs}

\newcommand{\surf}{{\tt surf}\xspace}

\newpage

\appendices
\input{appendixTex/appendix_lemmas}
%\input{appendixTex/appendix_duality}
\input{appendixTex/appendix_exact}

\input{appendixTex/appendix_exact_3D}
\input{appendixTex/appendix_noiseless_algebraic}
\input{appendixTex/appendix_noiseless_signConsistency2D}

\input{appendixTex/appendix_noiseless_errorBound2D}
\input{appendixTex/appendix_noiseless_signConsistency3D}
\input{appendixTex/appendix_noiseless_errorBound3D}
\input{appendixTex/appendix_noisy_algebraic}

\input{appendixTex/appendix_noisy_signConsistency2D}
\input{appendixTex/appendix_noisy_errorBound2D}
\input{appendixTex/appendix_algorithm}
\input{appendixTex/appendix_nesta}

\onecolumn

\clearpage
\input{supplementalTex/supp_gazeboImages}
\pagebreak

\clearpage
\input{supplementalTex/supp_zedImages}

\end{document}

%% file: symbols/preamble_packages.tex
\usepackage{floatrow}
\usepackage{amsthm}
\usepackage{graphicx}
\usepackage{epstopdf}
\usepackage{pdfpages}

\usepackage{comment}
\usepackage{siunitx}
\usepackage{relsize}
\usepackage{ifthen}

\usepackage[caption=false]{subfig}

\usepackage[vlined,ruled,linesnumbered]{algorithm2e}
\usepackage{graphics} 
\usepackage{rotating}
\usepackage{color}
\usepackage{enumerate}
\usepackage[T1]{fontenc}
\usepackage{psfrag}
\usepackage{epsfig} 
\usepackage{booktabs}
\usepackage{graphicx,url}
\usepackage{multirow}
\usepackage{array}
\usepackage{latexsym}
\usepackage{amsfonts}
\usepackage{amsmath}
\usepackage{amssymb}
\usepackage{xstring}
\usepackage[noend]{algorithmic}
\usepackage{multirow}
\usepackage{xcolor}
\usepackage{flexisym}
\usepackage{bigdelim}
\usepackage{breqn} 
\usepackage{listings}
\let\labelindent\relax
\usepackage{enumitem}
\usepackage{xspace}
\usepackage{bm}
\graphicspath{{./figures/}}
\usepackage{tikz}
\usetikzlibrary{matrix,calc}

\usepackage{blindtext}

\usepackage{mdwlist}

\let\itemize\undefined
\makecompactlist{itemize}{stditemize}

\usepackage{xr}
\usepackage{array,multirow}

\usepackage{caption} % to convert table captions to regular font

\usepackage{etoolbox}
\makeatletter
\patchcmd{\@makecaption}
  {\scshape}
  {}
  {}
  {}
\makeatother
\makeatletter
\let\NAT@parse\undefined
\makeatother
\usepackage[numbers,sort&compress,sectionbib]{natbib} 

%% file: symbols/preamble_symbols.tex
 \newcommand{\veryoptional}[1]{}  
\newcommand{\forFinalVersion}[1]{}
\newcommand{\cosparsity}{cosparsity\xspace} 
\newcommand{\spac}{\hspace{-0.27cm}}

\newrefformat{prob}{Problem\,\ref{#1}}
\newrefformat{def}{Definition\,\ref{#1}}
\newrefformat{sec}{Section\,\ref{#1}}
\newrefformat{sub}{Section\,\ref{#1}}
\newrefformat{prop}{Proposition\,\ref{#1}}
\newrefformat{app}{Appendix\,\ref{#1}}
\newrefformat{alg}{Algorithm\,\ref{#1}}
\newrefformat{cor}{Corollary\,\ref{#1}}
\newrefformat{thm}{Theorem\,\ref{#1}}
\newrefformat{lem}{Lemma\,\ref{#1}}
\newrefformat{fig}{Fig.\,\ref{#1}}
\newrefformat{tab}{Table\,\ref{#1}}
\newrefformat{proof}{Appendix\,\ref{#1}}
\newrefformat{line}{Line\,\ref{#1}}
\newrefformat{eq}{Equation\,\ref{#1}}

\newtheorem{theorem}{Theorem}

\newtheorem{corollary}[theorem]{Corollary}

\newtheorem{lemma}[theorem]{Lemma}

\newtheorem{definition}[theorem]{Definition}
\newtheorem{proposition}[theorem]{Proposition}

\newcommand{\cf}{\emph{cf.}\xspace}

\newcommand{\bdmath}{\begin{dmath}}
\newcommand{\edmath}{\end{dmath}}
\newcommand{\beq}{\begin{equation}}
\newcommand{\eeq}{\end{equation}}
\newcommand{\bdm}{\begin{displaymath}}
\newcommand{\edm}{\end{displaymath}}
\newcommand{\bea}{\begin{eqnarray}}
\newcommand{\eea}{\end{eqnarray}}
\newcommand{\beal}{\beq \begin{array}{ll}}
\newcommand{\eeal}{\end{array} \eeq}
\newcommand{\beas}{\begin{eqnarray*}}
\newcommand{\eeas}{\end{eqnarray*}}
\newcommand{\ba}{\begin{array}}
\newcommand{\ea}{\end{array}}
\newcommand{\bit}{\begin{itemize}}
\newcommand{\eit}{\end{itemize}}
\newcommand{\ben}{\begin{enumerate}}
\newcommand{\een}{\end{enumerate}}

\newcommand{\calA}{{\cal A}}

\newcommand{\calC}{{\cal C}}

\newcommand{\calE}{{\cal E}}

\newcommand{\calI}{{\cal I}}
\newcommand{\calJ}{{\cal J}}

\newcommand{\calM}{{\cal M}}

\newcommand{\calO}{{\cal O}}

\newcommand{\calQ}{{\cal Q}}

\newcommand{\calS}{{\cal S}}

\newcommand{\setal}{~\emph{et~al.}\xspace}

\newcommand{\hide}[1]{}

\newcommand{\hiddenText}{{\color{gray} hidden text.}}
\newcommand{\hideWithText}[1]{\hiddenText}

\newcommand{\kron}{\otimes}

\newcommand{\subject}{\text{ subject to }}

\DeclareMathOperator*{\argmin}{arg\,min}

\newcommand{\lzero}{\ell_{0}}
\newcommand{\lone}{\ell_{1}}
\newcommand{\linf}{\ell_{\infty}}

\newcommand{\tran}{^{\mathsf{T}}}

\newcommand{\inv}{^{-1}}
\newcommand{\pinv}{^\dag}

\newcommand{\ones}{{\mathbf 1}}
\newcommand{\zero}{{\mathbf 0}}
\newcommand{\eye}{{\mathbf I}}
\newcommand{\vect}[1]{\left[\begin{array}{c}  #1  \end{array}\right]}

\newcommand{\Real}[1]{ { {\mathbb R}^{#1} } }

\newcommand{\opt}{^{\star}}

\newcommand{\at}[1]{^{(#1)}}

\newcommand{\setdef}[2]{ \{#1 \; {:} \; #2 \} }

\renewcommand{\ij}{_{ij}}

\newcommand{\scenario}[1]{{\smaller \sf#1}\xspace}

\newcommand{\MOSEK}{\scenario{MOSEK}}
\newcommand{\NESTA}{\scenario{NESTA}}

\newcommand{\cvx}{{\sf cvx}\xspace}

 \newcommand{\meas}{\boldsymbol{z}}

%% file: symbols/shortcuts.tex
\newcommand*\widebar[1]{  \hbox{    \vbox{      \hrule height 0.5pt       \kern0.5ex      \hbox{        \kern-0.1em        \ensuremath{#1}        \kern-0.1em      }    }  }}

\newcommand{\bal}{\begin{align}}
\newcommand{\eal}{\end{align}}

\newcommand{\bals}{\begin{align*}}
\newcommand{\eals}{\end{align*}}

\renewcommand{\theenumi}{\Alph{enumi}}

\newcommand{\vare}{\varepsilon}

\renewcommand{\vec}[1]{\,\text{vec}{\left(#1\right)}}

\newcommand{\R}{\mathbb{R}}

\newcommand{\ra}{{\rangle}}
\newcommand{\la}{{\langle}}

\renewcommand{\argmin}{\mathrm{argmin}}

\newcommand{\supp}{\mathrm{supp}}

\newcommand{\kernel}{\mathrm{ker}}

\newcommand{\nam}{C_\text{er}}

\newcommand{\myParagraph}[1]{{\bf #1.}}
\renewcommand{\opt}[1]{{#1^\star}}
\newcommand{\ltwo}[1]{\|#1\|_2}
\renewcommand{\lone}[1]{\|#1\|_1}
\renewcommand{\linf}[1]{\|#1\|_\infty}
\newcommand{\linfM}[1]{\|#1\|_{\infty \rightarrow \infty}}

\renewcommand{\lzero}[1]{\|#1\|_0}

\newcommand{\Ind}{\chi}

\newcommand{\rtilde}{\tilde{r}}
\newcommand{\onen}{ \{1,\ldots,n\} }
\newcommand{\sign}{\text{sign}}
\newcommand{\ts}{\textsuperscript}

\newcommand{\second}{2\ts{nd}}
\newcommand{\red}[1]{{\color{red}#1}}

\newcommand{\toep}{Toeplitz\xspace}
\newcommand{\SC}{{\cal SC}}

\newcommand{\xhat}{\hat{\xvar}}

\newcommand{\CS}{CS\xspace}
\newcommand{\signal}{profile\xspace}
\newcommand{\signals}{profiles\xspace}

\newcommand{\xvar}{z}
\newcommand{\Xvar}{Z}
\newcommand{\xnaive}{\tilde{\xvar}}
\newcommand{\vareps}{\varepsilon}
\newcommand{\xtrue}{\xvar^\diamond} 
\newcommand{\Xtrue}{\Xvar^\diamond}

\newcommand{\xopt}{\opt{\xvar}}
\newcommand{\xoptInd}[1]{\opt{\xvar_{#1}}}
\newcommand{\Xopt}{\opt{\Xvar}}
\newcommand{\xopti}{\opt{\xvar_i}}
\renewcommand{\SS}{\opt{\calS}}
\renewcommand{\meas}{y}

\newcommand{\matU}{A}
\newcommand{\samples}{\calM}
\newcommand{\cosamples}{\overline{\calM}}
\newcommand{\mbar}{\bar{m}}
\newcommand{\tv}{D}
\newcommand{\TV}{\Delta}
\newcommand{\support}{\calI}
\newcommand{\cosupport}{\calJ}
\newcommand{\error}{\eta}

\newcommand{\ActDown}{{\cal A}_\downarrow}
\newcommand{\ActUp}{{\cal A}_\uparrow}

\newcommand{\twd}{2D\xspace}
\newcommand{\thd}{3D\xspace}

\newcommand{\algOne}{{\tt A1}\xspace}
\newcommand{\naive}{{\tt naive}\xspace}
\newcommand{\lmin}{{\tt L1}\xspace}
\newcommand{\lminDiag}{{\tt L1diag}\xspace}
\newcommand{\lminCart}{{\tt L1cart}\xspace}

\newcommand{\XvarMi}{\Xvar_{[\cosamples_i]}}  

\newcommand{\gazebo}{\texttt{Gazebo}\xspace}
\newcommand{\zed}{\texttt{ZED}\xspace}
\newcommand{\PL}{\texttt{PL}\xspace}
\newcommand{\kinect}[1]{\texttt{K#1}\xspace}

\newcommand{\somespace}{\vspace{3mm}}

\newcommand{\myIncludeTwoFigures}[6]
{
\newcommand{\minipageWidth}{0.48\linewidth}
\centering
\begin{minipage}{\textwidth}
\def\arraystretch{1}   \setlength\tabcolsep{1mm} {
\begin{tabular}{cc}
\begin{minipage}[b]{\minipageWidth}\centering\includegraphics[width=\linewidth, trim={#3 #4 #5 #6}, clip]{#1} \\
(a)
\end{minipage}
&  
\begin{minipage}[b]{\minipageWidth}\centering\includegraphics[width=\linewidth, trim={#3 #4 #5 #6}, clip]{#2} \\
(b)
\end{minipage}
\end{tabular}
}
\end{minipage}}

\newcommand{\omittedLC}[1]{}
\newcommand{\unsampled}{non-sampled\xspace}
\newcommand{\noisyEnvelop}{$\epsilon$-envelope\xspace}
\newcommand{\param}{\delta}
\newcommand{\WTCT}{\texttt{WT+CT}\xspace}
\newcommand{\CSR}{\texttt{CSR}\xspace}
\newcommand{\ccc}{g}

\newcommand{\xnesta}{\bar{q}}
\newcommand{\wnesta}{\bar{w}}

%% file: abstract.tex
%!TEX root = main.tex

\begin{abstract}
   We consider the case in which a robot 
 has to navigate in an unknown environment but 
 does not have enough on-board power or payload to carry a traditional depth sensor 
 (e.g., a 3D lidar) and thus can only acquire a few (point-wise) depth measurements. 
We address the following question: 
 \emph{is it possible to reconstruct the geometry of an unknown environment using 
 sparse and incomplete depth measurements?}
 Reconstruction from incomplete data is not possible in general, but 
 when the robot operates in man-made environments, the depth exhibits some regularity
 (e.g., many planar surfaces with only a few edges); 
 we leverage this regularity to infer depth from a small number of measurements.
 Our first contribution is a formulation of the depth reconstruction problem that bridges
   robot perception with the \emph{compressive sensing} 
 literature in signal processing. 
 The second contribution includes a set of formal results 
  that ascertain the exactness and stability of the depth reconstruction in 2D and 3D problems, 
 and completely characterize the geometry of the 
 \signals that we can reconstruct.
      Our third contribution is a set of practical algorithms for depth reconstruction: 
 our formulation directly translates into algorithms for depth estimation based on convex programming.
 In real-world problems, these convex programs are very large and general-purpose solvers are 
 relatively slow. For this reason, we discuss ad-hoc solvers that enable fast depth reconstruction 
 in real problems.  
The last contribution is an extensive experimental evaluation in 2D and 3D problems, including 
Monte Carlo runs on simulated instances and testing on multiple real datasets.   
Empirical results confirm that the proposed approach ensures accurate depth reconstruction, 
outperforms interpolation-based strategies, and performs well even when the 
assumption of structured environment is violated. 
    \end{abstract}

%% file: supplementalInfo.tex
%!TEX root = main.tex

\setlist[itemize]{leftmargin=*}

\section*{Supplemental Material}
\begin{itemize}
\item {Video demonstrations}: \\ {\small \url{https://youtu.be/vE56akCGeJQ}}
\item {Source code}: \\ {\small \url{https://github.com/sparse-depth-sensing}}

\end{itemize}

%% file: introduction.tex
%!TEX root = main.tex

\section{Introduction}
\label{sec:intro}

Recent years have witnessed a growing interest towards miniaturized robots, for instance the \emph{RoboBee}~\citep{Wood15robotbee}, \emph{Piccolissimo}~\citep{Piccolissimo16}, the \emph{DelFly}~\citep{de2014autonomous, de2012sub}, the \emph{Black Hornet Nano}~\citep{BlackHornet16}, \emph{Salto}~\citep{Haldane2016}.
 These robots are usually palm-sized (or even smaller), can be deployed in large volumes, 
 and provide a new perspective on societally relevant applications, including artificial pollination, 
 environmental monitoring, and disaster response.
Despite the rapid development and recent success in control, actuation, and manufacturing of miniature robots, 
%the study of 
on-board sensing and perception capabilities for such robots remain a relatively unexplored, challenging open problem. 
These small platforms have extremely limited payload, power, and on-board computational resources, 
 thus preventing the use of standard sensing and computation paradigms.

\input{figureTex/figureZED.tex}

In this paper we explore novel sensing techniques for miniaturized robots that 
cannot carry standard sensors. 
In the last two decades, a large body of robotics research focused on the development of techniques
 to perform inference 
 from data produced by ``information-rich'' sensors 
 (e.g., high-resolution cameras, 2D and 3D laser scanners).
 A variety of approaches has been proposed to perform geometric reconstruction using these 
 sensors, for instance see \citep{Newcombe11iccv,Mur-Artal15tro,Whelan15rss} and the references therein.
 \veryoptional{; different techniques have been also 
 engineered into reliable prototypes and products (e.g., \emph{Google Tango}~\citep{googleTango}, 
 \emph{PhotoSynth}~\citep{PhotoSynth}, \emph{ReconstructMe}~\citep{Reconstructme}).}
 On the other extreme of the sensor spectrum, applications and theories have been developed to 
 cope with the case of minimalistic sensing~\citep{Suri08ijrr,Derenick13icra,Tovar11ijrr,Tovar14tsn}.
 In this latter case, the sensor data is usually not metric (i.e., the sensor cannot measure distances or angles) but instead binary in nature (e.g., binary detection of landmarks), and the goal is to infer only the topology of the (usually planar) environment rather than its geometry. This work studies a relatively unexplored region between these two extremes of the sensor spectrum.

Our goal is to design algorithms (and lay the theoretical foundations)
to reconstruct a depth \signal (i.e., a laser scan in 2D, or a depth image in 3D, 
see~\prettyref{fig:zed}) 
from sparse and incomplete depth measurements. 
Contrary to the literature on minimalistic sensing, we provide tools to recover complete geometric information, while
requiring much fewer data points compared to 
standard information-rich sensors.
This effort complements recent work on hardware and sensor design, including the 
development of 
lightweight, small-sized depth sensors. For instance, a number of ultra-tiny laser range sensors are being developed as research prototypes (e.g., the dime-sized, 20-gram laser of~\cite{chen2016development}, and an even smaller lidar-on-a-chip system with no moving parts~\citep{LidarOnChip16}), while some other distance sensors have already been released to the market (e.g., the TeraRanger's single-beam, 8-gram distance sensor~\citep{TeraRanger16}, and the LeddarVu's 8-beam, 100-gram laser 
scanner~\citep{LeddarVu16}). 
These sensors provide potential hardware solutions for sensing on micro (or even nano) robots.
%, as well as other compact-sized mobile platforms.
Although these sensors meet the requirements of payload and power consumption of miniature robots, they only provide very sparse and incomplete depth data, in the sense that the raw depth measurements are extremely low-resolution (or even provide only a few beams). In other words, the output of these sensors cannot be utilized directly in high-level tasks (e.g., object recognition and mapping), and the need to reconstruct a complete depth profile from such sparse data arises.

\myParagraph{Contribution}
We address the following question:
\emph{is it possible to reconstruct a complete depth \signal from sparse and incomplete depth samples?}
In general, the answer is negative, since the environment can be very adversarial 
(e.g., 2D laser scan where each beam is drawn randomly from a uniform distribution), and it is impossible to recover the depth from a
 small set of measurements. However, when the robot operates in structured environments 
(e.g., indoor, urban scenarios) the depth data exhibits some regularity. 
For instance, man-made environments are characterized by the presence of many planar surfaces and a few 
edges and corners. This work shows how to 
  leverage this regularity to recover a depth \signal from a handful of sensor measurements.
Our overarching goal is two-fold: 
to establish theoretical conditions under which depth reconstruction from sparse and incomplete measurements is possible, 
and
to  develop practical inference algorithms for depth estimation. 

Our first contribution, presented in \prettyref{sec:cs}, is a 
general formulation of the depth estimation problem. 
Here we recognize that the ``regularity'' of a depth \signal is captured by 
a specific function (the $\ell_0$-norm of the \second-order differences of the depth \signal). 
We also show that by relaxing the $\ell_0$-norm to the (convex) $\ell_1$-norm, our problem 
falls within the \cosparsity model in \emph{compressive sensing} (\CS).
We review related work and give preliminaries on \CS in \prettyref{sec:relatedWork} and \prettyref{sec:preliminaries}.

The second contribution, presented in \prettyref{sec:sensingConstraints}, 
is the derivation of theoretical conditions for depth recovery. 
In particular, we provide conditions under which reconstruction of a \signal from incomplete measurements is possible, 
investigate the robustness of depth reconstruction in the presence of noise, and provide bounds on the reconstruction error.
Contrary to the existing literature in \CS, our conditions are geometric
(rather than  algebraic) and provide actionable information to guide sampling strategy.

Our third contribution, presented in \prettyref{sec:algorithms}, is algorithmic. 
We discuss practical algorithms for depth reconstruction, including different variants of the 
proposed optimization-based formulation, and solvers that enable fast depth recovery.
In particular, we discuss the application of a state-of-the-art solver for non-smooth convex programming, called \NESTA~\citep{BeckerBC11}.
%, to our depth estimation problem.

Our fourth contribution, presented in \prettyref{sec:experiments}, is an extensive experimental evaluation, 
including Monte Carlo runs on simulated data and testing with real sensors.
The experiments confirm our theoretical findings and show that our 
depth reconstruction approach is extremely resilient to noise and works well even when the 
regularity assumptions are partially violated. 
We discuss many applications for the proposed approach. 
Besides our motivating scenario of navigation with miniaturized robots, 
our approach finds application in several endeavors, including 
%multi-frame depth reconstruction, 
data compression and super-resolution depth estimation.

\prettyref{sec:conclusion} draws conclusions and 
discusses future research.
% For space reasons we omit the technical proofs. 
Proofs and extra visualizations are given in the appendix. 
%All proofs, together with extra visualizations, are given in the appendix. 
%are given in the appendix of the paper. 
% in Section~\ref{sec:smOverview}. 
% We use the notation ``\ref{sec:1D-polarCartesian}'' to recall a specific section (Section 2.1 in the example) in~\citep{Ma16iros-supplemental}.

This paper extend the preliminary results presented in~\citep{Ma16iros} in multiple directions. 
In particular, the error bounds in~\prettyref{sec:optimality} and~\prettyref{sec:stableRecovery}, 
the algorithms and solvers in~\prettyref{sec:algorithms}, and most of the experiments of~\prettyref{sec:experiments} 
are novel and have not been previously published.

%% file: figureTex/figureZED.tex
%!TEX root = ../main.tex

\begin{figure}[t]
\begin{minipage}{\textwidth}
\newcommand{\figWidth}{ 0.31\linewidth } 
\setlength\tabcolsep{1mm} \smaller
\begin{tabular}{ c c c }

\begin{minipage}[b]{\figWidth}\centering
\includegraphics[width=\linewidth]{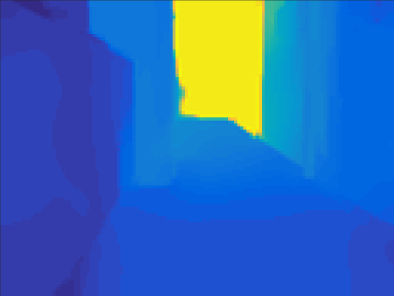}  
\\
(a) ground truth % \thd depth
\end{minipage}
& 
\begin{minipage}[b]{\figWidth}\centering
\includegraphics[width=\linewidth]{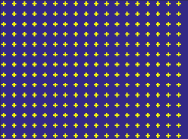} 
\\
(b) 2\% grid samples
\end{minipage}
& 
\begin{minipage}[b]{\figWidth}\centering
\includegraphics[width=\linewidth]{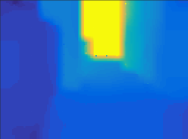}  
\\
(c) reconstruction
\end{minipage}
\\ 
%%%%%%%%%%%%%%%%%%%%%%%%%%%%%%%%%%%%%%%%%%%%%%%%%%%%%%%%%%%%%%%%
\begin{minipage}[b]{\figWidth}\centering
\includegraphics[width=\linewidth]{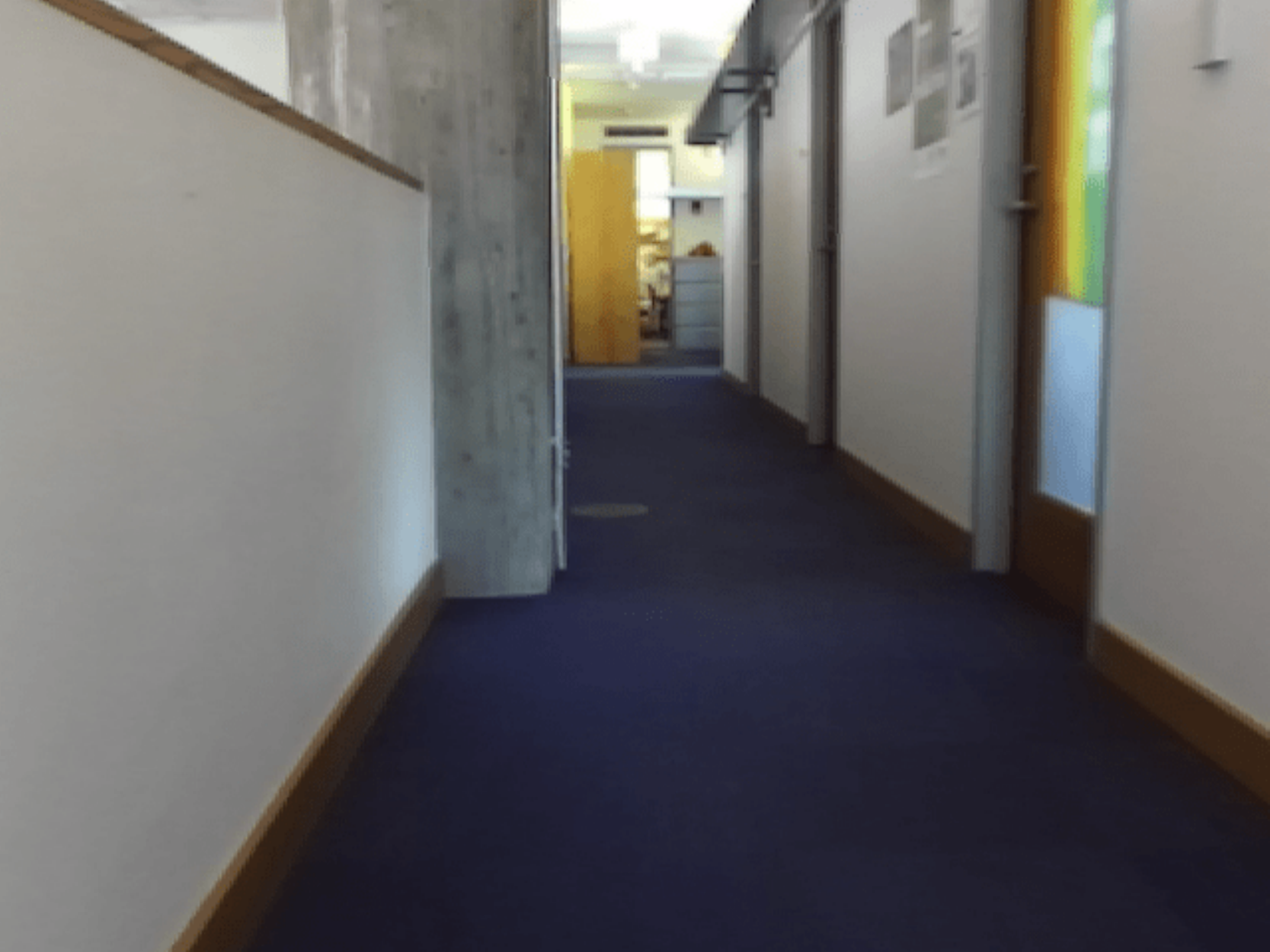}   
\\
(d) RGB image
\end{minipage}
&  
\begin{minipage}[b]{\figWidth}\centering
\includegraphics[width=\linewidth]{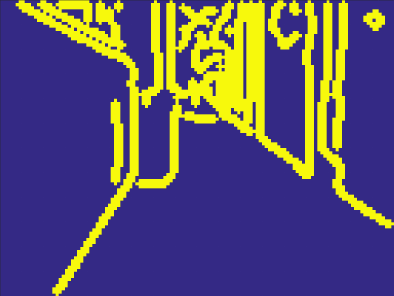}
\\
(e) sample along edges
\end{minipage}
& 
\begin{minipage}[b]{\figWidth}\centering
\includegraphics[width=\linewidth]{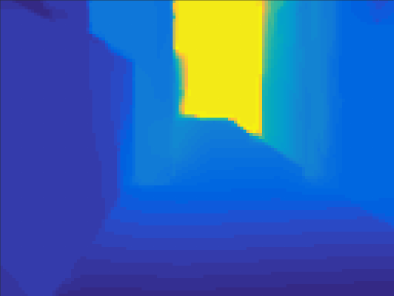}  
\\
(f) reconstruction
\end{minipage}

\end{tabular}
\end{minipage}

\caption{We show how to reconstruct an unknown depth \signal (a) 
from a handful of samples (b). Our reconstruction is shown in (c).
Our results also apply to traditional stereo vision and 
enable accurate reconstruction (f) from few depth measurements (e)
corresponding to the edges in the RBG image (d).
Figures (a) and (d) are obtained from a ZED stereo camera.}
\label{fig:zed}
\end{figure}

%% file: related_work.tex
%!TEX root = main.tex

\section{Related Work}
\label{sec:relatedWork} 
This work intersects several lines of research across fields.

\myParagraph{Minimalistic Sensing}
Our study of depth reconstruction from sparse sensor data is related to the literature on \emph{minimalistic sensing}. 
Early work on minimalistic sensing
 includes contributions on  sensor-less manipulation~\citep{Goldberg93algorithmica},  robot sensor design~\citep{Boning06ark,OKane06rss}, and target tracking~\citep{Shrivastava06enss}.
\citep{Suri08ijrr}, \citep{Derenick13icra}. 
\cite{Tovar07asm}
 use binary measurements of the presence of landmarks to infer the topology of the environment.
\cite{Marinakis07ccai,Marinakis08tro} reconstruct the topology of a sensor network from unlabeled observations from a mobile robot.
\cite{OKane07tro} and \cite{Erickson08icra} investigate a localization problem using contact sensors.
\cite{Tovar10icra} use depth discontinuities measurements to support exploration and search in 
unknown environments.
\cite{Tovar08wafr,Tovar14tsn} propose a combinatorial filter to estimate the path (up to homotopy class) 
of a robot from binary detections. 
\cite{Milford13ijrr} addresses minimality of information for vision-based place recognition.

\myParagraph{Sensing and perception on miniaturized robots}
A fairly recent body of work in robotics focuses on 
miniaturized robots and draws inspiration from small animals and insects. Most of the existing literature focuses on the control of such robots, either open-loop or based on information from external infrastructures. However, there has been relatively little work on onboard sensing and perception. For example, 
the \emph{Black Hornet Nano}~\citep{BlackHornet16} is a military-grade micro aerial vehicle equipped with three cameras but with basically no autonomy. 
\emph{Salto}~\citep{Haldane2016} is Berkeley's 100g legged robot with agile jumping skills. The jump behavior is open-loop due to lack of sensing capabilities, and the motion is controlled by a remote laptop. 
The \emph{RoboBee}~\citep{Wood15robotbee} is an 80-milligram, insect-scale robot % with independently actuated flapping wings, 
capable of hovering motion. The state estimation relies  on an external array of cameras. % at 500Hz.
\emph{Piccolissimo}~\citep{Piccolissimo16} is a tiny, self-powered drone with only two moving parts, completely controlled  by an external, hand-held infrared device. The \emph{DelFly} \emph{Explorer}~\citep{de2014autonomous, de2012sub} is a 20-gram flying robot with an onboard stereo vision system. % and limited processing power. 
It is capable of producing a coarse depth image at 11Hz and is thus one of the first examples of miniaturized flying robot with basic obstacle avoidance capabilities.

\myParagraph{Fast Perception and Dense 3D Reconstruction}
The idea of leveraging priors on the structure of the environment to improve or enable 
geometry estimation has been investigated in early work in computer vision for single-view 3D reconstruction and feature matching~\citep{Kanade81ai,Hong04ijcv}. 
Early work by \citep{faugeras1988motion} addresses \emph{Structure from Motion} by assuming the 
environment to be piecewise planar. More recently, \cite{Pillai16icra} propose an approach to speed-up stereo reconstruction 
by computing the disparity at a small set of pixels and considering the environment to be piecewise planar elsewhere. \cite{dame2013dense} combine live dense reconstruction with shape-priors-based 3D tracking and reconstruction.
\cite{estellers2015adaptive} propose a regularization based on the structure tensor to better capture the local geometry of images.
\cite{lu2015sparse} produce high-resolution depth maps from subsampled depth measurements by using segmentation based on both RGB images and depth samples. \cite{Pinies15icra} compute a dense depth map from a sparse point cloud.
This work is related to our proposal with three main differences.
First, the work~\citep{Pinies15icra} uses an energy minimization approach that requires parameter tuning 
(the authors use Bayesian optimization to learn such parameter); 
our approach is parameter free and only assumes bounded noise. 
Second, we use a \second-order difference operator to promote 
 depth regularity, while~\citep{Pinies15icra} considers alternative costs, including nonconvex 
 regularizers. Finally, by recognizing connections with the \emph{cosparsity} model in \CS, we provide theoretical foundations for the reconstruction problem.

\myParagraph{Map Compression}
Our approach is also motivated by the recent interest in \emph{map compression}.
\cite{Nelson15iros} propose a compression method for occupancy grid maps, 
based on the \emph{information bottleneck} theory.
\cite{Ramos15rss,OCallaghan14iser} use Gaussian processes 
to improve 2D mapping quality from smaller amount of laser data.
\cite{Im10dscc} investigate wavelet-based 
compression techniques for 3D point clouds.
\cite{Ruhnke13aaai,Ruhnke14icra} discuss point cloud compression techniques 
based on sparse coding.
\cite{mu2017two,mu2015two} propose a variable selection method to retain only an important subset of measurements during map building.

\myParagraph{Compressive Sensing (\CS)}
Finally, our work is related to the literature on \emph{compressive sensing}~\citep{Candes06tit-2,Donoho06tit,Candes08spm,FoR13}.
While Shannon's theorem states that to reconstruct a signal (e.g., a depth \signal) we need a sampling rate
(e.g., the spatial resolution of our sensor) which
must be at least twice the maximum frequency of the signal, 
\CS revolutionized signal processing by showing that a signal can be reconstructed from a 
much smaller set of samples if it is \emph{sparse} in some domain. 
\CS mainly invokes two principles. First, by inserting \emph{randomness} in the data acquisition, one can 
improve reconstruction. Second, one can use $\ell_1$-minimization to encourage sparsity of the 
reconstructed signal. Since its emergence, \CS impacted many 
research areas, including image processing (e.g., inpainting~\citep{Liang12eccv}, 
total variation minimization~\citep{NeedellWard13}), data compression and 3D reconstruction~\citep{Usevitch01spm,Reddy08icip,Du11iasp}, tactile sensor data acquisition~\citep{hollis2017compressed},
inverse problems and regularization~\citep{Vaiter13tit},  
matrix completion~\citep{Recht2010}, and single-pixel imaging techniques~\citep{phillips2016adaptive, duarte2008single, sun20133d}.  While most \CS literature assumes that the original signal $\xvar$ is sparse in a particular domain, i.e., $\xvar = \tv x$ for some matrix $\tv$ and a sparse vector $x$ (this setup is 
 usually called the \emph{synthesis model}), very recent work considers the case in which 
 the signal becomes sparse after a transformation is applied (i.e., given a matrix $\tv$, the vector $\tv \xvar$ 
 is sparse). The latter setup is called the \emph{analysis (or cosparsity) model}~\citep{Nam13acha,Kabanava15}. An important application of the analysis model in compressive sensing is total variation minimization, which is ubiquitous in image processing 
 \citep{NeedellWard13,Rudin92}.  In a hindsight we generalize total variation (which applies to piecewise constant signals) to 
 %deal with 
 piecewise \emph{linear} functions. 

\myParagraph{Depth Estimation from Sparse Measurements}
Few recent papers investigate the problem of reconstructing a dense depth image from sparse measurements. \cite{hawe2011dense} exploit the sparsity of the disparity maps in the Wavelet domain. The dense reconstruction problem is then posed as an optimization problem that simultaneously seeks a sparse coefficient vector in the Wavelet domain while preserving image smoothness. They also introduce a conjugate subgradient method for the resulting large-scale optimization problem.
Liu\setal~\citep{liu2015depth} empirically show that a combined dictionary of wavelets and
contourlets produces a better sparse representation of disparity maps, leading to more accurate reconstruction. 
% Their proposed ADMM algorithm also converges at a significantly faster rate than the conjugate subgradient method in~\citep{hawe2011dense}. 
In comparison with~\citep{hawe2011dense,liu2015depth}, our work has four major advantages. Firstly, our algorithm works with a remarkably small number of samples (e.g. 0.5\%), while both~\citep{hawe2011dense,liu2015depth} operate with at least 5\% samples, depending on the image resolution. Secondly, our algorithm significantly outperforms previous work in both reconstruction accuracy and computation time, 
hence pushing the boundary of achievable performance in depth reconstruction from sparse measurements. An extensive experimental comparison is presented in \prettyref{sec:exp-comparison}. 
Thirdly, the sparse representation presented in this work is specifically designed to encode depth \signals, while both~\citep{hawe2011dense,liu2015depth} use wavelet representations, which do not explicitly leverage the geometry of the problem.
%which treat the depth \signals as an \twd images.
%are more suitable to encode color (or grayscale) images. 
Indeed, our representation is derived from a simple, intuitive geometric model and thus has clear physical interpretation. Lastly, unlike previous work which are mostly algorithmic in nature, we provide theoretical guarantees and error bounds, as well as conditions under which the reconstruction is possible.

% Lastly, we explore application of our algorithm to beyond stereo reconstruction. We demonstrate the effectiveness of our algorithm in applications including depth image super-resolution and compression.
% \FMupdate{ Compressive sensing has also found applications in robotics. For instance, recent work on robotic tactile skins~\citep{hollis2017compressed} is able to compress tactile sensor signals to a fourth of its original size and still produce quality reconstruction using compressive sensing techniques.}

%% file: preliminaries.tex
%!TEX root = main.tex

\section{Preliminaries and Notation}
\label{sec:preliminaries}

We use  uppercase letters for matrices, e.g., $\tv \in \Real{p \times n}$, and lowercase letters 
for vectors and scalars, e.g, $\xvar \in \Real{n}$ and $a \in \Real{}$. Sets are denoted with calligraphic fonts, e.g., $\samples$. 
The cardinality of a set $\samples$ is denoted with $|\samples|$. 
For a set $\samples$, the symbol $\cosamples$ denotes its complement.
For a vector $\xvar \in \Real{n}$ and a set of indices $\samples \subseteq \{1,\ldots,n\}$, $\xvar_\samples$ is the sub-vector of $\xvar$ 
corresponding to the entries of $\xvar$ with indices in $\samples$. 
In particular, $\xvar_i$ is the $i$-th entry. 
%For two vector $\xvar$ and $\meas$, the symbols
% ``$=$'' and ``$\neq$'' denote component-wise equality and inequality, respectively.
 The symbols $\ones$ (resp. $\zero$) denote a vector of all ones (resp. zeros) of suitable dimension.
%A vector of all ones and all zeros (of suitable dimension) are denoted $\ones$ and $\zero$, respectively.

The \emph{support set} of a vector is denoted with
$$
\supp(\xvar) = \setdef{i \in \{1,\dots,n\}}{\xvar_i \neq 0 }.
$$
We denote with $\ltwo{\xvar}$ the Euclidean norm and we also use the following norms: 
\bea
\linf{\xvar} & \doteq &\max_{i=1,\ldots,n} \; |\xvar_i| \quad \text{($\ell_\infty$ norm)}  \\
\lone{\xvar} &\doteq &\sum_{i=1,\ldots,n} |\xvar_i|  \quad \text{($\ell_1$ norm)} \\ 
\lzero{\xvar} &\doteq &| \supp(\xvar)  | \quad \text{($\ell_0$ pseudo-norm)}
\eea
Note that $\lzero{\xvar}$ is simply the number of nonzero elements in $\xvar$.
The sign vector $\sign(\xvar)$ of $\xvar \in \R^n$ is a vector with entries: 
\begin{align*}
\sign(\xvar)_i \doteq 
\left\{
\ba{ll}
+1 & \text{if } \xvar_i > 0 \\
\;\;\;0 & \text{if } \xvar_i =0 \\
-1 & \text{if } \xvar_i < 0
\ea
\right.
\end{align*}

For a matrix $\tv$ and an index set $\samples$, 
let $\tv_\samples$ denote the sub-matrix of $\tv$ containing only the \emph{rows} of $\tv$ 
with indices in $\samples$; in particular, $\tv_i$ is the $i$-th row of $\tv$. 
Similarly, given two index sets $\support$ and $\cosupport$,
let $\tv_{\support,\cosupport}$ denote the sub-matrix of $\tv$ 
including only rows in $\support$ and columns in $\cosupport$.
Let $\eye$ denote the identity matrix. Given a matrix $\tv \in \Real{p\times n}$, we define the following matrix operator norm 
$$\linfM{\tv} \doteq \max_{i=1,\ldots,p} \lone{\tv_i}.$$
                                                                                             
 %Throughout the paper, 
In the rest of the paper we use the \emph{\cosparsity model} in \CS. 
In particular, we assume that the signal of interest is sparse under the application of an \emph{analysis operator}.
The following definitions formalize this concept.

\begin{definition}[{\bf Cosparsity}]
A vector $\xvar \in \Real{n}$ is said to be \emph{cosparse} with respect to a matrix $\tv \in \Real{p \times n}$
if $\| \tv \xvar \|_0 \ll p$. 
\end{definition}

\begin{definition}[{\bf $\tv$-support and $\tv$-cosupport}]
Given a vector $\xvar \in \Real{n}$ and a matrix $\tv \in \Real{p \times n}$, 
the \emph{$\tv$-support} of $\xvar$ 
is the set of indices corresponding to the nonzero entries of $\tv \xvar$, i.e., 
$\support = \supp(\tv \xvar)$. The \emph{$\tv$-cosupport} $\cosupport$ is the complement 
of $\support$, i.e., the indices of the zero entries of $\tv \xvar$.
\end{definition}

%% file: formulation/problemFormulation.tex
\section{Problem Formulation}
% Problem Formulation: \\Sparse Depth Sensing and Reconstruction
\label{sec:cs}

Our goal is to reconstruct \twd depth \signals (i.e., a scan from a 2D laser range finder) 
and \thd depth \signals (e.g., a depth image produced by a kinect or a stereo camera) from partial and incomplete depth measurements.
In this section we formalize the depth reconstruction problem,
by first considering the \twd and the \thd cases 
separately, and then reconciling them under a unified framework.

%% file: formulation/problemFormulation_2D.tex
%!TEX root = ../main.tex

%%%%%%%%%%%%%%%%%%%%%%%%%%%%%%%%%%%%%%%%%%%%%%%%%%%%%%%%%%%%%%%%%%%%%%%%%%%%%%%%%%%%%%%%%%%%%%%%%%%%%%%%%%%%%%%%%%%%%%%%%%%%%%%%%%
\subsection{\twd Depth Reconstruction}

In this section we discuss how to recover a \twd depth \signal $\xtrue \in \Real{n}$. One can imagine that the vector $\xtrue$ includes 
(unknown) depth measurements at discrete angles; this is what a standard planar 
range finder would measure. % (\prettyref{fig:nam1}(a)).

In our problem, due to sensing constraints, we 
do not have direct access to $\xtrue$, and we only observe a subset of its entries.
 In particular, we measure
\beq
\label{eq:measurements}
\meas = \matU \xtrue + \error \qquad \text{(sparse measurements)}
\eeq
where the matrix $\matU \in \Real{m\times n}$ with $m \ll n$ is the \emph{measurement matrix}, 
and $\error$ represents measurement noise. 
The structure of $\matU$ is formalized in the following definition.

\begin{definition}[{\bf Sample set and sparse sampling matrix}]
\label{def:samplingMatrix}
A sample set $\samples \subseteq \onen$ is the set of entries of the \signal that are measured.
A matrix $\matU \in \Real{m \times n}$ is called a \emph{(sparse) sampling matrix} (with sample set $\samples$), if 
$\matU = \eye_\samples$. % Note that $\matU \xvar = \xvar_\samples$.
\end{definition}

Recall that $\eye_\samples$ is a sub-matrix of the identity matrix, with only rows of indices in $\samples$. 
It follows that $\matU \xvar = \xvar_\samples$, i.e., the matrix $\matU$ selects a subset of entries from $\xvar$.
Since $m \ll n$, we have much fewer measurements 
than unknowns. Consequently, $\xtrue$ cannot be recovered from $\meas$, without further assumptions.

\input{figureTex/figureBP_exact}

In this paper we assume that the \signal $\xtrue$ is sufficiently regular, in the sense that 
it contains only a few ``corners'', e.g.,~\prettyref{fig:nam1}(a).
Corners are produced by changes of slope: considering 3 consecutive points at coordinates
 $(x_{i-1},\xvar_{i-1})$,  $(x_{i},\xvar_{i})$, and  $(x_{i+1},\xvar_{i+1})$,\footnote{Note that $x$ corresponds to the 
 horizontal axis in~\prettyref{fig:nam1}(a), while the depth $z$ is shown on the vertical axis in the figure.} there is a corner at $i$ 
 if 
 \beq \label{eq:slopeDiff}
   \frac{ \xvar_{i+1}-\xvar_{i} }{ x_{i+1}-x_{i} }  
- \frac{ \xvar_{i}-\xvar_{i-1} }{ x_{i}-x_{i-1} }  \neq 0.
\eeq
In the following we assume that $x_{i}-x_{i-1} = 1$ for all $i$: this comes without loss 
of generality since the full \signal is unknown and we can reconstruct it at arbitrary 
resolution (i.e., at arbitrary $x$); hence~\eqref{eq:slopeDiff}   simplifies to $\xvar_{i-1} - 2 \xvar_{i} + \xvar_{i+1} \neq 0$.
We formalize the definition of ``corner'' as follows.

\begin{definition}[{\bf Corner set}]
\label{def:cornerSet}
Given a \twd depth \signal $\xvar \in \Real{n}$, the \emph{corner set} $\calC \subseteq \{2,\ldots,n-1\}$ is the 
set of indices $i$ such that $\xvar_{i-1} - 2 \xvar_{i} + \xvar_{i+1} \neq 0$. 
\end{definition}

Intuitively, $\xvar_{i-1} - 2 \xvar_{i} + \xvar_{i+1}$ is the discrete equivalent of the 
\second-order derivative at $\xvar_{i}$.
We call $\xvar_{i-1} - 2 \xvar_{i} + \xvar_{i+1}$ the \emph{curvature} at sample $i$: 
if this quantity is zero, the neighborhood of $i$ is flat (the three points are collinear); if it is negative, the curve is locally concave;
if it is positive, it is locally convex.
To make notation more compact, 
we introduce the \emph{\second-order difference operator}:
\beq
\label{eq:tv}
\tv \doteq 
\left[
\ba{cccccc}
1 & -2  & 1 & 0 & \ldots & 0\\
0 & 1 & -2  & 1  & \ldots & 0 \\
\vdots & 0 & \ddots  & \ddots  & \ddots & 0 \\
0 &  \ldots & 0  & 1  & -2 & 1 \\
\ea
\right] 
\in \Real{(n-2) \times n}
\eeq

Then a \signal with only a few corners is one where $\tv \xtrue$ is sparse. In fact, the 
$\ell_0$-norm of $\tv \xtrue$ counts exactly the number of corners of a \signal: \beq
\label{eq:nrCorners}
\lzero{\tv \xtrue} = |\calC| \qquad \text{(\# of corners)}
\eeq
where $|\calC|$ is the number of corners in the \signal.

When operating in indoor environments, it is reasonable to assume that 
$\xtrue$ has only a few corners. Therefore, we want to exploit this regularity assumption 
and the partial measurements  
 $\meas$ in~\eqref{eq:measurements}  to reconstruct $\xtrue$.
 Let us start from the noiseless case in which $\error=\zero$ in~\eqref{eq:measurements}.
 In this case, a reasonable way to reconstruct the \signal $\xtrue$ is to solve the 
 following optimization problem:
\beq
\label{eq:l0formulation}
\min_\xvar \; \lzero{\tv \xvar} \quad \subject  \matU \xvar = \meas  \tag{L0}
\eeq
which seeks the \signal $\xvar$ that is consistent with the measurements~\eqref{eq:measurements} and 
contains the smallest number of corners. Unfortunately, problem~\eqref{eq:l0formulation} is NP-hard due to the 
nonconvexity of the $\ell_0$ (pseudo) norm.
In this work we study the following relaxation of problem~\eqref{eq:l0formulation}:
\beq
\label{eq:BP}
\min_\xvar \; \lone{\tv \xvar} \quad \subject   \matU \xvar = \meas  \tag{$\text{L1}_\tv$}
\eeq
which is a convex program (it can  be indeed rephrased as a linear program), and can be solved efficiently in practice.
\prettyref{sec:sensingConstraints}  
provides conditions under which~\eqref{eq:BP} recovers the solution of~\eqref{eq:l0formulation}.
Problem~\eqref{eq:BP} falls in the class of the cosparsity models in~\CS~\citep{Kabanava15}. 

In the presence of bounded measurement noise~\eqref{eq:measurements}, i.e., $\linf{\error} \leq \vareps$, the $\ell_1$-minimization problem becomes: 
\begin{align}
\label{eq:BPD}
\min_\xvar \; \lone{ \tv \xvar }  \quad \subject \; \linf{\matU \xvar - \meas} \leq \vareps \tag{$\text{L1}_\tv^\vareps$}
\end{align}
% 
% When considering the noisy problem~\eqref{eq:BPD}, we will be able to provide bounds on how far 
% is the solution $\xopt$ of~\eqref{eq:BPD} from the original \signal $\xtrue$.
Note that we assume that the $\ell_\infty$ norm 
%(rather than the 2-norm)
of the noise $\error$ is bounded, since this naturally reflects the sensor model in our robotic applications 
(i.e., bounded error in each laser beam). 
On the other hand, most \CS literature considers the $\ell_2$ norm 
of the error to be bounded and thus obtains an optimization problem with
the $\ell_2$ norm in the constraint. 
The use of the $\ell_\infty$ norm as a constraint in~\eqref{eq:BPD} resembles the \emph{Dantzig selector}
of Candes and Tao~\citep{candes2007dantzig}, with the main difference being the presence of the matrix $\tv$ in the objective.

%% file: figureTex/figureBP_exact.tex
%!TEX root = ../main.tex

\begin{figure}[bht]
\begin{minipage}{\textwidth}
\begin{tabular}{cc}\hspace{-0.6cm}
\begin{minipage}{4cm}\centering\includegraphics[height=2.8cm]{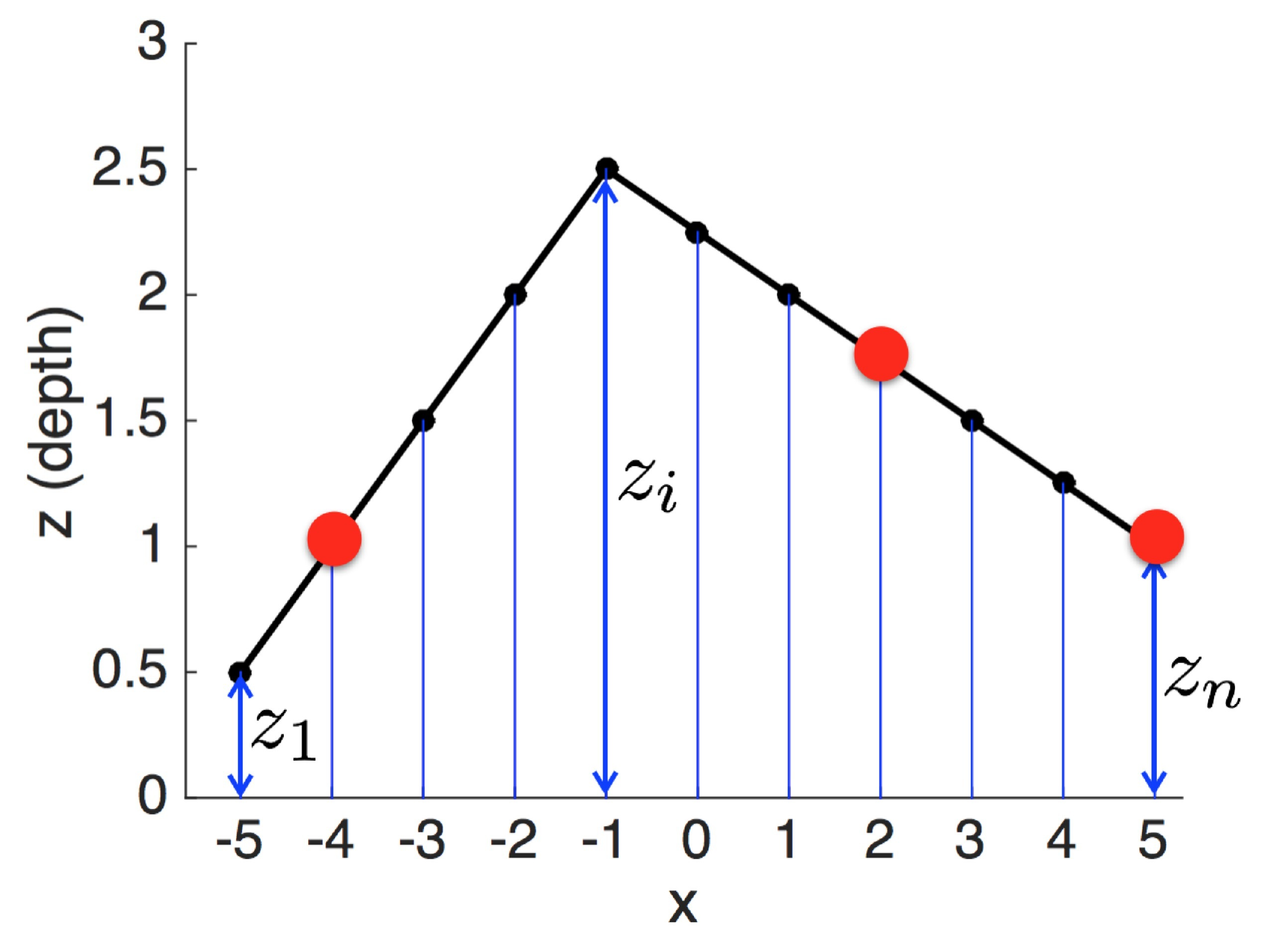}
\vspace{-3mm} \\
\hspace{-20mm} (a)
\end{minipage}
& 
 \hspace{-0.7cm}
\begin{minipage}{4cm}\centering \includegraphics[height=2.5cm]{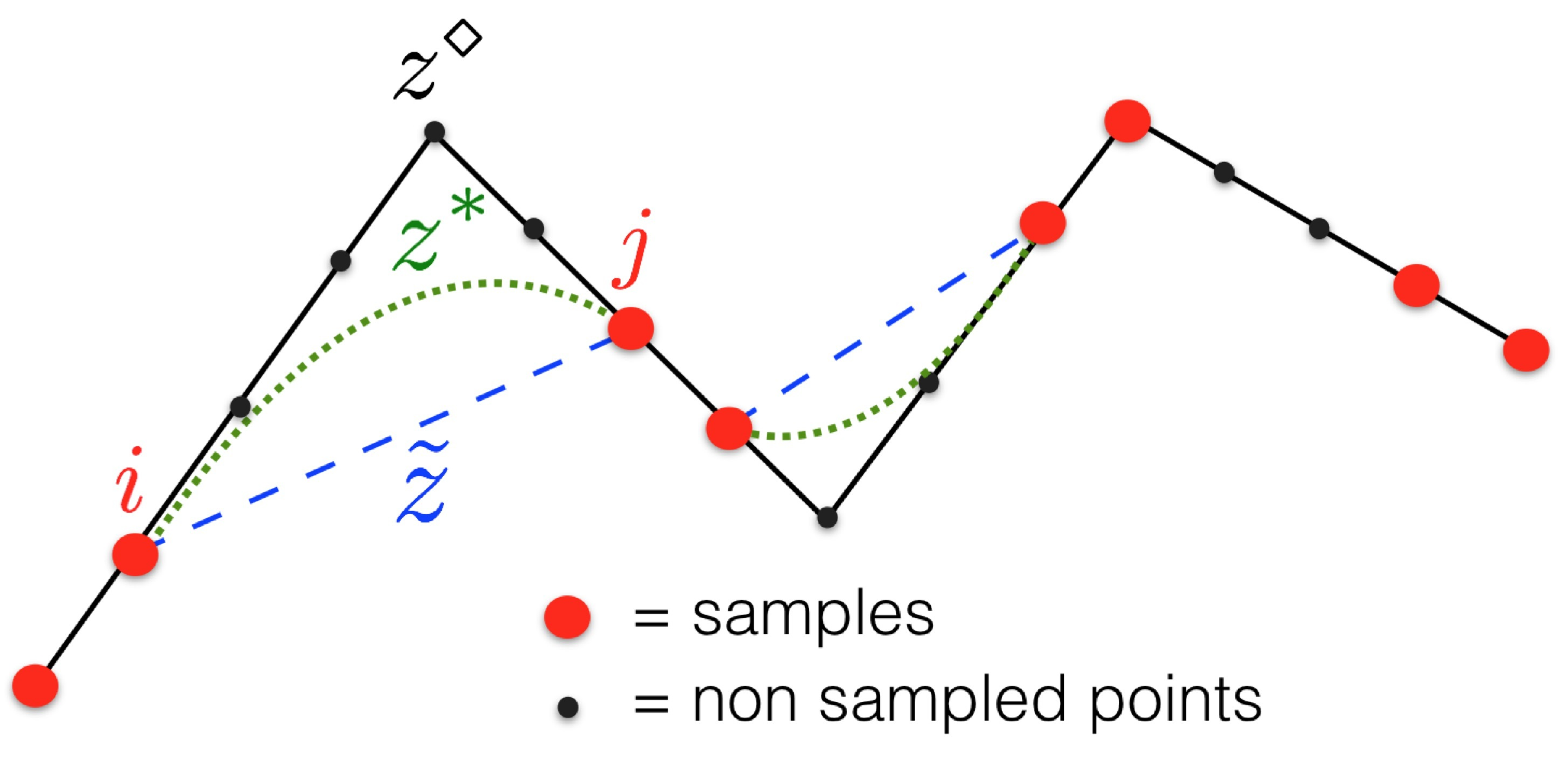} 
 \vspace{-7mm} \\
\hspace{-20mm} (b)
\end{minipage}
\end{tabular}
\vspace{-3mm}
\end{minipage}
\caption{
(a) Example of \twd depth \signal. Our goal is to reconstruct the full \signal (black solid line) 
from sparse samples (red dots); (b) When we do not sample the corners and the neighboring points,~\eqref{eq:BP}
admits multiple minimizers, which obey the conditions of~\prettyref{prop:1DrecoveryError}.
}
\label{fig:nam1}
\end{figure}

%% file: formulation/problemFormulation_3D.tex
\subsection{\thd Depth Reconstruction}

In this section we discuss how to recover a \thd depth \signal $\Xtrue \in \R^{r \times c}$ (a depth map, as the one in~\prettyref{fig:zed}(a)), 
using incomplete measurements.
As in the \twd setup, we 
do not have direct access to $\Xtrue$, but instead only have access to  $m \ll r \times c$ point-wise measurements in the form:
\beq
\label{eq:measurements2D}
\meas_{i,j} = \Xtrue_{i,j} + \error_{i,j}
\eeq
where $\error_{i,j} \in \Real{}$ 
 represents measurement noise. 
 Each measurement is 
   a noisy sample of the 
 depth of $\Xtrue$ at pixel $(i,j)$.

 We assume that  $\Xtrue$ is sufficiently regular, which intuitively means that 
the depth \signal contains mostly planar regions and only a few ``edges''. 
We define the edges as follows.

\begin{definition}[{\bf Edge set}]
\label{prop:edgeSet}
Given a \thd \signal $\Xvar \in \Real{r \times c}$, the \emph{vertical edge set} $\calE_V 
\subseteq \{2,\ldots,r-1\} \times \{1,\ldots,c\}$ is the 
set of indices $(i,j)$ such that $\Xvar_{i-1,j} - 2 \Xvar_{i,j} + \Xvar_{i+1,j} \neq 0$.
The \emph{horizontal edge set} $\calE_H 
\subseteq \{1,\ldots,r\} \times \{2,\ldots,c-1\}$ is the 
set of indices $(i,j)$ such that $\Xvar_{i,j-1} - 2 \Xvar_{i,j} + \Xvar_{i,j+1} \neq 0$. 
The \emph{edge set} $\calE$ is the union \veryoptional{(possibly with repeated elements)} 
of the two sets: $\calE \doteq \calE_V \cup \calE_H$.
\end{definition}

Intuitively, $(i,j)$ is \emph{not} in the edge set $\calE$ if 
the $3 \times 3$ patch centered at $(i,j)$ is planar, while 
$(i,j) \in \calE$ otherwise.
As in the \twd case we introduce \second-order difference operators $\tv_V$ and $\tv_H$ to compute the 
vertical differences $\Xvar_{i,j-1} - 2 \Xvar_{i,j} + \Xvar_{i,j+1}$ and the 
horizontal differences $\Xvar_{i-1,j} - 2 \Xvar_{i,j} + \Xvar_{i+1,j}$: 
\beq
\label{eq:DHDV}
\tv_V \Xtrue \in \Real{(r-2) \times c}, \qquad
\Xtrue \tv_H\tran \in \Real{r \times (c-2)}  
\eeq
where the matrices $\tv_V$ and $\tv_H$ are the same as the one defined~\eqref{eq:tv}, but 
with suitable dimensions; each entry of the matrix $\tv_V \Xtrue$ contains the vertical (\second-order) differences at a pixel, 
while $\Xtrue \tv_H\tran$ collects the horizontal differences.
\veryoptional{The sparsity of $\tv_V \Xtrue$ and $\Xtrue \tv_H\tran$ counts the edges in the 
depth \signal:
\beq
\lzero{\vec{\tv_V \Xtrue}} + \lzero{\vec{\Xtrue \tv_H\tran}} = |\calE|
\eeq
where $\vec{M}$ denotes the vectorization operator which stacks the columns 
of a matrix $M$ into a vector.
}

Following the same reasoning of the \twd case, we obtain the following 
$\ell_1$-norm minimization
\bea
\label{eq:l1matrix}
\min_\Xvar & \lone{ \vec{\tv_V \Xvar} } + \lone{ \vec{\Xvar \tv_H\tran} }  \\
\subject & \Xvar_{i,j} = \meas_{i,j}  \quad \text{for each measured pixel } (i,j) \nonumber
\eea
where $\vec{\cdot}$ denotes the (column-wise) vectorization of a matrix,  
and  we assume noiseless measurements. In the presence of measurement noise, 
the equality constraint in~\eqref{eq:l1matrix} is again replaced by  $|\Xvar_{i,j} - \meas_{i,j}| \leq \vareps $, $\forall (i,j)$, 
where $\vareps$ is an upper bound on the pixel-wise noise $\error_{i,j}$.

%% file: formulation/problemFormulation_unified.tex
%!TEX root = ../main.tex

%%%%%%%%%%%%%%%%%%%%%%%%%%%%%%%%%%%%%%%%%%%%%%%%%%%%%%%%%%%%%%%%%%%%%%%%%%%%%%%%%%%%%%%%%%%%%%%%%%%%%%%%%%%%%%%%%%%%%%%%%%%%%%%%%%

\subsection{Reconciling \twd and \thd Depth Reconstruction}
\label{sec:rec2d3d_unified}

In this section we show that the \thd depth reconstruction problem~\eqref{eq:l1matrix} can be reformulated 
to be closer to its \twd counterpart~\eqref{eq:BP}, if we vectorize the depth \signal (matrix $\Xvar$).
For a given \signal $\Xvar \in \Real{r \times c}$, we define the number of pixels $n \doteq r \times c$, and we call $\xvar$ the vectorized 
version of $\Xvar$, i.e., $\xvar \doteq \vec{\Xvar} \in \Real{n}$.
Using standard properties of the vectorization operator, we get
\bea
\label{eq:vectorProperties}
\vec{\tv_V \Xvar}
= (\eye_c \kron \tv_V) \xvar \nonumber  \\ 
\vec{ \Xvar \tv_H\tran }  = (\tv_H \kron \eye_r)  \xvar \\
\Xvar_{i,j} = \vec{e_i\tran \Xvar e_j}  = (e_i\tran \kron e_j\tran) \xvar \nonumber
\eea
where $\kron$ is the Kronecker product, $\eye_r$ is an identity matrix of size $r \times r$, and 
$e_i$ is a vector which is zero everywhere except the $i$-th entry which is $1$.
Stacking all measurements~\eqref{eq:measurements2D} in a vector $y \in \Real{m}$ and using~\eqref{eq:vectorProperties}, problem~\eqref{eq:l1matrix} 
can be written succinctly as follows: %in~\eqref{eq:BPmatrix}:
\beq
\label{eq:BPmatrix}
\min_\xvar \; \lone{\TV \xvar} \quad \subject  \matU \xvar = \meas   \tag{$\text{L1}_\TV$}
\eeq
where the matrix $\matU \in \Real{m \times n}$ (stacking rows in the form $e_i\tran \kron e_j\tran$) 
has the same structure of the sampling matrix introduced in Definition~\ref{def:samplingMatrix}, and the ``regularization'' matrix   $\TV$ is:
\beq
\label{eq:TVpartition}
\TV \doteq 
\vect{ \eye_c \kron \tv_V \\ \tv_H \kron \eye_r} \in \Real{2(n - r -c) \times n} \eeq
Note that~\eqref{eq:BPmatrix} is the same as~\eqref{eq:BP}, except for the fact that 
the matrix $\tv$ in the objective is replaced with a larger matrix $\TV$. 
It is worth noticing that the matrix 
 $\TV $ is also sparse, with only 3 non-zero entries ($1$, $-2$, and $1$) on each 
row in suitable (but not necessarily consecutive) positions.

In the presence of noise, we define an error vector $\error \in \Real{m}$ which stacks the noise terms in~\eqref{eq:measurements2D} for each pixel $(i,j)$, 
and assume pixel-wise bounded noise $\linf{\error} \leq \vareps$. The noisy \thd depth reconstruction problem then becomes: 
\beq
\label{eq:BPDmatrix}
\min_\xvar \; \lone{ \TV \xvar }  \quad \subject \; \linf{\matU \xvar - \meas} \leq \vareps \tag{$\text{L1}_\TV^\vareps$}
\eeq
Again, comparing~\eqref{eq:BPD} and~\eqref{eq:BPDmatrix}, 
it is clear that in \twd and \thd we solve the same optimization problem, with
the only difference lying in the matrices $\tv$ and $\TV$.

%% file: analysis/analysisIntro.tex
%!TEX root = ../main.tex
\section{Analysis: Conditions for Exact Recovery and Error Bounds for Noiseless and Noisy Reconstruction} % Sensing Constraints: Sparse Depth Measurements}
\label{sec:sensingConstraints}

\begin{table*}
  \centering
  \footnotesize
  \setlength\tabcolsep{5pt} % default value: 6pt
  \begin{tabular}{ c | c |  c | c  }
    \hline
    \twd / \thd & Sampling Strategy & Result & Remark \\ \hline \hline
    %%%%%%%%%%%%%%%%%%%%%%%%%%%%%%%%%%%%%%%%%%%%%%%%%%%%%%%%%%%%%%%%%%%%%%%%%%%%%%%%%%%%%%%%%%%%%%%%%%
    \twd\&\thd & noiseless & \prettyref{prop:nam} & sufficient condition for exact recovery (algebraic condition)\\ 
    \twd & noiseless, corners \& neighbors & \prettyref{prop:nam1D} & sufficient condition for exact recovery (geometric condition) \\
    \thd & noiseless, edges \& neighbors & \prettyref{prop:nam2D} & sufficient condition for exact recovery (geometric condition) \\ 
    \hline
    %%%%%%%%%%%%%%%%%%%%%%%%%%%%%%%%%%%%%%%%%%%%%%%%%%%%%%%%%%%%%%%%%%%%%%%%%%%%%%%%%%%%%%%%%%%%%%%%%%
    \twd & noiseless & \prettyref{prop:subdifferential} & necessary and sufficient condition for optimality (algebraic condition)\\
    \thd & noiseless & \prettyref{cor:subdifferential2D} & necessary and sufficient condition for optimality (algebraic condition)\\
    \twd & noiseless, twin samples \& boundaries & \prettyref{thm:1Doptimality} & necessary and sufficient condition for optimality (geometric condition)\\
    \twd & noiseless, twin samples \& boundaries & \prettyref{prop:1DrecoveryError} & reconstruction error bound\\
    \thd & noiseless, grid samples & \prettyref{thm:2Doptimality_xtrue} & sufficient condition for optimality (geometric condition)\\
    \thd & noiseless, grid samples & \prettyref{prop:2DrecoveryError} & reconstruction error bound\\
    \hline
    %%%%%%%%%%%%%%%%%%%%%%%%%%%%%%%%%%%%%%%%%%%%%%%%%%%%%%%%%%%%%%%%%%%%%%%%%%%%%%%%%%%%%%%%%%%%%%%%%%
    \twd & noisy & \prettyref{prop:robust_subdifferential} & necessary and sufficient condition for robust optimality (algebraic condition)\\
    \thd & noisy & \prettyref{cor:robust_subdifferential2D} & necessary and sufficient condition for robust optimality (algebraic condition)\\
    \twd & noisy & \prettyref{thm:1Doptimality_robust} & necessary condition for robust optimality (geometric condition)\\
    \twd & noisy, twin samples \& boundaries & \prettyref{prop:1DrecoveryError_robust} & reconstruction error bound\\
    \thd & noisy, grid samples & \prettyref{prop:2DrecoveryError_robust} & reconstruction error bound\\
    %%%%%%%%%%%%%%%%%%%%%%%%%%%%%%%%%%%%%%%%%%%%%%%%%%%%%%%%%%%%%%%%%%%%%%%%%%%%%%%%%%%%%%%%%%%%%%%%%%
    \hline \hline
  \end{tabular}
  \captionsetup{justification=centering}
  \caption{Summary of the key theoretical results.}
  \label{tab:summary}
\end{table*}

% In this section we consider the case in which the measurements matrix $\matU$ only selects few entries of the 
% \signal $\xtrue$. For instance, we measure only the red dots in the \twd scan on the \prettyref{fig:1Dsignal}(left), 
%  or we measure few pixels in the depth map on the right. 
%  %In the following we do not distinguish the \twd and the \thd case when unnecessary. 
%  To formalize this sparse sampling modality, we introduce the following definition. 
% We are now ready to delve in our technical results.

This section provides a comprehensive analysis on the quality of the depth \signals reconstructed by solving problems~\eqref{eq:BP} and~\eqref{eq:BPD} in the \twd case, and problems~\eqref{eq:BPmatrix} and~\eqref{eq:BPDmatrix} in \thd. A summary of the key technical results presented in this paper is given in \prettyref{tab:summary}.

In particular, \prettyref{sec:BP_exact} discusses \emph{exact recovery} and provides the conditions on the depth measurements such that the full depth \signal can be recovered exactly.
Since these conditions are quite restrictive in practice (although we will discuss an interesting application 
to data compression in \prettyref{sec:experiments}), \prettyref{sec:optimality} analyzes the reconstructed \signals under more general conditions.
More specifically, we derive error bounds that quantify the distance between the ground truth depth \signal and our reconstruction. 
\prettyref{sec:stableRecovery} extends these error bounds to the case in which the depth measurements are noisy.

%% file: analysis/BP_exactRecovery.tex
%!TEX root = ../main.tex

\subsection{Sufficient Conditions for Exact Recovery}
\label{sec:BP_exact}

In this section we provide sufficient conditions under which the full depth \signal can be reconstructed exactly from the given depth samples.

Recent results on \cosparsity in compressive sensing
provide sufficient conditions for exact recovery of a cosparse \signal $\xtrue$, 
from measurements $\meas = \matU \xtrue$ (where $\matU$ is a generic matrix). 
We recall this condition in \prettyref{prop:nam} below and, after presenting
 the result, we discuss why this condition is not directly amenable for roboticists to use.

\begin{proposition}[{\bf Exact Recovery~\citep{Nam13acha}}]
\label{prop:nam}
Consider a vector $\xtrue \in \Real{n}$ with $\tv$-support $\support$ and $\tv$-cosupport $\cosupport$. Define $\mbar \doteq n-m$.
Let $N \in \Real{\mbar \times n}$ be a matrix whose rows span the null space of 
the matrix $\matU$. Let $(\cdot)^\pinv$ denote the Moore-Penrose pseudoinverse of a matrix.
 If the following condition holds:
\beq
\label{eq:nam}
\nam \doteq \linfM{ (N (\tv_\cosupport)\tran)\pinv N (\tv_\support)\tran) } \; < \; 1
\eeq
then problem~\eqref{eq:BP} recovers $\xtrue$ exactly.             
\end{proposition}

Despite its generality,~\prettyref{prop:nam} provides only an \emph{algebraic} condition.
In our depth estimation problem, it would be more desirable to have \emph{geometric} conditions, which suggest the best sampling locations. % would be much more meaningful.
Our contribution in this section is a geometric interpretation of \prettyref{prop:nam}:

We first provide a result for the \twd case. The proof is given in \prettyref{proof:prop-nam1D}.
%We first consider the \twd case (proof in \prettyref{proof:prop-nam1D}).
\begin{proposition}[{\bf Exact Recovery of \twd depth \signals}]
\label{prop:nam1D}
Let $\xtrue \in \Real{n}$ be a \twd depth \signal with corner set $\calC$.
Assuming noiseless measurements~\eqref{eq:measurements}, the following hold:
\begin{enumerate}[label=(\roman*)]
\item if the sampling set $\samples$ is the union of the corner set and the first and last entries of $\xtrue$, then $\nam = 1$;
\item if the sampling set $\samples$ includes the corners and their neighbors (adjacent entries), then $\nam = 0$ 
and problem~\eqref{eq:BP} recovers $\xtrue$ exactly.
\end{enumerate}
\end{proposition}

\begin{figure}[hbt]
\begin{minipage}{\textwidth}
\hspace{-0.5cm}
\begin{tabular}{cc}\begin{minipage}{5cm}\centering\includegraphics[height=2.5cm]{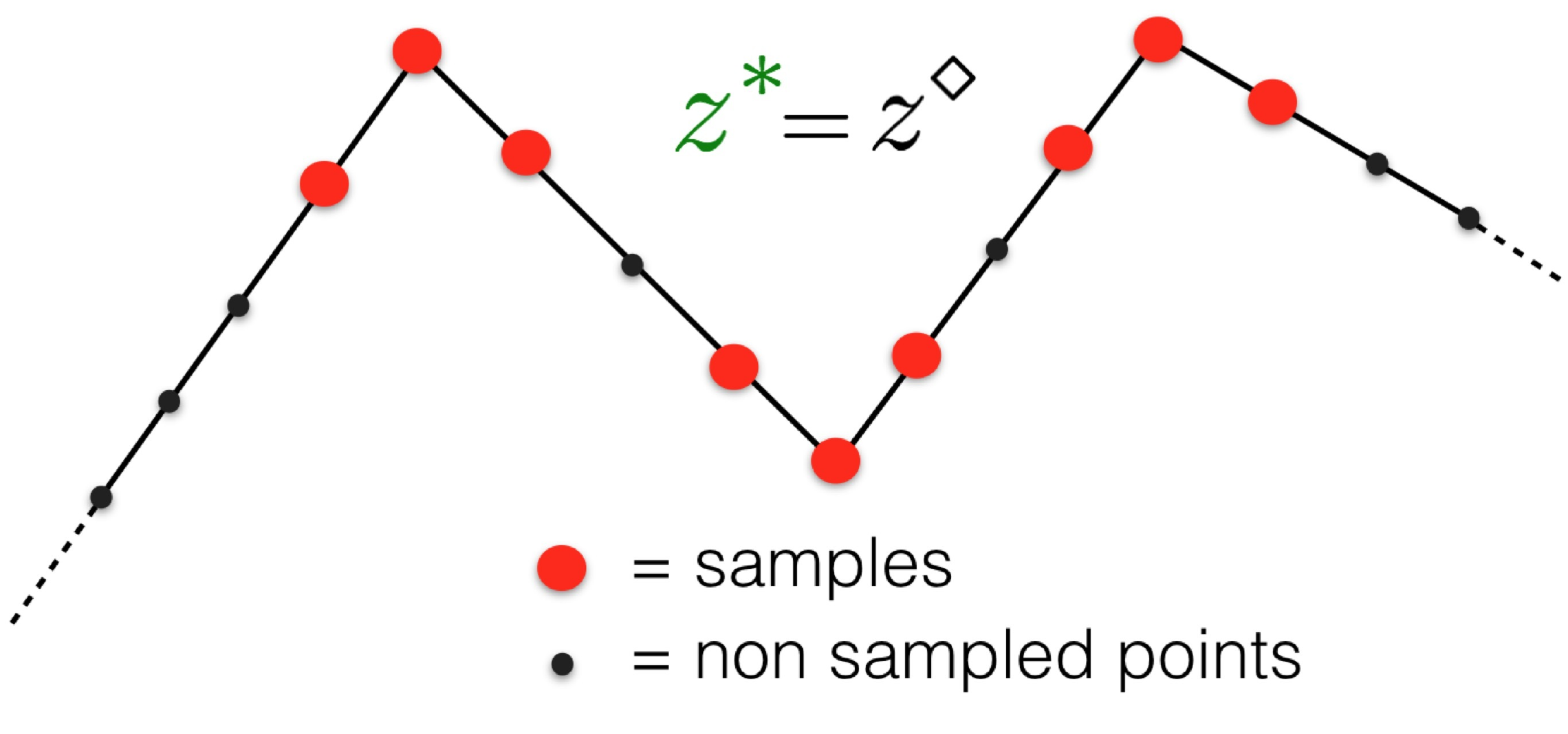}
\\
\hspace{-10mm} (a)
\end{minipage}
&  
\begin{minipage}{3cm}\centering\includegraphics[height=2.5cm]{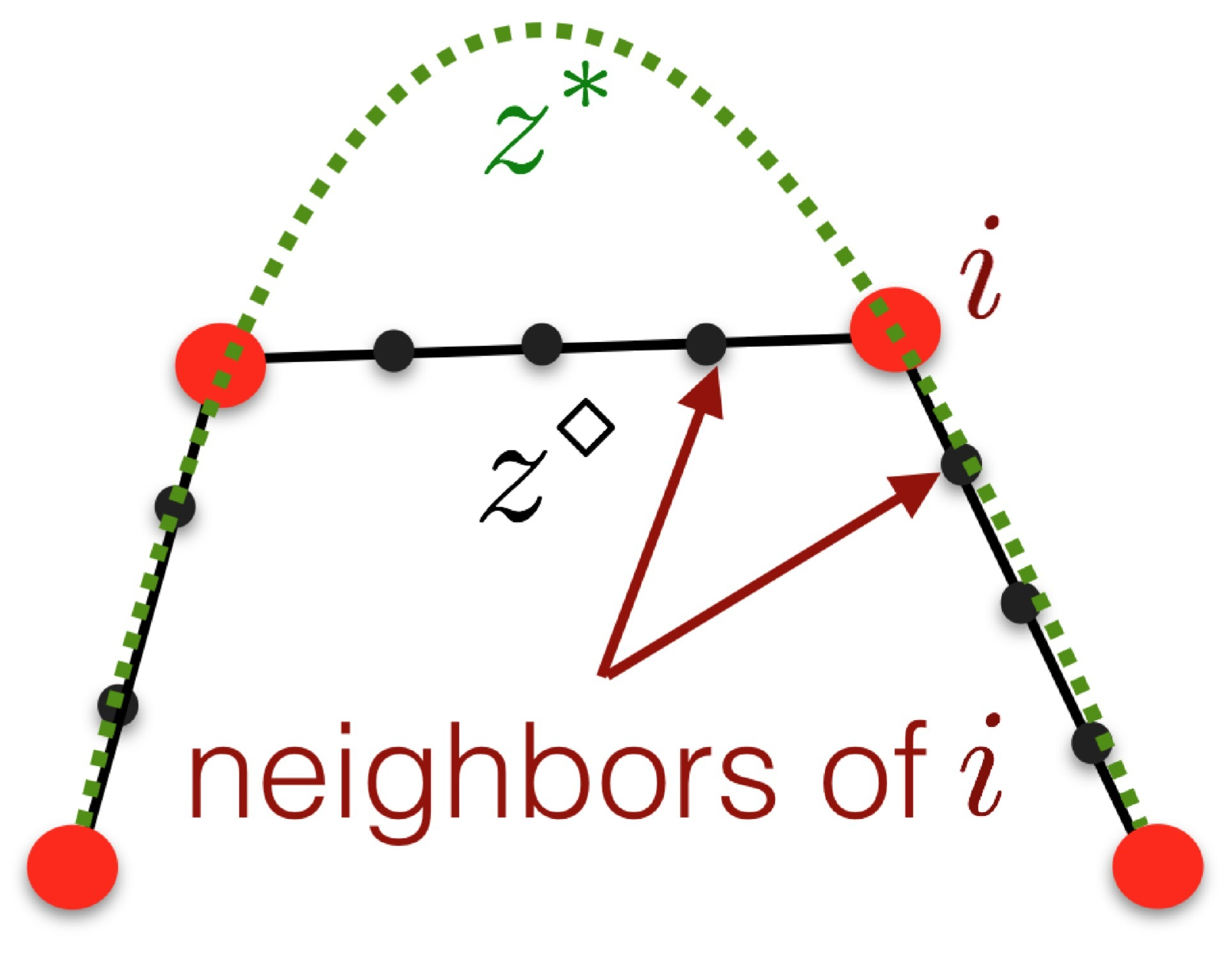} 
\\
\hspace{-20mm} (b)
\end{minipage}
\end{tabular}
\end{minipage}
\caption{
(a) Sampling corners and neighbors in \twd depth estimation guarantees exact recovery of 
the true \signal. (b) Sampling only the corners does not guarantee, in general, that 
the solution of the  $\ell_1$-minimization, namely $\xopt$, coincides with the true \signal $\xtrue$.
\vspace{-0.2cm}
}
\label{fig:samplingCorners} 
\end{figure}

\prettyref{prop:nam1D} implies that we can recover the original \signal exactly, if we measure the neighborhood of each corner. 
An example that satisfies such condition is illustrated in~\prettyref{fig:samplingCorners}(a). 
When we sample only the corners, however, \prettyref{prop:nam1D} states that $\nam = 1$;
in principle in this case one might still hope to recover the \signal $\xtrue$, since the 
condition $\nam < 1$ in Proposition~\ref{prop:nam} is only \emph{sufficient} for exact recovery. But it turns out that in our problem one can find counterexamples with $\nam = 1$ in which $\ell_1$-minimization fails to recover $\xtrue$. A pictorial example is shown in \prettyref{fig:samplingCorners}(b), where we show an optimal solution $\xopt$ which differs from the true \signal $\xtrue$.

We derive a similar condition for \thd problems. The proof is given in \prettyref{proof:prop-nam2D}.

\begin{proposition}[{\bf Exact Recovery of \thd depth \signals}]
\label{prop:nam2D}
Let $\Xtrue$ be a \thd depth \signal with edge set $\calE$.
Assume noiseless measurements. If 
the sampling set $\samples$ includes the edges and theirs (vertical and horizontal) neighbors (adjacent pixels), 
then $\nam = 0$, and~\eqref{eq:BPmatrix} recovers $\vec{\Xtrue}$ exactly.
\end{proposition}

In the experimental section, we show that these initial results already unleash interesting applications.
For instance, in stereo vision problems, we could locate the position of the edges from the RGB images and 
recover the depth in a neighborhood of the edge pixels.
Then, the complete depth \signal can be recovered 
(at arbitrary resolution) via~\eqref{eq:BPmatrix}.

%% file: analysis/BP_noiseless.tex
%!TEX root = ../main.tex

\subsection{Depth Reconstruction from Noiseless Samples}
\label{sec:optimality}

The exact recovery conditions of~\prettyref{prop:nam1D} and~\prettyref{prop:nam2D} are quite restrictive if we do not have prior knowledge of the position of the corners or edges. In this section we provide more powerful results that do not require sampling corners or edges. Empirically, we observe that when we do not sample all the edges, the optimization problems~\eqref{eq:BP} and~\eqref{eq:BPmatrix} admit multiple solutions, i.e., multiple \signals $\xvar$ attain the same optimal cost. The basic questions addressed in this section are: \emph{which \signals are in the solution set $\opt{\calS}$ of problems~\eqref{eq:BP} and~\eqref{eq:BPmatrix}? Is the ground truth \signal $\xtrue$ among these optimal solutions? How far can an optimal solution be from the ground truth \signal $\xtrue$?} In order to answer these questions, in this section we derive optimality conditions for problems~\eqref{eq:BP} and~\eqref{eq:BPmatrix}, under the assumption that all measurements are noise-free.

%% file: analysis/BP_noiseless_algebraic.tex
%!TEX root = ../main.tex

\somespace
\subsubsection{{\bf Algebraic Optimality Conditions (noiseless samples)}}
\label{sec:optimality_algebraic}
In this section, we derive a general algebraic condition for a \twd \signal (resp. \thd) to be in the solution set of~\eqref{eq:BP} (resp.~\eqref{eq:BPmatrix}). \prettyref{sec:BP_optimality1D} and~\prettyref{sec:BP_optimality2D} translate this algebraic condition into a geometric constraint on the curvature of the \signals in the solution set.

\begin{proposition}[{\bf \twd Optimality}]
\label{prop:subdifferential}
Let $\matU$ be the sampling matrix and 
$\samples$ be the sample set.  Given a \signal $\xvar \in \Real{n}$ which is feasible for~\eqref{eq:BP}, $\xvar$ is a
  minimizer of~\eqref{eq:BP} if and only if there exists a vector $u$ such that \beq
\label{eq:optimality}
(\tv\tran)_\cosamples \; u = \zero \; \text{ and } \; u_\support = \sign(\tv \xvar)_\support \; \text{ and } \; \linf{u} \leq 1
 \eeq
 where    $\support$ is the $\tv$-support of $\xvar$ 
(i.e., the set of indices of the nonzero entries of $\tv \xvar$)
  and $\cosamples$ is the set of entries of $\xvar$ that we do not sample (i.e., the complement of $\samples$).
\end{proposition}

The proof of \prettyref{prop:subdifferential} is based on the subdifferential of the $\ell_1$-minimization problem and is provided in \prettyref{proof:subdifferential}. An analogous result holds in \thd.

\begin{corollary}[{\bf \thd optimality}]
\label{cor:subdifferential2D}
 A given \signal $\Xvar$ is in the 
 set of minimizers of~\eqref{eq:BPmatrix} if and only if the conditions of~\prettyref{prop:subdifferential} 
 hold, replacing $\tv$ with $\TV$ in eqs.~\eqref{eq:optimality}.  
\end{corollary}

We omit the proof of \prettyref{cor:subdifferential2D} since it follows the same line of the proof of \prettyref{prop:subdifferential}.

%% file: analysis/BP_noiseless_2D.tex
%!TEX root = ../main.tex

\somespace
\subsubsection{{\bf Analysis of \twd Reconstruction (noiseless samples)}}  
\label{sec:BP_optimality1D}

In this section we 
derive \emph{necessary and sufficient geometric conditions}
for $\xtrue$ to be in the solution set of~\eqref{eq:BP}. 
Using these findings we obtain two practical results: (i) an upper bound on how far any solution $\xopt$ of~\eqref{eq:BP} can be from the ground truth \signal $\xtrue$; (ii) a general algorithm that recovers $\xtrue$ even when the conditions of~\prettyref{prop:nam1D} fail 
(the algorithm is presented in~\prettyref{sec:algorithmicVariants}).

To introduce our results, we need the following definition.

\begin{definition}[{\bf \twd Sign Consistency}]
\label{def:1DsignConsitency}
Let $s_k = \sign(\xvar_{k-1}-2\xvar_{k} + \xvar_{k+1})$ (sign of the curvature at $k$).
A \twd depth \signal $\xvar$ is \emph{sign consistent} if, 
for any two consecutive samples $i < j \in \samples$, one of the two conditions holds:
\begin{enumerate}[label=(\roman*)]
\item no sign change: for any two integers $k,h$, with $i \leq k,h \leq j$, if $s_k \neq 0$  
and  $s_h \neq 0$, then $s_k = s_h$;
\item sign change \emph{only} at the boundary: for any integer $k$, with $i < k < j$, $s_k = 0$; 
\end{enumerate}
\end{definition}

\begin{figure}[h]
\myIncludeTwoFigures{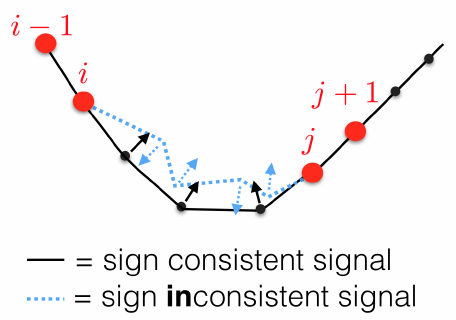}{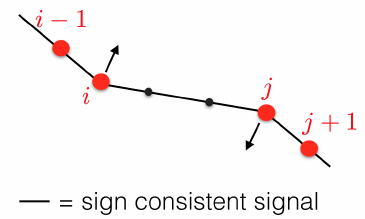}{0}{0}{0}{0}
\caption{Examples of sign-consistent (shown in black) and sign-inconsistent (shown in blue) \twd depth \signal.
}
\label{fig:signConsistency} 
\end{figure}

This technical definition has a clear geometric interpretation. In words, a \signal $\xvar$ is sign consistent, if its curvature does not change sign (i.e., it is either convex or concave) within each interval between consecutive samples. See~\prettyref{fig:signConsistency} for examples of sign consistency, alongside with a counter-example.

In the following we show that any optimal solutions for problem~\eqref{eq:BP} must be sign consistent.
In order to simplify the analysis for \prettyref{thm:1Doptimality} below, we assume that we pick pairs of consecutive samples (rather than individual, isolated samples). We formalize this notion as follows.  % ``twin samples'' 

\begin{definition}[{\bf Twin samples}]
\label{def:twinSamples}
A \emph{twin sample} is a pair of consecutive samples, i.e., $(i,i+1)$ with $i \in \{1,\ldots,n-1\}$. \end{definition}

\begin{theorem}[{\bf \twd Sign Consistency $\Leftrightarrow$ Optimality}] \label{thm:1Doptimality}
Let $\xvar$ be a \twd \signal which is feasible for problem~\eqref{eq:BP}.
Assume that the sample set includes only twin samples and we sample the ``boundary'' of the 
\signal, i.e., $\xvar_1$,  and $\xvar_n$. Then, $\xvar$ is optimal for 
\eqref{eq:BP} if and only if it is sign consistent. 
\end{theorem}

The proof of \prettyref{thm:1Doptimality} is given in \prettyref{proof:thm-1Doptimality}. This theorem provides a tight geometric condition for a \signal to be optimal.
More specifically, a \signal is optimal for problem~\eqref{eq:BP} if it passes through the 
given set of samples (i.e., it satisfies the constraint in~\eqref{eq:BP}) and does not 
change curvature between consecutive samples.
This result also provides insights into the conditions under which the ground truth \signal will be among the minimizers of~\eqref{eq:BP}, and how one can bound the depth estimation error, as stated in the following proposition.

\begin{proposition}[{\bf \twd Recovery Error - noiseless samples}] \label{prop:1DrecoveryError}
Let $\xtrue$ be the ground truth \signal generating noiseless measurements~\eqref{eq:measurements}.
Assume that we sample the boundary of $\xtrue$ and the sample set includes a twin 
sample in  each linear segment in $\xtrue$. Then, $\xtrue$ is in the set of minimizers of~\eqref{eq:BP}.
Moreover, denote with $\xnaive$ the naive solution 
obtained by connecting consecutive samples with a straight line (linear interpolation).
 Then, any optimal solution $\xopt$ lies between $\xtrue$ and $\xnaive$, i.e.,
for any index $i \in \onen$, it holds $\min(\xtrue_i, \xnaive_i) \leq \xopti \leq \max(\xtrue_i, \xnaive_i)$.
% that if $\xtrue_i \leq \xnaive_i$ then $\xtrue_i \leq \xopti  \leq \xnaive_i$, and if $\xnaive_i \leq \xtrue_i$ 
% then $\xnaive_i \leq \xopti\leq \xtrue_i$. 
Moreover, it holds
\beq
\label{eq:recoveryError}
\linf{\xtrue - \xopt} \leq \max_{i \in \samples} d_i \cos(\theta_i)
\eeq
where $d_i$ is the distance between the sample $i$ and the nearest corner in $\xtrue$, 
while $\theta_i$ is the angle that the line connecting $i$ with the nearest corner forms with the vertical. 
\end{proposition}

\begin{figure}[h]
\begin{minipage}{\textwidth}
\hspace{-3mm}
\begin{tabular}{cc}
\begin{minipage}{\textwidth}
\centering
\includegraphics[height=3.8cm]{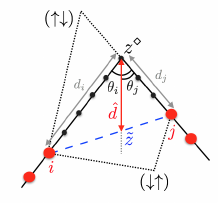} \\
\vspace{-1mm}
(a)
\end{minipage}
% &  \;\;
\\
\begin{minipage}{\textwidth}
\centering
\includegraphics[width=\textwidth]{envelop1D} \\
\vspace{-1mm}
(b)
\end{minipage}
\end{tabular}
\end{minipage}
\caption{
(a) Region between a pair of twin samples, with twin samples not including corners.
(b) Given a set of noiseless twin samples, all possible optimal solutions in \twd are contained in the envelope shown in gray. }
\label{fig:toyExample}
\end{figure}

A visualization of the parameters $d_i$ and $\theta_i$ is given in~\prettyref{fig:toyExample}(a). The proof of \prettyref{prop:1DrecoveryError}
 is given in \prettyref{proof:1DrecoveryError}.

\prettyref{prop:1DrecoveryError} provides two important results.
First, it states that any optimal solution $\xopt$ (e.g., the dotted green line in \prettyref{fig:nam1}(b)) 
should lie between the ground truth depth $\xtrue$ (solid black line) and the naive solution $\xnaive$ (dashed blue line).
In other words, any arbitrary set of twin samples defines an \emph{envelope} that contains all possible solutions. An example of such \emph{envelope} is illustrated in \prettyref{fig:toyExample}(b). The width of this envelope bounds the maximum 
distance between any optimal solution and the ground truth, and hence such envelope provides a \emph{point-wise quantification of the reconstruction error}. 
%(we used this point-wise error bound to derive the global bound in eq.\eqref{eq:recoveryError}). 
Second, \prettyref{prop:1DrecoveryError} provides an upper bound on the overall reconstruction error in eq.~\eqref{eq:recoveryError}. The inequality implies that the reconstruction error grows with the parameter $d_i$, the distance between our samples and the corners. In addition, the error also increases if the parameter $\theta_i$ is small, meaning that the ground truth \signals are ``pointy'' and there exist abrupt changes of slope between consecutive segments. An instance of such ``pointy'' behavior is the second corner from right in~\prettyref{fig:toyExample}(b).

We will further show in \prettyref{sec:algorithms} that \prettyref{prop:1DrecoveryError} has algorithmic implications. Based on \prettyref{prop:1DrecoveryError}, we design an algorithm that exactly recovers a \twd \signal, even when the sample set does not contain all corners. Before moving to algorithmic aspects, let us consider the \thd case.

\veryoptional{
This result is intuitive in the sense that closer we sample to the corners, smaller recovery error we achieve. Also the error decreasing at wider angled corners is useful from a 
robotic point of view since, one would care less about recovering very narrow corners truthfully when compared to wider ones. 

where $d_k$ is the distance between between sample $k$ and the nearby corner in $\xtrue$, 
while $\theta_k$ is the angle that the line passing through the twin samples (say $k-1$ and $k$) 
forms with the vertical, see~\prettyref{fig:toyExample}(right).

\marginpar{\red{Clear the proposition statement, angle etc}}

Informally,~\prettyref{prop:1DrecoveryError} says that any optimal solution if ``between'' 
the naive solution $\xnaive$ (obtained by connecting the dots, 
see the blue dashed line in~\prettyref{fig:example1D})  and the 
true solution $\xtrue$ (solid black line in~\prettyref{fig:example1D}).
}

%% file: analysis/BP_noiseless_3D.tex
%!TEX root = ../main.tex

\somespace
\subsubsection{{\bf Analysis of \thd Reconstruction (noiseless samples)}}  
\label{sec:BP_optimality2D}

In this section we provide a sufficient geometric condition for a \thd \signal to be in the solution set of \eqref{eq:BPmatrix}.
We start by introducing a specific sampling strategy (the analogous of the twin 
samples in \twd) to simplify the analysis.

\begin{definition}[{\bf Grid samples and Patches}] 
\label{def:2DgridSamples} 
Given a \thd \signal $\Xvar \in \R^{r \times c}$, a \emph{grid sample set} 
includes pairs of consecutive rows and columns of $\Xvar$, along with the boundaries (first and last two rows, first and last two columns). This sampling strategy divides the image in rectangular \emph{patches}, i.e., sets of \unsampled pixels enclosed by row-samples and column-samples.
\end{definition}

% \begin{figure}[bht]
% \centering
% \includegraphics[scale=1.5]{gridSample}
% \caption{Illustration of a grid sample set, along with 6 \unsampled patches in white.}
% \label{fig:gridSamples}
% \end{figure}

\begin{figure}[htbp]
\begin{minipage}{\textwidth}
\hspace{-3mm}
\begin{tabular}{cc}
\begin{minipage}{0.49\textwidth}
\centering
\includegraphics[width=\linewidth]{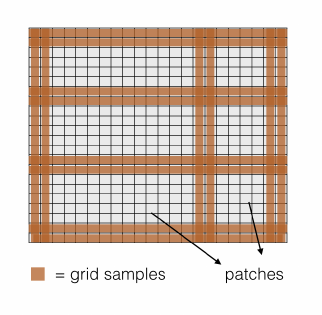} \\
\vspace{-3mm}
(a)
\end{minipage}
% &  \;\;
\begin{minipage}{0.5\textwidth}
\centering
\text{} \\ \vspace{10mm}
\includegraphics[width=\linewidth]{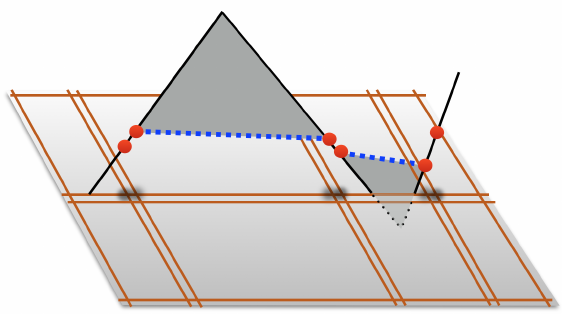} \\
\vspace{5mm}
(b)
\end{minipage}
\end{tabular}
\end{minipage}
\caption{(a) Illustration of a grid sample set, along with 6 \unsampled patches in white. (b) 
A cross section of the envelope in \thd.
}
\label{fig:envelop1D} 
\end{figure}

\prettyref{fig:envelop1D}(a) shows an example of grid samples and patches. If we have $K$ patches and we denote the set of \unsampled pixels in patch $i$ with $\cosamples_i$, 
then the union $\samples \cup \{\cosamples_i\}_{i=1}^K$ includes all the pixels in the depth image.
We can now extend the notion of sign consistency to the \thd case.
\veryoptional{
\begin{figure}[t]
\centering
\includegraphics[scale=0.4]{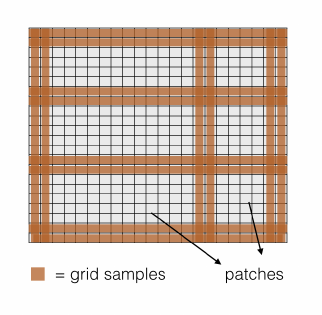}
 \caption{Grid samples and patches. \label{fig:gridSample}}
 \vspace{-0.5cm}
\end{figure}
}

\begin{definition}[{\bf \thd Sign Consistency}]
\label{def:2DsignConsitency} 
Let $\Xvar \in \R^{r \times c}$ be a \thd depth \signal. 
Let $\samples$ be a grid sampling set and $\{\cosamples_i\}_{i=1}^K$ be the \unsampled patches. 
Let $\XvarMi$ be the restriction of $\Xvar$ to its entries in $\cosamples_i$. Then, $\Xvar$ is called \emph{\thd sign consistent} if for all $i =\{1,\ldots,K\}$, the nonzero entries of 
$\sign(\text{vec}(\tv \XvarMi))$ are all $+1$ or $-1$, 
and the nonzero entries of $\sign(\text{vec}(\XvarMi \tv\tran))$ are all $+1$ or $-1$, where $\tv$ is the \second-order difference 
operator \eqref{eq:tv} of suitable dimension.
\end{definition}

Intuitively, \thd sign consistency indicates that the sign of the \signal's curvature does not change, either horizontally or vertically, within each \unsampled patch.
We now present a sufficient condition for $\Xtrue$ to be in the solution set of~\eqref{eq:BPmatrix}. \begin{theorem}[{\bf \thd Sign Consistency $\Rightarrow$ Optimality}] \label{thm:2Doptimality_xtrue}
Let $\Xvar \in \R^{r \times c}$ be a \thd \signal, feasible for problem \eqref{eq:BPmatrix}. Assume the sample set $\samples$ is a grid sample set. Then $\Xvar$ is in the set of minimizers of \eqref{eq:BPmatrix} if it is \thd sign consistent.
\end{theorem}

The proof is given in \prettyref{proof:thm-2Doptimality_xtrue}. 
\prettyref{thm:2Doptimality_xtrue} is weaker than \prettyref{thm:1Doptimality}, the \twd counterpart, since our definition of 
\thd sign consistency is only \emph{sufficient}, but not necessary, for optimality. Nevertheless, it can be used to 
bound the depth recovery error as follows.

\begin{proposition}[{\bf \thd Recovery Error - noiseless samples}] \label{prop:2DrecoveryError}
Let $\Xtrue \in \R^{r \times c}$ be the ground truth \signal generating noiseless measurements~\eqref{eq:measurements}.
Let $\samples$ be a grid sampling set and assume $\Xtrue$ to be \emph{\thd sign consistent} with respect to $\samples$. 
Moreover, 
let $\underbar{\Xvar} \in \R^{r \times c}$ and $\bar{\Xvar} \in \R^{r \times c}$ be the point-wise lower and upper bound of the 
\emph{row-wise envelope}, built as in~\prettyref{fig:toyExample}(b) by considering each row of the \thd depth \signal as a \twd \signal.
Then, $\Xtrue$ is an optimal solution of~\eqref{eq:BPmatrix},
and any other optimal solution $\Xopt$ of~\eqref{eq:BPmatrix} satisfies:
\beq
\label{eq:recoveryError2D}
|\Xtrue_{i,j} - \Xvar^\star_{i,j}| \leq \max(|\underbar{\Xvar}_{i,j} - \Xopt_{i,j}| , |\bar{\Xvar}_{i,j} - \Xopt_{i,j}|)
\eeq
\end{proposition}

Roughly speaking, if our grid sampling is ``fine'' enough to capture all changes in the sign of the curvature of $\Xtrue$, then $\Xtrue$ is among the solutions of \eqref{eq:BPmatrix}. 
Despite the similarity to~\prettyref{prop:1DrecoveryError}, the result in~\prettyref{prop:2DrecoveryError} is weaker. More specifically, \prettyref{prop:2DrecoveryError} is based on the fact that we can compute an envelope \emph{only} for the ground truth \signal (but \emph{not} for all the optimal solutions, as in~\prettyref{prop:1DrecoveryError}). 
Moreover, the estimation error bound in eq.~\eqref{eq:recoveryError2D} can be only computed \emph{a posteriori},
i.e., after obtaining an optimal solution $\Xopt$. Nevertheless, the result can be readily used in practical applications, in which one wants
to bound the depth estimation error.
%The resulting error bounds are still important in practice .
An example of the row-wise envelope is given in \prettyref{fig:envelop1D}(b).
% 
%Therefore, \prettyref{prop:2DrecoveryError} does not provide a geometric understanding as clear as the bound in eq.~\eqref{eq:recoveryError}.

%% file: analysis/BP_noisy.tex
\subsection{Depth Reconstruction from Noisy Samples}
\label{sec:stableRecovery}

In this section we analyze the depth reconstruction quality 
for the case where the measurements \eqref{eq:measurements} are noisy. 
In other words, we now focus on problems~\eqref{eq:BPD} and~\eqref{eq:BPDmatrix}.

%% file: analysis/BP_noisy_algebraic.tex
%!TEX root = ../main.tex

\somespace
\subsubsection{{\bf Algebraic Optimality Conditions (noisy samples)}}
\label{sec:optimality_algebraic_noisy}

In this section, we derive a general algebraic condition for a \twd \signal (resp. \thd) to be in the solution set of~\eqref{eq:BPD} (resp.~\eqref{eq:BPDmatrix}). This condition generalizes the optimality condition of~\prettyref{sec:optimality_algebraic} to the noisy case. In \prettyref{sec:BPD_optimality1D} and~\prettyref{sec:BPD_optimality2D}, we apply this algebraic condition to bound the depth reconstruction error.

\begin{proposition}[{\bf \twd robust optimality}]
\label{prop:robust_subdifferential}
Let $\matU$ be the sampling matrix, $\samples$ be the sample set and $\meas$ be the noisy measurements as in \eqref{eq:measurements}, with $\linf{\error} \leq \vareps$ and $\vareps > 0$. Given a \signal $\xvar$ which is feasible for~\eqref{eq:BPD}, 
define the active set $\calA \subset \samples$ as follows
\begin{align}
\label{eq:activeset}
\calA = \{i \, \in \samples: |\meas_i - \xvar_i| = \vareps \}.
\end{align}
We also define its two subsets
\begin{align}
\label{eq:activesubset}
\ActUp &= \{i \, \in \samples: \meas_i - \xvar_i = \vareps \}, \notag \\
\ActDown &= \{i \, \in \samples: \meas_i - \xvar_i = -\vareps \}.
\end{align}
Also denote $\bar{\calA} = \samples \setminus \calA$. Then $\xvar$ is a
  minimizer of~\eqref{eq:BPD} if and only if there exists a vector $u$ such that \begin{align}
&\hspace{-0.5cm}(\tv\tran)_{\cosamples \cup \bar{\calA}}\; u \!=\! \zero \; \text{ and } \; u_\support \!=\! \sign(\tv \xvar)_\support \; \text{ and } \; \linf{u} \leq 1 \label{eq:robust_optimality1}  \\
&(\tv\tran)_\ActUp \; u \geq \zero  \; \text{ and } \; (\tv\tran)_\ActDown \; u \leq \zero \label{eq:robust_optimality}
 \end{align}
 where $\support$ is the $\tv$-support of $\xvar$, 
 and $\cosamples$ is the set of un-sampled entries in $\xvar$ (i.e., the complement of $\samples$).
\end{proposition}

\begin{figure}[htbp]
\includegraphics[width=0.7\textwidth]{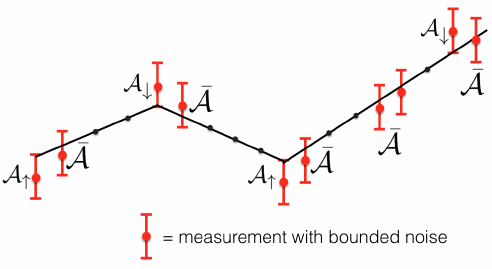}
\caption{A illustration of the \emph{active set} $\calA$ of samples in \prettyref{prop:robust_subdifferential}. The three corners, as well as the second sample from the right, are all within the active set. A measurement $y_i$ is \emph{active} if the reconstructed \signal \xvar{ }hits the boundary of $y_i$'s associated error bound.
}
\label{fig:activeSet}
\end{figure}

The proof is given in \prettyref{proof:prop-robust_subdifferential}. A visual illustration of the active set is given in \prettyref{fig:activeSet}. We will provide some geometric insights on the algebraic conditions in \prettyref{prop:robust_subdifferential} in the next two sections. Before moving on, we re-ensure that the robust optimality conditions straightforwardly extends to the \thd case.

\begin{corollary}[{\bf \thd robust optimality}]
\label{cor:robust_subdifferential2D}
 A given \signal $\Xvar$ is in the 
 set of minimizers of~\eqref{eq:BPDmatrix} if and only if the conditions of~\prettyref{prop:robust_subdifferential} 
 hold, replacing $\TV$ with $\tv$ in eqs.~\eqref{eq:robust_optimality1}-\eqref{eq:robust_optimality}.  
\end{corollary}
We skip the proof of \prettyref{cor:robust_subdifferential2D} since it proceeds along the same line of the proof of \prettyref{prop:robust_subdifferential}.

%% file: analysis/BP_noisy_2D.tex
%!TEX root = ../main.tex
% \somespace
\subsubsection{{\bf Analysis of \twd Reconstruction (noisy samples)}}  
\label{sec:BPD_optimality1D}

In this section we consider the \twd case and provide a geometric interpretation of the algebraic conditions in~\prettyref{prop:robust_subdifferential}. 
The geometric interpretation follows from a basic observation, which enables us to relate the noisy case with our noiseless analysis of~\prettyref{sec:BP_optimality1D}.
The observation is that if a \signal satisfies the robust optimality conditions~\eqref{eq:robust_optimality1}-\eqref{eq:robust_optimality} 
then it also satisfies the noiseless optimality condition~\eqref{eq:optimality}, hence being sign consistent, 
as per~\prettyref{thm:1Doptimality}.

\begin{theorem}[{\bf Robust optimality $\Rightarrow$ \twd Sign Consistent}] \label{thm:1Doptimality_robust}
Let $\xopt$ be a \twd \signal which is optimal for problem~\eqref{eq:BPD}, and 
assume that the sample set includes only twin samples and we sample the ``boundary'' of the 
\signal, i.e., $\xvar_1$,  and $\xvar_n$. Then, $\xopt$ is \twd sign consistent. 
\end{theorem} 
We present a brief proof for \prettyref{thm:1Doptimality_robust} below. 
\begin{proof}
Note that
$%\begin{equation*}
(\tv\tran)_{\cosamples \cup \overline{\calA}}\; u \!=\! \zero  \;\; \Rightarrow \;\; (\tv\tran)_{\cosamples}\; u \!=\! \zero 
$. %\end{equation*}
In other words, condition~\eqref{eq:robust_optimality1} implies condition~\eqref{eq:optimality}, which in turn is equivalent to sign consistency as per \prettyref{thm:1Doptimality}. 
Therefore, we come to the conclusion that any optimal solution of~\eqref{eq:BPD} must be sign consistent.
\end{proof}

\prettyref{thm:1Doptimality_robust} will help establish error bounds on the depth reconstruction. Before presenting these bounds, 
we formally define the \emph{\twd sign consistent \noisyEnvelop}.

\begin{definition}[{\bf \twd Sign Consistent \noisyEnvelop}]
\label{def:scEnvelop}
Assume that the sample set includes only twin samples and we sample the ``boundary'' of the 
\signal, i.e., $\xvar_1$,  and $\xvar_n$.
Moreover, for each pair of consecutive twin samples $i,i+1$ and $j,j+1$, 
define the following line segments for $k \in (i+1,j)$: 
%envelope contains the set of all \twd sign-consistent $\xvar$, such that for all samples $k \in (i+1, j)$, 
\begin{enumerate}[label*=(\arabic*)]
  % [label=(\roman*)]
\item $\frac {\xvar_k - (y_i - \vareps)} {x_k - x_i} = \frac {(y_{i+1} + \vareps) - (y_i - \vareps)} {x_{i+1} - x_i}$
\item $\frac {\xvar_k - (y_i + \vareps)} {x_k - x_i} = \frac {(y_{i+1} - \vareps) - (y_i - \vareps)} {x_{i+1} - x_i}$
\item $\frac {\xvar_k - (y_{j+1} - \vareps)} {x_k - x_{j+1}} = \frac {(y_{j} + \vareps) - (y_{j+1} - \vareps)} {x_j - x_{j+1}}$
\item $\frac {\xvar_k - (y_{j+1} + \vareps)} {x_k - x_{j+1}} = \frac {(y_{j} - \vareps) - (y_{j+1} + \vareps)} {x_j - x_{j+1}}$
\item $\frac {\xvar_k - (y_{i+1} + \vareps)} {x_k - x_{i+1}} = \frac {(y_{j} + \vareps) - (y_{i+1} + \vareps)} {x_j - x_{i+1}}$
\item $\frac {\xvar_k - (y_{i+1} - \vareps)} {x_k - x_{i+1}} = \frac {(y_{j} - \vareps) - (y_{j+1} - \vareps)} {x_j - x_{j+1}}$
\end{enumerate}
Further define the following \signals:
\begin{align*}
&\bar{\xvar} := \left\{ \begin{array}{ll} \mathrm{\max\{(1), (3), (5)\}},& \ \ \ \text{if (1) and (3) intersect} \\ \mathrm{(5)},& \ \ \ \text{otherwise} \end{array} \right. \\
&\text{and} \\
&\underbar{\xvar} := \left\{ \begin{array}{ll} \mathrm{\max\{(2), (4), (6)\}},& \ \ \ \text{if (2) and (4) intersect} \\ \mathrm{(6)},& \ \ \ \text{otherwise} \end{array} \right. \\
\end{align*}
where $\mathrm{\max\{(1), (3), (5)\}}$ denotes the point-wise maximum among the segments in eqs.~(1), (3), and (5).
We define the \twd \emph{sign consistent \noisyEnvelop} as the region enclosed between 
the upper bound $\bar{\xvar} \in \Real{n}$ and the lower bound $\underbar{\xvar} \in \Real{n}$. % as follows:
% Then we have two cases: 
% \begin{enumerate}[label=(\roman*)]
% \item if none of the lines (1) or (2) intersects with lines (3) or (4), then define \twd \emph{sign consistency \noisyEnvelop} as the set between lines (5) and (6).
% \item otherwise, define \twd \emph{sign consistency \noisyEnvelop} as the set between pointwise $\max\{(1), (3), (5)\}$ and pointwise $\min\{(2),(4),(6)\}$.  
% \end{enumerate}

\end{definition}

\begin{figure}[htb]
\begin{tabular}{c}
\begin{minipage}{\textwidth}
\hspace{-3mm}
\begin{tabular}{cc}
\begin{minipage}{0.6\textwidth}
\centering
\includegraphics[width=\textwidth]{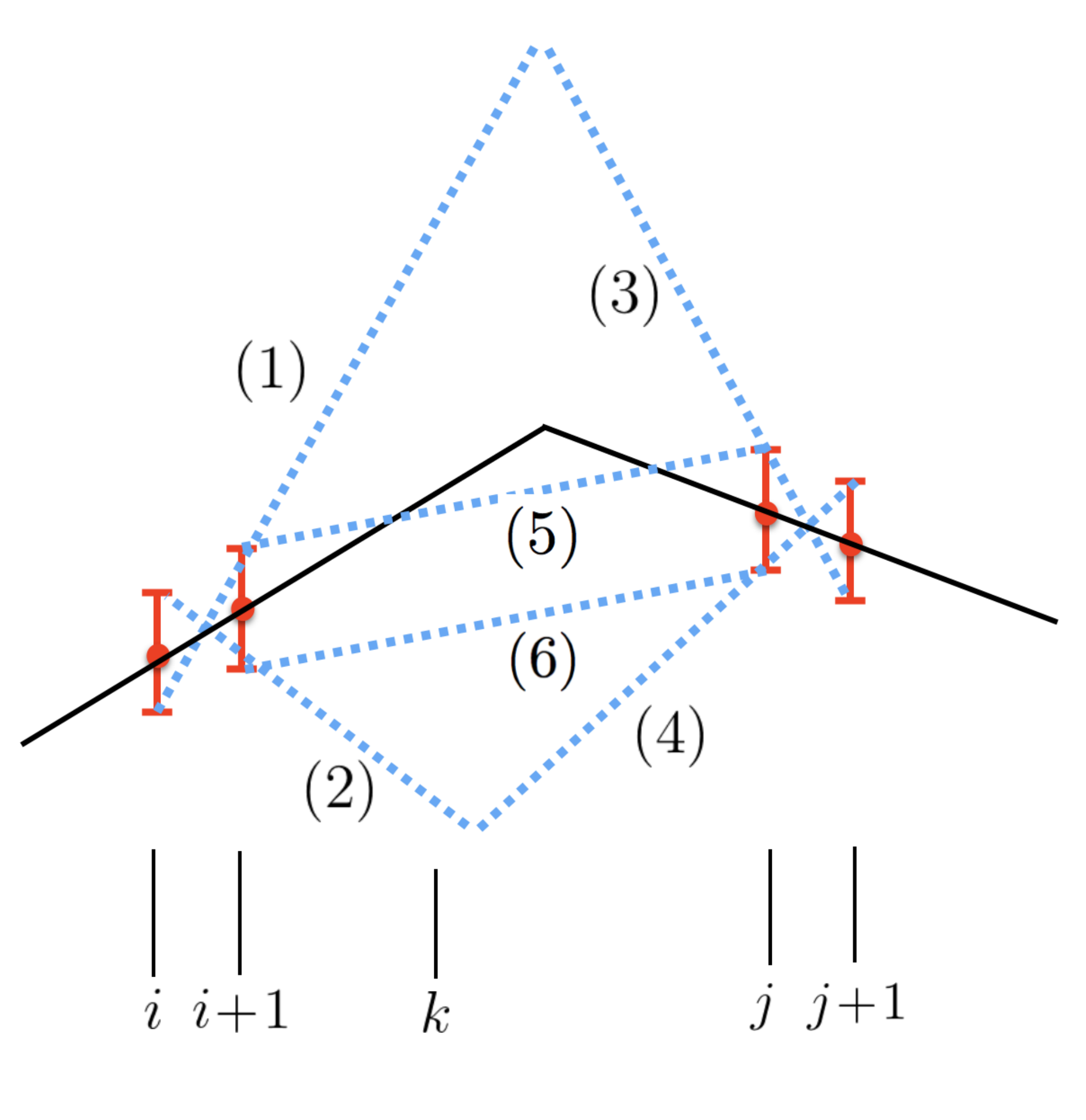} \\
\vspace{-1mm}
(a)
\end{minipage}
% &  \;\;
\begin{minipage}{0.5\textwidth}
\centering
\includegraphics[width=\textwidth]{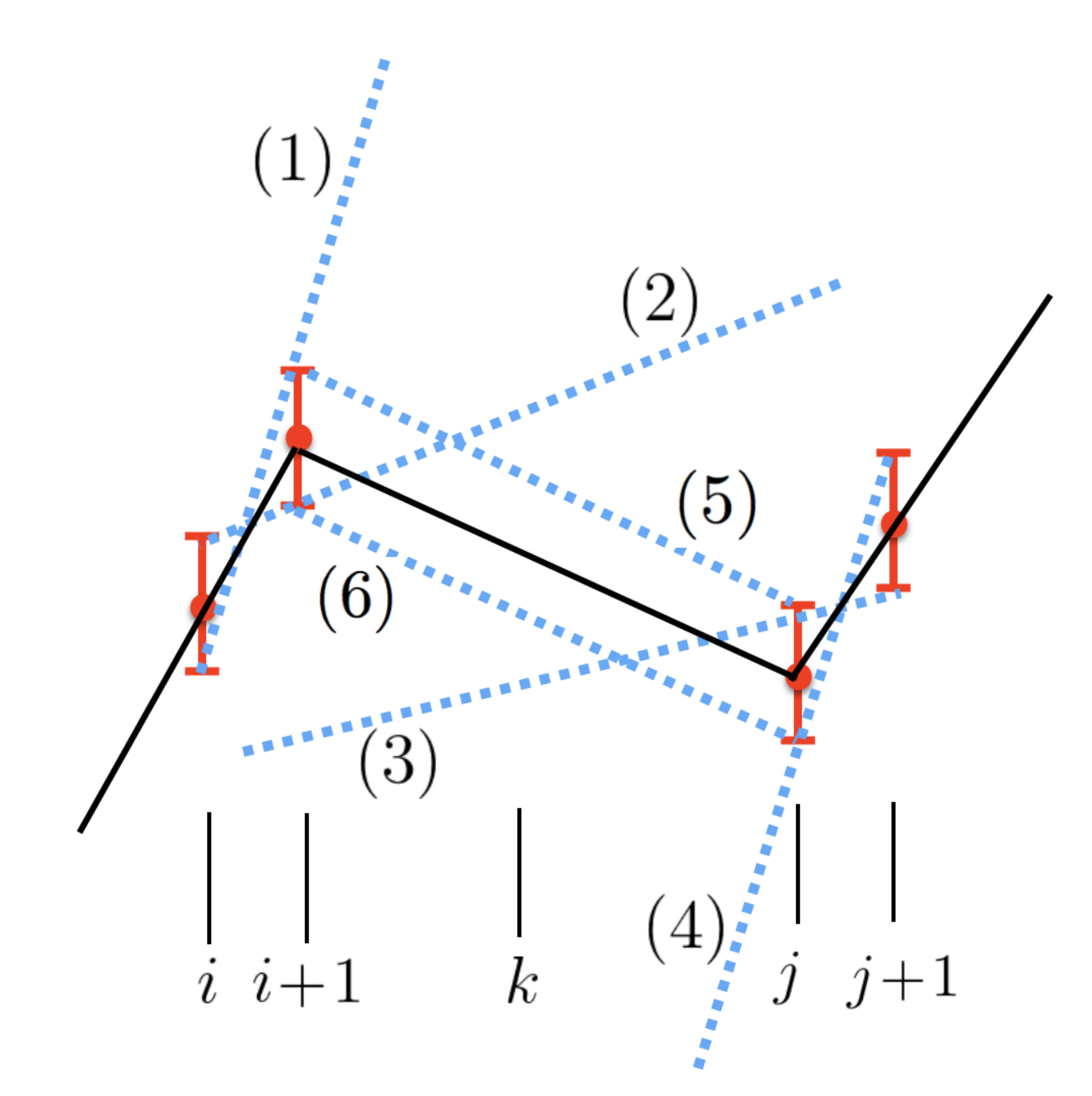} \\
\vspace{4mm}
(b)
\end{minipage}
\end{tabular}
\end{minipage}
\\
\begin{minipage}{\textwidth}
\centering
\includegraphics[width=0.9\textwidth]{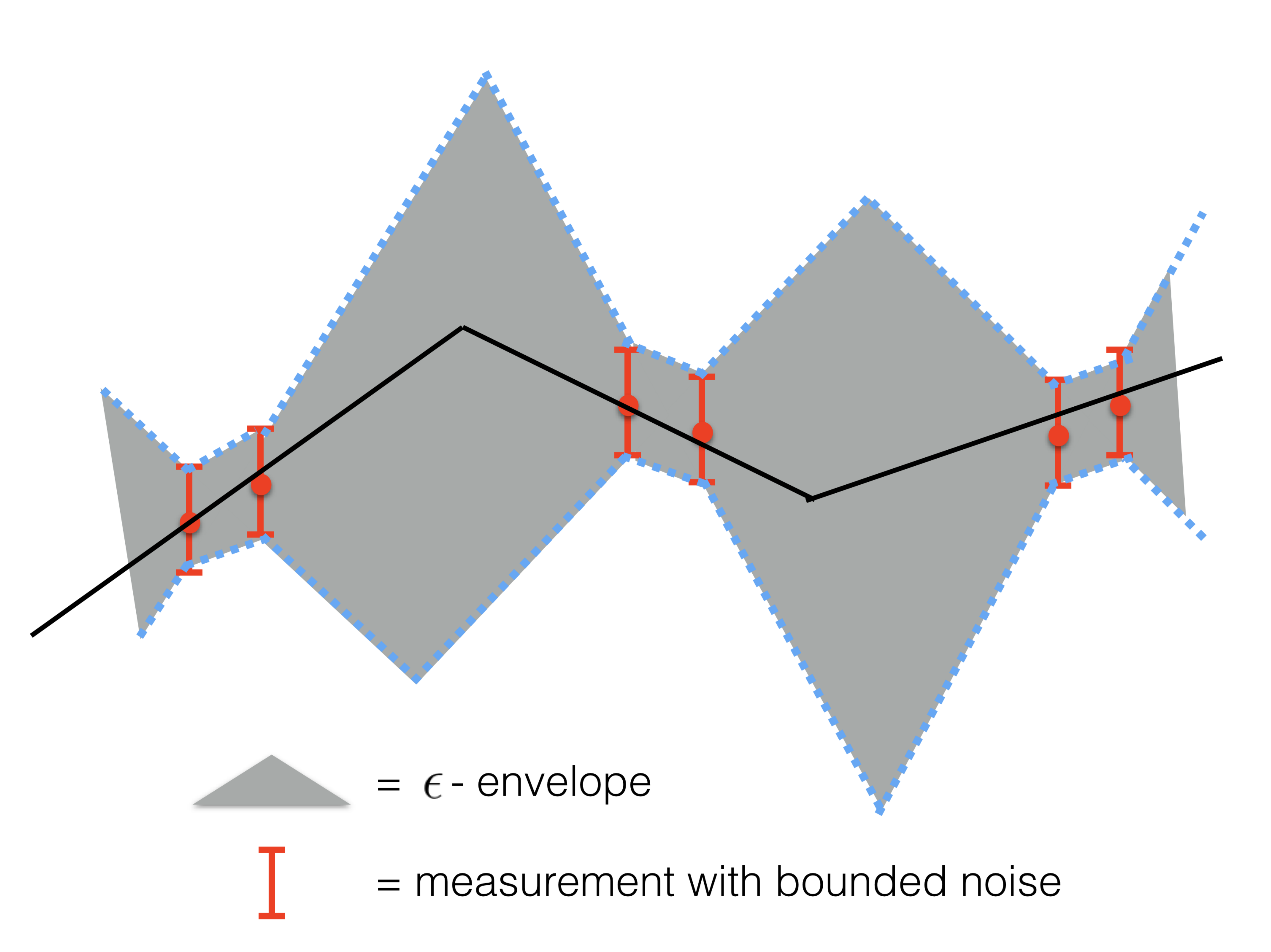} \\
\vspace{-1mm}
(c)
\end{minipage}
\end{tabular}
\caption{(a) Illustration of the line segments (1)-(6) in \prettyref{def:scEnvelop} with intersection. (b) Line segments (1)-(6) without intersection. (c) An example of \twd \emph{sign consistent \noisyEnvelop}.
% that contains all possible \twd sign consistent \signals, given noisy samples. 
% \LC{\noisyEnvelop in label}
% \FM{updated fig 8(b)}
}
\label{fig:envelopNoisy} 
\end{figure}

A pictorial representation of the line segments (1)-(6) in~\prettyref{def:scEnvelop} is given in~\prettyref{fig:envelopNoisy}(a)-(b). \prettyref{fig:envelopNoisy}(a) shows an example where line segment (1) intersects with (3) and line segment (2) intersects with (4). In \prettyref{fig:envelopNoisy}(b), these line segments do no intersect. % with each other. 
%Roughly speaking, every sign consistent \signal, between two samples, should be enclosed by a set of 4 half planes, as shown 
An example of the resulting \twd sign consistent \noisyEnvelop is illustrated in \prettyref{fig:envelopNoisy}(c).
 
Our interest towards the \twd sign consistent \noisyEnvelop is motivated by the following proposition.

\begin{proposition}[{\bf \twd Sign Consistent \noisyEnvelop}]
\label{prop:scEnvelop}
Under the conditions of \prettyref{def:scEnvelop}, any \twd sign-consistent \signal, %for all samples $k \in (i+1, j)$, 
belongs to the \twd \emph{sign consistent \noisyEnvelop}.
\end{proposition}

%An illustration of the inequalities (1)-(4) is shown in \prettyref{fig:envelopNoisy}(a).

The proof of~\prettyref{prop:scEnvelop} is given in \prettyref{proof:prop-noisyEnvelop}.

 Next we introduce a proposition that characterizes the depth reconstruction error bounds of an optimal solution.

\begin{proposition}[{\bf \twd Recovery Error - noisy samples}] \label{prop:1DrecoveryError_robust}
Let $\xtrue \in \Real{n}$ be the ground truth generating noisy measurements~\eqref{eq:measurements}.
Assume that we sample the boundary of $\xtrue$ and the sample set includes a twin 
sample in each linear segment in $\xtrue$. 
Then, $\xtrue$ belongs to the \twd sign consistent \noisyEnvelop, and any optimal solution $\xopt$ of~\eqref{eq:BPD} also lies in the  
\noisyEnvelop. Moreover, denoting with $\underbar{\xvar} \in \Real{n}$ and $\bar{\xvar} \in \Real{n}$, the point-wise lower and upper bound of the \noisyEnvelop (\prettyref{def:scEnvelop}), 
and considering any consecutive pairs of twin samples $i,i+1$ and $j,j+1$, 
for all $k \in (i+1, j)$, it holds: 
$$
|\xtrue_{k} - \xoptInd{k}| \leq \bar{\xvar}_k - \underbar{\xvar}_k. %\leq d_i \cos(\theta_i) + d_j \cos(\theta_j) + 2 \vare,
$$
%where a pictorial representation of the parameters $d_i, d_j, \theta_i, \theta_j$ is given in \prettyref{fig:envelopNoisy}(b), %and $d_i, d_j, \cos(\theta_i), \cos(\theta_j)$ are all monotonically increasing functions of $\vare$.
\end{proposition}

The proof of~\prettyref{prop:1DrecoveryError_robust} is given in \prettyref{proof:prop-1DrecoveryError_robust}.

%% file: analysis/BP_noisy_3D.tex
%!TEX root = ../main.tex

\somespace
\subsubsection{{\bf Analysis of \thd Reconstruction (noisy samples)}}  
\label{sec:BPD_optimality2D}
In this section we characterize the error bounds of an optimal solution $\Xopt$ of~\eqref{eq:BPDmatrix} in the noisy case. The result is similar to its noiseless counterpart in \prettyref{prop:2DrecoveryError}.

\begin{proposition}[{\bf \thd Recovery Error - noisy samples}] 
\label{prop:2DrecoveryError_robust}
Let $\Xtrue \in \R^{r \times c}$ be the ground truth generating noisy measurements~\eqref{eq:measurements}.
Let $\samples$ be a grid sample set and assume $\Xtrue$ to be \emph{\thd sign consistent} with respect to $\samples$. 
Moreover, 
let $\underbar{\Xvar} \in \R^{r \times c}$ and $\bar{\Xvar} \in \R^{r \times c}$ be the point-wise lower and upper bound of the 
row-wise \twd sign consistent \noisyEnvelop, built as in~\prettyref{fig:envelopNoisy}(b) by considering each row of the \thd depth \signal as a \twd \signal.
Then, given any optimal solution $\Xopt$ of~\eqref{eq:BPDmatrix}, it holds that
\beq
|\Xtrue_{i,j} - \Xvar^\star_{i,j}| \leq \max(|\underbar{\Xvar}_{i,j} - \Xopt_{i,j}| , |\bar{\Xvar}_{i,j} - \Xopt_{i,j}|)
\eeq
\end{proposition}

The proof for \prettyref{prop:2DrecoveryError_robust} follows the same line as the proof of \prettyref{prop:2DrecoveryError}, 
and we omit the proof for brevity.
%so we skip the details.

% \begin{proposition}[{\bf \thd Recovery Error - noiseless samples}] \label{prop:2DrecoveryError}
% Let $\Xtrue \in \R^{r \times c}$ be the ground truth \signal generating noiseless measurements~\eqref{eq:measurements}.
% Let $\samples$ be a grid sampling set and assume $\Xtrue$ to be \emph{\thd sign consistent} with respect to $\samples$. 
% Moreover, 
% let $\underbar{\Xvar} \in \R^{r \times c}$ and $\bar{\Xvar} \in \R^{r \times c}$ be the point-wise lower and upper bound of the 
% \emph{row-wise envelop}, built as in~\prettyref{fig:toyExample}(b) by considering each row of the \thd depth \signal as a \twd \signal.
% Then, $\Xtrue$ is an optimal solution of~\eqref{eq:BPmatrix},
% and any other optimal solution $\Xopt$ of~\eqref{eq:BPmatrix} satisfies:
% \beq
% \label{eq:recoveryError2D}
% |\Xtrue_{i,j} - \Xvar^\star_{i,j}| \leq \max(|\underbar{\Xvar}_{i,j} - \Xopt_{i,j}| , |\bar{\Xvar}_{i,j} - \Xopt_{i,j}|)
% \eeq
% \end{proposition}

%% file: algorithm/algorithms.tex
%!TEX root = ../main.tex

\section{Algorithms and Fast Solvers}
\label{sec:algorithms}

The formulations discussed so far, namely~\eqref{eq:BP}, \eqref{eq:BPD},~\eqref{eq:BPmatrix},~\eqref{eq:BPDmatrix},
directly translate into algorithms: each optimization problem can be solved using standard convex 
programming routines and returns an optimal depth \signal.

This section describes two algorithmic variants that further enhance the quality of the depth reconstruction (\prettyref{sec:algorithmicVariants}), and then presents a fast solver for the resulting $\ell_1$-minimization problems (\prettyref{sec:solvers}).

%% file: algorithm/algorithmicVariants.tex
%!TEX root = ../main.tex

\input{algorithm/algo1}

\somespace
\subsection{Enhanced Recovery in \twd and \thd} 
\label{sec:algorithmicVariants}

In this section we describe other algorithmic variants for the \twd and \thd case. 
\prettyref{sec:alg1} proposes a first algorithm that 
solves \twd problems and is inspired by~\prettyref{prop:1DrecoveryError}.
\prettyref{sec:alg2} discusses variants of~\eqref{eq:BPmatrix} for \thd problems.
%The second one, given in~\prettyref{sec:alg2}, solves \thd problems and is a variant of~\eqref{eq:BPmatrix}.  

%%%%%%%%%%%%%%%%%%%%%%%%%%%%%%%%%%%%%%%%%%%%%%%%%%%%%%%%%%%%%%%%%%%%%%%%
\subsubsection{{\bf Enhanced Recovery in \twd problems}} 
\label{sec:alg1}

\prettyref{prop:1DrecoveryError} dictates that any optimal solution of~\eqref{eq:BP} 
lies between the naive interpolation solution and the ground truth \signal $\xtrue$ (recall~\prettyref{fig:nam1}(b)).
\prettyref{alg:1} is based on a simple idea:
on the one hand, if the true \signal is concave between two consecutive samples (\cf with the 
first corner in~\prettyref{fig:nam1}(b)), then we should look for an optimal \signal having
 depth  ``as large as possible'' in that particular interval 
 (while still being within the optimal set of~\eqref{eq:BP}); on the other hand, if the shape is convex (second corner in~\prettyref{fig:nam1}(b)) 
we should look for an optimal \signal with depth as   ``as small as possible'', since this is the closest to $\xtrue$. 

\prettyref{alg:1} first solves problem~\eqref{eq:BP} and computes an optimal solution $\xopt$ and the corresponding optimal cost $f^\star$ (lines~\ref{line:create}-\ref{line:solve1}). 
Let us skip lines~\ref{line:signs1}-\ref{line:signs3} for the moment and take a look 
at line~\ref{line:solve2}: the constraints in this optimization problem include the same constraint 
of line~\ref{line:solve1} ($\matU \xvar = \meas$), plus an additional constraint in line~\ref{line:solve1} ($\lone{\tv \xvar} \leq f^\star$) that restricts $\xvar$ to stay within the optimal solution set of~\eqref{eq:BP}.
Therefore, it only remains to design a new objective function that ``encourages'' a solution that is close to $\xtrue$ while still being within this optimal set. To this end, we use a simple linear objective $s\tran \xvar$, where $s \in \{0, \pm 1\}^n$ is a vector of coefficients, such that the objective function penalizes large entries in the \signal $\xvar$ if $s_k = +1$, and rewards large entries when  
$s_k = -1$. 
More specifically, the procedure for choosing a proper coefficient $s_k$ is as follows. For any consecutive pairs of twin samples $(i-1,i)$ and ($j,j+1$), the algorithm looks at the slope difference between the second pair (i.e., $\xvar^\star_{j+1} -\xvar^\star_{j}$) 
and the first pair ($\xvar^\star_{i} -\xvar^\star_{i-1}$). If this difference is negative, 
then the function $\xopt$ is expected to be concave between the samples. 
In this case the sign $s_k$ for any point $k$ between the 
samples is set to $-1$. If the difference is positive, then the signs are set to $+1$. Otherwise the signs will be 0.
   We prove the following result.

\begin{corollary}[{\bf Exact Recovery of 2D \signals by \prettyref{alg:1}}]
\label{cor:algorithm}
Under the assumptions of~\prettyref{prop:1DrecoveryError}, \prettyref{alg:1} recovers the 
\twd depth \signal $\xtrue$ exactly. 
\end{corollary}
The proof is in \prettyref{proof:cor-algorithm}. Although \prettyref{alg:1} is designed for noiseless samples, in the experiments in \prettyref{sec:exp-2d} we also test a noisy variant by substituting the constraints in lines~\ref{line:solve1} and \ref{line:solve2}
with $\linf{\matU \xvar - \meas} \leq \vareps$.

%\somespace
%%%%%%%%%%%%%%%%%%%%%%%%%%%%%%%%%%%%%%%%%%%%%%%%%%%%%%%%%%%%%%%%%%%%%%%%
\subsubsection{{\bf Enhanced Recovery in \thd problems}} 
\label{sec:alg2}

In the formulations~\eqref{eq:BPmatrix} and~\eqref{eq:BPDmatrix} we 
used the matrix $\TV$ to encourage ``flatness'', or in other words, regularity of the 
depth \signals. In this section we discuss alternative objective functions 
which we evaluate experimentally in~\prettyref{sec:experiments}. 
%several variants of objective functions. 
These objectives simply adopt different definitions for the matrix $\TV$ in~\eqref{eq:BPmatrix} and~\eqref{eq:BPDmatrix}. 
For clarify, we denote the formulation introduced earlier in this paper (using the matrix $\TV$ defined in~\eqref{eq:TVpartition}) as the ``\lmin'' formulation (also recalled below), and 
we introduce two new formulations, denoted as ``\lminDiag'' and ``\lminCart'', which use different objectives. 

{\it \lmin formulation}: 
Although we already discussed the structure of the matrix $\TV$ in~\prettyref{sec:rec2d3d_unified}, % and in eq.~\eqref{eq:TVpartition},
 here we adopt a slightly different perspective that will make the presentation of the 
variants \lminDiag and \lminCart clearer.
In particular, rather than taking a matrix view as done in~\prettyref{sec:rec2d3d_unified}, we 
interpret the action of the matrices $\tv_V$ and $\tv_H$ in eq.~\eqref{eq:l1matrix} 
as the application of a kernel (or convolution filter) 
to the \thd depth \signal $\Xvar$. In particular, we note that:
\beq
\tv_V \Xvar = \Xvar * K_{xx} \qquad
(\Xvar \tv_H\tran)\tran = \Xvar * K_{yy}
\eeq
where ``$*$'' denotes the action of a discrete convolution filter and
the  kernels $K_{xx}$ and $K_{yy}$ are defined as
\begin{equation*}
K_{xx} = 
\begin{bmatrix}
    0       & 0 & 0  \\
    1       & -2 & 1  \\
    0       & 0 & 0 
\end{bmatrix}
, \quad
K_{yy} =
\begin{bmatrix}
    0       & 1 & 0  \\
    0       & -2 & 0  \\
    0       & 1 & 0 
\end{bmatrix}.
\end{equation*} 
Intuitively, $K_{xx}$ and $K_{yy}$ applied at a pixel return the \second-order 
differences along the horizontal and vertical directions at that pixel, respectively. 
The \lmin objective, presented in~\prettyref{sec:rec2d3d_unified}, can be then written as:
\begin{equation*}
f_{\text{\lmin}}(\Xvar)
%\lone{\TV \xvar}^{\text{\lmin}} 
\doteq \lone{\vec{\Xvar * K_{xx}}} + \lone{\vec{\Xvar * K_{yy}}}.
\end{equation*}
%where $z$ is the depth image to reconstruct, ``$*$'' is the convolution operation in image processing, and 

{\it \lminDiag formulation}: %The second objective function be denoted as \lminDiag. Compared with \lmin, 
While \lmin only penalizes, for each pixel, variations along the horizontal and vertical direction, 
the objective of the \lminDiag formulation includes an additional \second-order derivative, which penalizes changes along the diagonal direction. This additional term can be written as $\lone{\vec{\Xvar * K_{xy}}}$, where the kernel $K_{xy}$ is: % as follows:
\begin{equation*}
\begin{split}
K_{xy} &= \frac 1 4 * 
\begin{bmatrix}
    -1       & 0 & 1  \\
    0       & 0 & 0  \\
    1       & 0 & -1 
\end{bmatrix}.
\end{split}
\end{equation*}
 Therefore, the objective in the \lminDiag formulation is
$$
f_{\text{\lminDiag}}(\Xvar) \doteq f_{\text{\lmin}}(\Xvar) + \lone{\vec{\Xvar * K_{xy}}}.
$$
%The kernel $K_{xy}$ for the diagonal term is as follows.
% Note that with both \lmin and \lminDiag, the $\TV$ matrix corresponds to a $2^{nd}$ order difference operator on the image plane (i.e., the pixel space), but with different kernels. 

{\it \lminCart formulation}: 
When introducing the \lmin formulation in~\prettyref{sec:cs}, we assumed that we reconstruct the depth at 
uniformly-spaced point, i.e., the $(x,v)$\footnote{With slight abuse of notation here we  use $x$ and $v$ 
to denote the horizontal and vertical coordinates of a point with respect to the image plane.} 
coordinates of each point belong to a uniform grid; 
in other words, looking at the notion of curvature in~\eqref{eq:slopeDiff}, we assumed 
$x_{i+1}-x_{i} = x_{i}-x_{i-1} = 1$ (also $v_{i+1}-v_{i} = v_{i}-v_{i-1} = 1$ in the \thd case).
 While this comes without loss of generality, since the full profile is unknown and we can reconstruct it at arbitrary resolution, 
 we note that typical sensors, even in \twd, do not produce measurements with uniform spacing, see~\prettyref{fig:metricExample}.

\input{figureTex/figureMetricExample}

For this reason, in this section we generalize the \lmin objective to account for irregularly-spaced points.
If we denote with $x_{i,j}$ and $v_{i,j}$ the horizontal and vertical coordinates of the \thd point observed at pixel $(i,j)$,
a general expression for the horizontal and vertical \second-order differences is:
\beq
\Xvar * K_{xx}^{\text{\lmin-cart}} \qquad\qquad \Xvar * K_{yy}^{\text{\lmin-cart}}
\eeq
where the convolution kernels at pixel $(i,j)$ are defined as:
\beal
\label{eq:l1cart_kernel}
K_{xx}^{\text{cart}}(i,j) = \\
\begin{bmatrix}
    0       & 0 & 0  \\
    \frac 1 {x_{i,j} - x_{i-1,j}}       & -(\frac 1 {x_{i,j} - x_{i-1,j}} + \frac 1 {x_{i+1,j} - x_{i,j}}) &  \frac 1 {x_{i+1,j} - x_{i, j}} \\
    0       & 0 & 0 
\end{bmatrix}, \\
K_{yy}^{\text{cart}}(i,j) = 
\begin{bmatrix}
    0       & \frac 1 { v_{i,j} - v_{i, j-1}} & 0  \\
    0       & \frac 1 {v_{i,j} - v_{i, j-1} } - \frac 1 {v_{i, j+1} - v_{i, j}} & 0  \\
    0       & \frac 1 {v_{i, j+1} - v_{i, j}} & 0 
\end{bmatrix}
\eeal
% \begin{equation}
% \label{eq:l1cart_kernel}
% \begin{split}
% K_{xx}^{\text{cart}}(i,j) &= 
% \begin{bmatrix}
%     0       & \frac 1 {x_{i,j} - x_{i-1,j} + c} & 0  \\
%     0       & -(\frac 1 {x_{i,j} - x_{i-1,j} + c} + \frac 1 {x_{i+1,j} - x_{i,j} + c}) & 0 \\
%     0       & \frac 1 {x_{i+1,j} + c} & 0 
% \end{bmatrix}^T
% , \\
% K_{yy}^{\text{cart}}(i,j) &=
% \begin{bmatrix}
%     0       & \frac 1 { v_{i,j} - v_{i, j-1} + c } & 0  \\
%     0       & \frac 1 {v_{i,j} - v_{i, j-1} + c} - \frac 1 {v_{i, j+1} - v_{i, j} + c} & 0  \\
%     0       & \frac 1 {v_{i, j+1} - v_{i, j} + c} & 0 
% \end{bmatrix}
% \end{split}
% \end{equation}
 The kernels $K_{xx}^{\text{cart}}(i,j)$ and $K_{yy}^{\text{cart}}(i,j)$ simplify to $K_{xx}$ and $K_{yy}$ when the points are uniformly spaced, and can be used to define a new objective function:
% Moreover, 
% The ``Cartesian'' kernels $K_{xx}^{\text{cart}}(i,j)$ and $K_{yy}^{\text{cart}}(i,j)$  
% induce the objective function:
% The objective of the \lminCart formulation considers \second-order differences defined in the Cartesian space, 
% rather than in the image space: 
%In order to reconstruct the depth 
%The objective function is defined as follows
$$
f_{\text{\lminCart}}(\Xvar)
 \doteq \lone{\vec{\Xvar * K_{xx}^{\text{\lmin-cart}}}} + \lone{\vec{\Xvar * K_{yy}^{\text{\lmin-cart}}}}.
$$
% Where
% More, specifically, if we want to reconstruct a point at coordinates $x_{i,j}$ and $v_{i,j}$

% This \lminCart objective function is motivated by fact that the coordinates in the $x,v$ dimensions have non-uniform spacing, as illustrated in~\prettyref{fig:metricExample}. This phenomenon is due to the perspective projection of sensors such as a LiDAR and a camera. 
\lminCart may be used to query the depth at arbitrary points and in this sense it is more general than 
\lmin. On the downside, we notice that extra care should be taken to ensure that the denominators in the entries of the 
kernels $K_{xx}^{\text{cart}}(i,j)$ and $K_{yy}^{\text{cart}}(i,j)$ do not vanish, and small denominators (close to zero) may 
introduce numerical errors. For this reason, in our tests, we add a small positive constant $\param$ to all denominators. 
% make the corresponding optimization . 
% is expected to produce better reconstruction results, since it captures the actual geometry of the environment. In comparison, both \lmin and \lminDiag suffer from distortion due to perspective projection, and thus are can be viewed as simplified approximate of \lminCart. However, 
% additional technical issues arise with using \lminCart: (1) \lminCart requires a prior knowledge of all the $(x, v)$ coordinates. These coordinates might come from, for example, a predetermined grid of query points where the depth of each point is of interest. (2) In order to avoid the ill-conditioned case with division by zero (e.g. when $x_{i,j} - x_{i-1,j} = 0$), we have to add a small positive constant c to all denominators. 

%In \prettyref{sec:exp-objective-functions} we will have an empirical evaluation of these different objective functions.

%% file: algorithm/algo1.tex
%!TEX root = ../main.tex

\begin{algorithm}[t]
\label{alg:two-stage}
\footnotesize
\SetKwInOut{Input}{input}\SetKwInOut{Output}{output}
 \Input{Measurements $\meas$, and sample set $\samples$, including boundary and twin samples }
 \Output{ Original \signal $\xtrue$ }
 \BlankLine
 \tcc{solve $\ell_1$-minimization}
 create matrices $\matU$ (\prettyref{def:samplingMatrix}) and $\tv$~(eq. \eqref{eq:tv}) \label{line:create} \; 
 solve $(f^\star,\xopt) = \min_\xvar \; \lone{\tv \xvar} \quad \subject \matU \xvar= \meas$ \label{line:solve1}  \; 
  \BlankLine
 \tcc{populate  a vector of signs $s \in \{-1,0,+1\}^n$}  
 \For{consecutive twin samples $(i-1,i)$, ($j,j+1$)}{ \label{line:signs1}
 	\ForEach{$k \in \{i+1,\ldots,j-1\}$}{ \label{line:signs2}
 	 set $s_k = \sign( (\xvar^\star_{j+1} -\xvar^\star_{j}) - (\xvar^\star_{i} -\xvar^\star_{i-1}) )$ \label{line:signs3}
 	 }
 } 
  \BlankLine
 \tcc{recover $\xtrue$ within the solution set}
 $\xtrue = \argmin_\xvar \; s\tran \xvar \quad \subject  \matU \xvar= \meas \text{ and }  \lone{\tv \xvar} \leq f^\star $ \label{line:solve2} \;
  return $\xtrue$.
 \caption{ Exact recovery of \twd depth \signals. \label{alg:1}}
\end{algorithm}

%% file: figureTex/figureMetricExample.tex
\begin{figure}[htbp]

\includegraphics[width=0.8\linewidth]{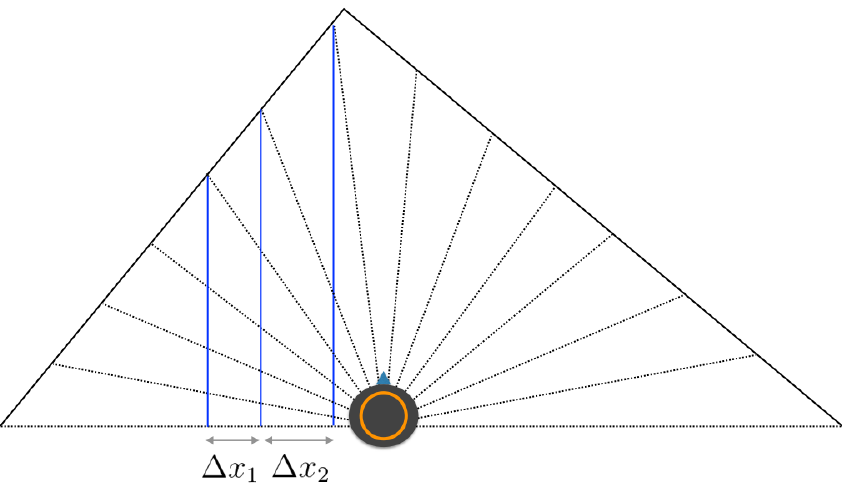}

\caption{
A toy example illustrating that while a \twd  lidar produces measurements at 
fixed angular resolution, the resulting Cartesian coordinates are not equally spaced, i.e., $\Delta x_1 \not= \Delta x_2$.
% due to perspectie projection. This occurs when using LiDAR and cameras, and thus motivating the \lminCart objective function.
This occurs in both lidars and perspective cameras, hence motivating the introduction of the \lminCart formulation.
}
\label{fig:metricExample} 
\end{figure}

%% file: algorithm/solvers.tex
%!TEX root = ../main.tex

\subsection{Fast Solvers}
\label{sec:solvers}

All the formulations presented in this paper, including the algorithmic variants proposed in \prettyref{sec:algorithmicVariants}, rely on solving the optimization problems 
\eqref{eq:BP},~\eqref{eq:BPmatrix},~\eqref{eq:BPDmatrix}, and \eqref{eq:BPD} efficiently. Despite the convexity of these problems, off-the-shelf solvers based on interior point methods tend to be slow and do not scale to very large problems. Recalling that in the \thd case, the number of unknown variables in our problems is equal to the number of \unsampled pixels in the depth map, these optimization problems   can easily involve more than $10^5$ variables.
Indeed, in the experiments in \prettyref{sec:exp-solvers} we show that off-the-self solvers such as \cvx/\MOSEK \citep{CVXwebsite,gb08} are quite slow and practically unusable for \thd \signals larger than $100\times100$ pixels.

For these reasons, in this section we discuss a more efficient first-order method to solve these minimization problems. This solver is a variant of \NESTA, an algorithm for fast $\ell_1$ minimization recently developed by \cite{BeckerBC11} and based on Nesterov's method for nonsmooth optimization \citep{Nesterov05,Nesterov83}.
We tailor \NESTA to our specific optimization problems with $\ell_\infty$-norm constraints, instead of the original $\ell_2$ norm used in~\citep{BeckerBC11}. In this section we focus on the \twd problem~\eqref{eq:BPD}, 
since the algorithm is identical in the \thd case (with the only exception that the matrix $\TV$ is used in place of $\tv$).  

\input{algorithm/nesta}

In this section, we provide an overview of  \NESTA, adapted to problem \eqref{eq:BPD}, while we leave technical details to \prettyref{app:app_nesta}. %In its most generality, \NESTA aims to solve 
\NESTA solves convex optimization problems with nonsmooth objectives, in the general form:
\begin{align}\label{eq:nonsmooth_min}
\min_\xvar \; f(\xvar)  \quad \subject \xvar \in \calQ
\end{align}
where $f(\xvar)$ is a nonsmooth convex function and $\calQ$ is a convex set.
The basic idea in \NESTA is to replace the original objective $f(\xvar)$ with a smooth approximation
$f_\mu(\xvar)$
\begin{align}\label{eq:smooth_min}
\min_\xvar \; f_\mu(\xvar)  \quad \subject \xvar \in \calQ
\end{align}
where $\mu$ is a parameter controlling the smoothness of $f_\mu(\xvar)$ and such that when $\mu$ goes to zero, 
 $f_\mu(\xvar)$ approaches $f(\xvar)$.
%where $f_\mu(\xvar)$ is a smooth approximation of a convex but nonsmooth function $f(\xvar)$ and $\calQ$ is a convex set. 

In our problem \eqref{eq:BPD}, we have $f(\xvar) = \lone{ \tv \xvar }$ and $\calQ = \{ \xvar: \linf{\matU \xvar - \meas} \leq \vareps \}$. 
Following~\citep{Nesterov05}, we first notice that our nonsmooth objective can be written as:
%An equivalent formulation for $f(\xvar)$ is given in terms of the dual $\ell_\infty$-ball
\beq
f(\xvar) \doteq \lone{ \tv \xvar } = \max_{u: \linf{u} \leq 1 } \la u, \tv \xvar \ra. 
\eeq
Then a convenient choice for $f_\mu(\xvar)$ is
\begin{align}\label{eq:fmu}
f_\mu(\xvar) = \max_{u: \linf{u} \leq 1 } \la u, \tv \xvar \ra - \mu \frac{\|u\|_2^2}{2}.
\end{align}
The function $f_\mu(\xvar)$ is differentiable, see~\citep{Nesterov05}, and its gradient is Lipschitz with constant $L_\mu$ 
(\prettyref{app:app_nesta} provides an explicit expression for the constant $L_\mu$). 
% The smooth approximation $f_\mu(\xvar)$ of the objective function $f(\xvar)$ is such that when $\mu$ goes to zero, 
%  $f_\mu(\xvar)$ approaches $f(\xvar)$, 
It can be readily noticed from eq.~\eqref{eq:fmu} that when $\mu$ goes to zero, 
 $f_\mu(\xvar)$ approaches our objective $f(\xvar)$.

\NESTA adopts a \emph{continuation} approach, in that it solves a sequence of optimization problems with decreasing values of 
 $\mu$, such that the result of the last optimization problem approximates closely the solution of $f(\xvar)$. 
 %This technique is usually referred to as \emph{continuation}
 The advantage in doing so is that,  instead of minimizing directly $f(\xvar)$ with nonsmooth optimization techniques which are generally slow, at each iteration \NESTA applies Nesterov's accelerated gradient method to the smooth function $f_\mu(\xvar)$, 
ensuring an optimal convergence rate of  $\calO(1/K^2)$ in the number of gradient iterations $K$. 

The pseudo-code of \NESTA, tailored to~\eqref{eq:BPD}, is given in \prettyref{alg:nesta}.
The outer iterations in line~\ref{line:accelerate} iterate for decreasing values of $\mu$, 
starting at an initial value $\mu_0$ (computed in line~\ref{line:mu0}) till a user-specified final 
value $\mu_f$. The user also specifies the numbers of outer iterations $T$, such that at each iteration 
the value of $\mu$ is decreased by an amount $\gamma < 1$, computed in line~\ref{line:gamma};
the value of $\mu$ is decreased after each outer iteration, as shown in line~\ref{line:update_mu}.
The choice of $\mu_f$ implies a trade-off between the speed of convergence 
(the convergence rate of solving~\eqref{eq:smooth_min} is proportional to the $\mu$ used in each iteration) and the accuracy of the smoothed approximation $f_\mu$, which consequently determines the \NESTA's overall accuracy. According to experiments in \citep{BeckerBC11}, decreasing $\mu_f$ by a factor of $10$ gives about $1$ additional digit of accuracy on the optimal value.

\NESTA  uses a warm start mechanism, such that the solution $\xvar\at{K}$ for a given $\mu$ is used as initial 
guess at the next iteration, as shown in line~\ref{line:warm}. Choosing a good initial guess for the first iteration 
(input $\xvar\at{0}$ in \prettyref{alg:nesta})
may also contribute to speed-up the solver. In our tests we used the naive solution (linear interpolation) as initial guess 
for \NESTA.
% $\xvar\at{0} = \matU\tran \meas$, as suggested in 
%As suggested in we 
% Solving \eqref{eq:smooth_min} is faster with larger $\mu$ and using intermediate solutions as a warm start for the next problem increases the overall speed significantly. 
% As initial values for the algorithm,  we use  $\xvar\at{0} = \matU\tran \meas$ and $\mu_0 = \linf{\tv\tran \meas}$.
%\label{line:accelerate}

For a given value of $\mu$, lines~\ref{line:nesterov1}-\ref{line:stopping} describe Nesterov's accelerated gradient 
method applied to the smooth problem with objective $f_\mu$. 
The accelerated gradient method involves $K$ inner iterations (line~\ref{line:nesterov1}) and terminates if the change in the depth estimate 
is small (stopping condition in lines~\ref{line:stopping0}-\ref{line:stopping}).
Nesterov's method updates the depth estimate $\xvar\at{k+1}$ (line~\ref{line:nesterov3}) 
using a linear combination of intermediate variables 
$\xnesta$ (line~\ref{line:xk})
and $\wnesta$ (line~\ref{line:yk}). We refer the reader to~\citep{Nesterov05} for more details.
We provide closed-form expressions for the gradient $\nabla f_\mu(\xvar)$ and for the vectors $\xnesta$ and $\wnesta$ (lines~\ref{line:xk}-\ref{line:yk}) in \prettyref{app:app_nesta}. 

% \NESTA also involves a continuation step of updating the smoothing parameter $\mu$ in order to substantially accelerate the algorithm, see lines~\ref{line:accelerate}-\ref{line:update_mu}. This step solves a succession of problems \eqref{eq:smooth_min} with decreasing smoothing parameters from an initial value $\mu_0$ to the desired value $\mu_f$. The number $T$ of continuation steps is of choice and it determines the step size $\gamma < 1$ in line~\ref{line:gamma}.  %We use the naive interpolation solution $\xnaive$ as an initial point $\xvar\at{0}$ for the algorithm.% 
%The initial value $\mu_0 = \linf{\tv\tran \meas}$.

Note that when $\vareps = 0$, \prettyref{alg:nesta} solves the noiseless problem \eqref{eq:BP}. 
This only affects the closed-form solutions for $\xnesta$ and $\wnesta$, but does not alter the overall structure of the algorithm. Similarly, \prettyref{alg:nesta} can be used to solve problems \eqref{eq:BPmatrix} and \eqref{eq:BPDmatrix}, after replacing the matrix $\tv$ with $\TV$ in the definition of $f(\xvar)$. As discussed earlier, the choice of a nonzero $\mu_f$ in \NESTA will result in an approximate solution to the optimal solution of \eqref{eq:BPD}. Consequently, \NESTA may produce slightly less accurate solutions, while being much faster than \cvx. 
Our experimental results show that the accuracy loss is negligible if the parameter $\mu_f$ is chosen appropriately, see
%A detailed comparison of the performances of two solvers is provided in our simulations 
\prettyref{sec:exp-solvers}.

%% file: algorithm/nesta.tex
%!TEX root = ../main.tex

\begin{algorithm}[t]
\footnotesize
\SetKwInOut{Input}{input}\SetKwInOut{Output}{output}
 \Input{Measurements $\meas$, sampling matrix $\matU$, noise level $\vareps$, initial guess $\xvar\at{0}$, desired final smoothing parameter value $\mu_f$, maximum Nesterov's iterations $K$, continuation steps $T$, stopping criterion $\tau$}
 \Output{ Approximate solution $\xvar\at{K} $ for \eqref{eq:BPD} }
 \BlankLine
 \tcc{initialize parameters}
 initialize $\mu_0 = \linf{\tv\tran \meas} $ \label{line:mu0} \;
 compute $\gamma = (\mu_f - \mu_0)^{1/T}$ \;   \label{line:gamma}
 \BlankLine
 \tcc{outer iterations with decreasing $\mu$}
\For{ $t = 1: T$}{  \label{line:accelerate}
set $\mu = \mu_{t-1}$ \;
\BlankLine
 \tcc{Nesterov's accelerated gradient} 
 	\For{$k = 0: K-1$}{  \label{line:nesterov1}
 	compute $\nabla f_\mu(\xvar\at{k})$ \; \label{line:nesterov2}
 	set $\alpha_k = \frac{k+1}{2}$ and $ \tau_k = \frac{2}{k+3} $ \;
 	\BlankLine
 	 \tcc{solve:} 
 	$\!\!\xnesta = \argmin_\xvar \; \frac{L_\mu}{2} \|\xvar - \xvar\at{k}\|_2^2 + \la \nabla f_\mu(\xvar\at{k}), \xvar - \xvar\at{k} \ra$ \label{line:xk} \vspace{0.2cm}
 	
 	$\hspace{0.75cm}\subject \linf{\matU \xvar - \meas} \leq \vareps$ \;
 	\BlankLine
 	% \tcc{solve:}
    $\!\!\wnesta \!\!= \!\argmin_\xvar \; \frac{L_\mu}{2} \|\xvar \!-\! \xvar\at{0}\|_2^2 + \! \displaystyle \sum_{i = 0}^k \alpha_i \la \nabla f_\mu(\xvar\at{i}), \xvar \!-\! \xvar\at{k} \ra \!\!$ \label{line:yk}
 	$ \hspace{0.62cm} \subject \linf{\matU \xvar - \meas} \leq \vareps $ \;
 	\BlankLine
 	 \tcc{update $\xvar$}
 	 $\xvar\at{k+1} = \tau_k \wnesta + (1 - \tau_k) \xnesta$ \; \label{line:nesterov3}
 	 \BlankLine
 	 \tcc{stopping criterion}
 	 \If {$\linf{ \xvar\at{k+1} - \xvar\at{k} } < \tau$ \label{line:stopping0}}{
 	 	$\xvar\at{K} = \xvar\at{k+1}$; \quad break loop \label{line:stopping}
 	 	}  
 	 }
 	 \tcc{decrease the value of $\mu$}
 	 set $\mu_t = \gamma \mu_{t-1}$ \; \label{line:update_mu}
 	 set $\xvar\at{0} = \xvar\at{K}$ \; \label{line:warm}
 } 
  \BlankLine
  return $\xvar\at{K}$.
\caption{ \NESTA for solving \eqref{eq:BPD} } 
\label{alg:nesta}
\end{algorithm}

%% file: experimentTex/experiments.tex
%!TEX root = ../main.tex

\section{Experiments}
\label{sec:experiments}
This section validates our theoretical derivations with experiments on synthetic, simulated, and real data.
Empirical evidence shows that our recovery techniques perform very well in practice, in both 2D and 3D environments. Our algorithm is also  more robust to noise than a naive linear interpolation, and outperforms previous work in both reconstruction accuracy and computational speed. We discuss a number of applications, including \twd mapping (\prettyref{sec:exp-2d}), \thd depth reconstruction from sparse measurements (\prettyref{sec:exp-single-frame-sparse-reconstruction}-\ref{sec:exp-multi-frame-sparse-reconstruction}), 
data compression applied to bandwidth-limited robot-server communication (\prettyref{sec:exp_compression}), 
and super-resolution depth imaging (\prettyref{sec:exp_superResolution}).
For the \thd case, we also provide a Monte Carlo analysis comparing the different solvers and choices of the objective functions 
(\prettyref{sec:exp_3Drec}).

In the following tests, we evaluate the accuracy of the reconstruction by the average pixel-wise depth error, i.e., $\frac{1}{n}\lone{\xopt - \xtrue}$, where $\xtrue$ is the ground truth and $\xopt$ is the reconstruction, unless otherwise specified.

%% file: experimentTex/experiments_2d.tex
%!TEX root = ../main.tex

\subsection{\twd Sparse Reconstruction and Mapping}
\label{sec:exp-2d}
In this section, %as a sanity check of our algorithm in the simplest settings, 
we apply our algorithm to reconstruct \twd depth \signals (e.g., the data returned by a \twd laser scanner). 
We provide both a statistical analysis on randomly generated synthetic \signals (Sections~\ref{sec:exp_2drec}-\ref{sec:exp_2drec_stats}), and 
a realistic example of application to \twd mapping (\prettyref{sec:exp_2drec_mapping}).

%%%%%%%%%%%%%%%%%%%%%%%%%%%%%%%%%%%%%%%%%%%%%%%%%%%%%%%%%%%%%%%%%%%%%%%%%%%%%%%%%%%%
\subsubsection{Typical Examples of \twd Reconstruction}
\label{sec:exp_2drec}

\begin{figure}[htbp]
\centering
\includegraphics[width=0.6\linewidth]{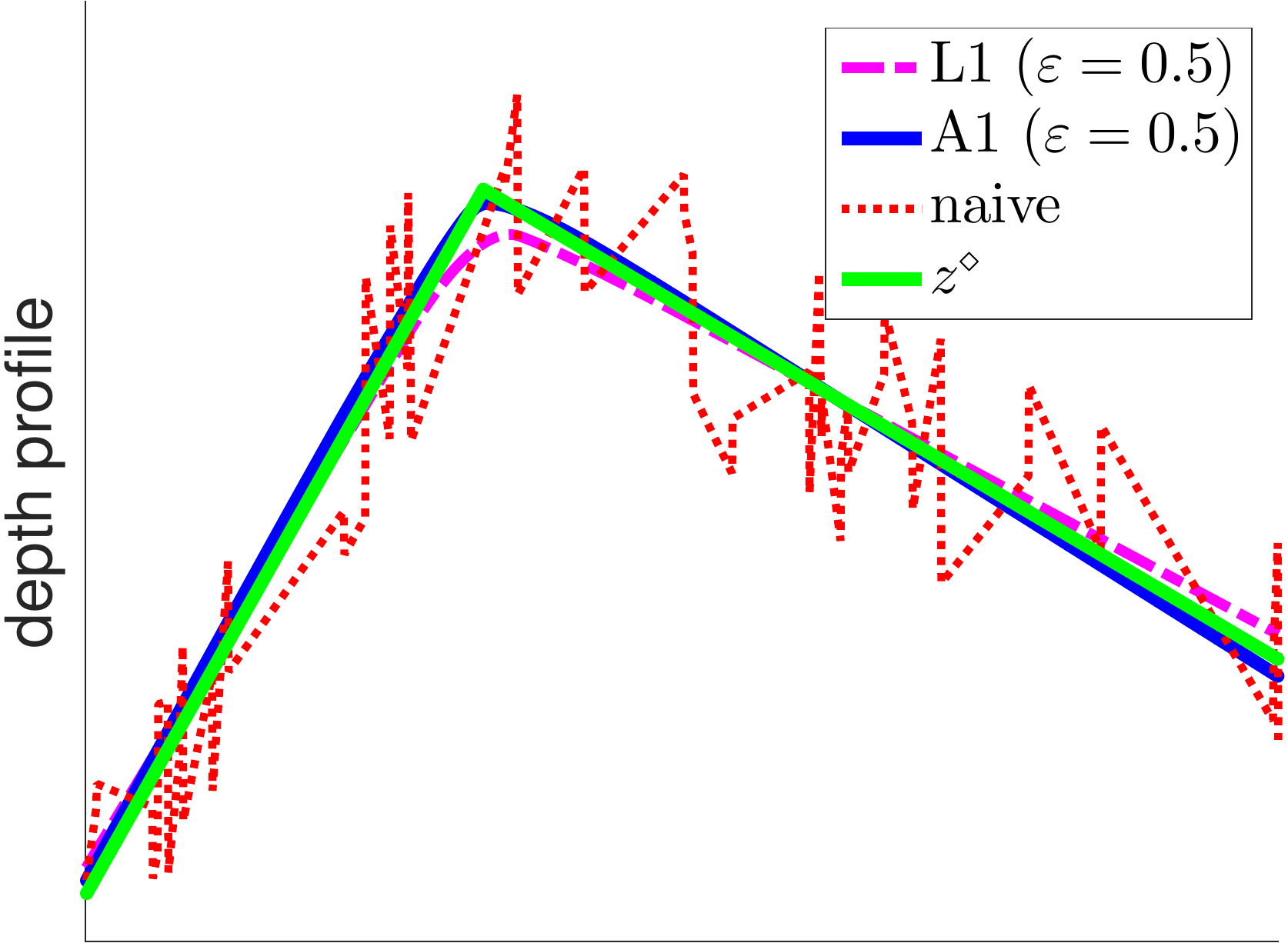}
\caption{
An example of synthetic \twd \signal and typical behavior of the compared techniques, 
\naive, \lmin, and \algOne, for noise level $\vareps=0.05\text{m}$.
}
\label{fig:example1D} 
\end{figure}

We create a synthetic dataset that contains random piecewise 
linear depth \signals of size $n\!=\! 2000$, with given number of corners. 
Since the number of variables is small, we use  \cvx/\MOSEK \citep{CVXwebsite,gb08} as solver in all \twd experiments.
When possible, we compare three different reconstruction algorithms: 
(i)~the linear interpolation produced by Matlab's command {\tt intep1}, denoted as \naive, 
(ii)~the estimate from~\eqref{eq:BP} (noiseless case) or~\eqref{eq:BPD} (noisy case), denoted as \lmin,
%the \lmin algorithm corresponding to ,
and 
(iii)~the estimate produced by~\prettyref{alg:1}, denoted as \algOne. 
%More specifically, the results produced by~\eqref{eq:BP} and~\eqref{eq:BPmatrix} are denoted with \lmin. The estimate produced by the noisy problems 
 %and~\eqref{eq:BPDmatrix} are denoted with $\lmin (\vareps=\cdot)$, where we specify the noise level in the parenthesis. 
 
An example of synthetic \twd \signal (with only one corner) is shown in~\prettyref{fig:example1D}. The green line is the ground truth \signal, while the others are reconstructed depth profiles from sparse and noisy measurements using the three different algorithms. 

\prettyref{fig:example1D} provides a typical example of \twd reconstruction results.
\naive  linearly interpolates 
the samples, hence even when measuring all depth data, it still produces a jagged line, due to measurement noise. 
It is easy to show that when measurement noise is uniformly distributed in $[-\vareps,+\vareps]$ 
(as in our tests), the average error committed by \naive converges to $\vareps/2$ for increasing number of samples.
In the figure, we consider $\vareps = 0.05\text{m}$.
On the other hand, \lmin and \algOne correctly smooth the noise out. 
In particular, while \lmin returns a (sign consistent) solution that typically has rounded corners,
\algOne is able to rectify these errors, producing an estimate that, even in the noisy case, is very close to the truth.

%%%%%%%%%%%%%%%%%%%%%%%%%%%%%%%%%%%%%%%%%%%%%%%%%%%%%%%%%%%%%%%%%%%%%%%%%%%%%%%%%%%%
\subsubsection{Statistics for \twd Reconstruction}
\label{sec:exp_2drec_stats}

\input{figureTex/figure2D}

This section presents a Monte Carlo analysis of the reconstruction errors and timing, comparing \naive, \lmin, and \algOne.
Results are averaged over 50 runs, and the synthetic \twd \signals are generated as specified in the previous section. 

% \prettyref{fig:recovery1D}(a)-(c) show how the reconstruction quality is influenced by the number of corners in the ground truth \signal 
% (i.e., the sparsity of the true \signal), the percentage of measurements, and the noise level. 
%Each data point in the plot is generated by averaging over 50 runs. 
\prettyref{fig:recovery1D}(a) shows how the depth reconstruction quality is influenced by the number of corners in the ground truth 
\signal 
 (i.e., the sparsity of the true \signal), comparing \naive, \lmin, and \algOne. 
These results consider noiseless measurements and sample set including a
twin sample in each linear region (these are the assumptions of \prettyref{prop:1DrecoveryError}).
 %reconstruction errors are shown for \signals with increasing number of corners.
As predicted by~\prettyref{cor:algorithm}, \algOne recovers the original \signal exactly (zero error). 
\naive has large errors, while the \lmin estimate 
falls between the two. 

\prettyref{fig:recovery1D}(b) considers a more realistic setup: since in practice we do not 
know where the corners are (hence we cannot guarantee to sample each linear segment of the true \signal), in this case we uniformly sample depth measurements 
and we consider noisy measurements with $\vareps = 0.1\text{m}$. 
The figure reports the estimation error for increasing number of samples.
As the percentage of samples goes to 1 (100\%), we sample all 
entries of the depth \signal. We consider \signals with 3 corners in this test. The figure shows that for increasing number of samples, our approaches largely outperform 
the \naive approach. \algOne improves over \lmin even in presence of noise, while the improvement is not as substantial as in the noiseless case of \prettyref{fig:recovery1D}(a).
\prettyref{fig:recovery1D}(b) also shows that the error committed by \naive does not improve when adding more samples. 
This can be understood from \prettyref{fig:example1D} and the discussion in~\prettyref{sec:exp_2drec}. 

\prettyref{fig:recovery1D}(c) considers a fixed amount of samples (5\%) and tests the three approaches for increasing measurement noise. 
Our techniques (\lmin, \algOne), are very resilient to noise and degrade gracefully in presence of large noise (e.g., $\vareps = 1$m).

\prettyref{fig:recovery1D}(d) shows the CPU times required by \lmin and \algOne in \twd reconstruction problems using the \cvx solver.
The CPU time for \naive is negligible (in the milliseconds).

%%%%%%%%%%%%%%%%%%%%%%%%%%%%%%%%%%%%%%%%%%%%%%%%%%%%%%%%%%%%%%%%%%%%%%%%%%%%%%%%%%%%
\subsubsection{\twd mapping from sparse measurements} 
\label{sec:exp_2drec_mapping}

% We also create a simulated dataset on the \emph{Stage} simulator~\cite{gerkey2003player}, where a robot equipped with a laser scanner with only 10 beams moves in a \twd scenario. This dataset includes both odometry information and \twd laser measurements. 

This section applies our approach to a \twd mapping problem from sparse measurements.
We use the \emph{Stage} simulator~\citep{gerkey2003player} to simulate a robot equipped with a laser scanner with 
only 10 beams, moving in a \twd scenario. The robot is in charge of mapping the scenario; we assume the trajectory to be given.
 Our approach works as follows: we feed the 10 samples measured by our ``sparse laser'' to algorithm \algOne; 
\algOne returns a full scan (covering 180 degrees with 180 scans in our tests), which we feed to a standard mapping routine 
(we use {\tt gmapping}~\citep{Grisetti07tro} in our tests).

\input{figureTex/figureMapping2D}

\prettyref{fig:mapping1D} compares the occupancy grid map produced by a standard mapping algorithm
 based on a conventional laser scan (\prettyref{fig:mapping1D}(a)), against the occupancy grid map reconstructed from our 10-beam laser. 
  \prettyref{fig:mapping1D}(b) shows the map produced from the scans estimated using \naive: the map has  multiple artifacts. 
 \prettyref{fig:mapping1D}(c) shows the map produced from the scans estimated using \algOne; 
 the proposed technique produces a fairly accurate reconstruction from 
 very partial information.  
 % As a technical remark, we note that, before feeding the 10 laser measurements to \algOne, we express the corresponding points in Cartesian coordinates: the original data given by the laser is in polar coordinates and piecewise linear 
 % \signals do not remain piecewise linear in polar coordinates (see~\prettyref{fig:metricExample}).

%% file: figureTex/figure2D.tex
%!TEX root = ../main.tex

\begin{figure}[htbp]

\begin{minipage}{\textwidth}
\newcommand{\figWidth}{ 0.48\linewidth } 
\def\arraystretch{3}		\setlength\tabcolsep{1mm}	\smaller
\begin{tabular}{ l  r }\begin{minipage}[b]{\figWidth}\centering
\includegraphics[width=\linewidth]{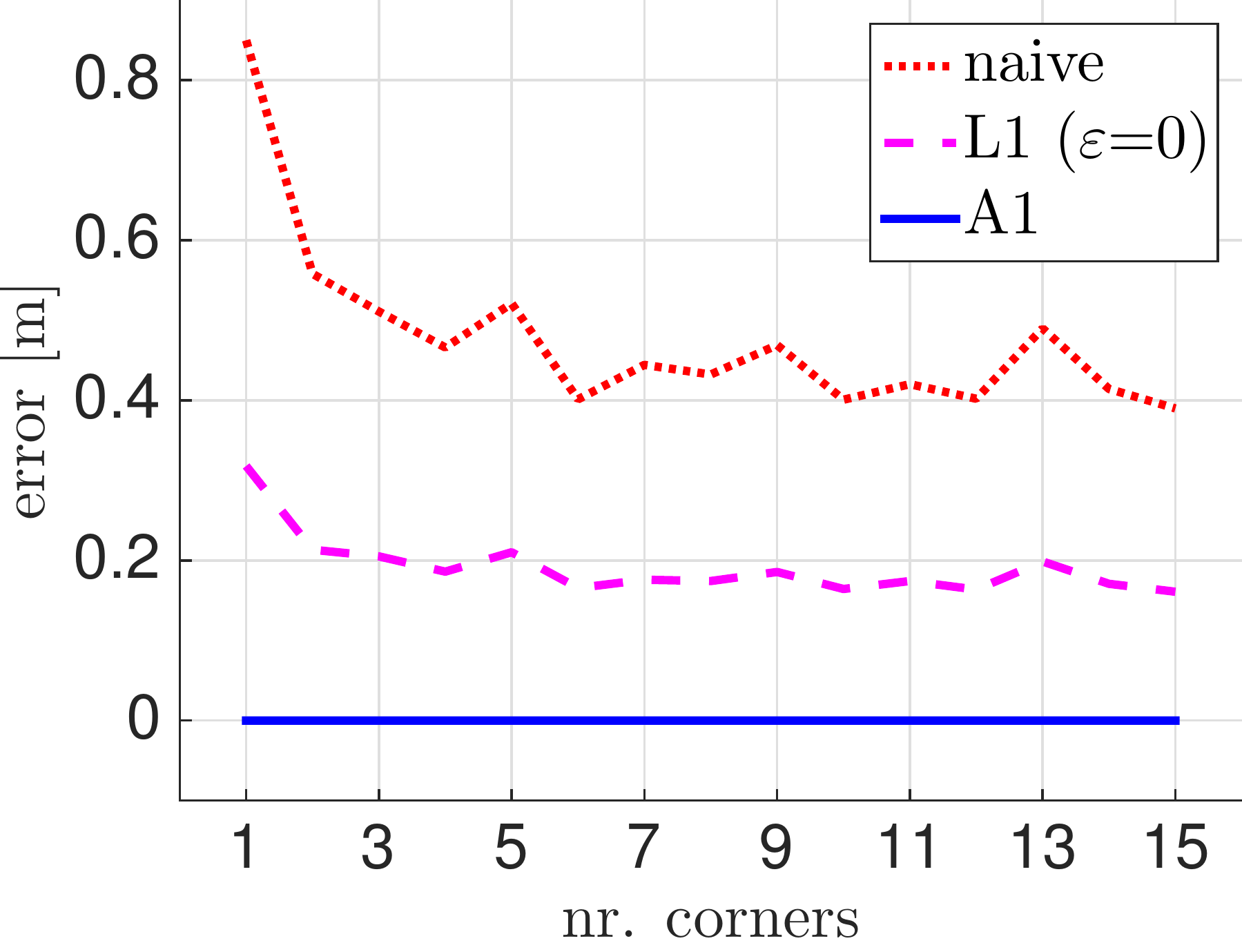} 
\\
(a) error vs. \# corners
\end{minipage}
& 
\begin{minipage}[b]{\figWidth}\centering
\includegraphics[width=\linewidth]{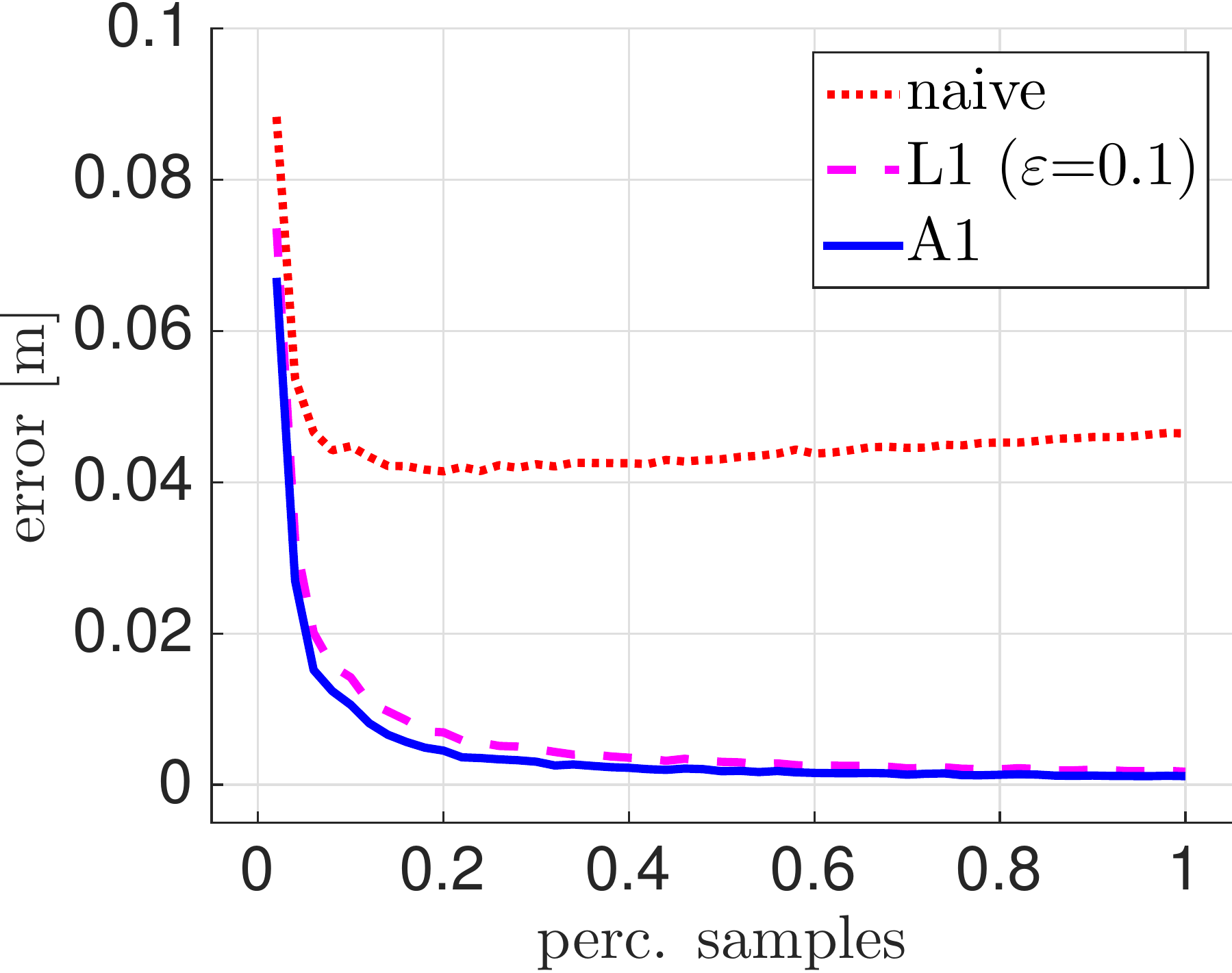}  
\\
(b) error vs. percentage of samples
\end{minipage}
\\ 
\begin{minipage}[b]{\figWidth}\centering
\includegraphics[width=\linewidth]{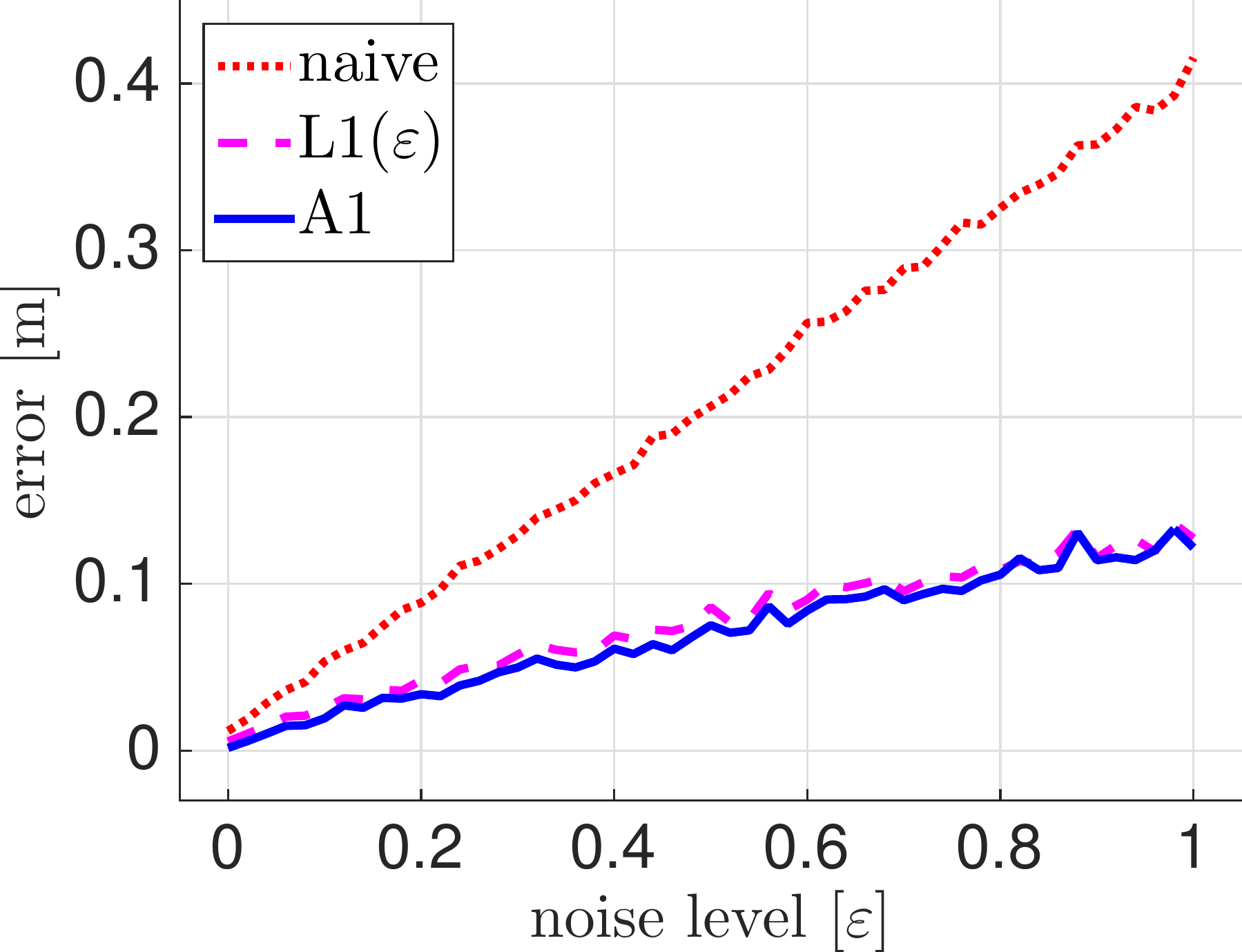}   
\\
(c) error vs. noise level
\end{minipage}
&  
\begin{minipage}[b]{\figWidth}\centering
\includegraphics[width=\linewidth]{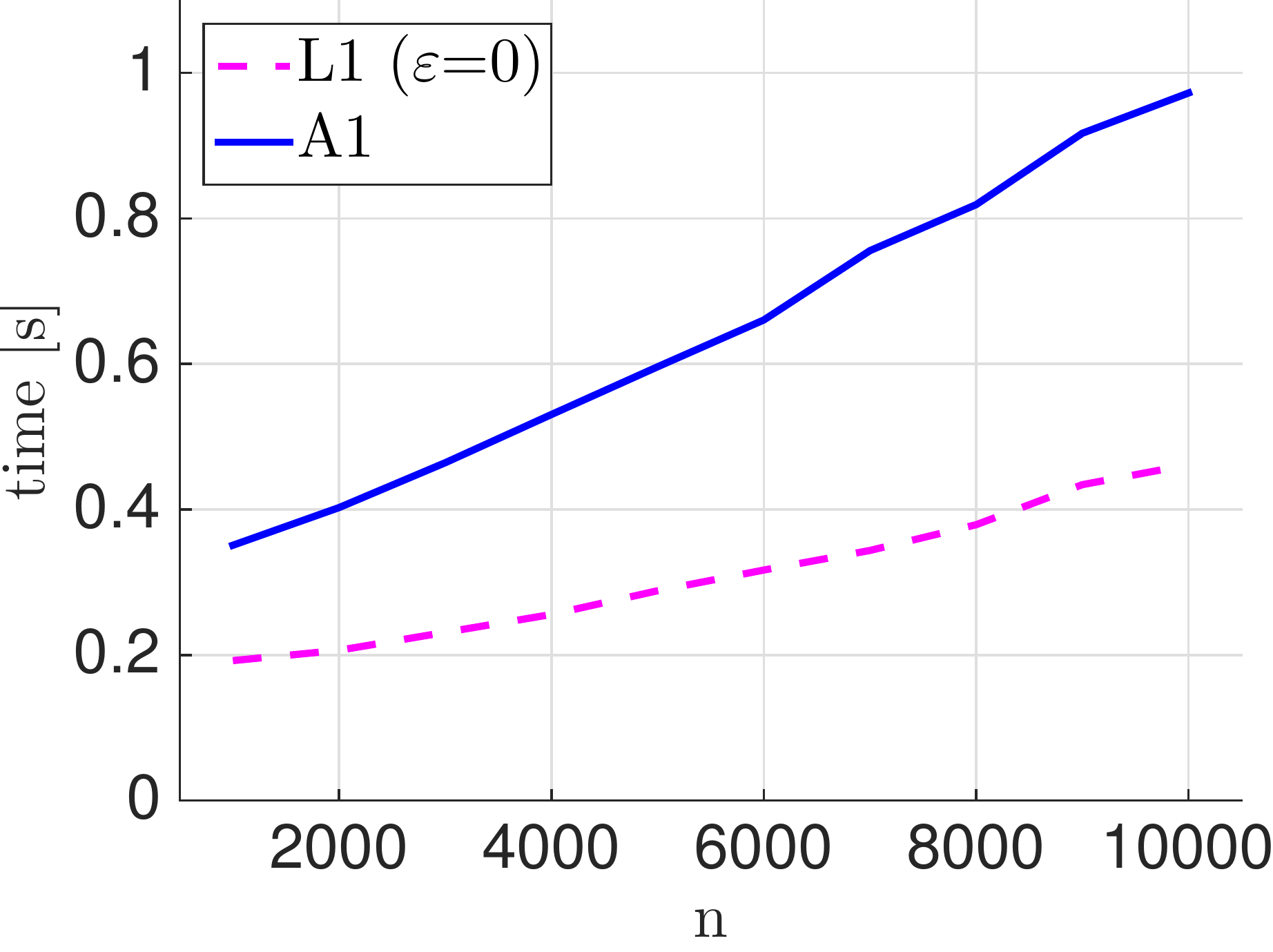}
\\
(d) CPU time required
\end{minipage}

\end{tabular}
\end{minipage}
\vspace{-3mm}
\caption{
(a)~Estimation errors committed by \algOne, \lmin, and \naive, for increasing number of corners in the ground truth \signal. A twin sample is acquired in each linear segment 
and measurements are noiseless. 
(b)~Estimation errors from uniformly sampled noisy measurements, with increasing number of samples.
(c)~Estimation errors from uniformly sampled noisy measurements ($5\%$ of the depth data points)
and increasing noise level.
(d)~CPU time required to solve \lmin and \algOne for increasing size $n$ of the \twd depth \signal.
}
\label{fig:recovery1D} 
\end{figure}

%% file: figureTex/figureMapping2D.tex
%!TEX root = ../main.tex

\begin{figure}[hbtp]

\begin{minipage}{\textwidth}
\newcommand{\figWidth}{ 0.325\linewidth } 
\def\arraystretch{1}		\setlength\tabcolsep{0.5mm}	\smaller
\begin{tabular}{ccc}\begin{minipage}[b]{\figWidth}\centering
\includegraphics[width=\linewidth]{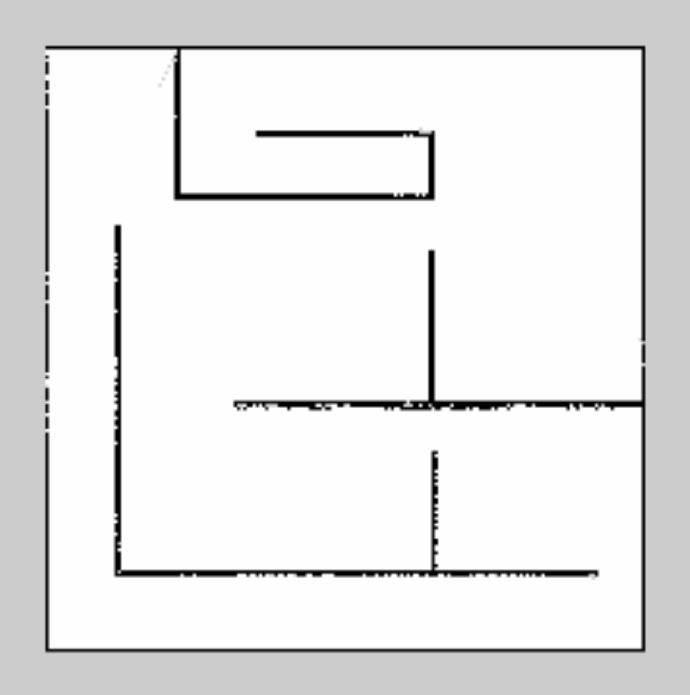}
\\
(a) full scan
\end{minipage}
& 
\begin{minipage}[b]{\figWidth}\centering
\includegraphics[width=\linewidth]{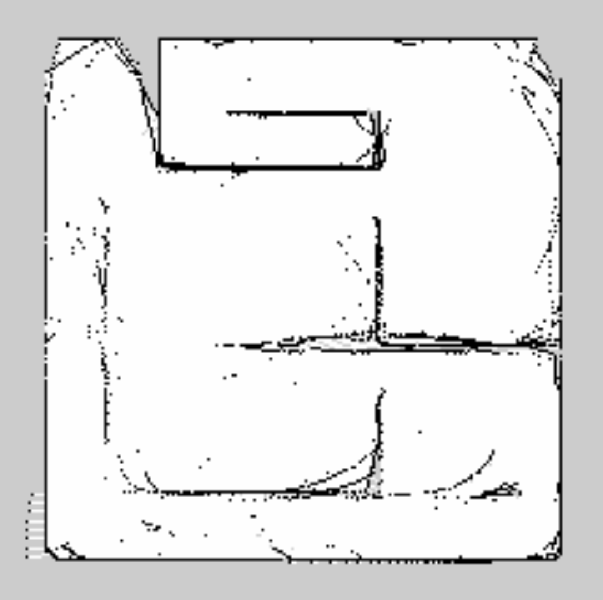} 
\\
(b) \naive
\end{minipage}
&
\begin{minipage}[b]{\figWidth}\centering
\includegraphics[width=\linewidth]{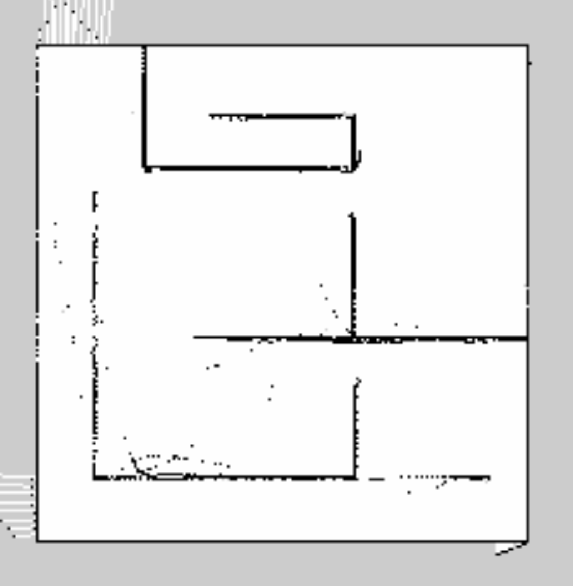} 
\\
(c) \algOne
\end{minipage}

\end{tabular}
\end{minipage}
\caption{
(a) Mapping using a conventional laser scanner,
(b-c) Mapping using scans reconstructed by \naive and \algOne using 10 depth measurements.
}
\label{fig:mapping1D}
\end{figure}

%% file: experimentTex/experiments_preliminary.tex
%!TEX root = ../main.tex
\subsection{\thd Reconstruction: Datasets, Objective Functions and Solvers}
\label{sec:exp_3Drec}

This section introduces the \thd datasets used for the evaluation in the following sections.
Moreover, it provides a statistical analysis of the performance obtained by the algorithmic variants 
presented in \prettyref{sec:alg2}, as well as the solvers presented in~\prettyref{sec:solvers}. 
The best performing variants and solvers will be used in the real-world examples and applications presented in 
Sections~\ref{sec:exp-single-frame-sparse-reconstruction}-\ref{sec:exp_superResolution}.

% In \prettyref{sec:cs} we introduced both the exact reconstruction problem~\ref{eq:BPmatrix} and the noisy reconstruction problem~\ref{eq:BPDmatrix} in 3D. There exist several variants of the problem with slightly different regularization matrices $\TV$ (and thus different objective functions to minimize), each with their own strength and weakness, as discussed in \prettyref{sec:alg2}. There also exist several different solvers. In this section, we will have an empirical evaluation of these objective functions and solvers. 

%%%%%%%%%%%%%%%%%%%%%%%%%%%%%%%%%%%%%%%%%%%%%%%%%%%%%%%%%%%%
\subsubsection{Datasets}
In this section we introduce the datasets we use to benchmark our \thd depth reconstruction approaches. In order to have a ground truth \signal, 
%for proper evaluation of the reconstruction algorithms, 
we collected several datasets with commonly-used high-resolution %, regular-sized 
depth sensors (including a \emph{Kinect} and a \emph{ZED} stereo camera) and use an heavily down-sampled depth image as our ``sparse'' depth measurements. Moreover, we created synthetic \signals for a more exhaustive evaluation.
%, instead of using actual sparse and low-resolution depth sensors. 

\input{figureTex/figureDataset}

Our testing datasets include a dataset of randomly-generated synthetic piecewise linear depth images (denoted as \PL), a simulated dataset from the \emph{Gazebo} simulator~\citep{koenig2004design} (denoted as \gazebo), a stereo dataset from a \emph{ZED} camera (denoted as \zed), 8 datasets from a \emph{Kinect} camera (denoted as \kinect{1} to \kinect{8}), and the Middlebury stereo datasets~\citep{scharstein2007learning,hirschmuller2007evaluation}. 
More specifically, \gazebo contains 20 full depth and RGB images rendered in an office-like environment from the \emph{Gazebo} simulator 
(\prettyref{fig:dataset}(a)). 
\zed includes 1000 full disparity and RBG images, collected from a \emph{ZED} stereo camera mounted on a dolly, 
 in the \emph{Laboratory of Information and Decision Systems} (LIDS) at MIT (\prettyref{fig:dataset}(b)). 
\kinect{1} to \kinect{8} contain odometry information, as well as depth and RGB images, collected from a \emph{Kinect} sensor mounted on a dolly with wheel odometers, moving in 8 different locations at MIT, including tunnels, offices, and corridors (\prettyref{fig:dataset}(c)-(d)). 
%Some examples are shown in \prettyref{fig:dataset}.
The Middlebury stereo dataset is used for the sake of benchmarking against the previous works~\citep{hawe2011dense,liu2015depth},
 which use a similar experimental setup, and includes  disparity images of size 256-by-256 (each down-sampled from the original 512-by-512 images).

\input{figureTex/figureMetricComparison}

%%%%%%%%%%%%%%%%%%%%%%%%%%%%%%%%%%%%%%%%%%%%%%%%%%%%%%%%%%%%
\subsubsection{Objective Functions}
\label{sec:exp-objective-functions}
In this section we compare the three objective functions discussed in \prettyref{sec:alg2} for the noiseless reconstruction problem~\ref{eq:BPmatrix}. We use the \cvx/\MOSEK \citep{CVXwebsite} solver in MATLAB in this section, to reduce numerical approximations, while we evaluate the use of other 
solvers (\NESTA) in the next section.
% such that the results are not 
% affected by the 
% %to compare the objectives without the 
% numerical approximations induced by \NESTA.

\prettyref{fig:metrics} compares the reconstruction errors of the three different objective functions on the datasets \PL, \zed, and \kinect{1}-\kinect{8}. The error bars show the reconstruction error for each objective functions (\lmin, \lminDiag, and \lminCart), averaged over all the images in the corresponding dataset. The depth measurements are sampled from a grid, such that only $4\%$ of the pixels in the depth \signals are used. 
The ground truth \signals have resolution 
%are down-sampled to 20\% of its original size, resulting in a resolution of 
85$\times$103 for the Kinect datasets, 96$\times$128 for the \zed dataset, and 40$\times$40 for the \PL dataset.
% \LC{resolution, size of the reconstruction}.
%\LC{(4\% samples, i.e., one measurement from each 5-by-5 image patch)}.

From \prettyref{sec:alg2}, we recall that \lminCart includes a parameter $\param$ which 
prevents 
%avoids degenerate cases in which 
the denominator 
of some of the entries in~\eqref{eq:l1cart_kernel} to become zero.
\prettyref{fig:metrics}(a) and \prettyref{fig:metrics}(b) show the reconstruction errors for 
$\param=0.1\text{m}$ and $\param=0.01\text{m}$, respectively.
% Recall the discussion in \prettyref{sec:alg2} that a parameter $\param$ is inserted into the denominator when computing the second-order derivative, such that the ill-conditioned case of division by zero can be avoided. 
From \prettyref{fig:metrics} it is clear that the accuracy of
\lminCart  heavily depends on the choice of $\param$, and degrades significantly for small values of $\param$. 
Moreover, even for a good choice of $\param$ (\prettyref{fig:metrics}(a)) the advantage of \lminCart over  \lmin and \lminDiag 
is minor in most datasets. 
% Therefore, it is in general not reliable. 
The \lminDiag objective, on the other hand, performs consistently better than \lmin across all datasets and is parameter-free. 

%From this analysis, 
We conclude that while the variants \lmin, \lminDiag, and \lminCart  do not induce large performance variations, 
 \lminDiag ensure accurate depth reconstruction and we focus our attention on this technique in the following  sections.

Extra visualizations for the \lmin and \lminDiag formulations are provided in Appendix~\ref{sec:gazeboImg} and Appendix~\ref{sec:zedImg}.

%%%%%%%%%%%%%%%%%%%%%%%%%%%%%%%%%%%%%%%%%%%%%%%%%%%%%%%%%%%%
\subsubsection{Solvers}
\label{sec:exp-solvers}

% \LC{unclear if we are using diag}
This section compares two solvers for $\ell_1$-minimization in terms of accuracy and speed.
The first solver is
%In \prettyref{sec:solvers} we discussed two different solvers for the $\ell_1$-minimization problems. One of the solvers is 
\cvx/\MOSEK \citep{CVXwebsite} (denoted as \cvx for simplicity), a popular general-purpose parser/solver for convex optimization.
The second is \NESTA \citep{BeckerBC11}, which we adapted to our problem setup in \prettyref{sec:solvers}.
We implemented \NESTA in Matlab, starting from the open-source implementation of~\citep{BeckerBC11}; 
our source code is also available at {\small \url{https://github.com/sparse-depth-sensing}}.
%a first-order method for sparse recovery problems. In this section, we empirically compare the two solvers on their accuracy and speed.

%\prettyref{fig:nesta_vs_cvx} 
We compare the two solvers on the synthetic dataset \PL, using the \lminDiag objective function.
% by running experiments on the piecewise linear synthetic datasets \PL. 
Each depth image in \PL is generated randomly with a fixed number of corners (3 in our tests) and is of size $100$-by-$100$, unless otherwise specified. All depth measurements are uniformly sampled at random from the ground truth \signal, and the 4 immediate neighbors 
(up, down, left, right) are also added into the sample sets. No noise is injected into the measurements ($\vareps = 0$). In all tests, we set the maximum number of inner iterations to $K = 10000$, the number of continuation steps to $T = 5$, and  the stopping criterion to $\tau = 10^{-5}$ for \NESTA. All data points in the plots are averaged from 50 random runs.
%Each data point on the plot is averaged out over 50 trials, using both \cvx/\MOSEK and \NESTA. 

% \LC{noise level?}
We start by evaluating the impact of the parameter $\mu_f$ on the accuracy and timing of \NESTA. 
% Note that \NESTA's accuracy is determined by the approximation factor $\mu$, and in 
 \prettyref{fig:nesta_mu} shows the trade-off between reconstruction error and computational time for different values of $\mu_f$. 
 The error is evaluated as the average mismatch between \NESTA and \cvx solutions.
 % Each data point is generated by averaging the results over 50 tests on piecewise linear synthetic data of size 100-by-100, with 5\% uniformly random depth samples. 
  In each test,  the depth samples include 5\% of the pixels, uniformly chosen at random.
% In all tests, 5\% uniformly random depth samples.
We note that the average error is in the order of millimeters in all cases. To obtain the best trade-off between accuracy and speed, we choose $\mu_f=0.001$, the ``elbow'' point in  \prettyref{fig:nesta_mu}. 
We use this value in all the following experiments.

\input{figureTex/figureNestaMu}

\prettyref{fig:nesta_vs_cvx} compares the performance of \cvx and \NESTA for increasing number of samples, noise, and size of the depth \signals.
\prettyref{fig:nesta_vs_cvx}(a)-(b) show the reconstruction error and computational time of the two solvers for increasing percentage of samples. Depth measurements are affected by entry-wise uniformly random measurement noise in $[-\vare, \vare]$; for this test we chose $\vare = 0.1$. 
% \LC{10 does not make sense}. 
\prettyref{fig:nesta_vs_cvx}(a) shows that the accuracy of \NESTA is close to the one of \cvx 
(the mismatch is in the order of few millimeters), while they both largely outperform Matlab's linear interpolation (\naive). 
 \prettyref{fig:nesta_vs_cvx}(b) shows that \NESTA is around 10x faster than \cvx (as in the \twd case, the computational time of \naive is
 negligible).
%are disturbed with noise with $\vare = 10$ (specifically, the noise is uniformly random in the range of $[-\vare, \vare]$ and is added to the measurements in an entry-wise manner). 

\prettyref{fig:nesta_vs_cvx}(c)-(d) show the reconstruction error and computational time for increasing noise level 
when sampling $5\%$ of the depth \signal. Also in this case the errors of  \NESTA and \cvx are very close, while 
\NESTA remains remarkably faster. % in all cases.
For both \NESTA and \cvx, the estimation error grows more gracefully with respect to the measurement noise $\vareps$, 
compared to the \naive approach.

\prettyref{fig:nesta_vs_cvx}(e)-(f) show the reconstruction error and computational time for increasing image size. We reconstruct random \signals  of size $N$-by-$N$ using 5\% of samples, without adding noise. \prettyref{fig:nesta_vs_cvx}(e) further confirms that the error curves for \cvx and \NESTA are almost indistinguishable, implying that they produce reconstructions of similar quality. However, the \NESTA solver entails a speed up of 3-10x, depending on the problem instance (\prettyref{fig:nesta_vs_cvx}(f)).

Given the significant advantage of \NESTA over \cvx, and since \cvx is not able to scale to large \signals, 
we use \NESTA in the
tests presented in the following sections.
%we will use \NESTA for the rest of the paper.

\input{figureTex/figureSolverComparison}

%% file: figureTex/figureDataset.tex
\begin{figure}[hbtp]
\begin{minipage}{\textwidth}
\newcommand{\figWidth}{ 0.48\linewidth } 
\def\arraystretch{3}		\setlength\tabcolsep{1mm}	\smaller
\begin{tabular}{ l  r }\begin{minipage}[b]{\figWidth}\centering
\includegraphics[width=\linewidth]{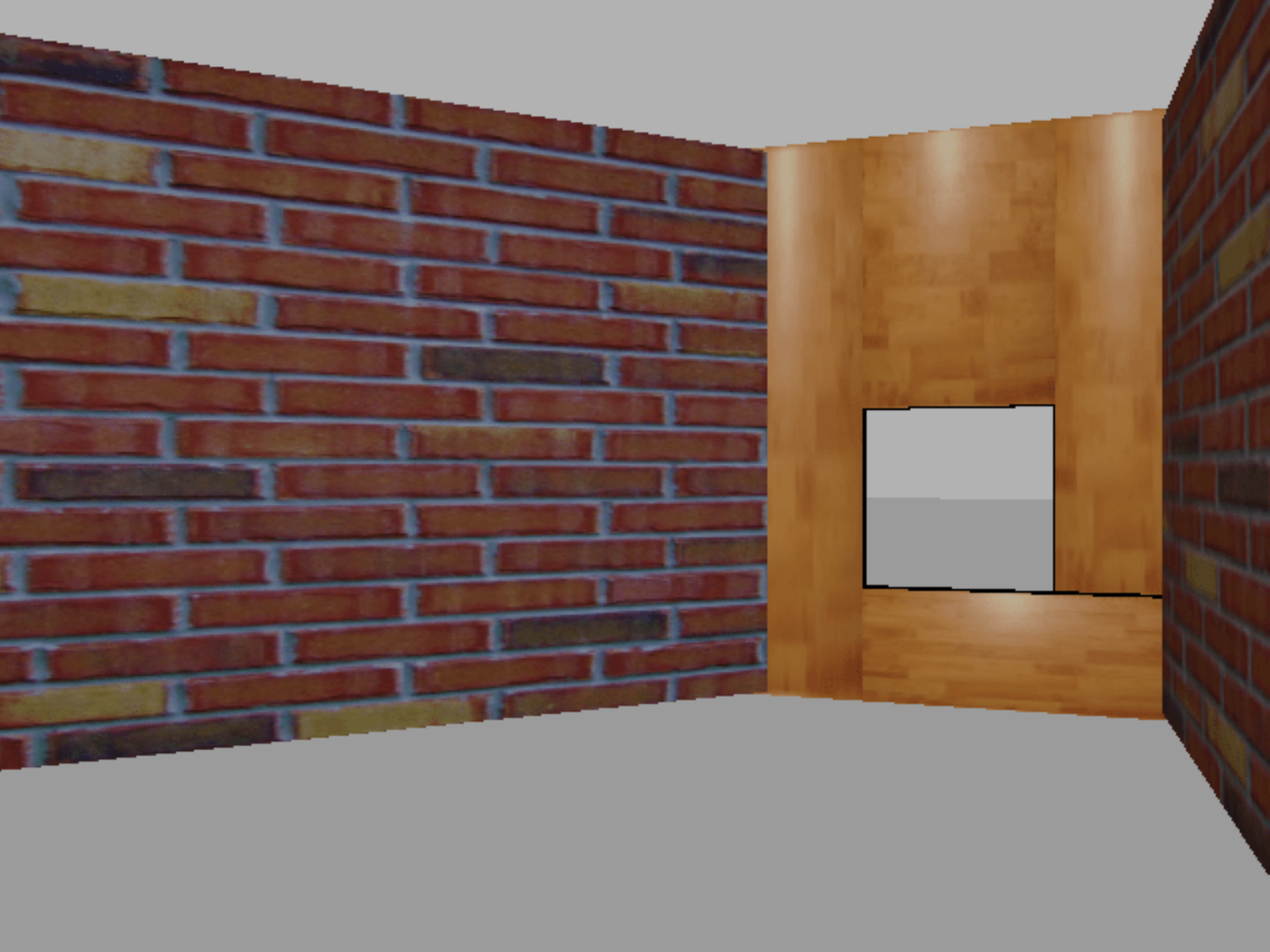} 
\\
(a) Gazebo
\end{minipage}
& 
\begin{minipage}[b]{\figWidth}\centering
\includegraphics[width=\linewidth]{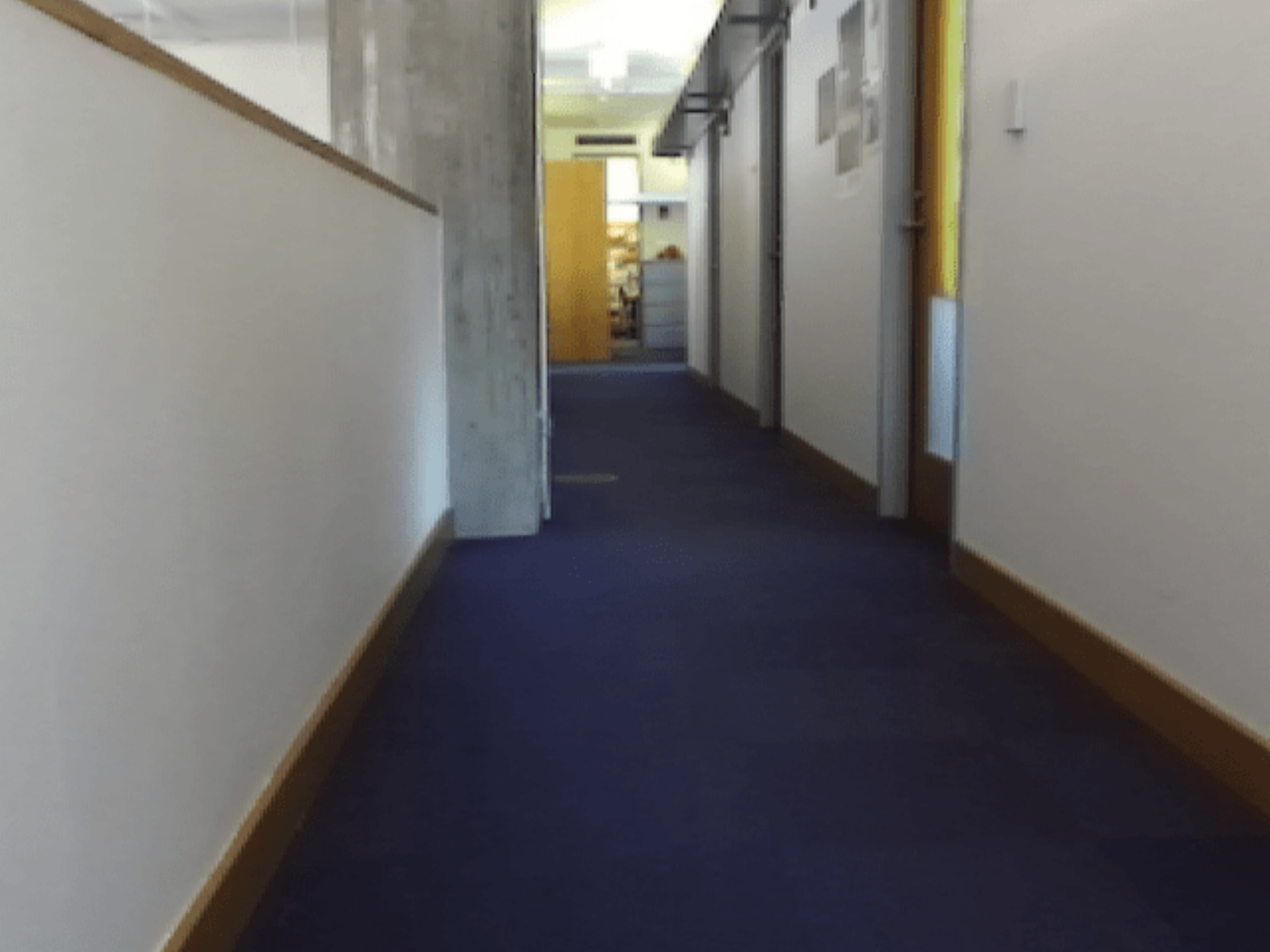} 
\\
(b) ZED
\end{minipage}
\\ 
\begin{minipage}[b]{\figWidth}\centering
\includegraphics[width=\linewidth]{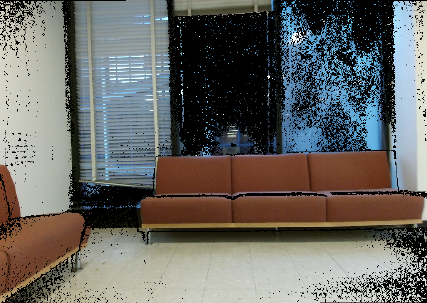} 
\\
(c) Kinect
\end{minipage}
&  
\begin{minipage}[b]{\figWidth}\centering
\includegraphics[width=\linewidth]{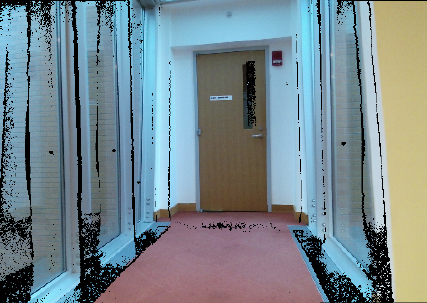} 
\\
(d) Kinect
\end{minipage}

\end{tabular}
\end{minipage}
\caption{
(a) \emph{Gazebo} Simulated data.
(b) \emph{ZED} Stereo data.
(c)-(d) \emph{Kinect} data. 
}
\label{fig:dataset}
\end{figure}

%% file: figureTex/figureMetricComparison.tex
%!TEX root = ../main.tex

\begin{figure}[htbp]

\myIncludeTwoFigures
{preliminary/{error-mean-stretch=0.1}.pdf}
{preliminary/{error-mean-stretch=0.01}.pdf}
{0}{0}{0}{0}
\caption{
Comparison between three different objective functions, \lmin, \lminDiag, and \lminCart, on 10 benchmarking datasets. 
(a)~constant in \lminCart is chosen as $\param=0.1\text{m}$, (b)~constant in \lminCart is chosen as $\param=0.01\text{m}$.
}
\label{fig:metrics} 
\end{figure}

%% file: figureTex/figureNestaMu.tex
%!TEX root = ../main.tex
\begin{figure}[t]

\centering\includegraphics[width=6cm]{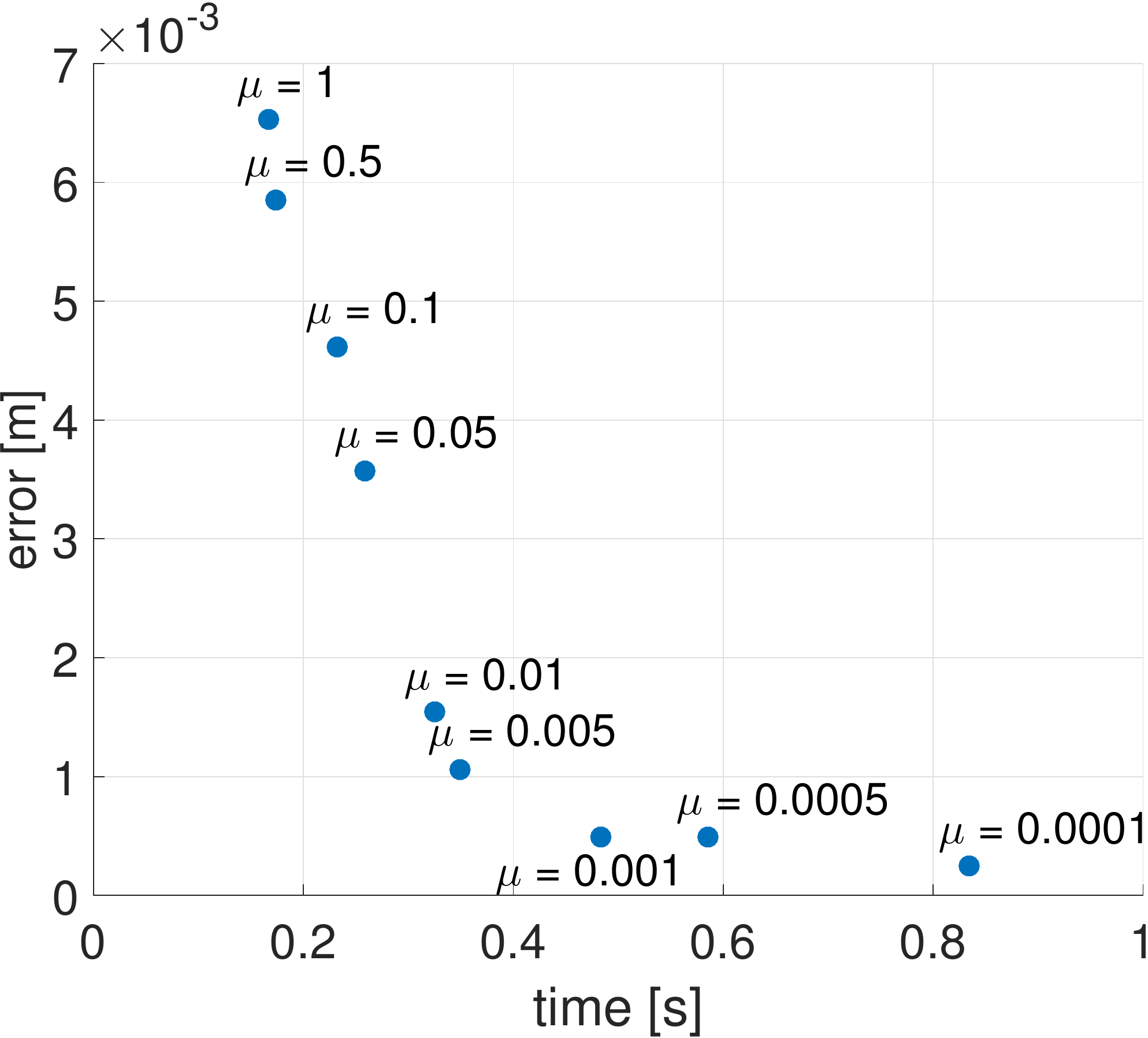}  
\vspace{-8mm} \\
\hspace{-20mm} (a)
\caption{
Trade-off between accuracy and speed for the \NESTA solver with different parameter values $\mu_f$. 
As $\mu_f$ decreases,  \NESTA  produces a more accurate solution at the cost of higher 
computational time. The error is computed as the average mismatch between the \NESTA and \cvx solutions.
}
\label{fig:nesta_mu} 
\end{figure}

%% file: figureTex/figureSolverComparison.tex
%!TEX root = ../main.tex
\begin{figure}[t]
\begin{minipage}{\textwidth}
\newcommand{\figWidth}{ 0.48\linewidth } 
\def\arraystretch{3}		\setlength\tabcolsep{1mm}	\smaller
\begin{tabular}{ l r }
\begin{minipage}[b]{4cm}\centering\includegraphics[width=\linewidth]{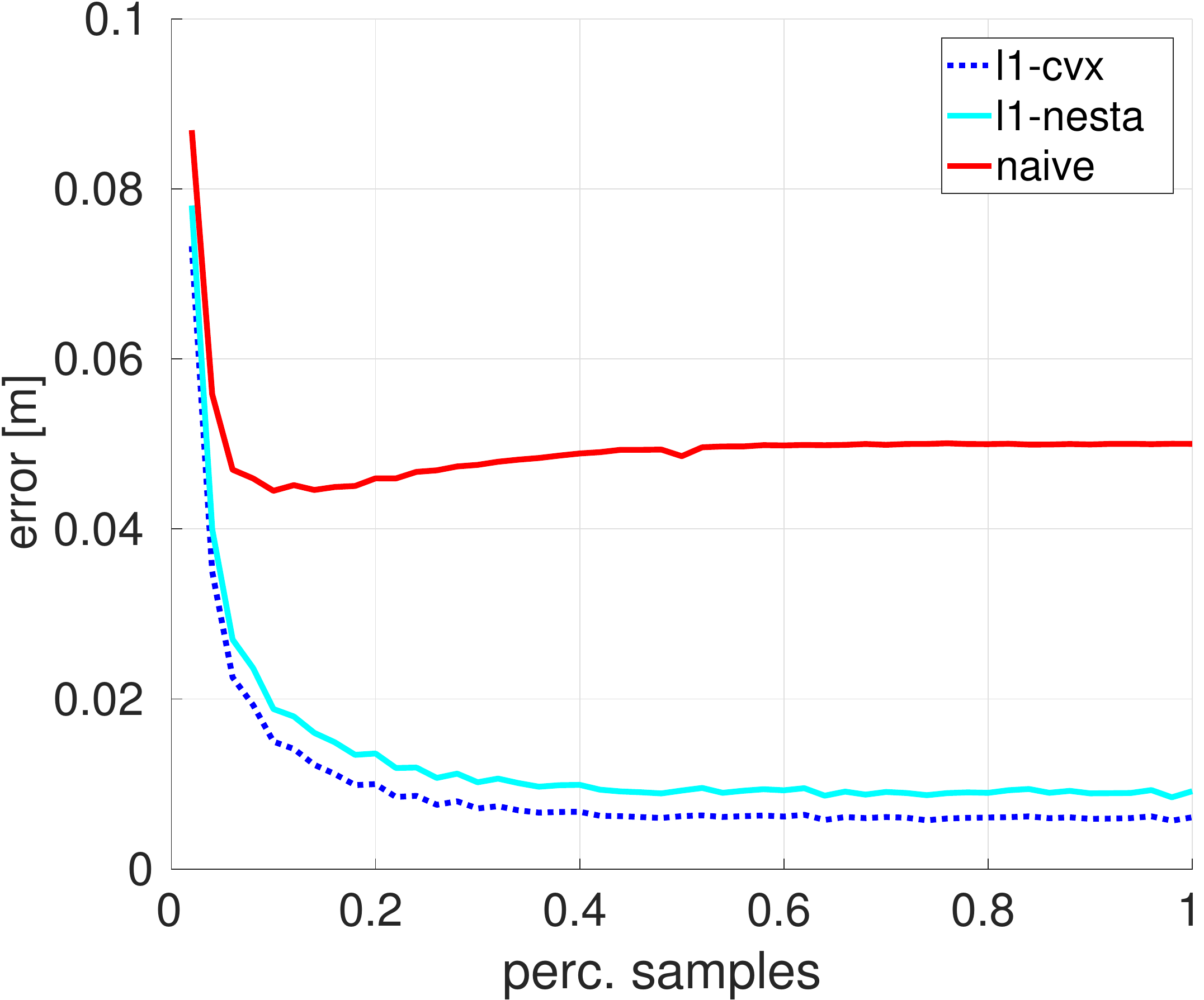} 
 \\
(a) error for increasing samples
\end{minipage}
&  
\begin{minipage}[b]{\figWidth}\centering\includegraphics[width=\linewidth]{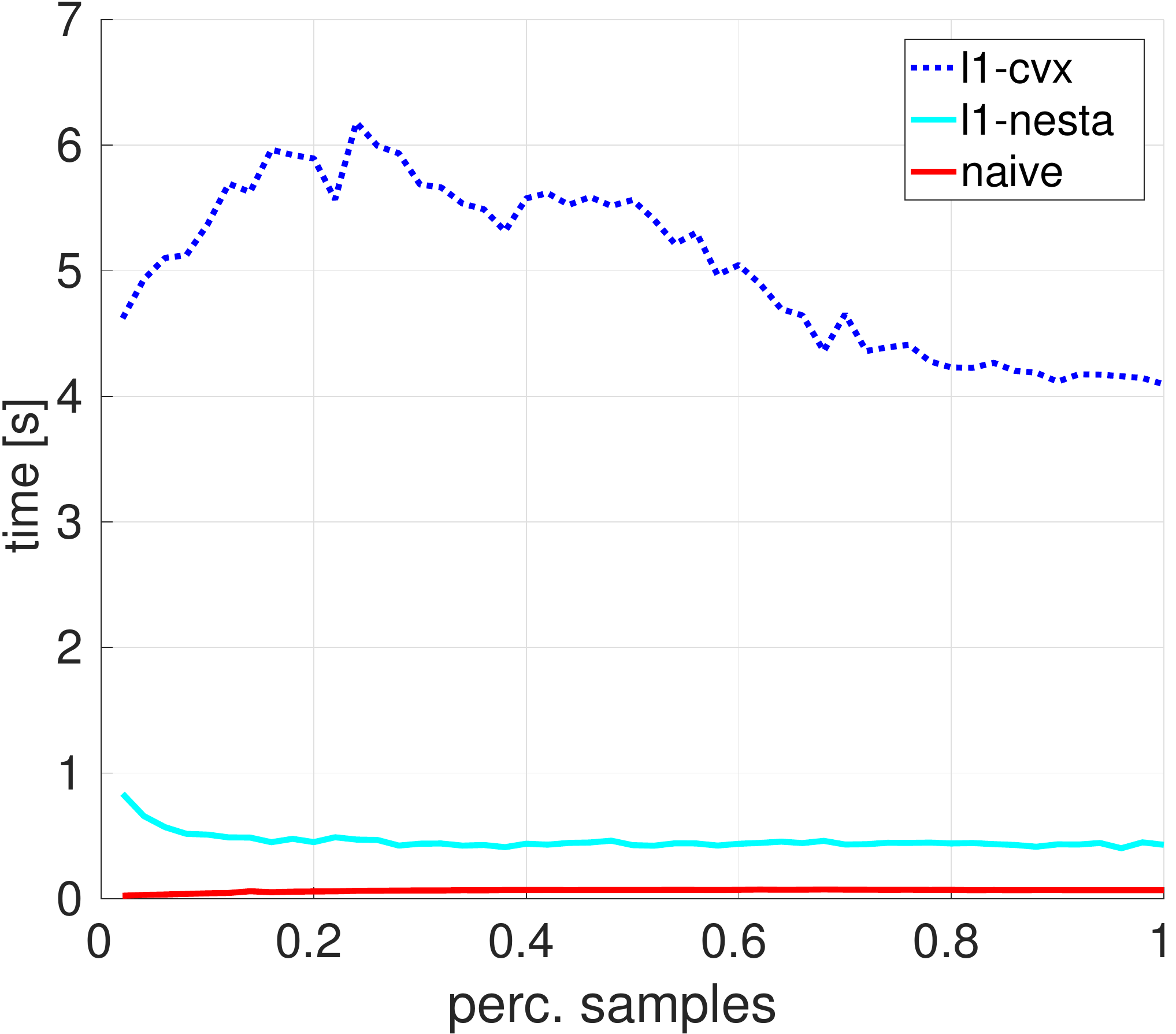}
\\
(b) time for increasing samples
\end{minipage}
\\
\begin{minipage}[b]{\figWidth}\centering\includegraphics[width=\linewidth]{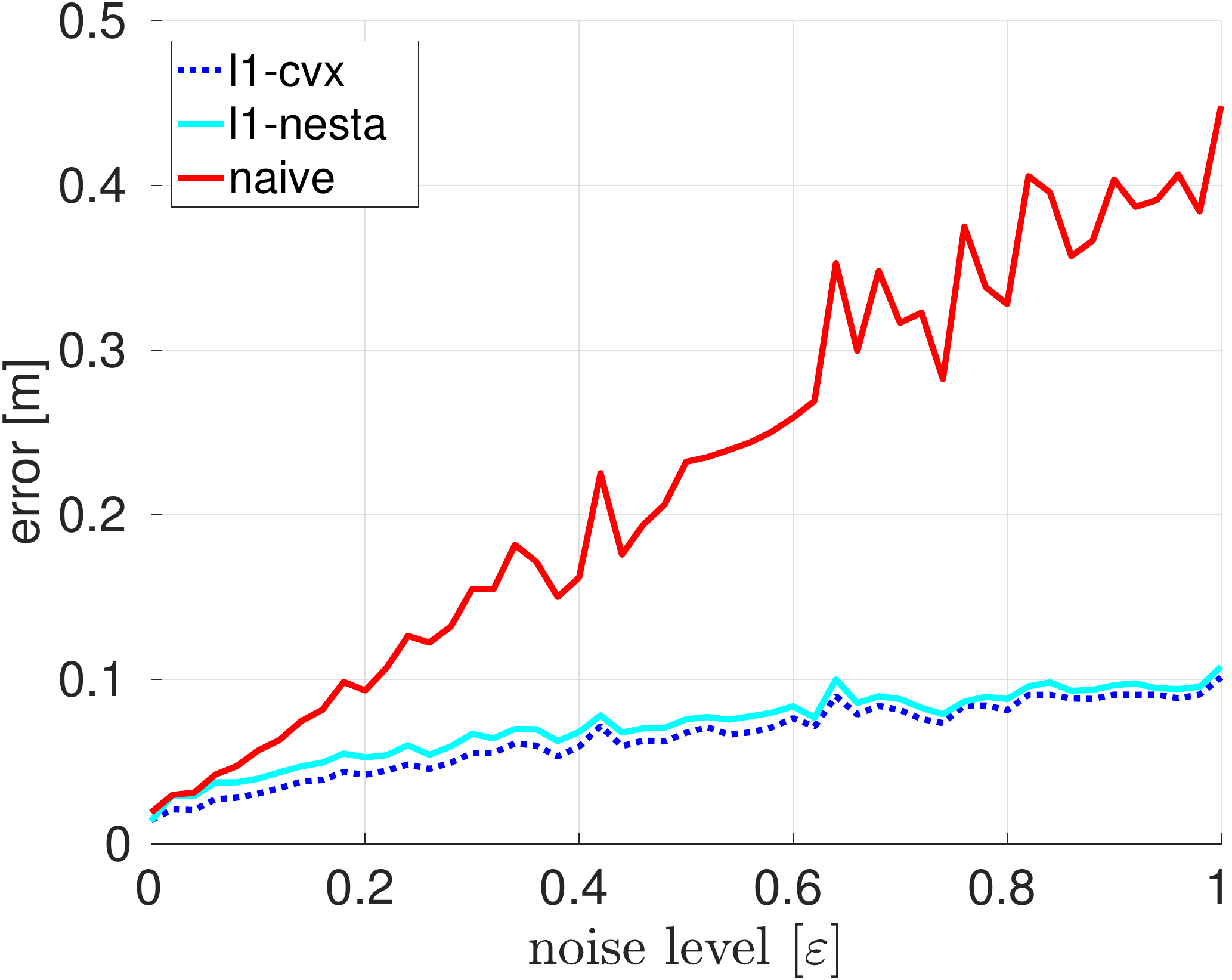} 
 \\
(c) error for increasing noise
\end{minipage}
&  
\begin{minipage}[b]{\figWidth}\centering\includegraphics[width=\linewidth]{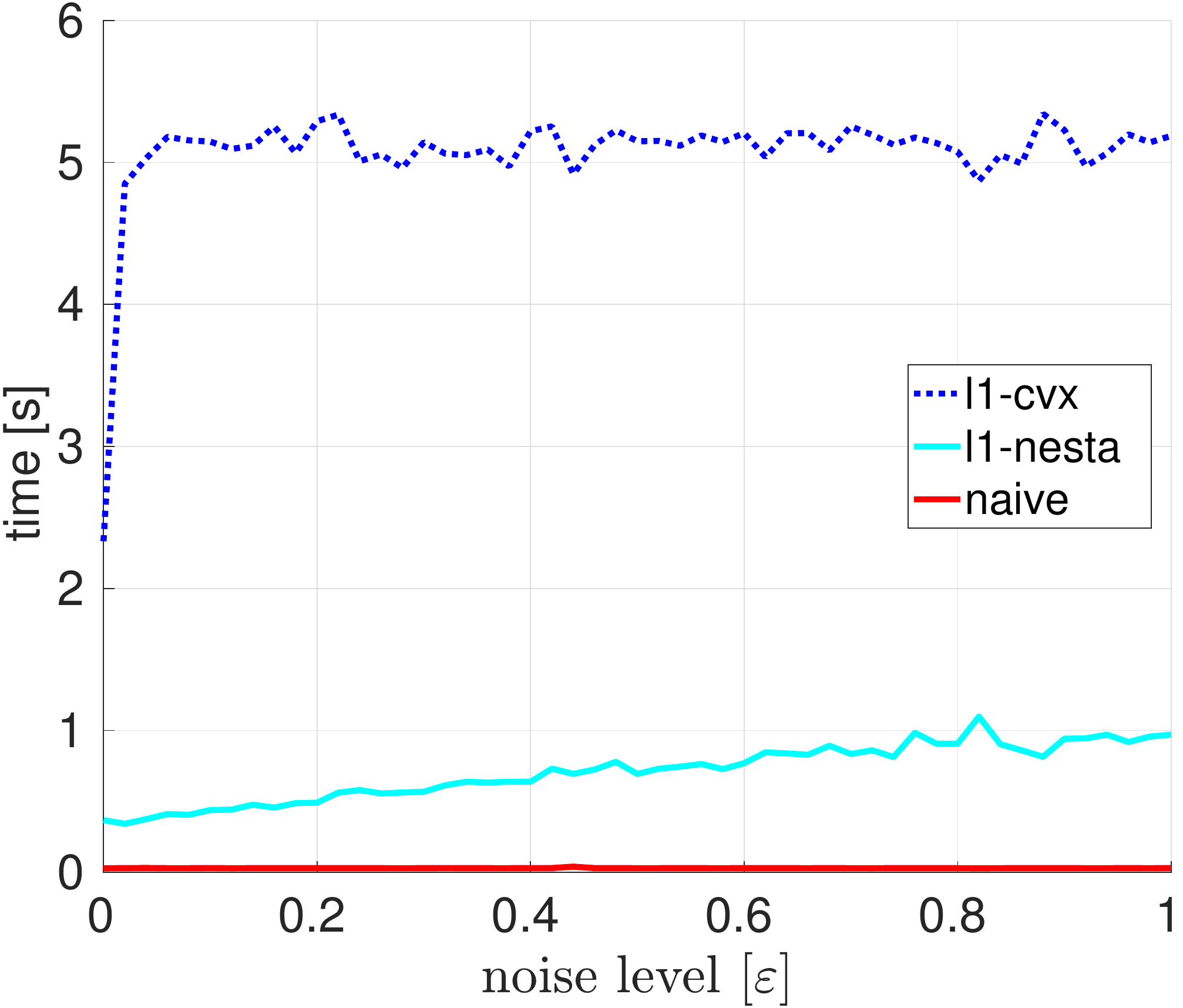} 
\\
(d) time for increasing noise
\end{minipage}
\\
\begin{minipage}[b]{\figWidth}\centering\includegraphics[width=\linewidth]{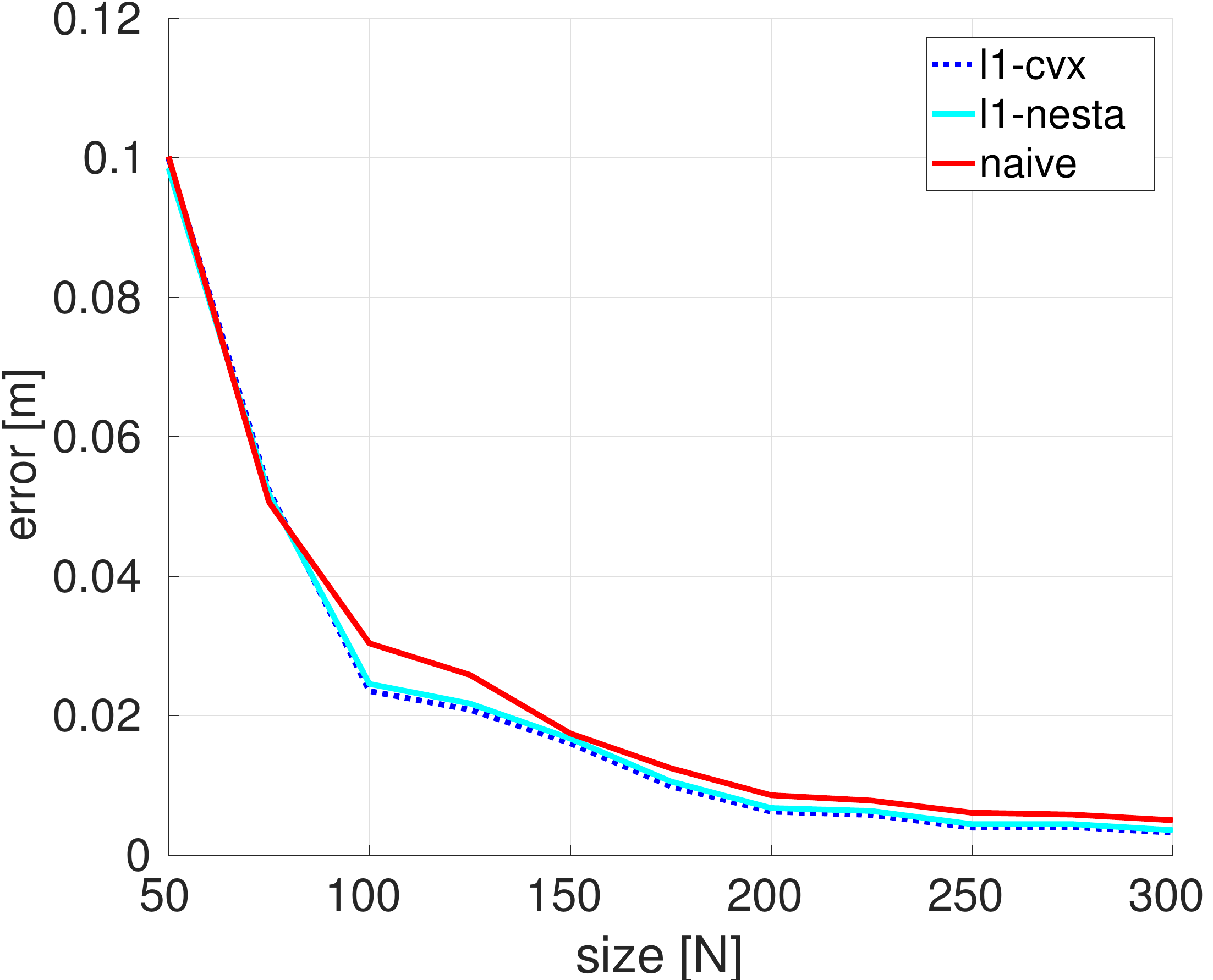}
 \\
(e) error for increasing size
\end{minipage}
&  
\begin{minipage}[b]{\figWidth}\centering\includegraphics[width=\linewidth]{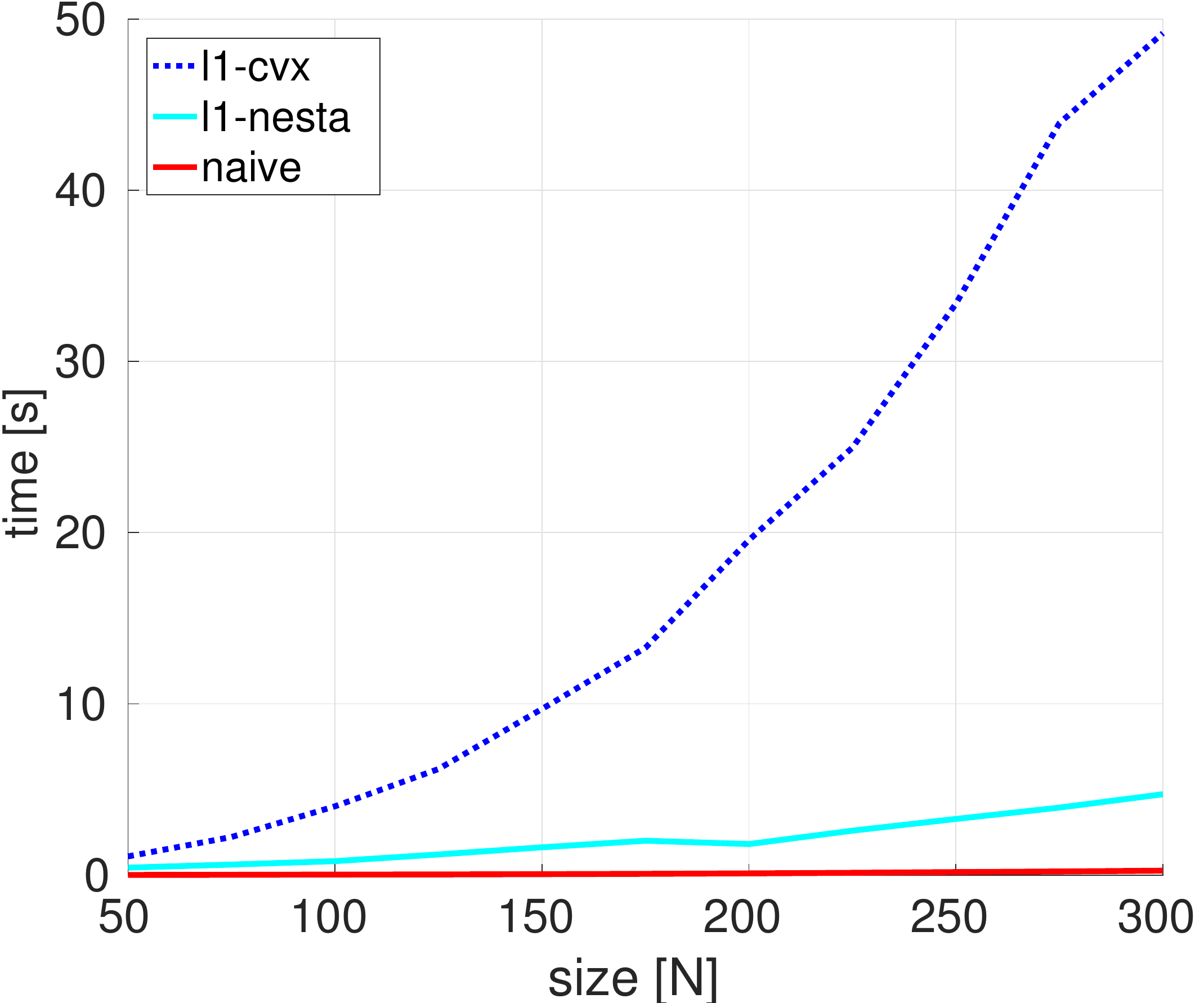}
 \\
(f) time for increasing size
\end{minipage}
\end{tabular}
\end{minipage}\caption{
Comparison between \NESTA ($\mu_f=0.001$) and \cvx. 
Estimation errors and timing are shown for (a-b)~increasing  number of samples, 
(c-d)~increasing measurement noise, (e-f)~increasing size of the $N\times N$ depth \signals.
\NESTA achieves comparable reconstruction errors, while offering a significant speedup. }
\label{fig:nesta_vs_cvx} 
\end{figure}

%% file: experimentTex/experiments_3dSparseReconstruction.tex
%!TEX root = ../main.tex

%%%%%%%%%%%%%%%%%%%%%%%%%%%%%%%%%%%%%%%%%%%%%%%%%%%%%%%%%%%%
\subsection{Single-Frame Sparse \thd Reconstruction}
\label{sec:exp-single-frame-sparse-reconstruction}

The previous section confirmed that choosing  \lminDiag as objective function and \NESTA (with $\mu_f=0.001$) as solver ensure the 
best performance.
This section extends the numerical evaluation to the other \thd datasets, including \gazebo, \zed, \kinect{1}-\kinect{8}.
For each dataset, we use \lminDiag to reconstruct the depth at each frame from a small subset of  samples,
and we compare our approach against the \naive linear interpolation.
% In this section we demonstrate how our algorithms can reconstruct single-frame, complete 3D depth \signals from sparse and incomplete depth measurements (e.g., low-resolution laser scanner or sparse stereo correspondences). 
% We evaluate the \lminDiag reconstruction by comparing its accuracy against the baseline approach, i.e., the \naive linear interpolation. The \lminDiag problem is solved with \NESTA with a fixed approximate parameter $\mu=0.001$. 
In the following, we discuss typical reconstruction results, provide 
%We show examples of the reconstruction on different datasets, as well as 
 error statistics for different percentages of samples and noise levels, 
 and compare \lminDiag against the state-of-the-art techniques proposed in 
 \citep{hawe2011dense, liu2015depth}.

%%%%%%%%%%%%%%%%%%%%%%%%%%%%%%%%%%%%%%%%%%%%%%%%%%%%%%%%%%%%%%%%%%
\subsubsection{Typical Examples of \thd Reconstruction} 
We start by showing reconstruction examples from sparse depth measurements on the \gazebo and \kinect{1} datasets.

\input{figureTex/figure3DExample}

\prettyref{fig:3D_Example}(a)-(c) show an example on the \gazebo simulated dataset with uniformly random depth measurements 
%(a sampling strategy that poses more serious challenges to the reconstruction algorithm)
and the reconstructed full depth \signal based on these samples. The reconstructed depth image reflects the true geometry of the scene, 
even when we are only using 2\% samples and their neighbors (total is roughly 8\%). The reconstruction error in this example is 5cm.

\prettyref{fig:3D_Example} (d)-(e) shows an example on the \kinect{1} dataset, where all depth measurements fall on a regular grid. This sampling strategy resembles the output of a low-resolution depth sensor. Note that even though only a total number of 42 measurements measurements is available, the reconstructed depth image still correctly identifies the corridor and the walls.
The reconstruction error in this example is 18cm.

Extra visualizations for the \gazebo and the \zed datasets are provided in Appendix~\ref{sec:gazeboImg} and Appendix~\ref{sec:zedImg}, 
respectively.

%%%%%%%%%%%%%%%%%%%%%%%%%%%%%%%%%%%%%%%%%%%%%%%%%%%%%%%%%%%%%%%%%%
\subsubsection{Statistics for \thd Reconstruction} 
In this section we rigorously benchmark the performance of \lminDiag against the \naive approach, in terms of both the reconstruction accuracy and the robustness to measurement noise.

\input{figureTex/figure3DError}

\prettyref{fig:Gazebo_ZED_percSamples} depicts the reconstruction errors for increasing percentages of uniformly-random samples on different datasets. \prettyref{fig:Gazebo_ZED_percSamples}(a) shows reconstruction from noiseless samples on the \gazebo simulated datasets, while \prettyref{fig:Gazebo_ZED_percSamples}(b) is the same plot except with additional pixel-wise independent Gaussian measurement noise $\vare = 0.1$. \prettyref{fig:Gazebo_ZED_percSamples}(c)-(d) show the experimental results on the \zed stereo dataset and \kinect{1} dataset. No additional noise is added to these two datasets, since the raw data is already affected by actual sensor noise. \prettyref{fig:Gazebo_ZED_percSamples}(e) shows the comparison between \naive and \lminDiag over all datasets for reconstructions from 10\% samples and their immediate neighbors.

From the figures it is clear that our approach consistently outperforms the \naive linear interpolation in both the noiseless and noisy settings and across different datasets. The gap between \lminDiag and \naive widens as the number of samples increases in the noisy setup, which demonstrates that our approach is more resilient to noise. In the noiseless setup, the gap shrinks as the percentage of samples converges to 100\%, since in this case we are sampling a large portion of the depth \signal, a regime in which the \naive interpolation 
 often provides a satisfactory approximation.
%since there's no loss of information. 
\lminDiag produces significantly more accurate reconstruction (20-50\% error reduction  compared with \naive) when operating below the 20\%-samples regime, which is the sparse sensing setup that motivated this work in the first place. 

%%%%%%%%%%%%%%%%%%%%%%%%%%%%%%%%%%%%%%%%%%%%%%%%%%%%%%%%%%%%%%%%%%
\subsubsection{Comparisons with Related Work} 
\label{sec:exp-comparison}
In this section we provide an empirical comparison of our algorithm against previous work on disparity image reconstruction from sparse measurements. 
%Contrarily to the previous section, We use a slightly different experimental setup, 
Following the experimental setup used in \citep{hawe2011dense, liu2015depth},
we benchmark our technique in the Middlebury\footnote{\url{http://vision.middlebury.edu/stereo/data/}} stereo datasets~\citep{scharstein2007learning,hirschmuller2007evaluation}. 
Six different disparity images of size 256-by-256 (each downsampled from the original 512-by-512 images) are selected from the Middlebury dataset. We evaluate both the reconstruction accuracy and computational times for 4 different algorithms, including 
\naive and \lminDiag (discussed earlier in this paper), 
Hawe's \CSR~\citep{hawe2011dense}, and Liu's \WTCT~\citep{liu2015depth}. The sparse measurements are uniformly sampled from the ground truth image without noise. The same set of sparse samples are used for all 4 methods in each set of experiments. 
In order to allow a closer comparison with \citep{hawe2011dense, liu2015depth} in this section we use 
the peak signal-to-noise ratio (PSNR) as a measure of 
 reconstruction accuracy, where a higher PSNR indicates a better reconstruction. The PSNR is defined as follows, where $\xvar$ is the reconstruction, $\xtrue$ is the ground truth, and $n$ is the dimension of the vectorized \signal:
$$
\text{PSNR} = \frac 1 {n} \sum_{i=1}^{n} [\xvar_i-\xtrue_{i}]^2
$$

\input{figureTex/figureLiteratureComparison}

To ensure a fair comparison, the initial setup (e.g., memory allocation for matrices, building a constant wavelet/contourlet dictionary) has been excluded from timing. All algorithms are initiated without warm-start, meaning that the sample image (rather than the result from \naive) is used as the initial guess to our optimization problems. 
For \lminDiag, we use \NESTA as solver with the same settings specified in~\prettyref{sec:exp-solvers}. 
For \WTCT, we set 100 as the maximum number of iterations, which strikes the best trade-off between accuracy and timing.
%To gain a balance of runtime and accuracy, we set $\mu=0.001$ for the \lminDiag (same as before) and a max iteration of 100 (default is 400) is used for Liu's \WTCT~\citep{liu2015depth}. 

\prettyref{tab:middlebury} reports the results of our evaluation, for each image in the Middlebury dataset (rows in the table), 
and for increasing number of samples (columns in the table).
%, with the best accuracy and computational time highlighted in bold for each set of experiments. 
For each cell, we report the PSNR in dB and the time in seconds.
A cell is marked as N/A if the PSNR falls below 20dB \citep{hawe2011dense, liu2015depth}, which indicates that either the algorithm fails to converge or that the reconstructed image is significantly different than the ground truth. Best accuracy and best timing are highlighted in bold (recall that 
the higher the PSNR the better).

\input{experimentTex/experiments_table_middlebury}

Our proposed algorithm \lminDiag consistently outperforms all other algorithms in terms of accuracy in every single experiments. In addition, \lminDiag is the only algorithm that ensures acceptable performance at aggressively low sampling rates (as low as 0.5\%), while both \citep{hawe2011dense} and \citep{liu2015depth} fail with 1\% samples or fewer. \lminDiag is  significantly faster than both \citep{hawe2011dense} and \citep{liu2015depth}. For instance, \lminDiag takes only 50\% to 10\%  of the computational time of \WTCT, depending on the number of samples. The \naive interpolation is very fast, but produces worse reconstruction than \lminDiag. 
%\CSR~\citep{hawe2011dense}, and Liu's \WTCT
We noticed that in these tests we can achieve even faster runtime for \lminDiag by using a larger parameter $\mu_f$ without suffering much loss in accuracy. For instance, for $\mu_f=0.1$,  the average computation time with 5\% samples reduces from around 3s to around 2s, while the PSNR remains at roughly the same level and still outperforms other approaches. 

For a visual comparison, \prettyref{fig:middlebury} reports some examples of the 
 reconstructed disparity images for each of the compared techniques. The proposed algorithm, \lminDiag, is able to preserve sharp boundaries and fine details, while avoiding the creation of  jagged edges as in \naive.

%% file: figureTex/figure3DExample.tex
\begin{figure}[hbtp]
\begin{minipage}{\textwidth}
\newcommand{\figWidth}{ 0.325\linewidth } 
\def\arraystretch{3}		\setlength\tabcolsep{0.5mm}	\smaller
\begin{tabular}{ccc}
\begin{minipage}[b]{\figWidth}\centering
\includegraphics[width=\linewidth]{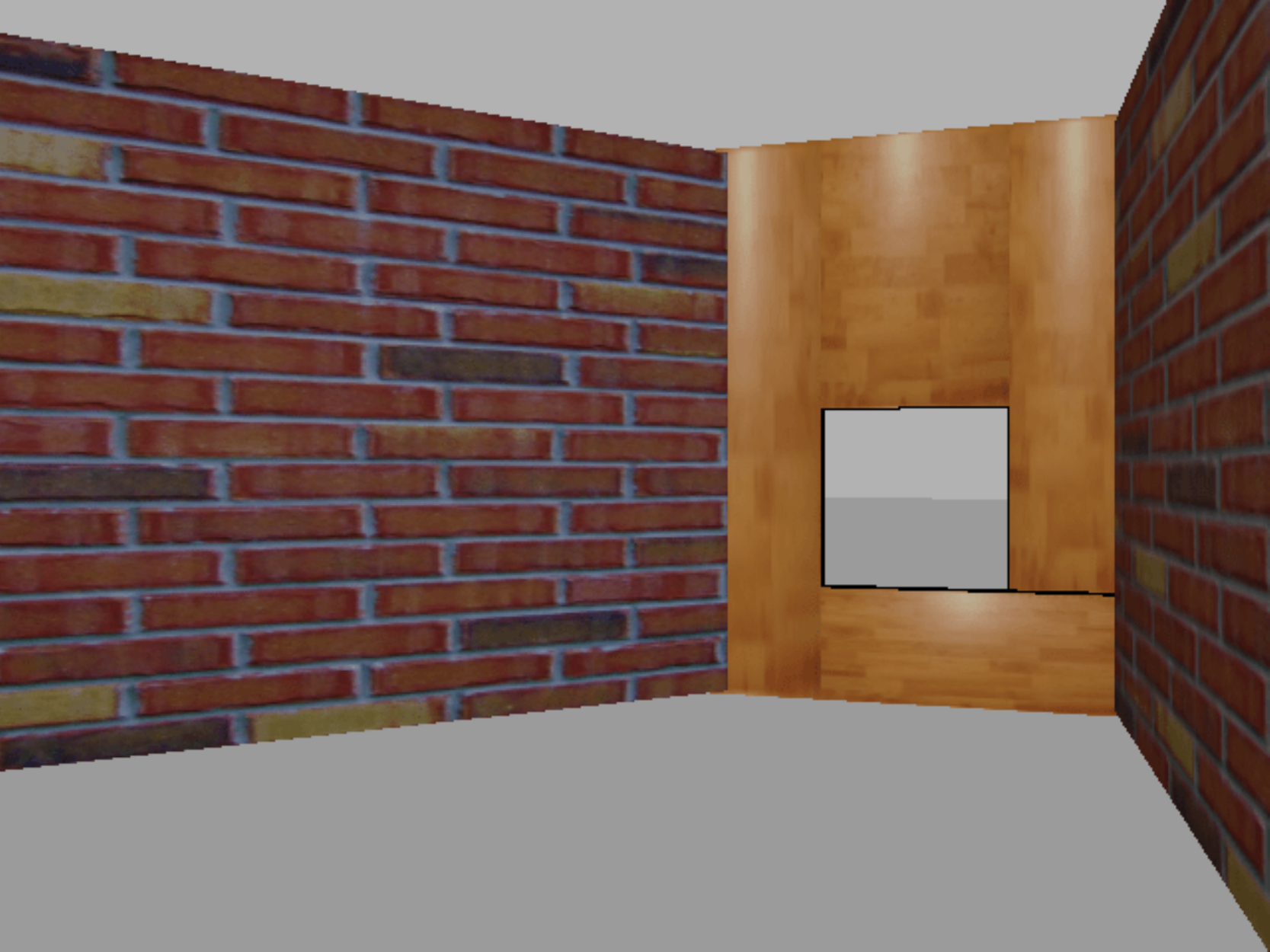} 
\\
(a) Gazebo RGB image
\end{minipage}
& 
\begin{minipage}[b]{\figWidth}\centering
\includegraphics[width=\linewidth]{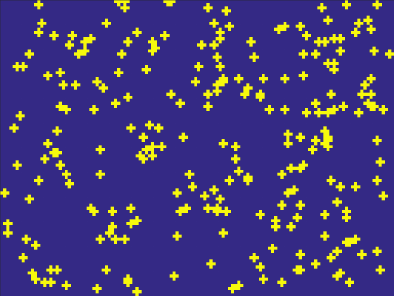}
\\
(b) random samples
\end{minipage}
&
\begin{minipage}[b]{\figWidth}\centering
\includegraphics[width=\linewidth]{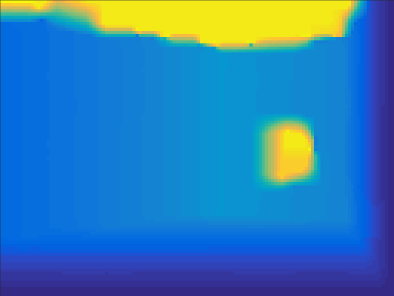}
\\
(c) reconstructed depth
\end{minipage}
\\ 
\begin{minipage}[b]{\figWidth}\centering
\includegraphics[width=\linewidth]{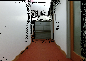} 
\\
(d) Kinect RGB image
\end{minipage}
&  
\begin{minipage}[b]{\figWidth}\centering
\includegraphics[width=\linewidth]{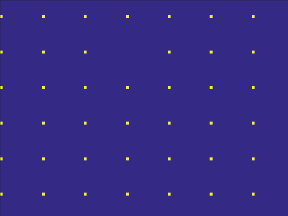}
\\
(e) grid samples
\end{minipage}
&
\begin{minipage}[b]{\figWidth}\centering
\includegraphics[width=\linewidth]{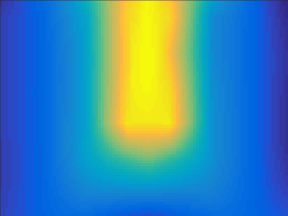}
\\
(f) reconstructed depth
\end{minipage}
\end{tabular}
\end{minipage}
\caption{
The first row is an example of sparse depth reconstruction on \gazebo{ }simulated data: (a) RGB image, (b) uniformly drawn sparse samples, and (c) reconstruction using \lminDiag.
The second row is an example on Kinect \kinect{1} data: (d) RGB image, (e) sparse samples on a grid, and (f) reconstruction using \lminDiag.}
\label{fig:3D_Example}
\end{figure}

%% file: figureTex/figure3DError.tex
%!TEX root = ../main.tex

\begin{figure}[hbtp]
\begin{minipage}{\textwidth}
\newcommand{\figWidth}{ 0.48\linewidth } 
\def\arraystretch{7}		\setlength\tabcolsep{1mm}	\smaller
\begin{tabular}{ l r }
\begin{minipage}{4cm}\centering\includegraphics[width=\linewidth]{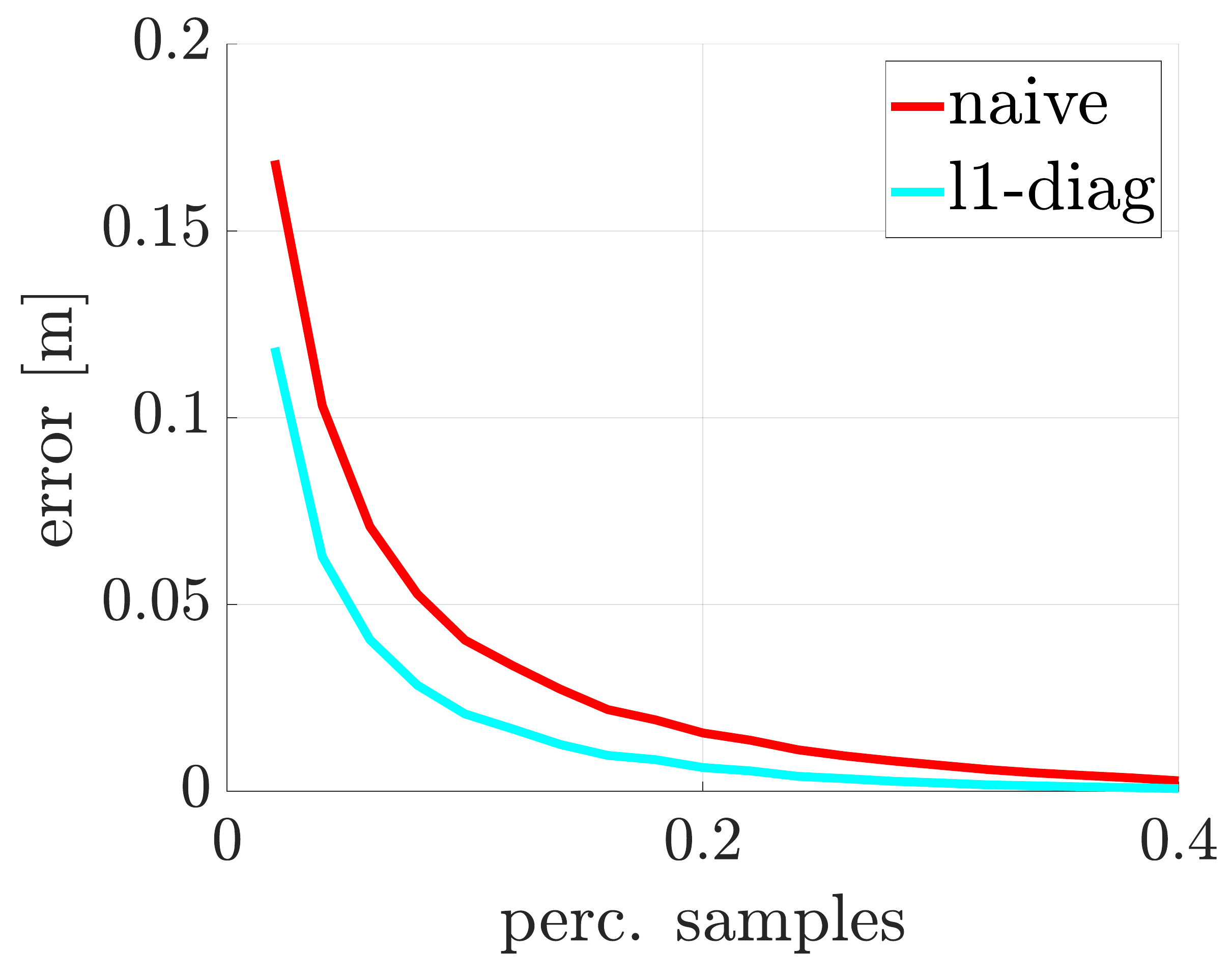}
 \\
(a) \gazebo
\end{minipage}
&  
\begin{minipage}{\figWidth}\centering\includegraphics[width=\linewidth]{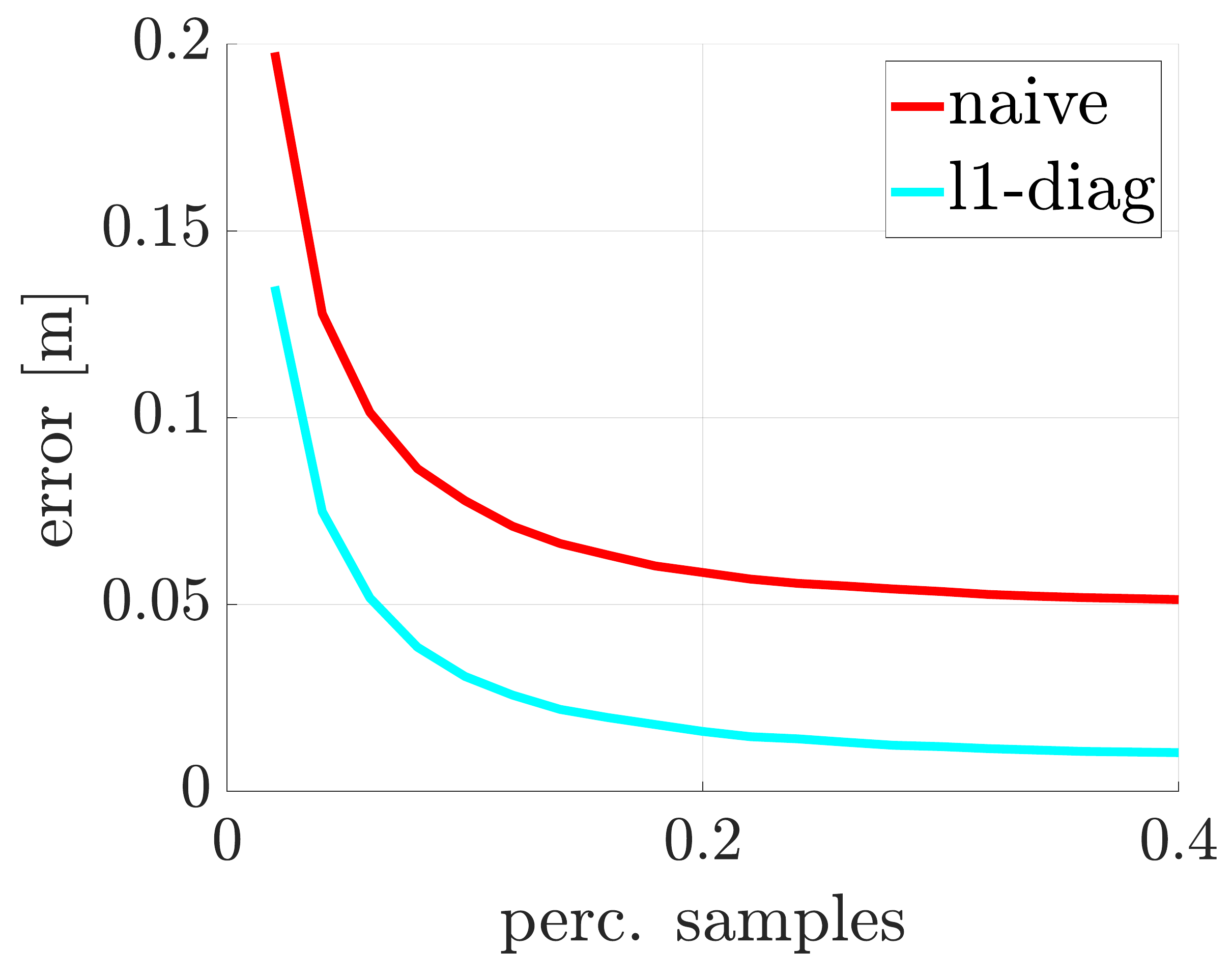}
\\
(b) \gazebo ($\epsilon=0.1$)
\end{minipage}
\\
\begin{minipage}{\figWidth}\centering\includegraphics[width=\linewidth]{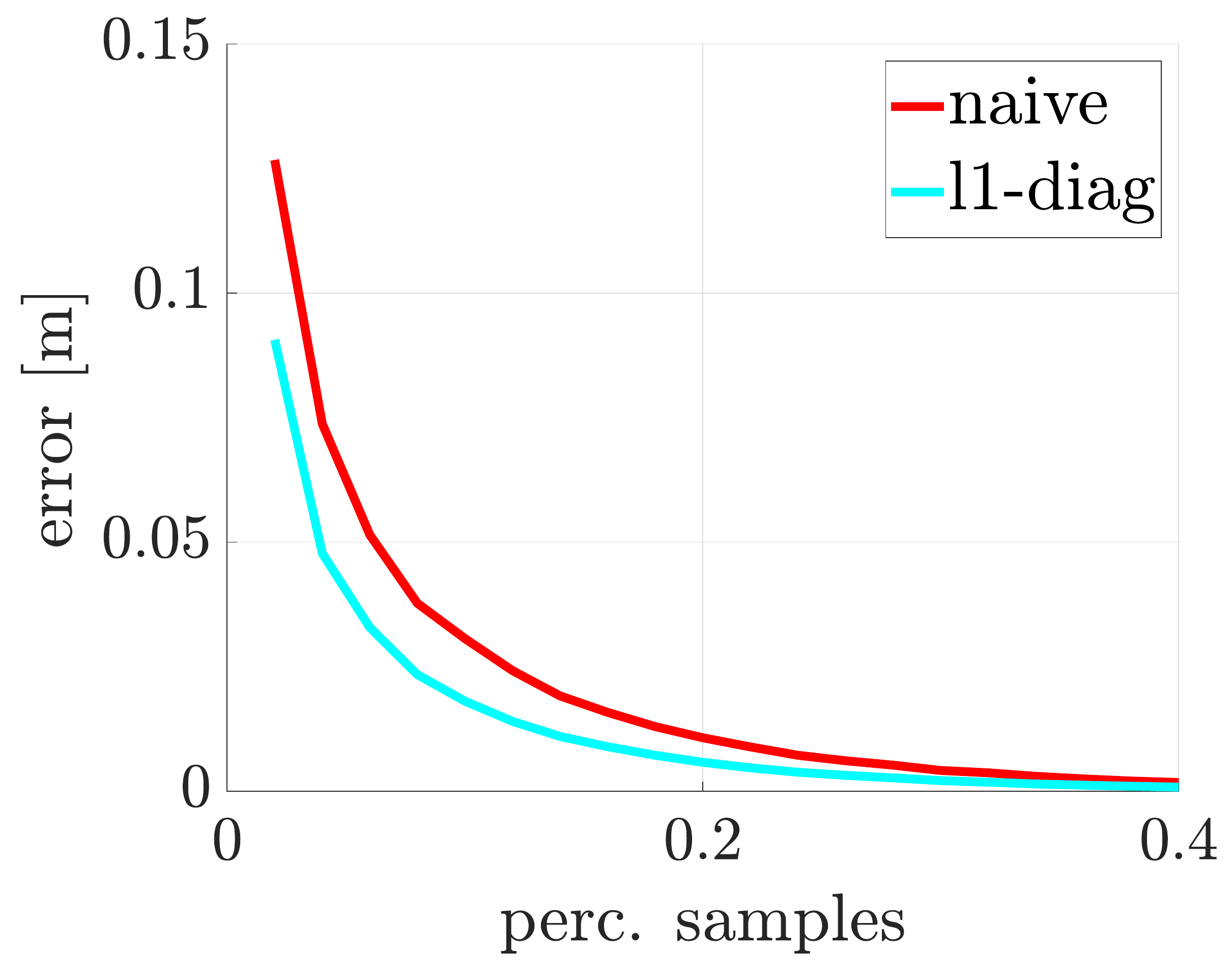}
 \\
(c) \zed
\end{minipage}
&  
\begin{minipage}{\figWidth}\centering\includegraphics[width=\linewidth]{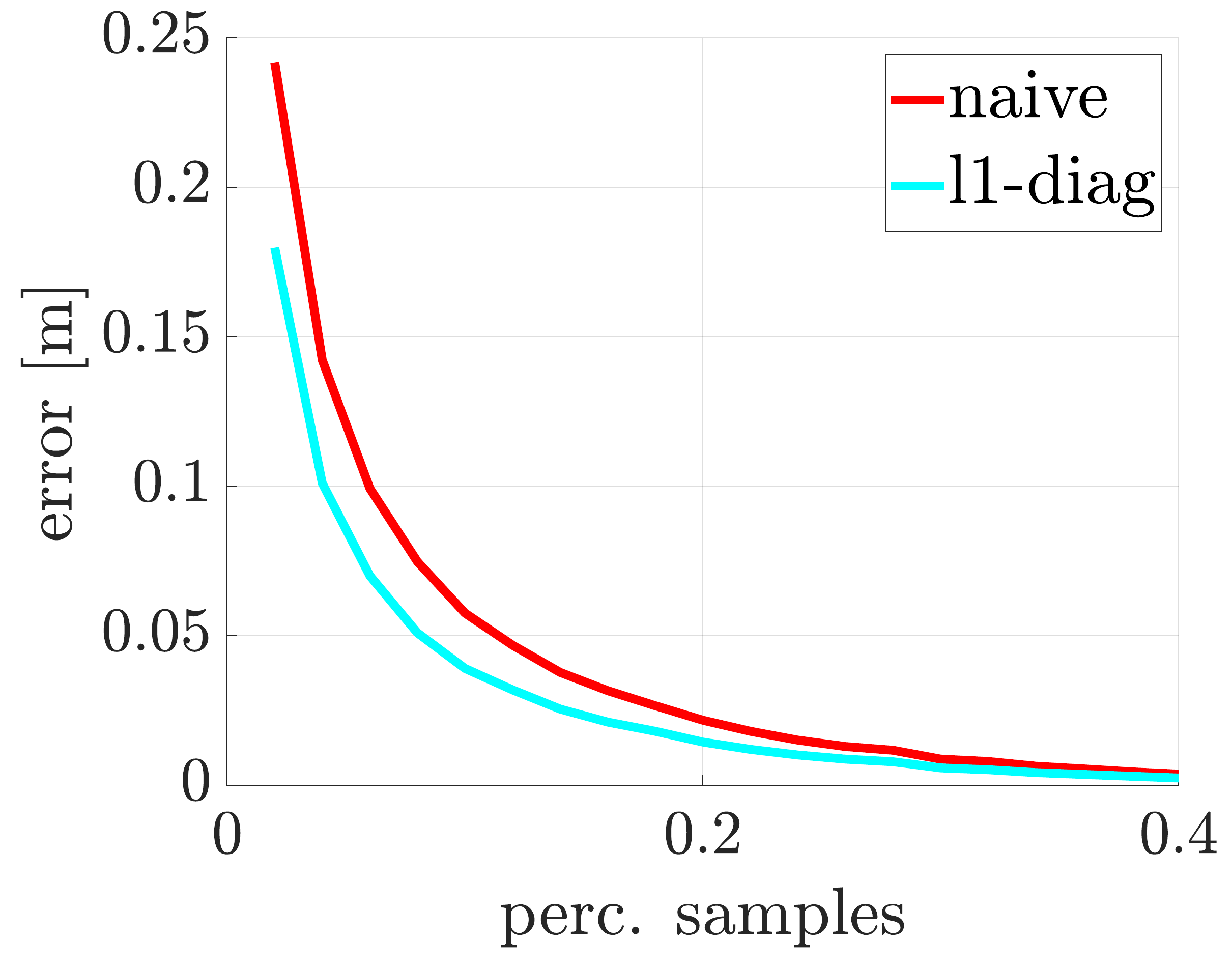}
 \\
(d) \kinect{1}
\end{minipage}
\end{tabular}

\begin{minipage}{\linewidth}
\centering
\includegraphics[width=0.9\linewidth]{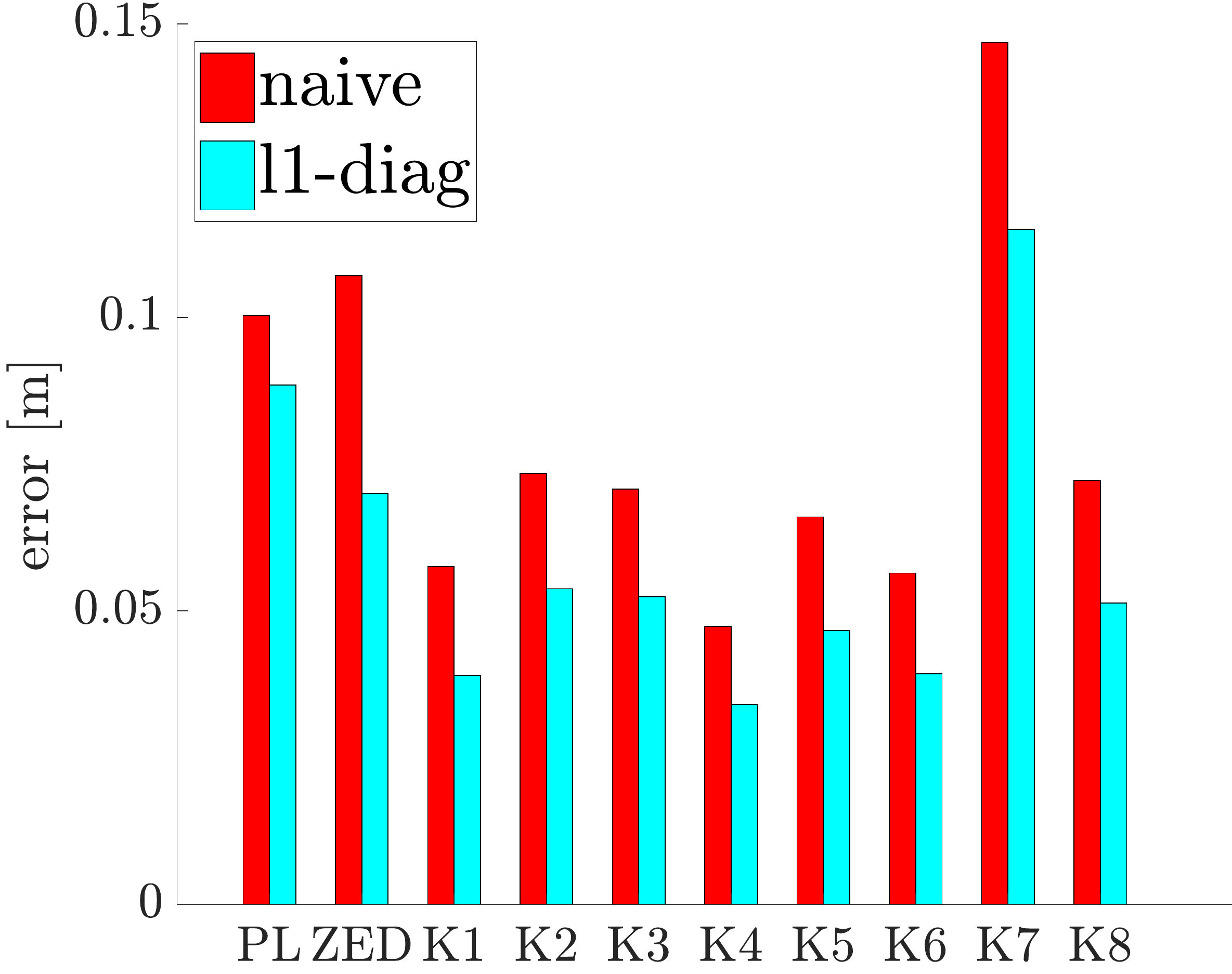}
 \\
(e) comparison on all datasets
\end{minipage}

\end{minipage}

\caption{Reconstruction errors for increasing percentage of uniform samples, and 
for different datasets. (a) and (b) are reconstructions on the \gazebo dataset, using noiseless and noisy (noise bounded by $\vareps = 0.1$m) samples, respectively. (c) reconstruction on the \zed dataset, (d) reconstruction on the \kinect{1} dataset. (e) comparison on all datasets.
}
\label{fig:Gazebo_ZED_percSamples}
\end{figure}

%% file: figureTex/figureLiteratureComparison.tex
%!TEX root = ../main.tex

\begin{figure*}[t]
\centering
\begin{minipage}{\textwidth}
\newcommand{\figWidth}{ 0.15\linewidth } 
\def\arraystretch{2}    \setlength\tabcolsep{0.3mm} \smaller
\begin{tabular}{ p{0.5cm} c c c c c c }
  %%%%%%%%%%%%%%%%%%%%%%%%%%%%%%%%%%%%%%%%%%%%%%%%%%%%%%%%%%%%%%%%%%%%%%
  \rotatebox[origin=c]{90}{Aloe \hspace{-2.5cm}}
  & 
  \begin{minipage}[b]{\figWidth}\centering
  \includegraphics[width=\linewidth]{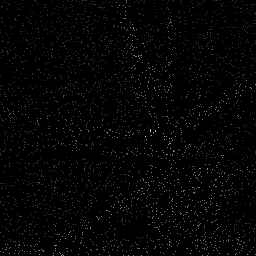} 
  \end{minipage}
  & 
  \begin{minipage}[b]{\figWidth}\centering
  \includegraphics[width=\linewidth]{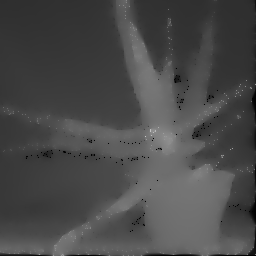} 
  \end{minipage}
  & 
  \begin{minipage}[b]{\figWidth}\centering
  \includegraphics[width=\linewidth]{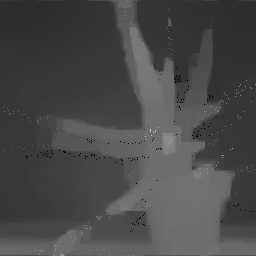} 
  \end{minipage}
  & 
  \begin{minipage}[b]{\figWidth}\centering
  \includegraphics[width=\linewidth]{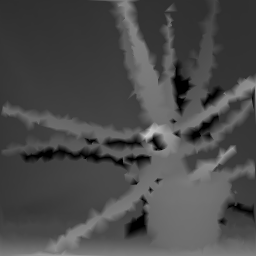} 
  \end{minipage}
  & 
  \begin{minipage}[b]{\figWidth}\centering
  \includegraphics[width=\linewidth]{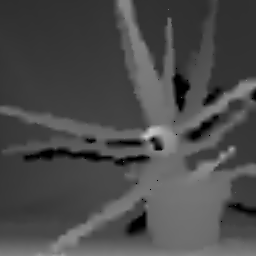} 
  \end{minipage}
  & 
  \begin{minipage}[b]{\figWidth}\centering
  \includegraphics[width=\linewidth]{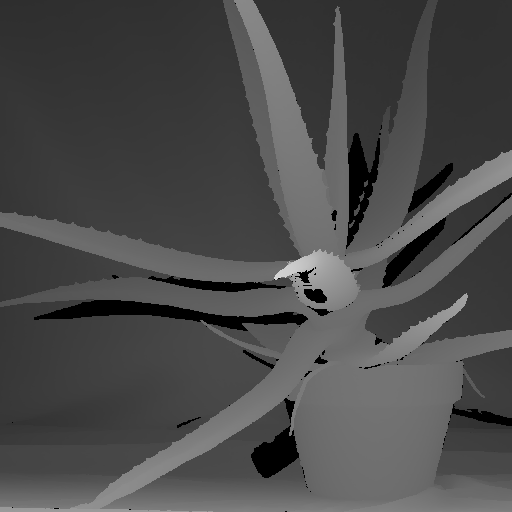} 
  \end{minipage}
  \\ 
  %%%%%%%%%%%%%%%%%%%%%%%%%%%%%%%%%%%%%%%%%%%%%%%%%%%%%%%%%%%%%%%%%%%%%%
  \rotatebox[origin=c]{90}{Baby \hspace{-2.5cm}}
  &
  \begin{minipage}[b]{\figWidth}\centering
  \includegraphics[width=\linewidth]{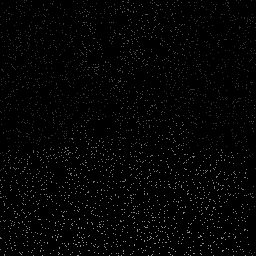} 
  \end{minipage}
  & 
  \begin{minipage}[b]{\figWidth}\centering
  \includegraphics[width=\linewidth]{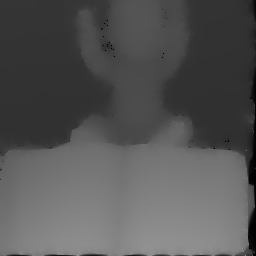} 
  \end{minipage}
  & 
  \begin{minipage}[b]{\figWidth}\centering
  \includegraphics[width=\linewidth]{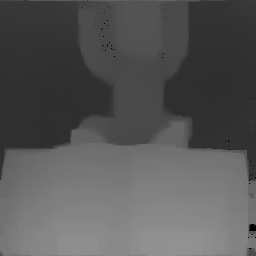} 
  \end{minipage}
  & 
  \begin{minipage}[b]{\figWidth}\centering
  \includegraphics[width=\linewidth]{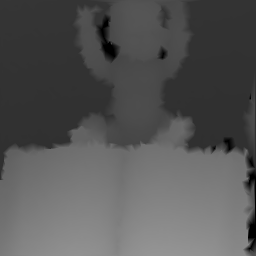} 
  \end{minipage}
  & 
  \begin{minipage}[b]{\figWidth}\centering
  \includegraphics[width=\linewidth]{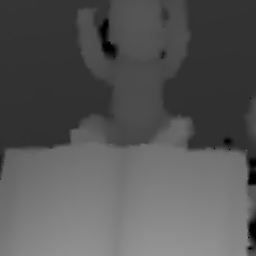} 
  \end{minipage}
  & 
  \begin{minipage}[b]{\figWidth}\centering
  \includegraphics[width=\linewidth]{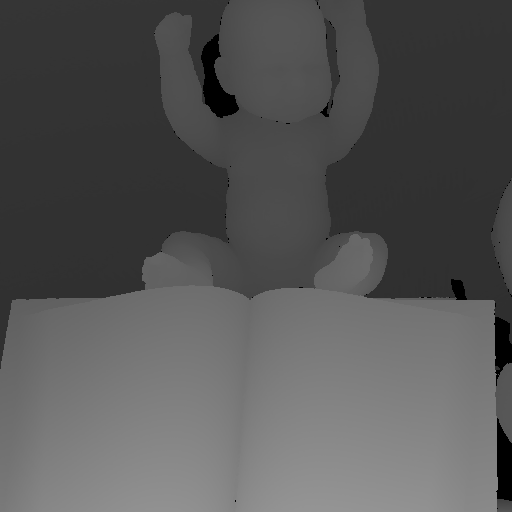} 
  \end{minipage}
  \\ 
  %%%%%%%%%%%%%%%%%%%%%%%%%%%%%%%%%%%%%%%%%%%%%%%%%%%%%%%%%%%%%%%%%%%%%%
  \rotatebox[origin=c]{90}{Art \hspace{-2.5cm}}
  &
  \begin{minipage}[b]{\figWidth}\centering
  \includegraphics[width=\linewidth]{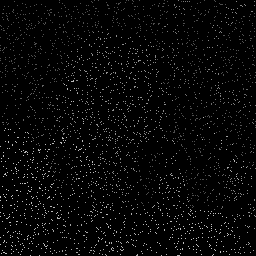} \\
  5\% uniform samples
  \end{minipage}
  & 
  \begin{minipage}[b]{\figWidth}\centering
  \includegraphics[width=\linewidth]{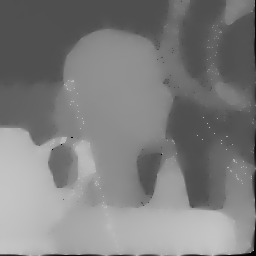} \\
  \CSR
  \end{minipage}
  & 
  \begin{minipage}[b]{\figWidth}\centering
  \includegraphics[width=\linewidth]{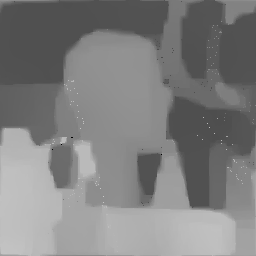} \\
  \WTCT
  \end{minipage}
  & 
  \begin{minipage}[b]{\figWidth}\centering
  \includegraphics[width=\linewidth]{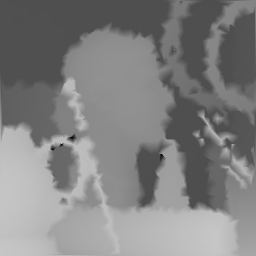} \\
  \naive
  \end{minipage}
  & 
  \begin{minipage}[b]{\figWidth}\centering
  \includegraphics[width=\linewidth]{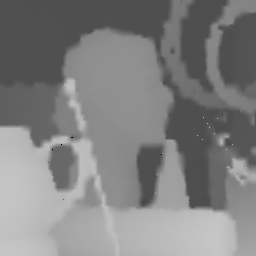} \\
  \lminDiag
  \end{minipage}
  & 
  \begin{minipage}[b]{\figWidth}\centering
  \includegraphics[width=\linewidth]{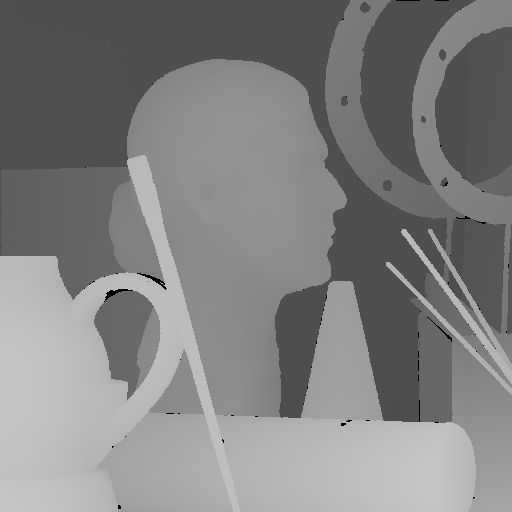} \\
  ground truth
  \end{minipage}
  \\ 
\end{tabular}
\end{minipage}
\caption{Examples of reconstruction from 5\% uniformly random samples on the Middlebury disparity dataset, using 4 different algorithms:
 \naive, \lminDiag,  \CSR~\citep{hawe2011dense}, and \WTCT~\citep{liu2015depth}. The proposed algorithm, \lminDiag, is able to preserve sharper boundaries and finer details, while not creating jagged edges as in \naive.\label{fig:middlebury}}
\end{figure*}

%% file: experimentTex/experiments_table_middlebury.tex
%!TEX root = ../main.tex

\begin{table}[h]
  \centering
  \footnotesize
  \newcommand{\sep}{~/~}
  \setlength\tabcolsep{3pt} % default value: 6pt
  \begin{tabular}{ c | c | *{4}{c}  }
    \hline
    \multirow{3}{*}{\shortstack[l]{Name}} &\multirow{3}{*}{Method}&\multicolumn{4}{c}{PSNR (dB) / Time (s) }  \\ % \cline{3-6}
     && \multicolumn{4}{c}{(Percentage of Samples)} \\
     &&0.5\%&1\%&5\%&10\% \\ %&20\% \\
    \hline \hline
    \multirow{4}{*}{Aloe} 
    &\CSR        &N/A&N/A&21.4\sep10.6&24.1\sep8.65 \\ %&27.5\sep7.05 \\
    &\WTCT         &N/A&N/A&21.9\sep19.4&24.3\sep19.5 \\ %&27.7\sep19.5 \\
    &\naive      &N/A 
                   &21.6\sep\textbf{0.17} 
                   &24.7\sep\textbf{0.14} 
                   &26.0\sep\textbf{0.22} 
                    \\ %&27.8\sep\textbf{0.22} \\
    &\lminDiag{ }&\textbf{20.6}\sep\textbf{14.5}   
                   &\textbf{21.7}\sep7.02 
                   &\textbf{24.9}\sep3.12 
                   &\textbf{26.4}\sep2.06 
                    \\ %&\textbf{28.4}\sep2.01 \\
    \hline
    \multirow{4}{*}{Art} 
    &\CSR        &N/A&N/A&23.1\sep11.9&25.3\sep9.80 \\ %&28.3\sep7.62 \\
    &\WTCT         &N/A&N/A&25.0\sep19.8&26.7\sep19.5 \\ %&28.4\sep19.5 \\
    &\naive      &21.9\sep\textbf{0.15} 
                   &23.5\sep\textbf{0.16} 
                   &26.3\sep\textbf{0.17} 
                   &27.7\sep\textbf{0.18} 
                    \\ %&29.0\sep\textbf{0.22} \\
    &\lminDiag{ }&\textbf{22.5}\sep11.1  
                   &\textbf{23.8}\sep8.86 
                   &\textbf{26.6}\sep3.78 
                   &\textbf{27.8}\sep2.23 
                    \\ %&\textbf{29.4}\sep1.92 \\
    \hline
    \multirow{4}{*}{Baby} 
    &\CSR        &N/A&N/A&26.6\sep10.0&31.1\sep9.11 \\ %&34.7\sep8.16 \\
    &\WTCT         &N/A&24.1\sep19.6&27.7\sep19.4&31.5\sep19.5 \\ %&\textbf{36.3}\sep19.7 \\
    &\naive      &27.6\sep\textbf{0.15} 
                   &27.4\sep\textbf{0.16} 
                   &31.3\sep\textbf{0.16} 
                   &33.3\sep\textbf{0.18} 
                    \\ %&35.2\sep\textbf{0.23} \\
    &\lminDiag{ }&\textbf{27.8}\sep12.1  
                   &\textbf{28.4}\sep10.5 
                   &\textbf{32.5}\sep3.21 
                   &\textbf{33.9}\sep2.06 
                    \\ %&36.1\sep1.49 \\
    \hline
    \multirow{4}{*}{Dolls} 
    &\CSR        &N/A&N/A&24.3\sep13.2&26.5\sep11.0 \\ %&28.8\sep10.2  \\
    &\WTCT         &N/A&20.6\sep19.5&27.5\sep19.6&28.2\sep20.3 \\ %&29.5\sep19.6  \\
    &\naive      &25.8\sep\textbf{0.13} 
                   &24.5\sep\textbf{0.16} 
                   &27.8\sep\textbf{0.16} 
                   &28.5\sep\textbf{0.18} 
                    \\ %&29.5\sep\textbf{0.23}  \\
    &\lminDiag{ }&\textbf{26.9}\sep7.07  
                   &\textbf{27.5}\sep5.49 
                   &\textbf{28.3}\sep2.24 
                   &\textbf{28.9}\sep3.03 
                    \\ %&\textbf{30.0}\sep1.31  \\
    \hline
    \multirow{4}{*}{Moebius} 
    &\CSR        &N/A&N/A&23.6\sep11.9&26.1\sep10.5 \\ %&27.4\sep9.34 \\
    &\WTCT         &N/A&22.4\sep19.3&26.3\sep19.5&27.6\sep19.4 \\ %&28.6\sep19.4 \\
    &\naive      &25.7\sep\textbf{0.14} 
                   &24.7\sep\textbf{0.16} 
                   &26.8\sep\textbf{0.15} 
                   &27.8\sep\textbf{0.18} 
                    \\ %&28.6\sep\textbf{0.23}  \\
    &\lminDiag{ }&\textbf{25.8}\sep6.91  
                   &\textbf{26.4}\sep7.03 
                   &\textbf{27.5}\sep2.90 
                   &\textbf{28.6}\sep2.59 
                    \\ %&\textbf{29.4}\sep2.02  \\
    \hline
    \multirow{4}{*}{Rocks} 
    &\CSR        &N/A&N/A&23.1\sep11.5&25.0\sep9.15 \\ %&27.6\sep8.08  \\
    &\WTCT         &N/A&N/A&23.2\sep19.3&25.6\sep19.2 \\ %&28.1\sep19.4  \\
    &\naive      &21.7\sep\textbf{0.15} 
                   &23.8\sep\textbf{0.15} 
                   &25.8\sep\textbf{0.15} 
                   &27.2\sep\textbf{0.19} 
                    \\ %&28.7\sep\textbf{0.23}  \\
    &\lminDiag{ }&\textbf{22.7}\sep12.0  
                   &\textbf{24.3}\sep9.71 
                   &\textbf{25.9}\sep3.22 
                   &\textbf{27.3}\sep2.68 
                    \\ %&\textbf{29.1}\sep2.00  \\
    \hline
  \end{tabular}
  \caption{Reconstruction accuracy and computational time comparing \naive, \lminDiag, 
  \CSR~\citep{hawe2011dense}, and \WTCT~\citep{liu2015depth}. \lminDiag consistently outperforms all other methods in accuracy, and performs robustly even with aggressively low number of measurements.}
  \label{tab:middlebury}
\end{table}

%% file: experimentTex/experiments_3dSparseReconstruction_multi.tex
%!TEX root = ../main.tex
%%%%%%%%%%%%%%%%%%%%%%%%%%%%%%%%%%%%%%%%%%%%%%%%%%%%%%%%%%%%
\subsection{Multi-Frame Sparse \thd Reconstruction} 
\label{sec:exp-multi-frame-sparse-reconstruction}

In the previous Section~\ref{sec:exp-single-frame-sparse-reconstruction} we focused on depth reconstruction from sparse measurements 
of a subset of pixels in a single frame. However, when odometry information is available (e.g., from a wheel odometer on ground vehicles, or from inertial measurement units on aerial robots), it is possible to combine sparse measurements across multiple consecutive frames in a time window in order to improve the reconstruction. 
More precisely, at every frame $t$, we use the samples collected at frames $t-H, t-H+1, \ldots, t$ (where $H$ is a given horizon) to improve the quality of the \thd reconstruction. Since each depth sample collected at time $t' < t$ can be associated to a \thd point 
in the reference frame of the sensor at time $t'$, we use the relative pose between $t'$ and $t$ to express the \thd point in the 
reference frame at time $t$, hence obtaining an extra measurement at time $t$.
% for every past frame $t$ in a given time window, we computate its associated sensor position (and pose) relative to the current frame by accumulating the odometry data. With this information, we project the set of measurements acquired at every past frames to the current frame.
 This way, we accumulate all measurements from the past frames in the time window, and we leverage this larger set of measurements 
 to improve the  depth reconstruction at time $t$. 
 %We reconstruct the dense depth image based on this set of measurements. 
 Note that we assume the environment is static and the odometry is accurate within a short time window. 
 To some extent, we can model odometric errors by associating larger noise levels $\vareps$ to samples acquired at older frames.

In this section, we demonstrate this idea on the \kinect{1}-\kinect{8} datasets, where the odometry information is available and can be considered reliable over a short temporal window.

%%%%%%%%%%%%%%%%%%%%%%%%%%%%%%%%%%%%%%%%%%%%%%%%%%
\input{figureTex/figureTemporalExample}

\subsubsection{Typical Examples of Multi-Frame \thd Reconstruction} 

\prettyref{fig:exampleTemporal} shows an example of multi-frame reconstruction. \prettyref{fig:exampleTemporal}(a) is an RGB image (registered with the depth image, and thus missing some pixels for which the depth is not available). \prettyref{fig:exampleTemporal}(b) shows the depth measurements collected over a temporal window of 10 frames. 
% With odometry information, measurements collected at different time frames (and thus different locations and orientations) are projected to the last frame. 
\prettyref{fig:exampleTemporal}(c) shows the reconstructed depth \signal from all the measurements collected in a receding time horizon. When compared against the single-frame counterpart in \prettyref{fig:3D_Example}(e), it can be observed that the additional measurements 
contribute to making the depth \signal sharper and more accurate. The reconstruction error with data from 10 frames is 26cm, as opposed to 39cm when using only one single frame.
 % \LC{(reconstruction error of XX, while the other YY).}
%the reconstructed depth image has sharper edges.

%%%%%%%%%%%%%%%%%%%%%%%%%%%%%%%%%%%%%%%%%%%%%%%%%%
\subsubsection{Statistics for Multi-Frame \thd Reconstruction} 

\input{figureTex/figureTemporalStatistics}

\prettyref{fig:temporal_statistics} reports the results of the comparison between the 
% reports the results of a Monte Carlo analysis 
% aimed at evaluating the
 performance of the multi-frame reconstruction (labeled as \lminDiag, with $\vareps=0$) and the baseline \naive (linear interpolation). 
 % \LC{\lminDiag is noisY?}
 In both approaches, we use all samples collected over a receding horizon as input to the reconstruction.
The results in the figure are obtained on the 
\kinect{1} dataset, and each data point in the plot is averaged over 910 images.
% 43 \times 52 = 
From each frame only 18 depth measurements are collected (the full image has $2236$ pixels), and the samples fall on a regular grid. 

\prettyref{fig:temporal_statistics}(a) reports the estimation errors for \naive and \lminDiag for increasing time windows. It can be observed that \lminDiag consistently outperforms \naive.
%  for different temporal window sizes. 
The best performance is achieved for a temporal window of 15 frames (approximately 2 seconds), with an average reconstruction error of $25\text{cm}$, while the average error for the single-frame reconstruction (i.e., 1-frame window) is $39\text{cm}$, which is 56\% higher.
% \LC{The performance boost of \lminDiag  is connected to the fact that this technique can correctly model the measurement noise 
% (described by the parameter $\vareps$), hence being able to compensate for small odometry noise which affects samples from past frames.} \FM{(here we are using the exact solver. the noisy version produces very curvy reconstruction.)}
The error curve of \prettyref{fig:temporal_statistics}(a)  has a minimum (15-frame horizon), and then starts increasing for longer horizons.
%Note that both error curves are convex, in the sense that the lowest error is achieved at window size of 15 and the error rises again with increasing length of temporal horizon. 
This phenomenon can be attributed to the odometry drift (i.e., the accumulation of odometry errors) over time. 
The odometric error cannot be considered negligible over a long time horizon, hence inducing 
larger noise in the samples and degraded reconstruction performance.
%More specifically, the drift leads to incorrect point correspondence between frames, and thus point measurements in the past frames are projected to wrong pixel locations, resulting in overall performance degradation.

\prettyref{fig:temporal_statistics}(b) shows the computational time for our algorithm \lminDiag using \NESTA. The runtime decreases  with the length of the temporal horizon. This is due to reduction of the search space in the optimization problem, thanks to additional constraints 
induced by the measurements at the past frames.

%% file: figureTex/figureTemporalExample.tex
%!TEX root = ../main.tex

\begin{figure}[H]
\begin{minipage}{\textwidth}
\newcommand{\figWidth}{ 0.325\linewidth } 
\def\arraystretch{3}		\setlength\tabcolsep{0.5mm}	\smaller
\begin{tabular}{ccc}\begin{minipage}[t]{\figWidth}\centering
\includegraphics[width=\linewidth]{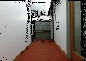} 
\\
(a) environment 
\end{minipage}
& 
\begin{minipage}[t]{\figWidth}\centering
\includegraphics[width=\linewidth]{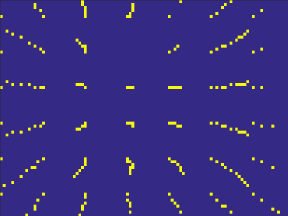}
\\
(b) samples collected from
10 frames
\end{minipage}
&
\begin{minipage}[t]{\figWidth}\centering
\includegraphics[width=\linewidth]{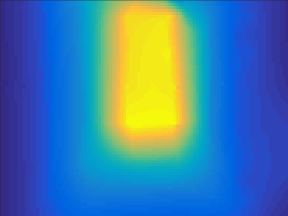}
\\
(c) reconstructed depth
\end{minipage}

\end{tabular}
\end{minipage}

\caption{
(a)~registered RGB image from the Kinect sensor,
(b)~depth samples collected across multiple time frames,
(c)~depth reconstruction using samples collected over a receding time horizon.
With more samples collected from several frames, the temporal reconstruction becomes more accurate (e.g., sharper at edges) 
compared to the reconstruction using samples from a single frame.
}
\label{fig:exampleTemporal}
\end{figure}

%% file: figureTex/figureTemporalStatistics.tex
%!TEX root = ../main.tex

\begin{figure}[htb]
\begin{minipage}{\textwidth}
\newcommand{\figWidth}{ 0.48\linewidth } 
\def\arraystretch{1}		\setlength\tabcolsep{1mm}	\smaller
\begin{tabular}{ l  r }\begin{minipage}[b]{\figWidth}\centering
\includegraphics[width=\linewidth]{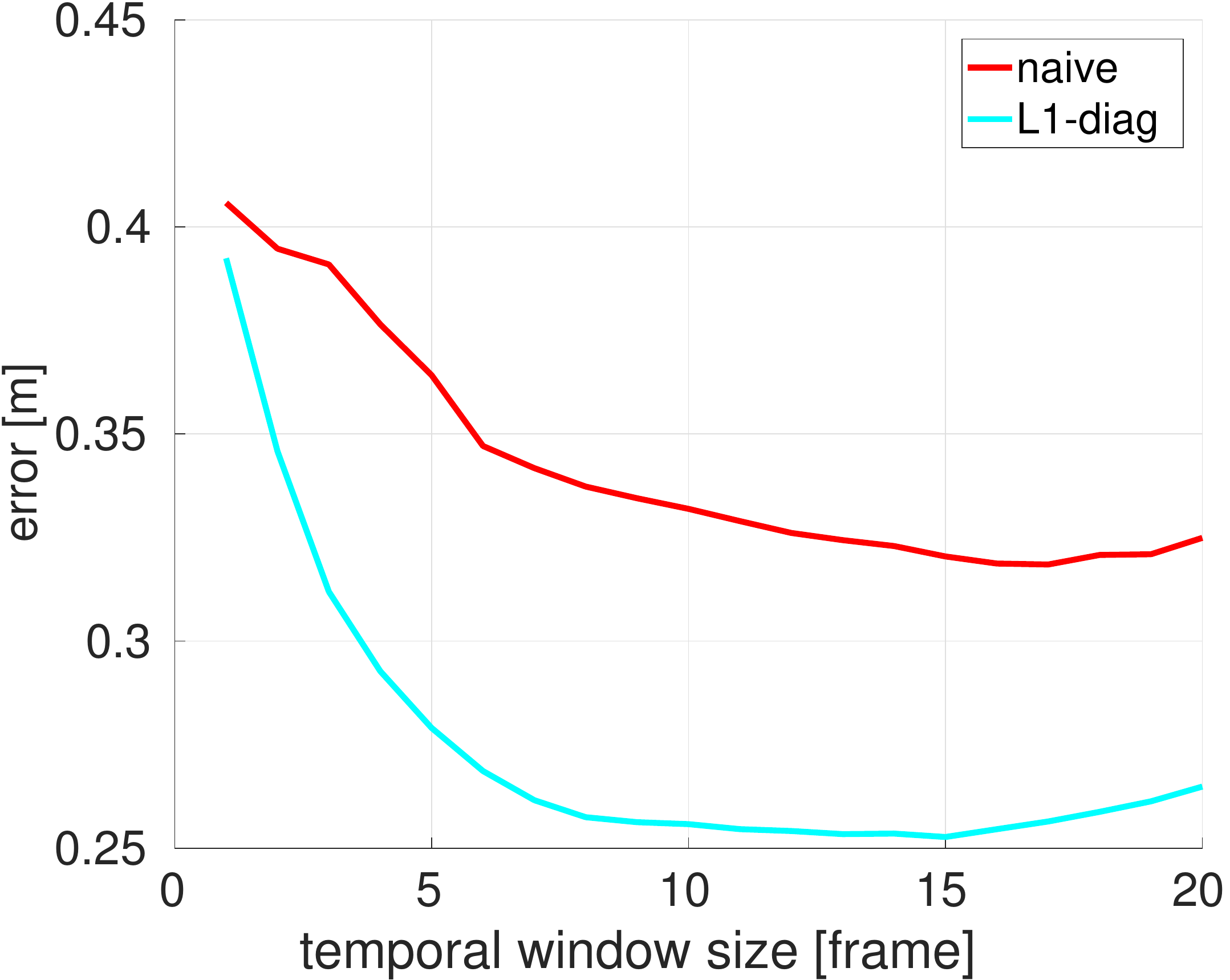} 
\\
(a) \end{minipage}
& 
\begin{minipage}[b]{\figWidth}\centering
\includegraphics[width=\linewidth]{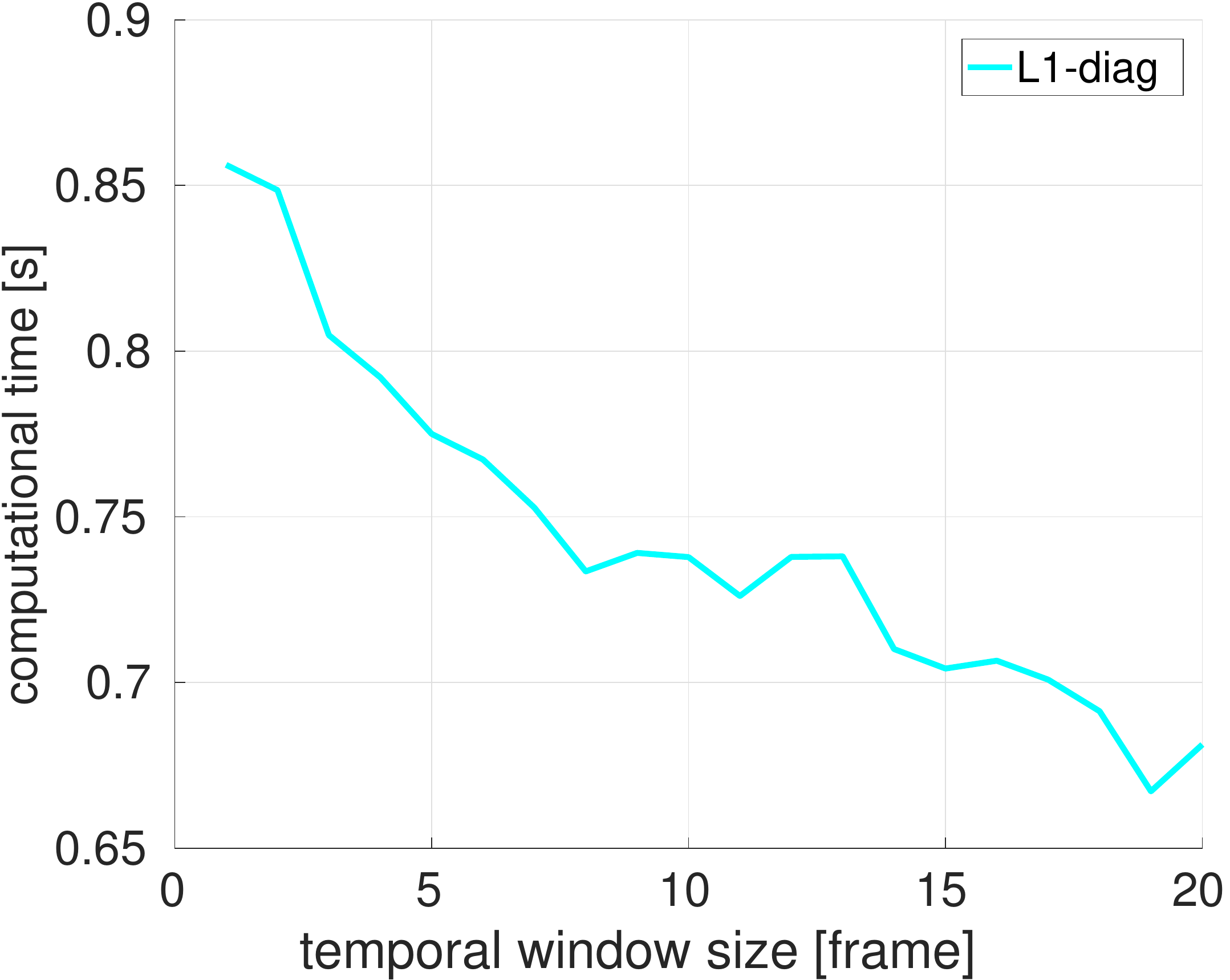}
\\
(b) \end{minipage}
\end{tabular}
\end{minipage}\caption{
The two figures show error and timing statistics against increasing horizon size for temporal reconstruction. 
With more samples collected, the accuracy increases and the computational decreases (due to more constraints in the optimization problem).
However, when the horizon size is very large, the error rises due to accumulated odometry error.
}
\label{fig:temporal_statistics} 
\end{figure}

%% file: experimentTex/experiments_dataCompression.tex
%!TEX root = ../main.tex

\subsection{Data Compression of \thd Profiles}
\label{sec:exp_compression}

Another major application of the proposed algorithms lies in bandwidth-limited robot-server communication. 
Instead of having to transmit the entire depth \signal from the robot to a remote server, the basic idea is that 
the robot can transmit a subset of the pixels and the server can then use the reconstruction algorithms 
discussed in this paper to retrieve the full \signal. 
A major difference with respect to the setup discussed in the previous sections, is that 
 in compression problems, the robot has access to the full \signal, hence it can 
use a more clever sampling strategy and improve the reconstruction results.
For instance, the robot can sample the edges (and theirs neighbors) in the depth \signal, 
which, according to \prettyref{prop:nam2D}, are sufficient to  
%ensures that using only the depth along the edges (and theirs neighbors) one can 
reconstruct the original \thd depth \signal exactly.

In this section, we show empirically that by sending only the depth data along the edges (extracted either from the RGB 
images or the depth \signals), we can significantly reduce the required communication bandwidth, at a minor loss of accuracy. 
We measure the amount of data compression using the \emph{data rate saving}, defined as
\beq
\label{eq:dataSaving}
\text{data rate saving} = 1 - \frac {\text{compressed data rate}} {\text{uncompressed data rate}}.
\eeq

We demonstrate the compression technique on both the \gazebo and the \zed datasets. We separate the discussions regarding edge extraction from RGB images and from the depth \signals. This is due to the fact that they have different pros and cons, and thus can be applied in different scenarios. 

%%%%%%%%%%%%%%%%%%%%%%%%%%%%%%%%%%%%%%%%%%%%%%%%%%%%%%%%%%%%%%%%
\subsubsection{Sampling Depth Edges}

\input{figureTex/figureEdgeDepthExample}

\prettyref{fig:EdgeDepthExample} presents an example of reconstruction based solely on samples along the depth edges (and their neighbors).
%, as defined in \prettyref{prop:edgeSet}. 
In this case, the robot only transmits the neighborhood of the pixels for which the depth curvature (along the horizontal or vertical directions) is larger than a given threshold; a large threshold implies that less pixels are transmitted as edges, at a potential loss of 
reconstruction accuracy. We use the Canny edge detector~\citep{Canny86acomputational} (implemented in Matlab, with  
default parameters) to extract the edges from the depth image.
%For our numerical tests we consider a pixel to belong to a depth edge if its 
%which correspond to non-trivial pixels in the second-order derivative of the depth images. 
We compare the reconstructed depth \signal (at the server) with the full \signal from the ZED stereo camera, and show the statistics in \prettyref{fig:EdgeDepthStatistics}. 

\prettyref{fig:EdgeDepthStatistics}(a) compares the reconstruction errors of the \naive linear interpolation and the proposed \lminDiag approach with respect to the full ZED depth \signal.
\lminDiag achieves almost half of the reconstruction error of \naive. In addition, the error is very small (in the order of few centimeters), implying almost exact recovery from samples along depth edges. On the other hand, \prettyref{fig:EdgeDepthStatistics}(b) shows that 
the data rate saving, defined in~\eqref{eq:dataSaving}, is around 70-85\%. This implies that the same bandwidth is now able to accommodate 
3x-6x more communication channels.

\input{figureTex/figureEdgeDepthStatistics}

Note that the full depth \signal from stereo cameras suffers from distortion and is error-prone in regions with small intensity gradients, resulting in many spurious edges in the original ZED image. For instance, in \prettyref{fig:EdgeDepthExample}(b), 
 unnecessary edges appear on the (flat) ground as well as on the walls. These unwanted edges 
 %will be treated as ``important'' information and thus 
 will be sent to the server,
 %along the communication channel, 
 thus being the result of  stereo reconstruction errors 
 rather than actual depth discontinuities, hence 
preventing further data compression. This motivates us to consider samples along the RGB (instead of the depth) edges, 
as discussed below.

%%%%%%%%%%%%%%%%%%%%%%%%%%%%%%%%%%%%%%%%%%%%%%%%%%%%%%%%%%%%%%%%
\subsubsection{Sampling RGB Edges}

\input{figureTex/figureEdgeRGBExample}

In this section, we discuss the advantage of using depth sampled along the RGB edges  
(pixels with large image intensity gradients), 
 %of edge extraction from the RGB images 
 over the depth edges used in the previous section. 
% \LC{RGB edges can be computed using standard computer vision methods, as the Canny algorithm ref?}.
Similar to the depth edge extraction, we use the Canny algorithm, this time applied to the RBG image.
 In most scenarios, depth discontinuities are reflected in appearance discontinuities. This implies that the RGB edges are usually  a superset of the real edges in the scene. By extracting RGB edges, our goal is to avoid unnecessary and erroneous edges as seen in \prettyref{fig:EdgeDepthExample}(b), and thus improve accuracy and data rate saving.
 We remark that these considerations are due the fact that the full depth \signal (collected by the robot) is noisy,  
 making tricky to distinguish actual depth discontinuities from pixel noise.
 %hence causing the presence of edges that do not 

\input{figureTex/figureEdgeRGBStatistics}

\prettyref{fig:EdgeRGBExample} shows the same example as in \prettyref{fig:EdgeDepthExample}, but with edges extracted from the RGB image. More specifically, \prettyref{fig:EdgeRGBExample}(b) shows the depth measurements along the RGB edges extracted from \prettyref{fig:EdgeRGBExample}(a), and \prettyref{fig:EdgeRGBExample}(c) is the reconstruction result. We observe a smaller and cleaner set of edges. 
%\LC{data saving, error compared to other example?}

The reconstruction error and data rate saving statistics, for the case in which only RBG edges are transmitted to the server, are shown in \prettyref{fig:EdgeRGBStatistics}. Note that \lminDiag still consistently outperforms the baseline \naive on every single datasets, although they both perform poorly on the ZED stereo datasets. As discussed before, the disparity images from the ZED stereo camera suffer from distortion (see \prettyref{fig:EdgeDepthExample}(b) for example). In other words, the ground truth depth itself is highly noisy. Therefore, even if the reconstructed depth may match more accurately the actual geometry of the \thd scene (since \lminDiag is capable of filtering our some of the noise), we would not expect a decline in the error metric, which is computed with 
respect to the ZED \signal.
%as the average difference between reconstruction and the ground truth.
\prettyref{fig:EdgeRGBStatistics}(b) shows that, as expected, the use of RBG edges implies a slightly larger data rate saving, 
compared to the depth edges of \prettyref{fig:EdgeDepthExample}(b).

Extra visualizations, comparing reconstruction from sparse samples and from RGB edges for the \gazebo and the \zed datasets, are provided in Appendix~\ref{sec:gazeboImg} and Appendix~\ref{sec:zedImg}.

%% file: figureTex/figureEdgeDepthExample.tex
%!TEX root = ../main.tex

\begin{figure}[hbtp]
\begin{minipage}{\textwidth}
\newcommand{\figWidth}{ 0.325\linewidth } 
\def\arraystretch{1}		\setlength\tabcolsep{0.5mm}	\smaller
\begin{tabular}{ccc}\begin{minipage}[b]{\figWidth}\centering
\includegraphics[width=\linewidth]{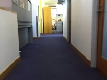} 
\\
(a) RGB image
\end{minipage}
& 
\begin{minipage}[b]{\figWidth}\centering
\includegraphics[width=\linewidth]{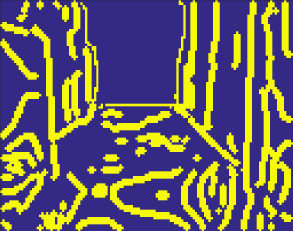}
\\
(b) depth edges
\end{minipage}
&
\begin{minipage}[b]{\figWidth}\centering
\includegraphics[width=\linewidth]{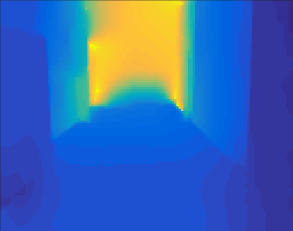}
\\
(c) reconstruction
\end{minipage}

\end{tabular}
\end{minipage}
\caption{An example of data compression using the edges in the depth \signal:
(a) RBG image, (b) edges in the depth \signal, (c) depth \signal reconstructed using \lminDiag.
%The edges extracted from rgb images match the geometry of the environments.
Spurious edges appear due to distortions in the full stereo disparity images.
}
\label{fig:EdgeDepthExample}
\end{figure}

%% file: figureTex/figureEdgeDepthStatistics.tex
%!TEX root = ../main.tex

\begin{figure}[hbtp]
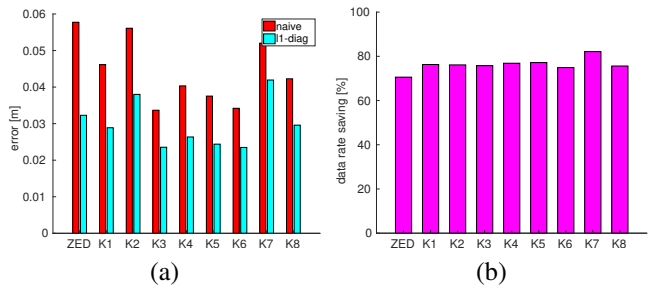

\myIncludeTwoFigures
{bandwidth/depth-edge/error-mean}
{bandwidth/depth-edge/datarate-mean}
{0}{0}{0}{0}
\caption{
(a) reconstruction errors and (b) data rate saving for the case when only 
depth edges are transmitted to the server.
%When only depth data along depth edges 
%(extracted from depth images) 
%are transmitted, the loss of accuracy is negligible and the data rate saving is around 80\%. 
}
\label{fig:EdgeDepthStatistics} 
\end{figure}

%% file: figureTex/figureEdgeRGBExample.tex
%!TEX root = ../main.tex

\begin{figure}[hbtp]
\begin{minipage}{\textwidth}
\newcommand{\figWidth}{ 0.325\linewidth } 
\def\arraystretch{3}		\setlength\tabcolsep{0.5mm}	\smaller
\begin{tabular}{ccc}\begin{minipage}[b]{\figWidth}\centering
\includegraphics[width=\linewidth]{bandwidth/rgb} 
\\
(a) RGB image
\end{minipage}
& 
\begin{minipage}[b]{\figWidth}\centering
\includegraphics[width=\linewidth]{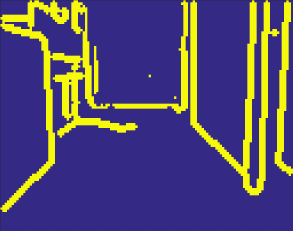} 
\\
(b) RGB edges
\end{minipage}
&
\begin{minipage}[b]{\figWidth}\centering
\includegraphics[width=\linewidth]{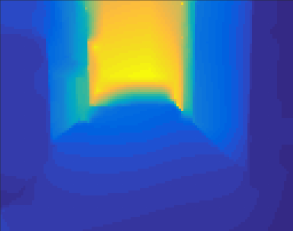} 
\\
(c) reconstruction
\end{minipage}

\end{tabular}
\end{minipage}
% An example of depth image compression with data only along RGB images:
% (a) RBG image, (b) edges in the RGB image, (c) \lminDiag reconstructed depth image.
% Undesirable edges appear due to distortion of stereo disparity images.
\caption{An example of data compression using the depth along the edges in the RGB image: (a) RBG image, (b) edges in the RGB image, (c) depth profile reconstructed using \lminDiag.
}
\label{fig:EdgeRGBExample}
\end{figure}

%% file: figureTex/figureEdgeRGBStatistics.tex
%!TEX root = ../main.tex

\begin{figure}[hbtp]
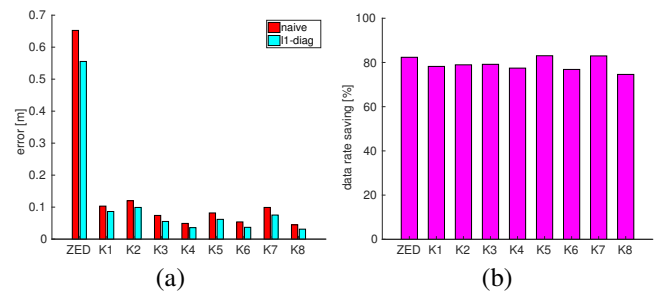

\myIncludeTwoFigures
{bandwidth/rgb-edge/error-mean}
{bandwidth/rgb-edge/datarate-mean}
{0}{0}{0}{0}
\caption{
When only depth data along RGB edges are transmitted, the reconstruction error is minimal except for the ZED dataset. This is because of the fact that disparity images provided by the ZED stereo camera are noisy and have undesirable edges, due to inherent depth distortion of stereo-vision. In other words, edges extracted from the depth images and from the RGB images are not consistent. (a) reconstruction error. (b) data rate saving.
}
\label{fig:EdgeRGBStatistics} 
\end{figure}

%% file: experimentTex/experiments_superResolution.tex
%!TEX root = ../main.tex

\input{figureTex/figureSuperResolution}

\subsection{Super-dense 3D Reconstruction and Super-resolution Depth Imaging}
\label{sec:exp_superResolution}

In this section, we demonstrate that our algorithm can also be applied to super-resolution depth imaging. 
Super-resolution imaging attempts to algorithmically increase the resolution of a given depth \signal.
This is fundamentally the same as viewing an input full depth \signal as measurements sampled from a higher-resolution ``ground truth'' and do reconstruction based on such measurements. An example is shown in \prettyref{fig:superresolution} using Kinect data. 
The original \signal, in \prettyref{fig:superresolution}(a), has a resolution of 72-by-103  and many missing pixels 
(due to Kinect sensor noise). \prettyref{fig:superresolution}(b) shows the reconstructed, or in other words \emph{up-scaled}, depth image. The size of the reconstructed depth image is 359-by-512, and thus has an up-scale factor of 24.79. Roughly speaking, 1 depth pixel in the input \signal translates to a 5-by-5 patch in the up-scaled depth \signal. Note that all missing pixels 
(including the legs of the chair in \prettyref{fig:superresolution}(a)) are smoothed out.
%, which in the example 
%but there is also loss of information, especially small details (e.g. legs of the chairs are gone). 

%% file: figureTex/figureSuperResolution.tex
%!TEX root = ../main.tex

\begin{figure}[htbp!]

\begin{minipage}{\textwidth}
\newcommand{\figWidth}{ 0.48\linewidth } 
\def\arraystretch{1}		\setlength\tabcolsep{1mm}	\smaller
\begin{tabular}{ l  r }\begin{minipage}[b]{\figWidth}\centering
\includegraphics[width=\linewidth]{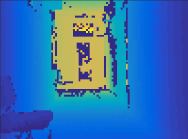}
\\
(a) 72-by-103 depth image
\end{minipage}
& 
\begin{minipage}[b]{\figWidth}\centering
\includegraphics[width=\linewidth]{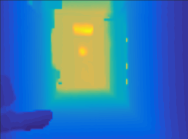}
\\
(b) up-scaled 359-by-512 image
\end{minipage}
\end{tabular}
\end{minipage}
\caption{
Super-resolution depth imaging. The up-scale factor is 24.79 in this example.
}
\label{fig:superresolution} 
\end{figure}

%% file: conclusion.tex
%!TEX root = main.tex

\section{Conclusion and Future Work} 
\label{sec:conclusion}

In this work, we propose a new approach to recover dense \twd and \thd depth \signals from sparse and incomplete depth measurements. 
As a first contribution, we formulate depth reconstruction as the problem of finding a \signal that has the sparsest second-order derivative, i.e., 
the least amount of corners and edges, while matching the given measurements. The problem itself is NP-hard, 
hence we relax it to a convex $\ell_1$-minimization problem with $\ell_\infty$-norm constraints. 
% convex relaxation and reformulated the problem as constrained $\ell_1$-minimization. We provided theoretical conditions under which the ground truth signal can be recovered, in both the noiseless case and when the measurements are corrupted by bounded noise. 

Our second contribution is a theoretical analysis that establishes precise conditions under which the dense depth \signal can be recovered 
from sparse samples. Even in the case in which exact recovery is not possible, we provide error bounds on the estimated \signal 
and discuss its sensitivity to measurement noise in both \twd and \thd  problems.

As a third contribution, we present several algorithmic variants to recover the depth \signal, each one resulting in a convex optimization problem. 
%The implementation of the presented approach is simple and straightforward, since they require solving only a convex minimization problem. 
%In the noiseless case, the problem can be further reduced to linear programming. 
To further accelerate these algorithms, we discuss how to adapt \NESTA, a first-order method for nonsmooth optimization, to our problem setup. 
%This solver significantly reduces the computational time by solving a smoothed, approximate version of the original problem.

The fourth contribution is an extensive experimental evaluation on both synthetic and real data. The experimental results show that our algorithms are able to reconstruct a dense depth \signal from an extremely low number of measurements (e.g., we can recover a 
$100$-by-$100$ depth \signal from
40 measurements), are robust to measurement noise, and are able to scale to large \signals. The capability of properly modeling measurement noise enables a performance boost with respect to interpolation-based approaches. We demonstrated the proposed approach in many applications, including \twd mapping, single-frame and multi-frame  \thd depth reconstruction from sparse measurements, \thd depth \signal compression and decompression, as well as super-resolution depth imaging.

As future work, we plan to further accelerate our algorithms using parallel computing (e.g., GPU). We would also like to apply the proposed algorithm to distributed mapping in bandwidth-limited multi-robot systems. In addition, we are interested in developing motion planning algorithms that can pro-actively guide the depth sampling process and further improve the reconstruction results. 

%that take into account the sampling strategy of sparse depth sensors. With more intelligent motion policies, the quality of reconstructed depth images can be further improved.

%% file: appendixTex/appendix_lemmas.tex
%!TEX root = ../main.tex

\section{Some Useful Lemmas}

\veryoptional{
\begin{lemma}\label{lem:edgeMap}
... there is a biunivocal mapping between edge set and support set \marginpar{Complete the lemma}
\end{lemma}
}

%%%%%%%%%%%%%%%%%%%%%%%%%%%%%%%%%%%%%%%%%%%%%%%%%%%%%%%%%%%%%%%%
We introduce some technical lemmas which simplify the derivations in the following sections.
\begin{lemma}[Null space of $\matU$]
\label{lem:nullSpaceA}
Consider the sparse sampling matrix $\matU \in \Real{m \times n}$. 
The null space of $\matU$ is spanned by the rows of the matrix 
$N \doteq \eye_\cosamples \in \Real{\mbar \times n} $ (with $\mbar \doteq n - m$).
Moreover, the action of the matrix $N$ on a vector $v$ and on a matrix $V$ of suitable dimensions
  is such that $N v = v_\cosamples$ and $N V = V_\cosamples$.  
\end{lemma} 

\proof
Denote the $i$-th standard basis vector as $e_i$.
Each row of $\matU \in \Real{m \times n}$ 
  is equal to $e_i\tran$ for some $i \in \samples$, hence $\matU$ has rank $m$. 
  Since the sets $\samples$ and $\cosamples$ 
  are disjoint and are such that $\samples \cup \cosamples = \{1,\ldots,n\}$, it follows that 
  $\matU N\tran = 0$ (entries of $\matU N\tran$ have the form $e_i\tran e_j$ which is zero for $\samples \ni i\neq j \in \cosamples$) 
  and $N\tran$ has rank $\mbar = n - m$. This proves that the rows of $N$ span the null space of $\matU$.
  Since each row of $N$ is $e_j\tran$ for some $j \in \cosamples$, the claims $N v = v_\cosamples$ 
  and $N V = V_\cosamples$  easily follow.
\qed

%%%%%%%%%%%%%%%%%%%%%%%%%%%%%%%%%%%%%%%%%%%%%%%%%%%%%%%%%%%%%%%%
\begin{lemma}[Symmetric Tridiagonal \toep matrix]
\label{lem:toep}
Let $T$ denote a symmetric tridiagonal \toep matrix with diagonal entries equal to 
$-2$ and off-diagonal entries equal to $1$:
\beq
\label{eq:toeplitz}
T \doteq 
\left[
\ba{cccccc}
-2 & 1  & 0 & 0 & \ldots & 0\\
1 & -2 & 1  & 0  & \ldots & 0 \\
0 & 1 & -2  & -1  & \ldots & 0 \\
\vdots & \ddots  & \ddots  & \ddots  & \ddots & 0 \\
0 &  \ldots & 0  & 0  & 1 & -2 \\
\ea
\right]
\eeq
Then the following claims hold:
\begin{enumerate}[label=(\roman*)] 
\item $T$ is invertible;
\item all the entries in the first and in the last column of $T\inv$ are negative and have absolute value smaller than 1;
\item let $v \in \Real{n}$ be defined as $v \doteq [1 \; 0 \; \ldots \; 0 \; 1]\tran$,
then $T\inv v = -\ones$.
\end{enumerate}
\end{lemma} 

\proof
Invertibility follows from~\citep[Corollary 4.2]{Fonseca01laa}, which also reports the explicit 
form of the inverse of a \toep matrix. We report the inverse here, tailoring it to our matrix. 
For the $n\times n$ \toep matrix $T$ in eq.~\eqref{eq:toeplitz},
the entry in row $i$ and column $j$ of $T\inv$ is: \beq
(T\inv)\ij = 
\left\{
\ba{ll}
(-1)^{2i-1} \frac{i(n-j+1)}{n+1} & \text{if } i \leq j \\
(-1)^{2j-1} \frac{j(n-i+1)}{n+1} & \text{if } i > j
\ea
\right.
\eeq
By inspection one can see that the first column ($j=1$) and the last column $j=n$ are all negative 
and have absolute value smaller than 1. 
The last claim can be proven by observing that $T \ones = -v$ and the matrix is invertible.
\qed

\begin{lemma}[Null Space of \tv]
\label{lem:nullSpacetv}
Given a \second-order difference operator $\tv \in \Real{(n-2) \times n}$, defined as in~\eqref{eq:tv}, 
the null space of $\tv$ is spanned by the following vectors:
\beq
v_1 = \ones_n  \qquad v_2 = [1 \; 2\; \ldots \; n]\tran
\eeq
\end{lemma}

\proof
By inspection one can see that $\tv v_1 = \tv v_2 = \zero$. 
Moreover, the rank of $\tv$ is $n-2$ and $v_1$ and $v_2$ are two linearly independent vectors, 
which proves the claim.
\qed

%% file: appendixTex/appendix_exact.tex
%!TEX root = ../main.tex

\section{Proof of \prettyref{prop:nam1D}}
\label{proof:prop-nam1D}

In this appendix we prove that in \twd depth reconstruction problems,
if we sample only the corners of the \signal (and the first and the last entry), then $\nam = 1$.
Moreover, we prove that if we sample the corners and their neighbors we have $\nam = 0$.
We start by rewriting~\prettyref{eq:nam} in a more convenient form.
Using~\prettyref{lem:nullSpaceA}, we know that $N (\tv_\cosupport)\tran = [\tv\tran]_{\cosamples,\cosupport}$.
In words, $\tv_\cosupport$ selects the rows of $\tv$ at indices $\cosupport$, or equivalently the columns of $\tv\tran$.
Similarly the multiplication by $N$ selects the rows of $\tv\tran$ at indices $\cosamples$. 
Similarly, $N (\tv_\support)\tran = [\tv\tran]_{\cosamples,\support}$.
Using these relations,~\prettyref{eq:nam} simplifies to:  
\beq
\nam = \linfM{ ([\tv\tran]_{\cosamples,\cosupport})\pinv [\tv\tran]_{\cosamples,\support} } < 1 
\eeq

Since $\support \cup \cosupport = \onen$, it is clear that $[\tv\tran]_{\cosamples,\cosupport}$
and $[\tv\tran]_{\cosamples,\support}$ are disjoint sets of columns of the matrix $[\tv\tran]_{\cosamples}$.

Let us start with the first claim: $\nam = 1$ whenever we sample the corners, the first, and the last entry of a \signal. 
We will make an extensive use of the structure of the matrix
 $\tv\tran$ which is the transpose of~\prettyref{eq:tv}. 
To give a more intuitive understanding of the proof we provide a small example of 
$\tv\tran$ with $n = 12$:
%\vspace{-0.5cm}
\beq
\label{eq:tvTran}
\hspace{-0.5cm}
\includegraphics[scale=1.3]{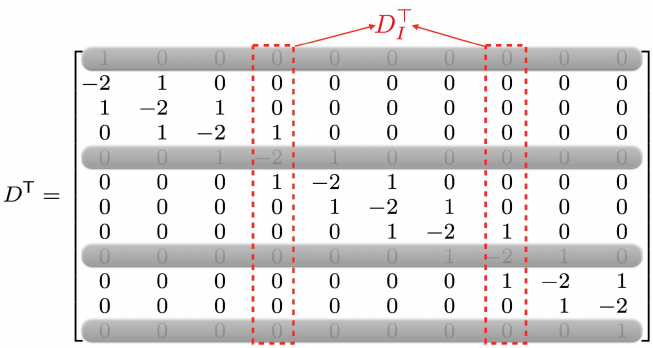}
\eeq
The matrix $[\tv\tran]_{\cosamples}$ is obtained from $\tv\tran$ after removing the 
 rows at indices in the sample set $\samples$: these ``deleted'' rows are shown in gray 
 in~\prettyref{eq:tvTran}. In particular, according to the assumptions of the first claim of~\prettyref{prop:nam1D}, 
 $\samples$ contains the first and the last sample (first and last row in $\tv\tran$) plus intermediate rows 
 corresponding to corners (two intermediate gray rows in the figure). 
 Now we note that the matrix $[\tv\tran]_{\cosamples,\support}$ selects the columns with indices in $\support$ 
 from $[\tv\tran]_\cosamples$. In figure, the columns that form $[\tv\tran]_{\cosamples,\support}$ are shown 
in dashed red boxes. The position of these columns is dictated by the position of the corners, 
hence if the $i$-th row corresponds to a corner, than column $i\!-\!1$ belongs to $\support$.

Three considerations are in order now. First, the matrix $[\tv\tran]_{\cosamples,\cosupport}$ is a block-diagonal
square matrix with diagonal blocks being \toep matrices (\cf with~\prettyref{eq:toeplitz}). Second, the matrix is 
invertible (follows from the first claim of~\prettyref{lem:toep}). Third, the matrix 
$[\tv\tran]_{\cosamples,\support}$ only contains 0 and 1 in suitable positions.
Therefore, the matrix $([\tv\tran]_{\cosamples,\cosupport})\pinv [\tv\tran]_{\cosamples,\support} = 
([\tv\tran]_{\cosamples,\cosupport})\inv [\tv\tran]_{\cosamples,\support}$ has the following block structure:
\beq
\label{eq:blockToep}
\left[
\ba{c}
T_1\inv \left[ \ba{ccccc} 0 & 0 &  \ldots & 0 & 0 \\ \vdots & \vdots & \vdots & \vdots & \vdots \\ 1 & 0 &  \ldots & 0 & 0\ea \right] \\
T_2\inv \left[ \ba{ccccc} 1 & 0 &  \ldots & 0 & 0 \\ \vdots & \vdots & \vdots & \vdots & \vdots \\ 0 & 1 &  \ldots & 0 & 0\ea \right] \\
\vdots \\
T_K\inv \left[ \ba{ccccc} 0 & 0 &  \ldots & 0 & 1 \\ \vdots & \vdots & \vdots & \vdots & \vdots \\ 0 & 0 &  \ldots & 0 & 0\ea \right] \\
\ea
\right]
=
\left[
\ba{c}
T_1\inv R_1 \\
T_2\inv R_2 \\
\vdots \\
T_K\inv R_K \\
\ea
\right]
\eeq
where $T_1,\ldots,T_K$ are \toep matrices of suitable dimensions and each $R_i$ contains at most 
two nonzero elements (equal to 1) in the first and the last row. 
Since for a matrix $M$, 
$\linfM{M}$ is the maximum of the $\ell_1$-norm 
of each row, we only need to demonstrate that the maximum $\ell_1$-norm of the rows of $T_i\inv R_i$ 
is no larger than 1 for all $i$. 
The action of $R_i$ on $T\inv_i$ is to select the first and/or the last column of $T_i\inv$ (depending on where the 1 appears).
For instance, $T\inv_1 R_1$ is zero everywhere, except the fist column which is equal to the last 
column of $T\inv_1$. From~\prettyref{lem:toep}(ii) we know that the entries in the first column 
have magnitude smaller than 1 hence it follows that $\linfM{T\inv_1 R_1 }<1$. 
A similar argument holds for the last column, hence $\linfM{T\inv_K R_K }<1$.
For the intermediate blocks $T\inv_i R_i$, $1<i<K$, it can be seen that 
$\linfM{T\inv_i R_i} = \linf{T\inv_i v}$, with $v \doteq [1 \; 0 \; \ldots \; 0 \; 1]\tran$; this follows from the fact that $R_i$ selects the 
first and the last columns of $T\inv_i$ which have negative entries due to \prettyref{lem:toep}(ii).
Using~\prettyref{lem:toep}(iii) we know that $T\inv v = -\ones$, from which it follows 
$\linfM{T\inv_i R_i} = \linf{T\inv_i v} = 1$. This proves the first claim.

The proof of the second claim ($\nam =0$ when we sample the corners and their neighbors) is much simpler.
Sampling the neighbors corresponds to deleting the rows contiguous to each ``corner'' 
from $\tv\tran$. From~\prettyref{eq:tvTran} the reader can easily see that this choice makes 
$[\tv\tran]_{\cosamples,\support} = \zero$, which in turns implies 
$\nam \doteq \linfM{ ([\tv\tran]_{\cosamples,\cosupport})\pinv [\tv\tran]_{\cosamples,\support} } = 0$.
\qed

%% file: appendixTex/appendix_exact_3D.tex
%!TEX root = ../main.tex

\section{Proof of~\prettyref{prop:nam2D}}
\label{proof:prop-nam2D}

In this appendix we prove that in \thd depth reconstruction
problems, if we sample the edges and the corresponding 
vertical and horizontal neighbors, then:
\beq
\label{eq:namRep}
\nam \doteq \linfM{ (N (\TV_\cosupport)\tran)\pinv N (\TV_\support)\tran) } = 0
\eeq
which implies exact recovery of the original depth \signal according to~\prettyref{prop:nam}.
As in Appendix~\ref{proof:prop-nam1D} we rewrite the condition~\eqref{eq:namRep} as:
\beq
\nam = \linfM{ ([\TV\tran]_{\cosamples,\cosupport})\pinv [\TV\tran]_{\cosamples,\support} } = 0
\eeq

The proof proceeds along the same line of the proof of the second claim in~\prettyref{prop:nam1D}. 
By observing the structure of $\TV\tran$, we realize that sampling the edges and the corresponding 
vertical and horizontal neighbors, makes $[\TV\tran]_{\cosamples,\support} = \zero$, which in turns implies $\nam = 0$. 
\qed

%% file: appendixTex/appendix_noiseless_algebraic.tex
%!TEX root = ../main.tex

\section{Proof of~\prettyref{prop:subdifferential}}
\label{proof:subdifferential}

In this appendix we establish necessary and sufficient conditions for an estimate 
$\xopt$ to be in the set $\SS$ of optimal solutions of problem~\eqref{eq:BP}.
The proof is identical for the \thd case in \prettyref{cor:subdifferential2D} (substituting $\tv$ with $\TV$), hence we 
restrict ourselves to the \twd case. We rewrite~\eqref{eq:BP} as: 
\beq
\label{eq:BPindicator}
\min_\xvar \lone{\tv \xvar} + \Ind_{\{\matU \xvar = \meas\}} \doteq \min_\xvar f(\xvar)
\eeq
where $\Ind_{\{\matU \xvar = \meas\}}$ is the indicator function of the set $\{\xvar:\matU \xvar = \meas\}$, 
which is zero whenever $\matU \xvar = \meas$ and $+\infty$ otherwise.
Since $\matU \xvar = \meas$ defines a convex (affine) set, the problem~\eqref{eq:BPindicator} is convex.
In the following we make extensive use of the notion of subgradients of convex functions. We refer the reader 
to~\citep[\S 4]{Bertsekas89book} for a comprehensive treatment and to~\citep{Boyd06notes} 
for a quick introduction.

A point $\xopt$ is a minimizer of a convex function $f$ if and only if $f$
is subdifferentiable at $\xopt$ and the zero vector belongs to the set of subgradients 
of $f$, i.e., $\zero \in \partial f(\xopt)$. The set of subgradients is also called the \emph{subdifferential}.
 The subdifferential of a sum of functions is the sum of the subdifferentials, therefore
  \beq
 \label{eq:sumSubdifferential}
\partial f(\xopt) = \partial(\lone{\tv \xvar})(\xopt) + \partial(\Ind_{\{\matU \xvar = \meas\}})(\xopt)
 \eeq 
 In the following we compute each subdifferential in~\eqref{eq:sumSubdifferential}. 
 Let us call $\support$ the support set of the vector $\tv \xopt$, and 
 recall that, given a vector $v$, we denote with $v_\support$ the subvector of $v$ including 
  the entries of $v$ at indices in $\support$.
  Using~\citep[Page 5]{Boyd06notes}:   \begin{align}
  \label{eq:subL1}
 &\partial(\lone{\tv \xvar})(\xopt) = \notag \\
 & \qquad \quad \setdef{ \tv\tran u \in \Real{n} }{  u_\support = \sign(\tv \xopt)_\support, \linf{u_\cosupport} \leq 1 }
  \end{align} 
  The second subdifferential in~\eqref{eq:sumSubdifferential} is~\citep[Page 254]{Bertsekas03book}:
  \beq
  \label{eq:subIndicator}
  \partial(\Ind_{\{\matU \xvar = \meas\}})(\xopt) = \setdef{ g \in \Real{n} }{ g\tran \xopt \geq g\tran r, \; \forall  r \text{ s.t. } \matU r = \meas  }
  \eeq
  To get a better understanding of the set in~\eqref{eq:subIndicator}, we note that every solution $r \in \Real{n}$ of the overdetermined linear system
  $\matU r = y$ can be written as a vector that satisfies the linear system, plus a vector that is in the null space of $\matU$.
  Now    we know that $\xtrue$, the vector that generated the data $y$, satisfies $\matU \xtrue = y$. Therefore, we 
  rewrite~\eqref{eq:subIndicator} as: 
 \begin{align} \label{eq:subIndicator_2}
  & \partial(\Ind_{\{\matU \xvar = \meas\}})(\xopt) = \notag \\
  & \qquad \quad \setdef{ g \in \Real{n} }{ g\tran \xopt \geq g\tran (\xtrue + \rtilde), \; \forall  \rtilde \in \kernel(\matU)}
  \end{align}
  where $\kernel(\matU)$ denotes the kernel of $\matU$. From Lemma~\ref{lem:nullSpaceA} we know that 
  the kernel  $\matU$ is spanned by the matrix $N$ (defined in the lemma), hence~\eqref{eq:subIndicator_2} further 
  simplifies to:
 \begin{align}
  \label{eq:subIndicator_3}
  &\partial(\Ind_{\{\matU \xvar = \meas\}})(\xopt) = \notag \\
  & \qquad \quad \setdef{ g \in \Real{n} }{ g\tran \xopt \geq g\tran (\xtrue + N w), \; 
  \forall  w \in \Real{\mbar} }
  \end{align}  
  Rearranging the terms:
  \begin{align}
  \label{eq:subIndicator_4}
  & \partial(\Ind_{\{\matU \xvar = \meas\}})(\xopt) = \notag \\
  & \qquad \quad  \setdef{ g \in \Real{n} }{ g\tran (\xopt-\xtrue) \geq (N\tran g)\tran w, \; 
  \forall  w \in \Real{\mbar}
   }
  \end{align}
  From the second claim of Lemma~\ref{lem:nullSpaceA} we know that $N\tran g = g_\cosamples$; moreover, 
  we observe that if an element of $g_\cosamples$ is different from zero, then we can pick an arbitrarily large 
  $w$ that falsifies the inequality, therefore, it must hold $g_\cosamples = \zero$.
  Therefore, we rewrite~\eqref{eq:subIndicator_4} as:
 \beq
  \label{eq:subIndicator_5}
  \partial(\Ind_{\{\matU \xvar = \meas\}})(\xopt) = \setdef{ g \in \Real{n} }{ g\tran (\xopt-\xtrue) \geq 0, \ g_\cosamples = \zero 
   }
  \eeq 
  Now we split the product $g\tran (\xopt-\xtrue)$ as $g_\cosamples\tran (\xopt-\xtrue)_\cosamples + g_\samples\tran (\xopt-\xtrue)_\samples$ 
and note that $g_\cosamples = 0$. Moreover, for any feasible $\xopt$, the $i$-th entry of $\xopt-\xtrue$ is zero for all $i \in \samples$, 
which implies $g_\samples\tran (\xopt-\xtrue)_\samples = \zero$. 
Therefore, the inequality $g\tran (\xopt-\xtrue) \geq 0$ 
vanishes and we
remain with: 
 \beq
  \label{eq:subIndicator_6}
  \partial(\Ind_{\{\matU \xvar = \meas\}})(\xopt) = \setdef{ g \in \Real{n} }{ g_\cosamples = \zero 
   }
  \eeq 
  Substituting~\eqref{eq:subIndicator_6} and~\eqref{eq:subL1} back into~\eqref{eq:sumSubdifferential}, we obtain:
  \bea
 \label{eq:subdiff_f}
\partial f(\xopt) = \{   \tv\tran u + g &:& u_\support = \sign(\tv \xopt)_\support, \ \linf{u_\cosupport} \leq 1, \nonumber \\
&&  g_\cosamples = \zero  \}
 \eea 
We can now use the subdifferential~\eqref{eq:subdiff_f} to describe the optimal solution set $\SS$ of~\eqref{eq:BP}; as mentioned earlier in this 
section, $\xopt$ is optimal if and only if zero is a subgradient, therefore $\SS$ is defined as:
\bea
\label{eq:appSS}
\SS =  \{ \xopt &:& \exists u \in \Real{n-2}, g \in \Real{n}, \text{ such that } \nonumber \\
&& \tv\tran u + g = \zero, \nonumber \\
&& u_\support = \sign(\tv \xopt)_\support, \;\; \linf{u_\cosupport} \leq 1, \nonumber \\
&&  g_\cosamples = \zero  \}
\eea
We note that the constraints $\tv\tran u + g = \zero$ and $g_\cosamples = \zero$ 
can be written compactly as $[\tv\tran u]_\cosamples = \zero$, which is the same as $(\tv\tran)_\cosamples \; u = \zero$.
This allows rewriting~\eqref{eq:appSS} as:
\bea
\label{eq:appSS1}
\SS =  \{ \xopt &:& \exists u \in \Real{n-2}, \text{ such that } \nonumber \\
&& (\tv\tran)_\cosamples \; u = \zero,  \\
&& u_\support = \sign(\tv \xopt)_\support, \;\; \linf{u_\cosupport} \leq 1 \} \nonumber
\eea
which coincides with the optimality condition of~\prettyref{prop:subdifferential}, proving the claim. \qed

%% file: appendixTex/appendix_noiseless_signConsistency2D.tex
%!TEX root = ../main.tex
\newcommand{\nK}{K}

\section{Proof of~\prettyref{thm:1Doptimality}}
\label{proof:thm-1Doptimality}

\prettyref{thm:1Doptimality} says that sign consistency of $\xvar$ is a necessary and sufficient condition 
for $\xvar$ to be in the solution set of~\eqref{eq:BP}. In the following we denote with $\SC$ the set 
of sign consistent \signals which are feasible for~\eqref{eq:BP}. Moreover, we denote with $\SS$ 
the set of optimal solutions of~\eqref{eq:BP}. The proof relies on the optimality conditions 
of~\prettyref{prop:subdifferential}, which we recall here: a \signal $\xvar$ is in the solution 
set $\SS$ if and only if 
there exists 
a $u \in \Real{n-2}$ such that 
\beq
\label{eq:opt1}
(\tv\tran)_\cosamples \; u = 0 \; \text{ and } \; u_\support = \sign(\tv \xvar)_\support \; \text{ and } \; \linf{u_\cosupport} \leq 1
\eeq
where $\support$ is the support set of the vector $\tv \xvar$. Before proving that $\xvar \in \SC \Leftrightarrow \xvar \in \SS$, we 
need a better understanding of the structure of the matrix $(\tv\tran)_\cosamples$.
We note that, when taking twin samples, the matrix $(\tv\tran)_\cosamples$ 
is obtained from $\tv\tran$ by removing pairs of consecutive rows.
For instance, considering a problem with $n = 12$, the product $(\tv\tran)_\cosamples u$ 
becomes:
\beq
\label{eq:tvTwin}
\hspace{0.2cm}
\includegraphics[scale=1.5]{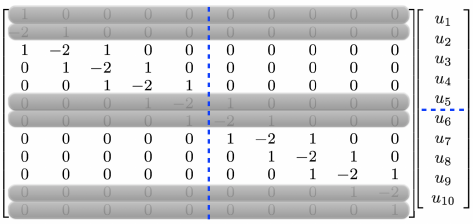}
\eeq
where gray rows are the ones we ``removed'' from $\tv\tran$ to obtain $(\tv\tran)_\cosamples$.
 By observing~\eqref{eq:tvTwin},
the reader can verify that the 
resulting matrix  $(\tv\tran)_\cosamples$ is block diagonal (in general will have more than 2 diagonal blocks), and each block
is a \second-order difference operator like~\eqref{eq:tv} of suitable size. 
We denote the diagonal blocks as $\tv\at{1}, \tv\at{2}, \ldots, \tv\at{\nK}$ ($\nK=2$ in the example 
of eq.~\eqref{eq:tvTwin}).
This also induces a partition in the vector $u$, which can be split vertically as $u = [u_{S_1} \; u_{S_2} \; \ldots \; u_{S_\nK}]$.
Geometrically, the block diagonal structure that arises means that each segment between consecutive twin samples can 
be studied independently.
Therefore, the condition~\eqref{eq:opt1} can be written as:
\beq
\tv\at{k} \; u_{S_k} = 0 \; \text{ and } \; u_{\support_k} = \sign(\tv \xvar)_{\support_k} \; \text{ and } \; \linf{u_{S_k}} \leq 1
\eeq
for $k= 1,\ldots,\nK$, where $\support_k \subseteq S_k$ are entries in the support set $\support$ that fall within the set $S_k$.
With this machinery we are ready to prove~\prettyref{thm:1Doptimality}.

Let us start with the implication $\xvar \in \SC \Rightarrow \xvar \in \SS$.
We first consider the case in which all signs are consistent, i.e., \prettyref{def:1DsignConsitency}(i).
This means that within each set $S_k$, $\sign(\tv \xvar)_{\support_k} = +\ones$ or 
$\sign(\tv \xvar)_{\support_k} = -\ones$. Without loss of generality, 
assume $\sign(\tv \xvar)_{\support_k} = +\ones$. Then, we can see that 
selecting $u_{S_k} = +\ones$ satisfies $u_{\support_k} = \sign(\tv \xvar)_{\support_k}$ and 
$\linf{u_{S_k}} \leq 1$. Moreover, since $\ones$ is in the null space of $\tv\at{k}$ 
(see~\prettyref{lem:nullSpacetv}), it follows $\tv\at{k}  u_{S_k} = \zero$, proving the claim.
To complete the demonstration of $\xvar \in \SC \Rightarrow \xvar \in \SS$ we
consider the case in which there is there is a sign change at the boundary of each segment, 
while all signs are zero in the interior (\prettyref{def:1DsignConsitency}(ii)). In this case the condition $u_{\support_k} = \sign(\tv \xvar)_{\support_k}$ 
imposes that the first and the last elements of $u_{\support_k}$ are $+1$ and $-1$ (or $-1$ and $+1$) respectively.
Without loss of generality, assume that the signs are $+1$ and $-1$.
Then the linear system $\tv\at{k} \; u_{S_k} = \zero$ becomes $T_k \hat{u}_{S_k} = \pm [+1 \; 0 \ldots \; 0 \; -1]\tran$, 
where $T_k$ is a \toep matrix of suitable dimension and $\hat{u}_{S_k}$ is the vector $u_{S_k}$ without the first and 
the last entry which we fixed to $+1$ and $-1$, respectively.
The existence of a suitable solution to the linear system $T_k \hat{u}_{S_k} = \pm [+1 \; 0 \ldots \; 0 \; -1]\tran$, which is such that $\linf{T_k \hat{u}_{S_k}} \leq 1$ 
follows from~\prettyref{lem:toep}(ii).

Let us prove the reverse implication, i.e., $\xvar \in \SS \Rightarrow \xvar \in \SC$. 
Without loss of generality, we consider a single segment $S_k$ and we re-label the corresponding entries from 
$1$ to $n_k$. Let us assume that $\xvar \in \SS$, which means that there exists $u_{S_k}$ such that 
$\tv\at{k}  u_{S_k} = \zero$, $u_{\support_k} = \sign(\tv \xvar)_{\support_k}$, and $\linf{u_{S_k}} \leq 1$.
Any solution of 
 $\tv\at{k}  u_{S_k} = \zero$ is in the null space of 
$\tv\at{k}$, which is spanned by the vectors $v_1$ and $v_2$ defined in~\prettyref{lem:nullSpacetv}.
Therefore, we write $u_{S_k}$ as $u_{S_k} = \alpha v_1 + \beta v_2$ with $\alpha,\beta \in \Real{}$.
Assume that there are indices $i,j \in S_k$,  such that 
$\sign(\tv \xvar)_{i} = +1$ and $\sign(\tv \xvar)_{j} = -1$, therefore it must hold:
\beq
\label{eq:ab_1}
[\alpha v_1 + \beta v_2]_i = +1  \;\; \text{ and  } \;\; [\alpha v_1 + \beta v_2]_j = -1
\eeq
which, recalling the definitions of $v_1$ and $v_2$ in~\prettyref{lem:nullSpacetv}, becomes:
$\alpha + \beta i = +1  \text{ and  } \alpha + \beta j = -1$.
It follows that:
\beq
\label{eq:beta}
\beta = \frac{2}{j-i}  
\eeq 
Now, since $\xvar \in \SS$ it must also hold $\linf{u_{S_k}} \leq 1$ which can be written as:
\bea
-\ones \leq &  \alpha v_1 + \beta v_2 & \leq \ones \Leftrightarrow \\ 
1\leq & \alpha + \beta h & \leq 1, \forall h =1,\ldots,n_k
\eea
Combining the inequalities for $h=1$ and $h = n_k$ we get:
\beq
\label{eq:betaIneq}
\beta (n_k-1) \leq 2
\eeq 
Substituting~\eqref{eq:beta} into~\eqref{eq:betaIneq}, we get:
\beq
\frac{n_k-1}{j-i} \leq 1
\eeq
which is satisfied if and only if $i=1$ and $j=n_k$. 
Hence, we proved that sign changes of the curvature of $\xvar \in \SS$  
 can only happen at the boundary 
of each segment, which agrees with our definition of $\SC$ (\prettyref{def:1DsignConsitency}),
proving the claim.
\qed

%% file: appendixTex/appendix_noiseless_errorBound2D.tex
%!TEX root = ../main.tex

\section{Proof of~\prettyref{prop:1DrecoveryError}}
\label{proof:1DrecoveryError}

Let us start by proving that if we sample the boundary of $\xtrue$ and the sample set includes a twin 
sample in  each linear segment of $\xtrue$, then, $\xtrue$ is in the set of minimizers of~\eqref{eq:BP}.
The claim is a direct consequence of~\prettyref{thm:1Doptimality}.
By construction, if we have a twin sample per segment, only two cases are possible:
either both samples fall inside the linear segment (i.e., none of them corresponds to a corner), 
or one of the samples corresponds to a corner.
In both cases it is easy to see that $\xtrue$ is sign consistent with respect to this choice of samples, which implies that 
$\xtrue \in \SS$ according to~\prettyref{thm:1Doptimality}. 

\begin{figure}[htbp]
\centering
\hspace{-3mm}
\includegraphics[height=4cm]{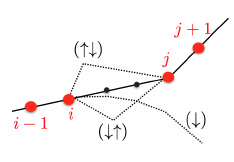}
\caption{Region between a pair of twin samples, with a twin sample including a corner.}
\label{fig:app-toyExample}
\end{figure}

The second claim states that any optimal solution $\xopt$ lies between the 
naive estimate $\xnaive$ and the true \signal $\xtrue$.
Before proving this claim
we note that the presence of twin samples makes the objective  of~\eqref{eq:BP} separable. 
To see this we note that, in a noiseless case, if we sample a point $i$, 
then its value is fixed by the corresponding linear constraint and $\xvar_i$ is no longer a variable. Therefore, by sampling, we are essentially fixing pairs of consecutive entries in $\xvar$.
Using this property, we see that the objective separates as:
\bea
\label{eq:separability}
\lone{\tv \xvar} = \lone{\tv\at{1} \xvar_{S_1}} + \lone{\tv\at{2} \xvar_{S_2}} + \ldots + \lone{\tv\at{K} \xvar_{S_K}} \nonumber \hspace{-5mm}\\
\eea
where $\tv\at{k}$ is a \second-order difference matrix of suitable dimensions,
 and  $\xvar_{S_k}$ is the subvector of $\xvar$ including the entries corresponding to 
consecutive twin samples (say $\xvar_{i-1}$, $\xvar_{i}$ and $\xvar_{j}$, $\xvar_{j+1}$, 
which are fixed to know values) and all the entries between those (i.e., $\xvar_{i+1}, \ldots, \xvar_{j-1}$);
we used the symbol $K$ in~\eqref{eq:separability} to denote the number of regions between 
consecutive twin samples.
From the separability of the objective, it follows that we can study each region (between twin samples)
independently (the optimization splits in $K$ independent optimizations).
We now prove the second claim: for any optimal solution $\xopt \in \SS$ and any index $i \in \onen$, it holds 
that  $\min(\xtrue_i, \xnaive_i) \leq \xopti \leq \max(\xtrue_i, \xnaive_i)$.
%if $\xtrue_i \leq \xnaive_i$ then $\xtrue_i \leq \xopti \leq \xnaive_i$, and if $\xnaive_i \leq \xtrue_i$ 
%then $\xnaive_i \leq \xopti\leq \xtrue_i$. 
As mentioned before, this means that any optimal solution is ``between'' 
the naive solution $\xnaive$ (obtained by connecting the dots, see the blue dashed line in \prettyref{fig:toyExample}(a))  and the 
true solution $\xtrue$ (black solid line in \prettyref{fig:toyExample}(a)). We show that the claim must hold true in all regions 
$S_1, \ldots, S_K$. 
First, let us get rid of the ``degenerate'' regions: these 
are the ones in which $\xnaive_i = \xtrue_i$ for all $i \in S_k$. This happens when we 
sample a corner and there are 3 collinear samples as in~\prettyref{fig:app-toyExample}.
In this case for any index $i \in S_k$, we prove $\xnaive_i = \xopti = \xtrue_i$, i.e., 
all optimal solutions must reduce to a straight line between the collinear samples.
We prove this with the visual support of~\prettyref{fig:app-toyExample}.
Our goal is to show that any 
$\xhat$ that deviates from linearity is not sign consistent (\SC), hence cannot be optimal.
If a sample $k \in \{i+1, \ldots, j-1\}$ has a curvature different from zero, then, to be sign consistent 
it cannot change curvature. This case is shown with the label ($\downarrow$) in~\prettyref{fig:app-toyExample}.
Clearly, if $\xhat$ cannot change curvature, after deviating from the straight line, it cannot 
reach the sample $j$, leading to contradiction. 
Similarly, contradiction occurs when the curvature 
is different from zero at $i$: a positive curvature at $i$ (case ($\uparrow\downarrow$) in figure), must be compensated by 
a negative curvature before $j$ for the curve to intersect $j$: this again violates sign consistency; 
analogous argument holds for a negative curvature at $i$ (case ($\downarrow \uparrow$) in figure). Therefore we proved 
that in these straight segments it holds $\xnaive_i = \xopti = \xtrue_i$. 
If only remains to discuss the case in which $\xnaive_i \neq \xtrue_i$ which occurs 
whenever consecutive double samples do not include corners. This situation is pictured 
in~\prettyref{fig:toyExample}(a); in this case $\xnaive_i \leq \xtrue_i$. We only prove that whenever 
$\xnaive_i \leq \xtrue_i$ then $\xnaive_i \leq \xopti \leq \xtrue_i$; the proof 
for the case $\xtrue_i \leq \xnaive_i$ is practically identical (the corner points downwards).
In~\prettyref{fig:toyExample}(a) we show two cases (dashed lines): in case ($\uparrow\downarrow$) we show a \signal $\xhat$ above $\xtrue$;
in ($\downarrow\uparrow$) we show a \signal below $\xnaive$. One can easily realize than any \signal as in case ($\uparrow\downarrow$)  has a 
positive curvature at sample $i$ and a negative curvature at the top corner: this violates sign consistency hence 
$\xhat$ is not in the solution set. A similar argument holds in the case ($\downarrow\uparrow$), which concludes the proof 
that $\xopt$ must be ``between'' $\xnaive$ and $\xtrue$.

%{\bf Extra result: error bounds for $\xopt$}. 
We conclude the proof by deriving the error bound in eq.~\eqref{eq:recoveryError}, 
 repeated below for the reader's convenience:
% prove the following error bound 
% which relates any solution $\xopt$ of the $\ell_1$-minimization problem~\eqref{eq:BP} to the true \signal $\xtrue$: 
\beq
\label{eq:recoveryError2}
\linf{\xtrue - \xopt} \leq \max_{i \in \samples} d_i \cos(\theta_i)
\eeq
where $d_i$ is the distance between the sample $i$ and the nearest corner in $\xtrue$, 
while $\theta_i$ is the angle that the line connecting $i$ with the nearest corner forms with the vertical, see~\prettyref{fig:toyExample}(a).The bound relates any solution $\xopt$ of the $\ell_1$-minimization problem~\eqref{eq:BP} to the true \signal $\xtrue$.

To prove~\eqref{eq:recoveryError2}, we note that $\xopt$ can deviate
from $\xtrue$ only in the ``corner cases'' as the one in~\prettyref{fig:toyExample}(a) 
(proven above in this section). Moreover, we note that the maximum error $|\xtrue_i - \xnaive_i|$ 
is attained at the corner 
of $\xtrue$ and is denoted with $\hat{d}$ in the figure.
From basic trigonometry we conclude:
\beq
\hat{d} \leq \max_{k \in \{i,j\}} d_k \cos{\theta_k}
\eeq
(in~\prettyref{fig:toyExample}(a) this becomes: $\hat{d} \leq d_i \cos(\theta_i)$), 
where $d_k$ is the distance between sample $k$ and the nearest  corner, while $\theta_k$ 
is the angle that the line connecting sample $k$ with the nearest corner forms with the vertical.
The bound~\eqref{eq:recoveryError2} follows by extending this inequality to 
all linear segments in $\xtrue$. \qed

%% file: appendixTex/appendix_noiseless_signConsistency3D.tex
%!TEX root = ../main.tex

\begin{figure*}[t]
\hspace{-0.5cm}
\centering
\includegraphics[scale=2.5]{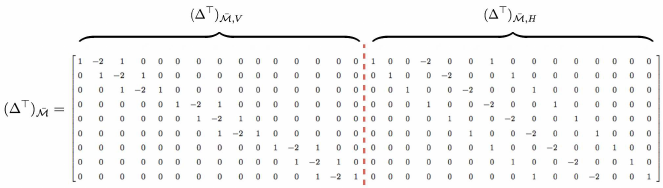}
\caption{Example of matrix $(\TV\tran)_\cosamples$ and corresponding column partition $(\TV\tran)_{\cosamples,V}$, $(\TV\tran)_{\cosamples,H}$.}
\label{fig:TVtran2D}
\end{figure*}

\section{Proof of~\prettyref{thm:2Doptimality_xtrue}}
\label{proof:thm-2Doptimality_xtrue}

In this section we prove that if a \thd \signal $\Xvar$ is feasible for problem \eqref{eq:BPmatrix} and it 
is \thd sign consistent, then it is also a minimizer of \eqref{eq:BPmatrix}.
The proof is similar to the \twd case in~\prettyref{prop:1DrecoveryError}, 
and relies on \prettyref{cor:subdifferential2D}, which we restate as follows:
given a \signal $\Xvar \in \Real{r \times c}$ which is feasible for problem~\eqref{eq:BPmatrix}, 
 $\Xvar$ is in the optimal set of~\eqref{eq:BPmatrix} if there exists  a vector $u \in \Real{2(n-r-c)}$, with $n = r \times  c$, such that 
\beq
\label{eq:opt2}
(\TV\tran)_\cosamples \; u = \zero \; \text{ and } \; u_\support = \sign(\TV \xvar)_\support \; \text{ and } \; \linf{u_\cosupport} \leq 1,
\eeq
where $\samples$ denotes a grid sample set. Let $\cosamples_i$ be patches defined in Definition~\ref{def:2DgridSamples}. We show that \thd sign consistency of $\Xvar$ with respect to grid samples $\samples$ implies \eqref{eq:opt2}. 
We start by noting that when using grid samples with $\nK$ patches (see \prettyref{fig:envelop1D}(b)), 
 problem \eqref{eq:BPmatrix} separates into
$\nK$ independent optimization subproblems (similarly to the \twd case of eq.~\eqref{eq:separability}).
Therefore, without loss of generality in the following we focus on a single patch and we assume that 
the grid samples include the boundaries of the patch (first and last two rows and columns). 
With slight abuse of notation we denote this patch with $\Xvar$ and we use $\xvar = \vec{\Xvar}$.

Before proving the claim we need some insight on the structure of the matrix $(\TV\tran)_\cosamples$.
 This matrix is obtained by deleting rows of $\TV\tran$ indexed by $\samples$.  Since we are assuming to sample the boundaries of the patch, the resulting $(\TV\tran)_\cosamples$ 
 has the structure described in~\prettyref{fig:TVtran2D}. 
Now we remain to show that when $\Xvar$ is \thd sign consistent, we can find a vector $u$
that satisfies the three conditions in~\eqref{eq:opt2}.
Towards this goal, we recall that $\TV$ is obtained by stacking two submatrices that compute the vertical and horizontal 
differences as in~\eqref{eq:TVpartition}. Therefore, we split $\sign(\TV \xvar)_\support$ accordingly as:
\beq
\sign(\TV \xvar)_\support = \vect{\sign(\TV \xvar)_{\support_V} \\ \sign(\TV \xvar)_{\support_H}}
\eeq
where $\support_V$ includes indices in the support set $\support$ corresponding to nonzero vertical 
differences, while $\support_H$ includes indices in $\support$ corresponding to nonzero horizontal 
differences.

Now, we note that the row partition of $\TV$ induces a column partition of 
 $(\TV\tran)_\cosamples$.
We call the corresponding submatrices $(\TV\tran)_{\cosamples,V}$ and $(\TV\tran)_{\cosamples,H}$,  as shown in~\prettyref{fig:TVtran2D}.  
This also partitions the vector $u$ into two subvectors $u_V$ and $u_H$.
Therefore, we rewrite the condition $(\TV\tran)_\cosamples u = \zero$ as:
\beq
\label{eq:conditionNullHV}
(\TV\tran)_{\cosamples,V} \; u_V + (\TV\tran)_{\cosamples,H} \; u_H = \zero
\eeq
Since $\Xvar$ is \thd sign consistent, then $\sign(\TV \xvar)_{\support_V}$ is either $+\ones$ or $-\ones$ 
and $\sign(\TV \xvar)_{\support_H}$ is either $+\ones$ or $-\ones$. 
Assume without loss of generality that $\sign(\TV \xvar)_{\support_V} = +\ones$ and 
$\sign(\TV \xvar)_{\support_H} = -\ones$. Now if we choose $u_V = +\ones$ and $u_H = -\ones$, 
it holds that $u_{\support_V} = \sign(\TV \xvar)_{\support_V}$, $u_{\support_H} = \sign(\TV \xvar)_{\support_H}$, 
and $\linf{u_\cosupport} \leq \linf{u} \leq 1$, hence the last two conditions in~\eqref{eq:opt2} are satisfied. 
Moreover, since each row of $(\TV\tran)_{\cosamples,V}$ 
and $(\TV\tran)_{\cosamples,H}$ includes only three nonzero entries with values $+1, -2, +1$, 
it follows that:
\bea
(\TV\tran)_{\cosamples,V} \; u_V = (\TV\tran)_{\cosamples,V} \; \ones = \zero \nonumber \\
(\TV\tran)_{\cosamples,H} \; u_H = -(\TV\tran)_{\cosamples,H} \; \ones = \zero
\eea
 which implies~\eqref{eq:conditionNullHV}, concluding the proof.
\qed

%% file: appendixTex/appendix_noiseless_errorBound3D.tex
%!TEX root = ../main.tex

\section{Proof of~\prettyref{prop:2DrecoveryError}}
\label{proof:prop-2DrecoveryError}

By the assumptions of \prettyref{prop:2DrecoveryError}, $\Xtrue$ is the ground truth generating noiseless measurements~\eqref{eq:measurements} and $\Xtrue$ is \emph{\thd sign consistent} with respect to $\samples$.
Then each row of $\Xtrue$, namely $\Xtrue_i$, is a sign consistent \twd depth \signal 
and, given the samples, we can build a \emph{row-wise envelope} for $\Xtrue_i$ as prescribed in~\prettyref{thm:1Doptimality} (pictorial explanation in~\prettyref{fig:envelop1D}(b)).  
Repeating this procedure for all rows $i$ and calling $\bar{\Xvar}$ and $\underbar{\Xvar}$ the point-wise upper bound an lower bound for 
the envelope of each row, then $\Xtrue$ is within the \emph{row-wise envelope}, i.e., $\underbar{\Xvar}_{i,j} \leq \Xtrue_{i,j} \leq \bar{\Xvar}_{i,j}$ for all $i=1,\ldots,r$ and $j=1,\ldots,c$.
Therefore, given an optimal solution $\Xopt$ of~\eqref{eq:BPmatrix} it holds that
\begin{equation*}
|\Xtrue_{i,j} - \Xvar^\star_{i,j}| \leq \max(|\underbar{\Xvar}_{i,j} - \Xopt_{i,j}| , |\bar{\Xvar}_{i,j} - \Xopt_{i,j}|).
\end{equation*}
which trivially follows from the chain of inequalities $\underbar{\Xvar}_{i,j} \leq \Xtrue_{i,j} \leq \bar{\Xvar}_{i,j}$, 
concluding the proof.
\qed

%% file: appendixTex/appendix_noisy_algebraic.tex
%!TEX root = ../main.tex

\section{Proof of~\prettyref{prop:robust_subdifferential}}
\label{proof:prop-robust_subdifferential}

This appendix proves the necessary and sufficient conditions 
for an estimate $\xopt$ to be in the set $\SS$ of optimal solutions of problem \eqref{eq:BPD}, 
as stated in~\prettyref{prop:robust_subdifferential}. We follow similar arguments as the proof in Appendix~\ref{proof:subdifferential}. 

We start by rewriting  \eqref{eq:BPD} as:
\beq
\label{eq:BPDindicator}
\min_\xvar \lone{\tv \xvar} + \Ind_{\{\xvar: \linf{\matU \xvar - \meas} \leq \vareps \}} \doteq \min_\xvar f(\xvar)
\eeq
As discussed earlier in Appendix~\ref{proof:subdifferential}, a \signal $\xopt$ is optimal for problem \eqref{eq:BPD} 
%we will establish the conditions so that 
if the zero vector belongs to the set of subgradients of $f$ at $\xopt$, i.e., $\zero \in \partial f(\xopt)$. The subdifferential of $\lone{\tv \xvar} $ was given in \eqref{eq:subL1}. The subdifferential of the indicator function in \eqref{eq:BPDindicator} is given, 
similarly to \eqref{eq:subIndicator}, as
\begin{align}
  \label{eq:robustsubIndicator}
  & \partial(\Ind_{\{\xvar: \linf{\matU \xvar - \meas} \leq \vareps \}})(\xopt) = \notag \\
  & \qquad \setdef{ g \in \Real{n} }{ g\tran \xopt \geq g\tran r, \; \forall  r \text{ s.t. } \linf{\matU r - \meas} \leq \vareps  }
\end{align}
Recall that the matrix $\matU$ restricts a vector $r$ to its entries in the sample set $\samples$, therefore we have
\begin{align}\label{eq:rewrite_r}
\{ r: \ \linf{\matU r - \meas} \leq \vareps \} = \{ r: \ \linf{ r_\samples - \meas} \leq \vareps \}.
\end{align}
Obviously $\linf{ r_\samples - \meas} \leq \vareps$ implies for all $i \in \samples$ that
\begin{align}\label{eq:rMcases}
r_i \leq \meas_i + \vareps \ \ \text{or} \ \  r_i \geq \meas_i - \vareps.
\end{align}
We decompose 
\begin{align}\label{eq:decompose_g}
g\tran \xopt \geq g\tran r = g\tran_\samples r_\samples + g\tran_\cosamples r_\cosamples.
\end{align}
A vector $g$ that satisfies \eqref{eq:decompose_g}, for all $r$ obeying \eqref{eq:rewrite_r}, must satisfy $g\tran_\cosamples  = \zero$ since $r_\cosamples$ is a free variable. With this at hand, we proceed as
\begin{align*}
g\tran \xopt &\geq \max_{\linf{ r_\samples - \meas} \leq \vareps} g\tran_\samples r_\samples = \max_{\linf{ r_\samples - \meas} \leq \vareps} \sum_{i \in \samples} g_i r_i \\
& = \max_{\linf{ r_\samples - \meas} \leq \vareps} \left( \sum_{\small g_i \geq 0, \ i \in \samples } |g_i| r_i - \sum_{\small g_i < 0, \ i \in \samples } |g_i| r_i \right) \\
& = \sum_{\small g_i \geq 0, \ i \in \samples } |g_i| (\meas_i + \vareps) - \sum_{\small g_i < 0, \ i \in \samples } |g_i| (\meas_i - \vareps) \\
& = \sum_{\small g_i \geq 0, \ i \in \samples } |g_i| \meas_i  - \sum_{\small g_i < 0, \ i \in \samples } |g_i| \meas_i  + \vareps \sum_{i \in \samples} |g_i| \\
& = g\tran_\samples \meas + \vareps \|g_\samples\|_1.
\end{align*}
Rearranging terms yields 
\begin{align}\label{eq:gCondition}
\|g_\samples\|_1 \leq \frac{g\tran_\samples (\xvar_\samples^\star - \meas)}{\vareps}.
\end{align}
We can then write the solution set as 
\begin{align}
\label{eq:appSS_robust}
\SS =  \{ \xopt &:& \exists u \in \Real{n-2}, g \in \Real{n}, \text{ such that } \nonumber \\
&& \tv\tran u + g = \zero, \nonumber \\
&& u_\support = \sign(\tv \xopt)_\support, \;\; \linf{u_\cosupport} \leq 1, \nonumber \\
&& \|g_\samples\|_1 \leq \frac{g\tran_\samples (\xvar_\samples^\star - \meas)}{\vareps}, \nonumber \\
&& g\tran_\cosamples  = \zero  \}.
\end{align}
The conditions $\tv\tran u + g = \zero$ and $g\tran_\cosamples  = \zero$ can be cast as $(\tv\tran)_\cosamples \; u = \zero$ and $g_\samples = -(\tv\tran)_\samples \; u$. Then inequality in \eqref{eq:gCondition} must be an equality since $|\xvar_\samples^\star - \meas| \leq \vareps$ 
(hence $\frac{\linf{\xvar_\samples^\star - \meas}}{\vareps} \leq 1$) and $\|g_\samples\|_1 \doteq \max_{v:\linf{v}\leq 1} g\tran_\samples v$ by definition of the $\ell_1$-norm; 
% $\|g_\samples\|_1 = g\tran_\samples \sign(g_\samples)$ 
moreover if $|\xvar_i^\star - \meas_i| = \vareps$ for some $i \in \samples$ it must hold that 
\begin{align}\label{eq:equalsigns}
\sign((\tv\tran)_i \; u) = (\xvar_i^\star - \meas_i)/ \vareps. 
\end{align}
Here we used the assumption that $\vareps > 0$. Denote the set of such $i \in \samples$ as the active set $\calA$ as in \eqref{eq:activeset}. For $i \in \samples \setminus \calA$, it holds that $(\tv\tran)_i \; u = \zero$. Splitting $\calA$ into two subsets as in \eqref{eq:activesubset} and using \eqref{eq:equalsigns} yields the conditions \eqref{eq:robust_optimality} of \prettyref{prop:robust_subdifferential}.
\qed

%% file: appendixTex/appendix_noisy_signConsistency2D.tex
%!TEX root = ../main.tex

\section{Proof of~\prettyref{prop:scEnvelop}}
\label{proof:prop-noisyEnvelop}

\begin{figure}[htb]
\begin{tabular}{c}
\begin{minipage}{\textwidth}
\hspace{-3mm}
\begin{tabular}{cc}
\begin{minipage}{0.5\textwidth}
\centering
\includegraphics[width=\textwidth]{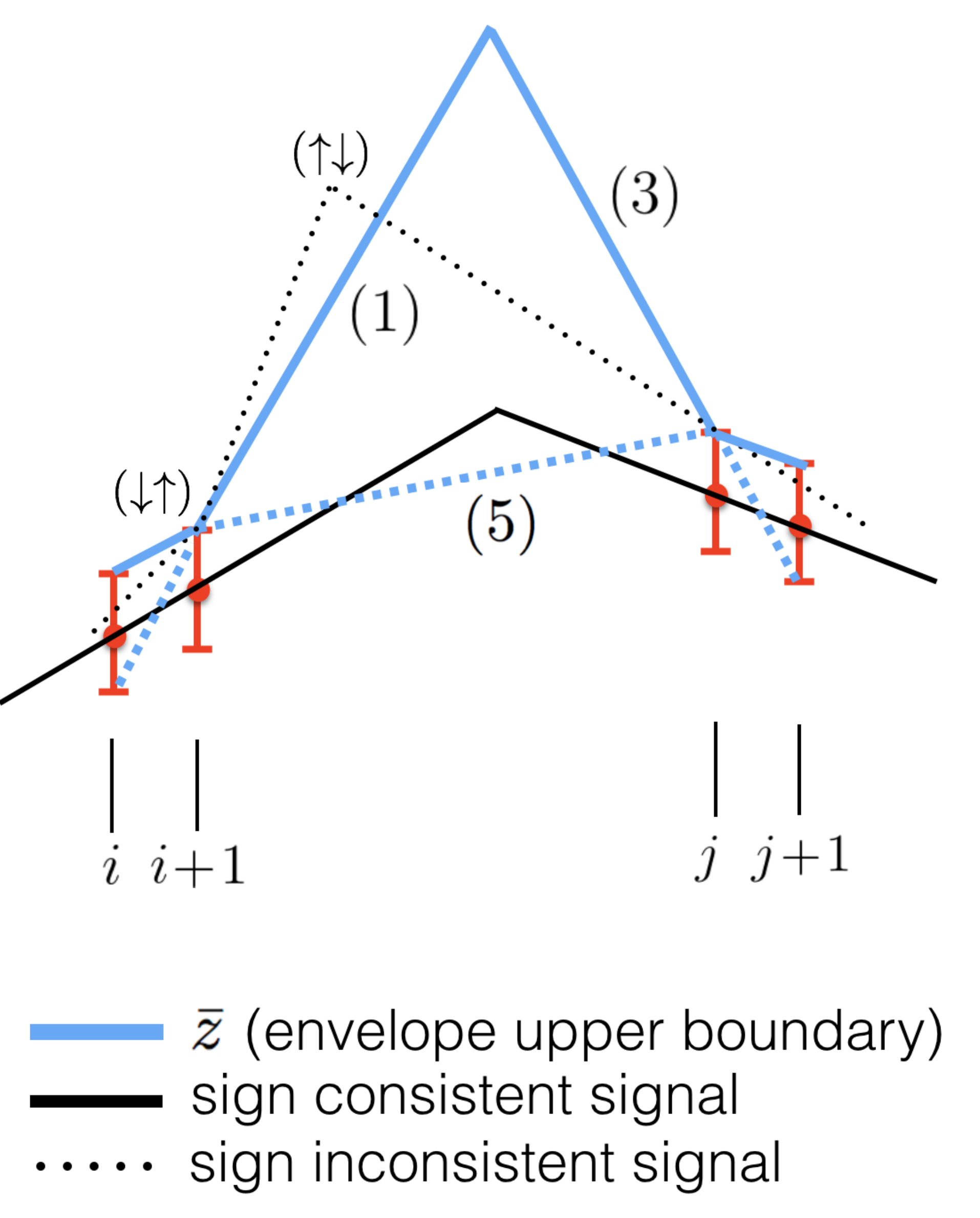} \\
\vspace{-1mm}
(a)
\end{minipage}
% &  \;\;
\begin{minipage}{0.5\textwidth}
\centering
\includegraphics[width=\textwidth]{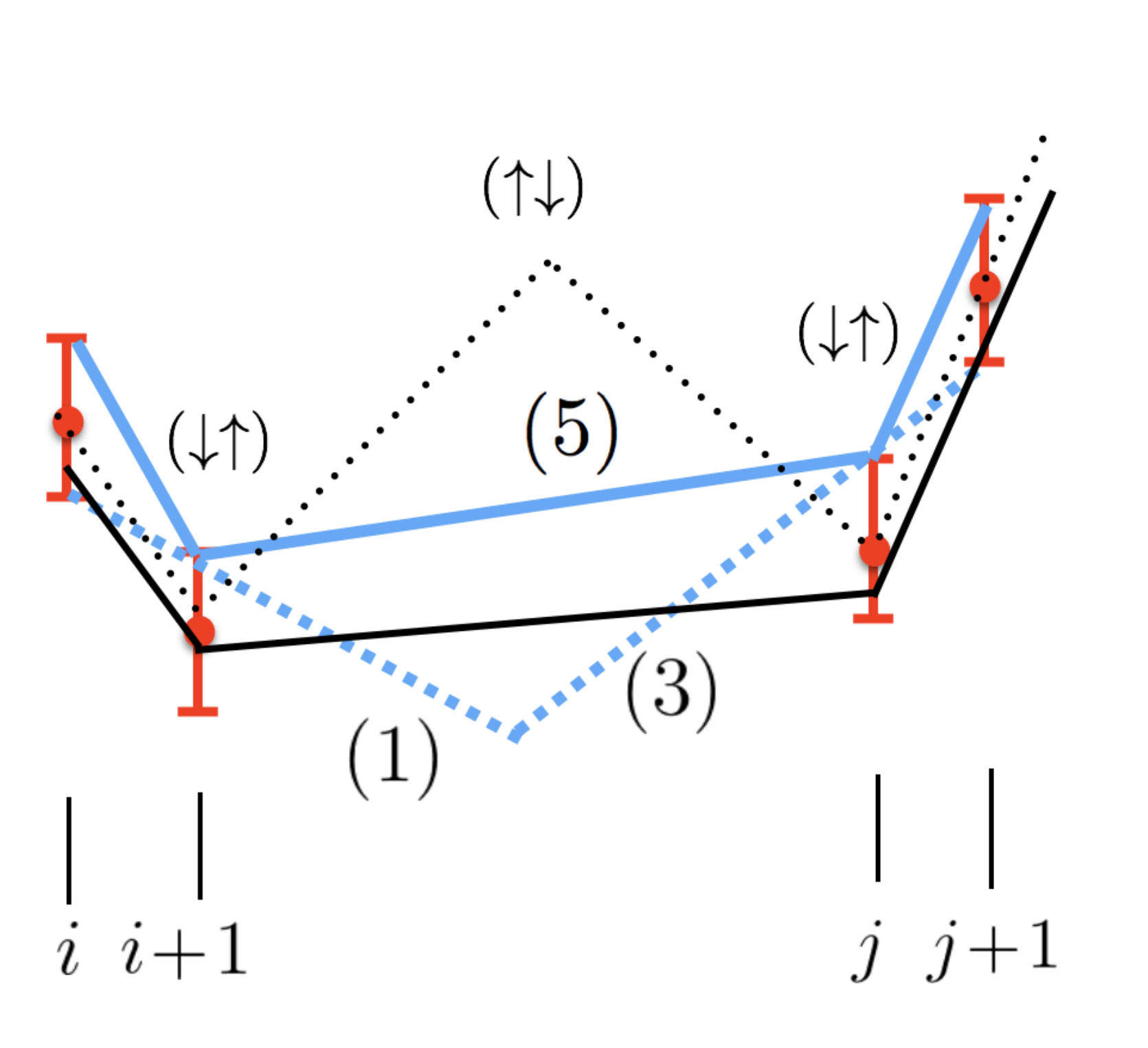} \\
\vspace{11mm}
(b)
\end{minipage}
\end{tabular}
\end{minipage}
\\
\begin{minipage}{\textwidth}
\hspace{-3mm}
\begin{tabular}{cc}
\begin{minipage}{0.5\textwidth}
\centering
\includegraphics[width=\textwidth]{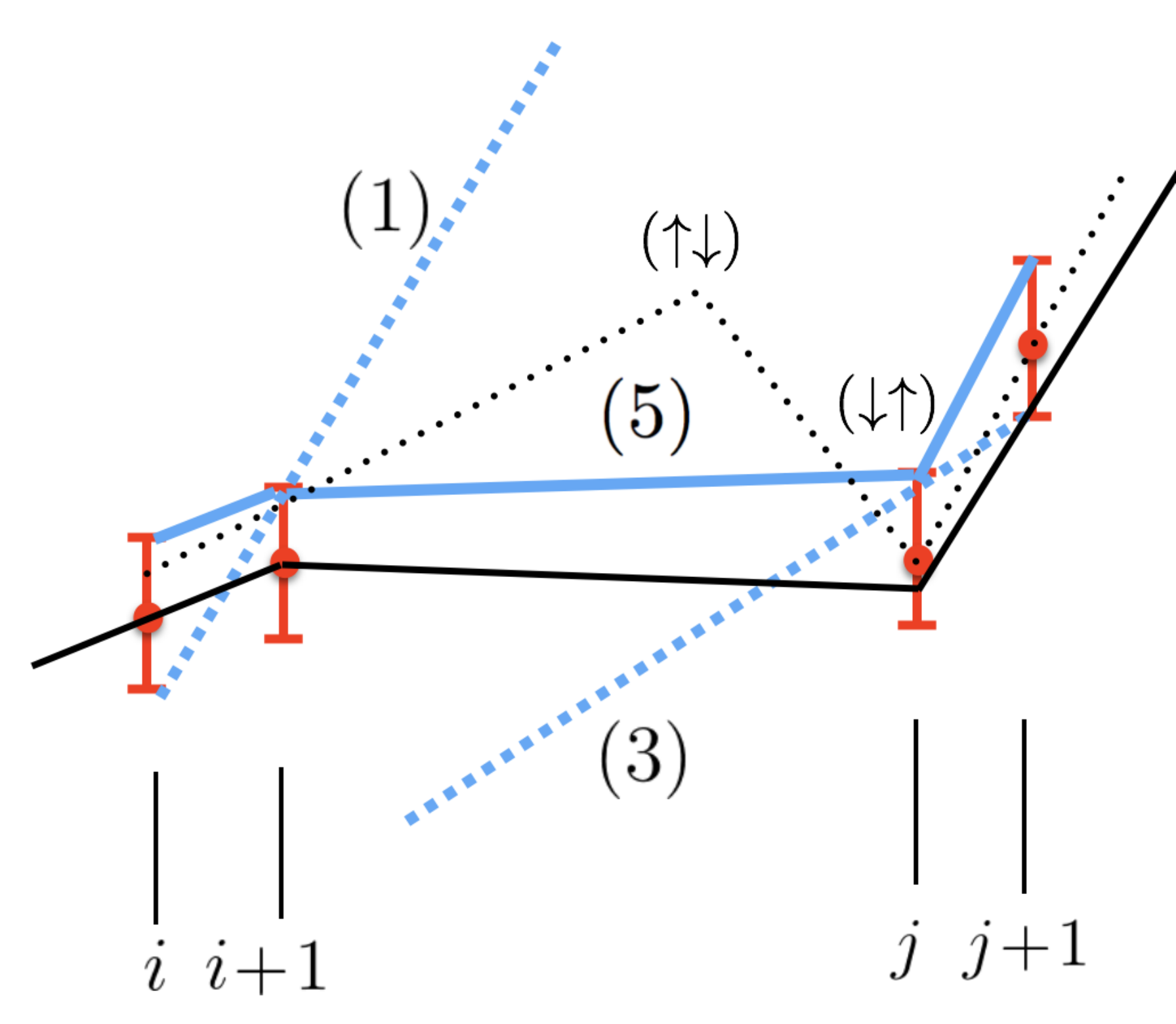} \\
\vspace{2mm}
(c)
\end{minipage}
% &  \;\;
\begin{minipage}{0.5\textwidth}
\centering
\includegraphics[width=\textwidth]{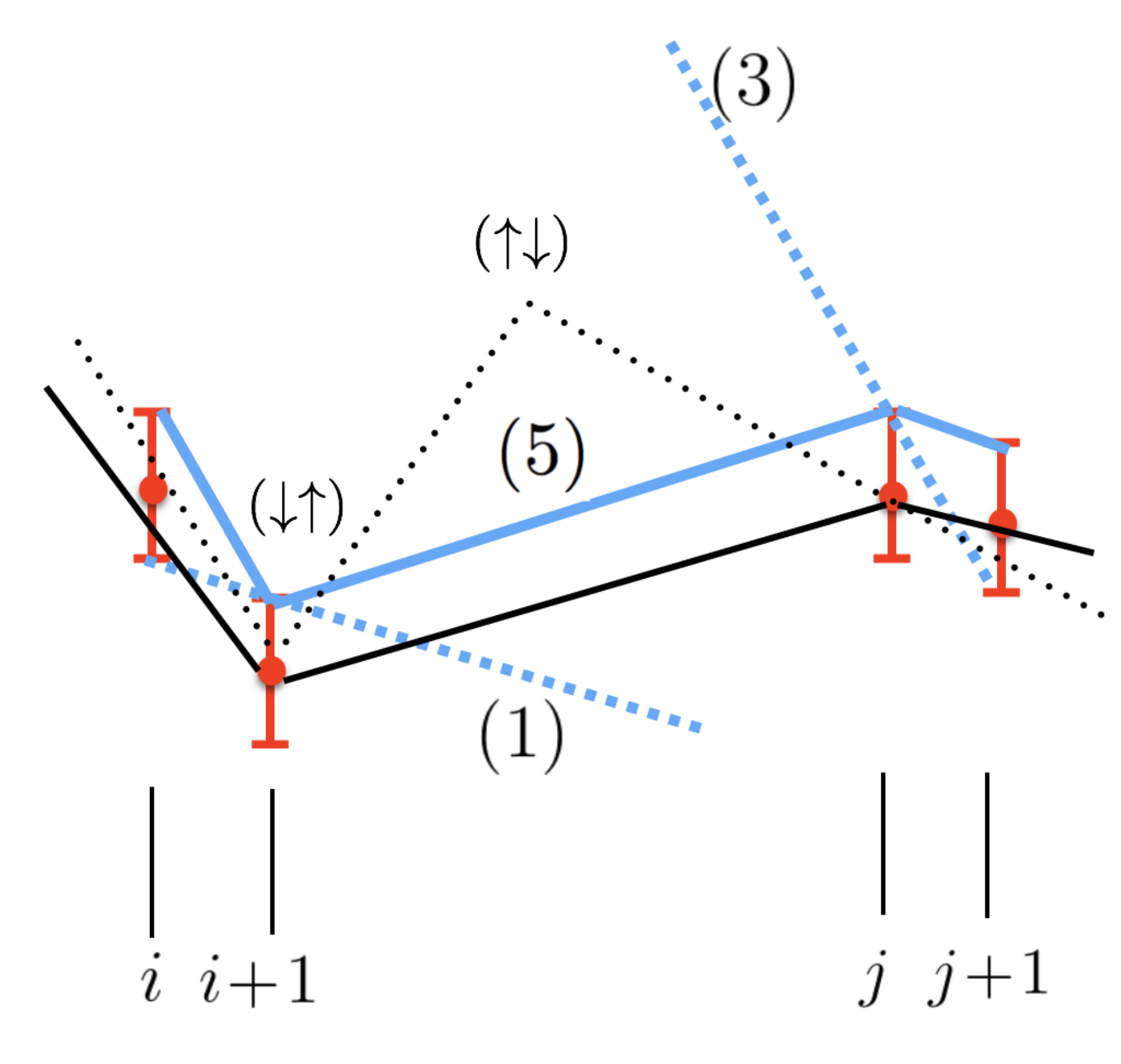} \\
\vspace{-1mm}
(d)
\end{minipage}
\end{tabular}
\end{minipage}
\end{tabular}
\caption{Illustration of the upper bound $\bar{\xvar}$ of \twd sign consistent \noisyEnvelop for all possible orientations of the line segments (1), (3) and (5) defined in \prettyref{def:scEnvelop}. 
% \LC{\noisyEnvelop in label}
% \FM{updated fig 8(b)}
}
\label{fig:envelopNoisyExamples} 
\end{figure}

To prove the claim of~\prettyref{prop:scEnvelop}, we  show that any \twd sign consistent \signal is bounded above by $\bar{\xvar}$ and below by $\underbar{\xvar}$ as defined in \prettyref{def:scEnvelop}. We restrict the proof to demonstrate the validity of the upper bound, since the argument for the lower bound follows similarly. To this end, we focus on line segments (1), (3) and (5) that determine $\bar{\xvar}$. \prettyref{fig:envelopNoisyExamples}(a)-(d) illustrate 4 possible orientations of these line segments and the resulting upper bounds (in solid blue line). 

Let us start with the case of \prettyref{fig:envelopNoisyExamples}(a). We use a contradiction argument similarly to the proof of \prettyref{prop:1DrecoveryError}. We show that a \signal which is not upper bounded by the solid blue line, that is $\max\{(1),(3)\}$ in this case, has to be sign inconsistent (\prettyref{def:1DsignConsitency}). Indeed such a \signal needs to have a nonzero curvature at sample either $i+1$ or $j$ since the line segments (1) and (3) represent the extreme slopes that a \signal can have with zero curvature at the end points. Assume without loss of generality, a \signal (dashed black line in \prettyref{fig:envelopNoisyExamples}(a)) that has positive curvature at $i+1$ (label ($\downarrow \uparrow$) in figure) and violates the upper bound. It is clear that this \signal cannot reach the $\epsilon$-interval (red bar in figure) at sample $j$ without having a negative curvature between $i+1$ and $j$; this violates sign consistency (case (i) in \prettyref{def:1DsignConsitency}), leading to contradiction.

Next we prove the claim for \prettyref{fig:envelopNoisyExamples}(b). In this case the upper bound $\bar{\xvar}$ is given by the line segment (5). Any \signal that is not upper-bounded by (5) needs to have a positive curvature ($\downarrow \uparrow$) between $i+1$ 
and $j$. It is obvious that such a \signal also needs to have a negative curvature at some other sample $k \in (i+1, j)$ to reach the $\epsilon$-interval at sample $j$, which leads to contradiction. 

Finally, observe that \prettyref{fig:envelopNoisyExamples}(c) and \prettyref{fig:envelopNoisyExamples}(d) are exactly symmetrical cases with different orientations for line segments (1) and (3). Therefore we consider only \prettyref{fig:envelopNoisyExamples}(c) here. Similarly to before, a \signal (dashed black line in figure) that is not upper bounded by (5) needs to have a positive curvature between $i+1$ 
and $j$. However in order to reach the $\epsilon$-bar at sample $i+1$, this \signal must have a negative curvature at some other sample $k \in (i+1, j)$, which contradicts sign consistency. 

We proved the claim of the proposition for all possible cases in \prettyref{fig:envelopNoisyExamples} which ends our proof. \qed

%% file: appendixTex/appendix_noisy_errorBound2D.tex
%!TEX root = ../main.tex 

\section{Proof of~\prettyref{prop:1DrecoveryError_robust}}
\label{proof:prop-1DrecoveryError_robust}

From \prettyref{thm:1Doptimality_robust} we learn that any optimal solution $\xopt$ must be \twd sign consistent, 
implying that it lies within the \twd sign consistent \noisyEnvelop. By the assumptions of \prettyref{prop:1DrecoveryError_robust} that we sample the boundary of $\xtrue$ and the sample set includes a twin sample in  each linear segment in $\xtrue$, it follows that the ground truth \signal $\xtrue$ is also sign consistent (see discussion in Appendix~\ref{proof:1DrecoveryError}), hence belongs to the \twd sign consistent \noisyEnvelop as a result of  \prettyref{prop:scEnvelop}. 

Since both $\xopt$ and $\xtrue$ lie inside the same \twd envelope, the difference between $\xoptInd{k}$ and $\xtrue_k$ for arbitrary $k \in (i+1, j)$ is bounded by the difference between the top and the bottom of the envelope, $\bar{\xvar}_k - \underbar{\xvar}_k$. 
\qed
% Therefore, 
% \beq
% |\xtrue_{k} - \xoptInd{k}| \leq \bar{\xvar}_k - \underbar{\xvar}_k \leq d_i \cos(\theta_i) + d_j \cos(\theta_j) + 2 \vare,
% \eeq
% Where the second inequality is derived using simple trigonometric arguments as shown in \prettyref{fig:envelopNoisy}(b).
% Monotonicity of the parameters $d_i, d_j, \cos(\theta_i), \cos(\theta_j)$ also follows from simple geometric arguments.

%% file: appendixTex/appendix_algorithm.tex
%!TEX root = ../main.tex

\section{Proof of~\prettyref{cor:algorithm}}
\label{proof:cor-algorithm}
In the proof of \prettyref{prop:1DrecoveryError} (Appendix~\ref{proof:1DrecoveryError})
 we have seen that the objective of \eqref{eq:BP} is separable as in 
\eqref{eq:separability}, where $S_k$ includes all indices falling in the region between two consecutive twin samples (e.g., 
if the twin samples are $(i-1,i)$ and $(j,j+1)$, then $S_k = \{i-1,i,i+1,\ldots,j-1,j,j+1\}$). This allows us to study the performance of \prettyref{alg:1} independently for each region $S_k$. As a result of \prettyref{prop:1DrecoveryError}, line~\ref{line:solve1} in \prettyref{alg:1} produces a solution $\xopt$ between $\xnaive$ and $\xtrue$. Assume without loss of generality that $\xopt$ is concave in the region $S_k$, e.g., the leftmost corner of~\prettyref{fig:nam1}(b),
%\citep[Fig.~\ref{fig:nam1}(b)]{ma2016sparse}, 
which yields $s_k = -1$ in line~\ref{line:signs3}. Then  $\xnaive_i \leq \xopti \leq \xtrue_i $ for all $i \in S_k$ which implies
\begin{align}\label{eq:sumIneq}
\sum_{i \in S_k} \xopti \leq \sum_{i \in S_k} \xtrue_i
\end{align} 
Now we note that maximizing $\sum_{i \in S_k} \xvar_i$ is the same as minimizing $- \sum_{i \in S_k} \xvar_i$.
Therefore, when $s_k=-1$, line~\ref{line:solve2} maximizes the objective $\sum_{i \in S_k} \xvar_i$ subject to $\xvar \in \SS$ where $\SS$ is the optimal set of 
\eqref{eq:BP}.
Since \eqref{eq:sumIneq} holds for all $\xvar \in \SS$, and since $\xtrue \in \SS$, \prettyref{alg:1} returns $\xtrue$. 
\qed

%% file: appendixTex/appendix_nesta.tex
%!TEX root = ../main.tex

\section{Computation for \NESTA}
\label{app:app_nesta}

In this appendix we provide some technical details as well as closed-form expressions related to \prettyref{alg:nesta}. 

The function $f_\mu$ in \eqref{eq:fmu} 
is shown to be differentiable by Nesterov \citep{Nesterov05} and has gradient 
\begin{align}\label{eq:nabla_fmu}
\nabla f_\mu(\xvar) = \tv\tran u^\star(\xvar)
\end{align}
where $u^\star(\xvar)$ is the optimal solution of the maximization in \eqref{eq:fmu} and, for any given $\xvar$, 
can be computed as
\begin{align}\label{eq:ustar}
u^\star(\xvar) = \left\{ \ba{ll} \mu^{-1} (\tv\xvar)_i, & \text{if } |(\tv\xvar)_i| < \mu \\
\sign(\tv\xvar)_i, & \text{otherwise}. \ea \right.
\end{align}
The gradient $\nabla f_\mu(\xvar)$ is said to be Lipschitz with Lipschitz constant $L_\mu$ if it obeys 
$$
\|\nabla f_\mu(x) - \nabla f_\mu(\xvar)\|_2 \leq L_\mu \|x - \xvar\|_2.
$$
The constant $L_\mu$, used in \prettyref{alg:nesta}, is shown to be $L_\mu = \frac{\|\tv\|^2}{\mu}$ in \citep[eq (3.4)]{BeckerBC11}, where $\|\cdot\|$ denotes the spectral norm of a matrix. 

Next we provide closed-form solutions for the optimization problems in
 lines~\ref{line:xk}-\ref{line:yk} of \prettyref{alg:nesta}. We start with $\xnesta$, which is the 
 solution of the optimization problem in line~\ref{line:xk}. 
 %A bit of algebra yields 
 Eliminating constant terms in the objective in line~\ref{line:xk} and completing the squares yields:
\begin{align}\label{eq:xk1}
\xnesta =& \argmin_{\xvar \in \calQ} \; \frac{L_\mu}{2} \|\xvar - \xvar\at{k}\|_2^2 + \la \ccc, \xvar \ra  \notag \\
=& \argmin_{\xvar \in \calQ} \| \xvar - (\xvar\at{k} - L_\mu^{-1} \ccc )\|_2^2 
\end{align}
where $\ccc \doteq \nabla f_\mu(\xvar\at{k})$ and $ \calQ \doteq \{\xvar: \linf{\matU \xvar - \meas} \leq \vareps \}$.
 For simplicity we introduce the vector $\hat{\xvar}\at{k} \doteq \xvar\at{k} - L_\mu^{-1} \ccc$, and recall that the sampling matrix $\matU$ restricts a vector to its entries at indices in $\samples$, hence $\calQ = \{\xvar: \linf{\xvar_\samples - \meas} \leq \vareps \}$. 
We then notice that the objective in~\eqref{eq:xk1} separates as:
\bea\label{eq:xk2}
\xnesta =
\displaystyle \argmin_{\xvar_\samples \in \calQ} \| [ \xvar - \hat{\xvar}\at{k} ]_\samples \|_2^2 %\nonumber \\
%+& \displaystyle \argmin_{\xvar_\cosamples} 
+ \| [ \xvar - \hat{\xvar}\at{k} ]_\cosamples \|_2^2 
\eea
which further separates into two independent optimization problems involving 
subvectors of $\xnesta$:
%allows to write~\eqref{eq:xk1} as two separate optimization problems: 
\bea
\xnesta_\samples = \argmin_{\xvar_\samples \in \calQ} \| [ \xvar - \hat{\xvar}\at{k} ]_\samples \|_2^2 \label{eq:xk3a} \\
\xnesta_\cosamples = \argmin_{\xvar_\cosamples} \| [ \xvar - \hat{\xvar}\at{k} ]_\cosamples \|_2^2  \label{eq:xk3b} 
\eea
Problem~\eqref{eq:xk3b} is an unconstrained minimization, since $\calQ$ only constraints $\xvar_\samples$.
By inspection, problem~\eqref{eq:xk3b} admits the trivial solution
$\xnesta_\cosamples = \hat{\xvar}\at{k}_\cosamples$.
 % This suggests that the optimal solution $\xnesta$ must agree with $\xvar\at{k}$ on $\cosamples$, i.e., $\xnesta_\cosamples = \hat{\xvar}\at{k}_\cosamples $. 
 It remains to solve~\eqref{eq:xk3a}: 
\begin{align}\label{eq:xk4}
\xnesta_\samples = \argmin_\zeta \| \zeta - \hat{\xvar}\at{k}_\samples \|_2^2 \quad \subject \linf{\zeta - \meas} \leq \vareps
\end{align}
where we ``renamed'' the optimization variable to $\zeta$ to simplify notation.
 Problem~\eqref{eq:xk4}  is nothing but the Euclidean projection of the vector $\hat{\xvar}\at{k}_\samples $ onto the $\ell_\infty$-ball with radius $\vareps$ centered at $\meas$. %Assuming that $\samples = \{1,2,\ldots,|\samples|\}$ without loss of generality, 
This projection can be explicitly calculated as:
\begin{align*}
\xnesta_i = \min\{ \max\{\hat{\xvar}\at{k}_i, \meas_i - \vareps \}, \meas_i + \vareps \} , \ \ \forall i \in \samples.
\end{align*}
% \begin{align*}
% \xnesta_i = \meas_i + \left\{ \ba{cl} \vareps, & \text{if } r_i > \vareps \\
% -\vareps, & \text{if } r_i < -\vareps \\
% r_i, & \text{if } |r_i| \leq \vareps \ea \right.  , \ \ \  i \in \samples
% \end{align*}
% where $r= \hat{\xvar}\at{k}_\samples - \meas$. 
which completes the derivation of the closed-form solution for $\xnesta$.
 The reader may notice that the optimization problem in line \ref{line:yk} of \prettyref{alg:nesta}, whose solution is $\wnesta$, 
 is identical to the one in \eqref{eq:xk1}, after replacing $\ccc = \sum_{i = 0}^k \alpha_i \nabla f_\mu(\xvar\at{i})$ and using $\xvar\at{0}$ instead of $\xvar\at{k}$. 
 %Note that $\xvar\at{0}$ is reset to a different value at every iteration of the inner \textit{for} loop in \NESTA solver. 
 For this reason the closed-form solution $\wnesta$ can be derived in full similarity with the one of $\xnesta$, 
 and can be explicitly written as:
 \bea
 \wnesta_\cosamples &=& \breve{\xvar}\at{k}_\cosamples \nonumber \\
 \wnesta_i &=& \min \{ \max \{ \breve{\xvar}\at{k}_i, \meas_i - \vareps \}, \meas_i + \vareps \} , \ \ \forall i \in \samples,
\nonumber
 \eea
 where $\breve{\xvar}\at{k} \doteq \xvar\at{0} - L_\mu^{-1} \sum_{i = 0}^k \alpha_i \nabla f_\mu(\xvar\at{i})$.

%% file: supplementalTex/supp_gazeboImages.tex
\section{Extra Visualizations: Gazebo Depth Images}
\label{sec:gazeboImg}

\renewcommand{\spac}{\hspace{-0.0cm}}

\newcommand{\imgIDrgb}{13}
\newcommand{\imgID}{13}
\begin{figure*}[h!]
\begin{minipage}{\textwidth}
\centering
\begin{tabular}{cccccc}\input{supplementalTex/supp_gazeboTool}
\end{tabular}
\end{minipage}\end{figure*}

\renewcommand{\imgIDrgb}{06}
\renewcommand{\imgID}{6}
\begin{figure*}[h!]
\begin{minipage}{\textwidth}
\centering
\begin{tabular}{cccccc}\input{supplementalTex/supp_gazeboTool}
\end{tabular}
\end{minipage}\end{figure*}

\renewcommand{\imgIDrgb}{03}
\renewcommand{\imgID}{3}
\begin{figure*}[h!]
\begin{minipage}{\textwidth}
\centering
\begin{tabular}{cccccc}\input{supplementalTex/supp_gazeboTool}
\end{tabular}
\end{minipage}\caption{
Gazebo: 3 examples of reconstructed depth \signals using the proposed approaches (\lmin, \lminDiag) and a naive 
linear interpolation (\naive). For each example we show the reconstruction from 2\% uniformly drawn depth measurements.
We also show the reconstruction for the case in which we can access the depth corresponding to (appearance) edges in the RGB images. 
}
% \label{fig:imagesGAZEBO}
\end{figure*}

%% file: supplementalTex/supp_gazeboTool.tex
&  &  & \multicolumn{3}{c}{Reconstructed \signals}
\\
\cline{4-6}
\vspace{-3mm}
\\
&  Depth \signal & samples  & \naive  & \lmin  & \lminDiag
\\
\rotatebox[origin=c]{90}{uniform \hspace{-1.8cm}}
& 
\spac
\includegraphics[height=2.15cm]{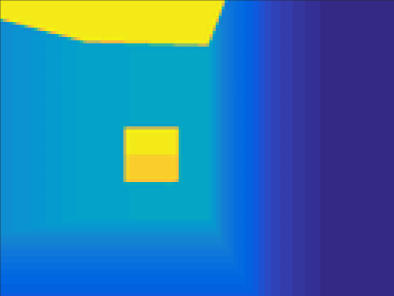}
& \spac
\includegraphics[height=2.15cm]{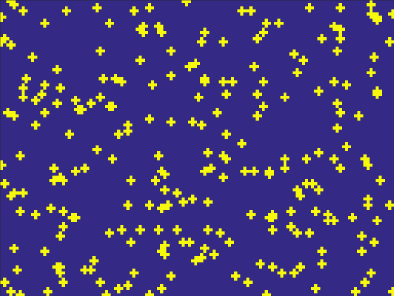}  
& \spac
\includegraphics[height=2.15cm]{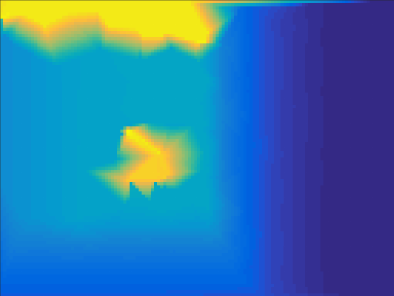} 
& \spac
\includegraphics[height=2.15cm]{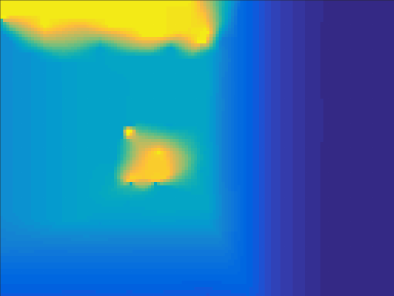}  
& \spac
\includegraphics[height=2.15cm]{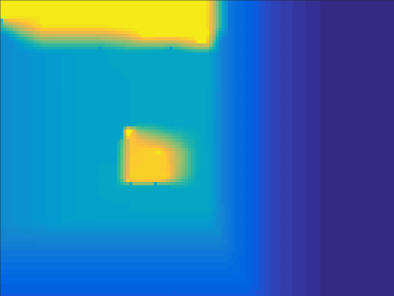}   
\vspace{-1mm}
\\
& \thd depth & 2\% samples &  &  & 
\\
\rotatebox[origin=c]{90}{RBG edges \hspace{-1.8cm}}
& 
\spac
\includegraphics[height=2.15cm]{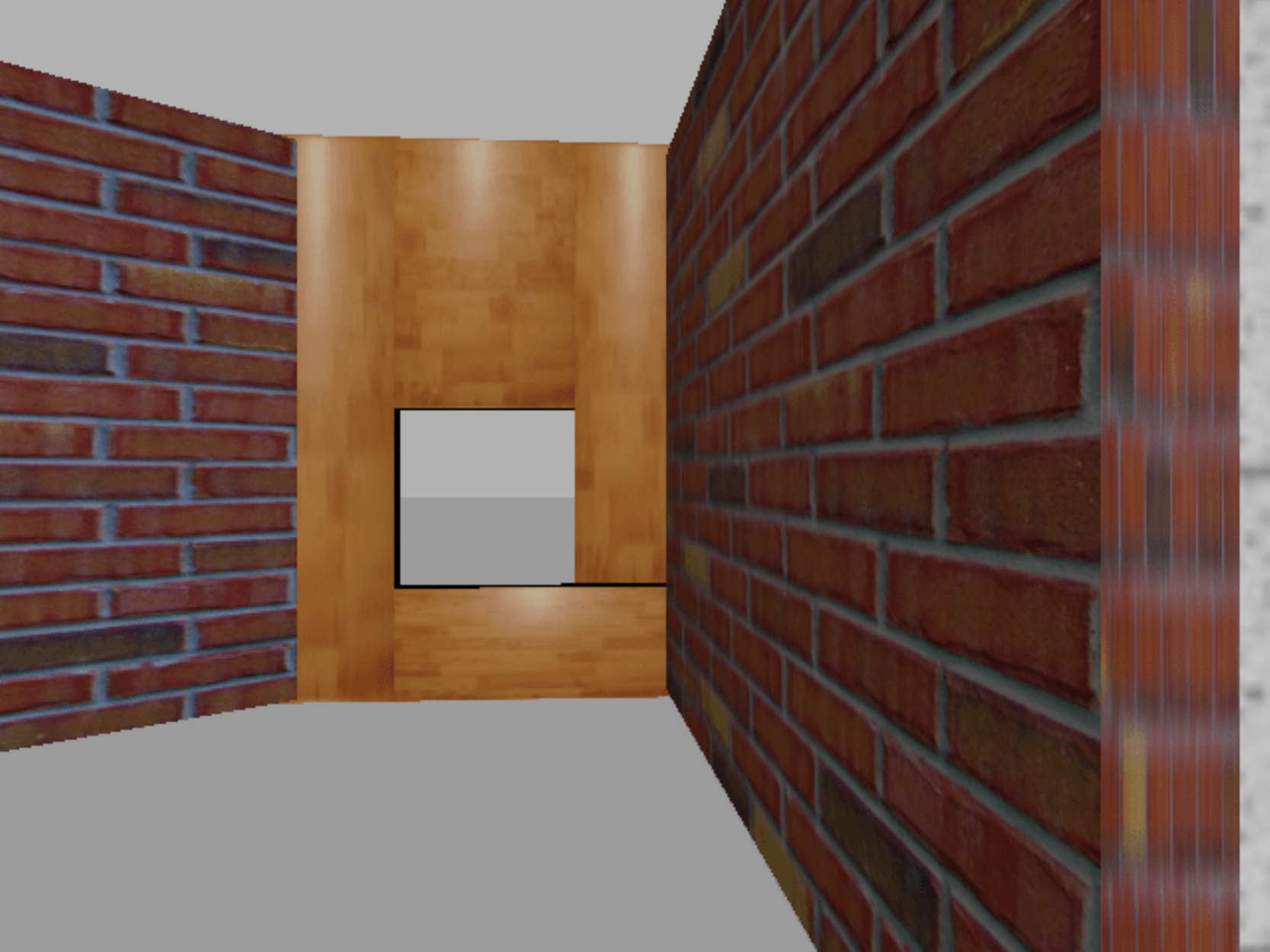} 
& \spac
\includegraphics[height=2.15cm]{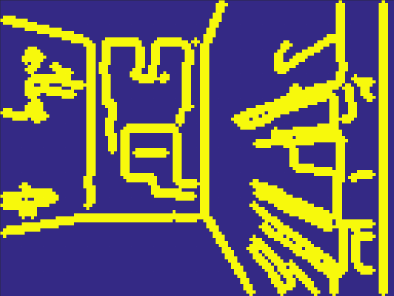} 
& \spac
\includegraphics[height=2.15cm]{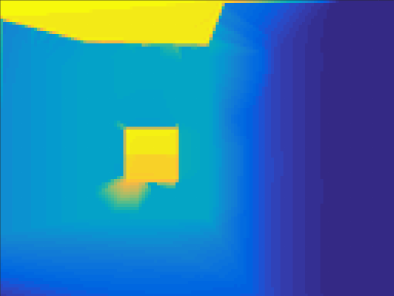}  & \spac
\includegraphics[height=2.15cm]{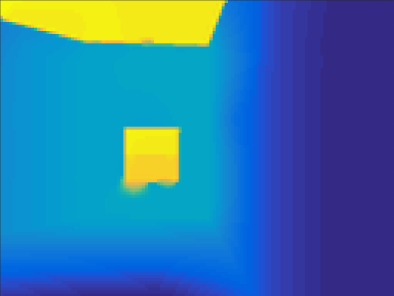}  & \spac
\includegraphics[height=2.15cm]{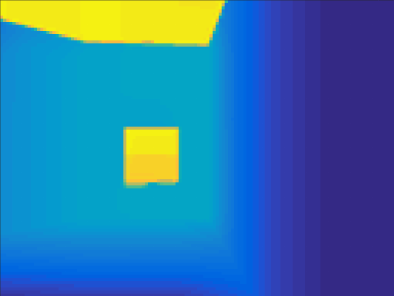}  \vspace{-1mm}
\\
& RGB image & RGB edges &  &  & 

%% file: supplementalTex/supp_zedImages.tex
\section{Extra Visualizations: ZED Depth Images}
\label{sec:zedImg}

\renewcommand{\spac}{\hspace{-0.0cm}}

\renewcommand{\imgID}{400}
\renewcommand{\imgIDrgb}{0\imgID}
\begin{figure*}[ht!]
\begin{minipage}{\textwidth}
\centering
\begin{tabular}{cccccc}\input{supplementalTex/supp_zedTool}
\end{tabular}
\end{minipage}\end{figure*}

\renewcommand{\imgID}{750} \renewcommand{\imgIDrgb}{0\imgID}
\begin{figure*}[ht!]
\begin{minipage}{\textwidth}
\centering
\begin{tabular}{cccccc}\input{supplementalTex/supp_zedTool}
\end{tabular}
\end{minipage}\end{figure*}

\renewcommand{\imgID}{550} \renewcommand{\imgIDrgb}{0\imgID}
\begin{figure*}[ht!]
\begin{minipage}{\textwidth}
\centering
\begin{tabular}{cccccc}\input{supplementalTex/supp_zedTool}
\end{tabular}
\end{minipage}\end{figure*}

\renewcommand{\imgID}{1100} \renewcommand{\imgIDrgb}{\imgID}
\begin{figure*}[ht!]
\begin{minipage}{\textwidth}
\centering
\begin{tabular}{cccccc}\input{supplementalTex/supp_zedTool}
\end{tabular}
\end{minipage}\end{figure*}

\renewcommand{\imgID}{1050} \renewcommand{\imgIDrgb}{\imgID}
\begin{figure*}[ht!]
\begin{minipage}{\textwidth}
\centering
\begin{tabular}{cccccc}\input{supplementalTex/supp_zedTool}
\end{tabular}
\end{minipage}\end{figure*}

\renewcommand{\imgID}{880} 
\renewcommand{\imgIDrgb}{0\imgID}
\begin{figure*}[ht!]
\begin{minipage}{\textwidth}
\centering
\begin{tabular}{cccccc}\input{supplementalTex/supp_zedTool}
\end{tabular}
\end{minipage}
\caption{
ZED: 6 examples of reconstructed depth \signals using the proposed approaches (\lmin, \lminDiag) and a naive 
linear interpolation (\naive). For each example we show the reconstruction from 2\% uniformly drawn depth measurements.
We also show the reconstruction for the case in which we can access the depth corresponding to (appearance) edges in the RGB images. 
}
% \label{fig:imagesZED}
\end{figure*}

%% file: supplementalTex/supp_zedTool.tex
&  &  & \multicolumn{3}{c}{Reconstructed \signals}
\\
\cline{4-6}
\vspace{-3mm}
\\
&  Depth \signal & samples  & \naive  & \lmin  & \lminDiag
\\
\rotatebox[origin=c]{90}{uniform \hspace{-1.8cm}}
& 
\spac
\includegraphics[height=2.15cm]{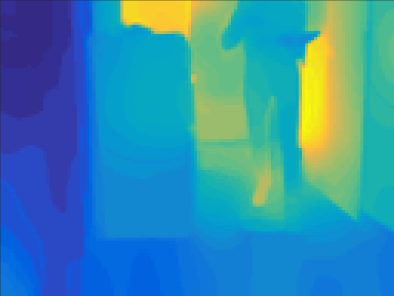}
& \spac
\includegraphics[height=2.15cm]{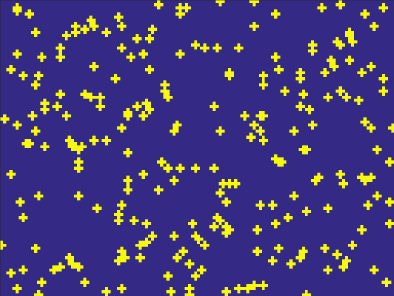}  
& \spac
\includegraphics[height=2.15cm]{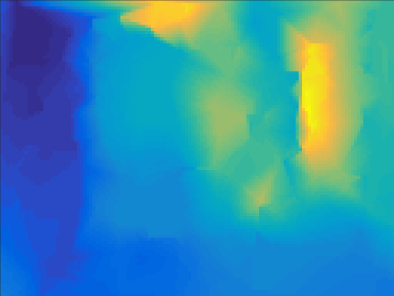} 
& \spac
\includegraphics[height=2.15cm]{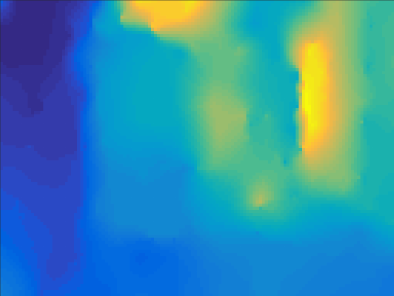}  
& \spac
\includegraphics[height=2.15cm]{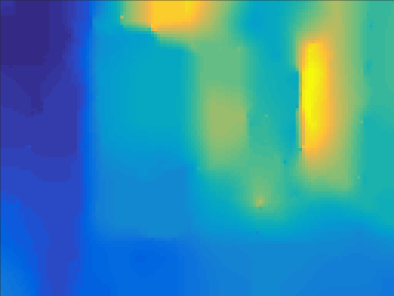}   
\vspace{-1mm}
\\
& \thd depth & 2\% samples &  &  & 
\\
\rotatebox[origin=c]{90}{RBG edges \hspace{-1.8cm}}
& 
\spac
\includegraphics[height=2.15cm]{rgb-zed/rgb_\imgIDrgb} 
& \spac
\includegraphics[height=2.15cm]{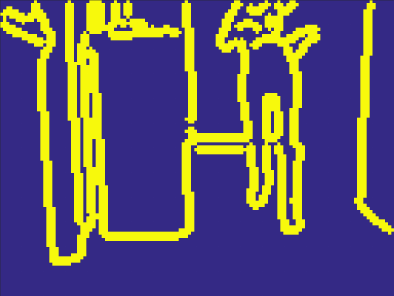} 
& \spac
\includegraphics[height=2.15cm]{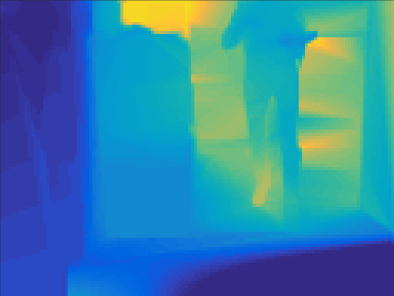}  & \spac
\includegraphics[height=2.15cm]{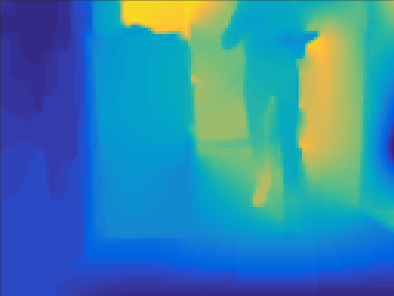}  & \spac
\includegraphics[height=2.15cm]{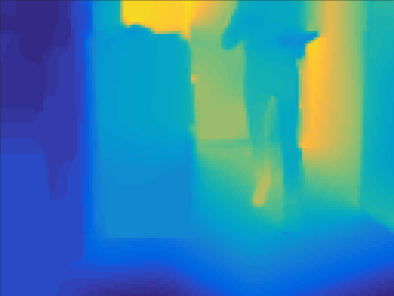}  \vspace{-1mm}
\\
& RGB image & RGB edges &  &  & 